\pdfoutput=1

\documentclass[11pt]{article}

\usepackage[final]{acl}

\usepackage{times}
\usepackage{latexsym}

\usepackage[T1]{fontenc}

\usepackage[utf8]{inputenc}
\usepackage{microtype}
\usepackage{inconsolata}
\usepackage{graphicx}
\usepackage{microtype}
\usepackage{amsmath}
\usepackage{graphicx}
\usepackage{amsfonts}
\usepackage{multirow}
\usepackage{listings}
\usepackage{url}
\usepackage{CJKutf8}
\usepackage{graphicx}
\usepackage{supertabular}
\usepackage{dcolumn,booktabs}
\usepackage{tikz}
\usepackage{xspace}
\usepackage{authblk}
\usepackage{graphicx}
\usepackage{subcaption} 
\usepackage{arydshln}
\usepackage{xcolor,colortbl}
\definecolor{forestgreen}{RGB}{34, 139, 34} 
\definecolor{darkmagenta}{RGB}{139, 0, 139}
\definecolor{yellowgreen}{RGB}{154, 205, 50}
\definecolor{melon}{RGB}{253, 188, 180}
\definecolor{skyblue}{RGB}{135, 206, 235}
\definecolor{dodgerblue}{RGB}{30, 144, 255}

\title{On Relation-Specific Neurons in Large Language Models}

\author[1,2,*]{\bf Yihong Liu}
\author[3,*]{\bf Runsheng Chen}
\author[1,2]{\bf Lea Hirlimann}
\author[1,2]{\\ \bf Ahmad Dawar Hakimi}
\author[1,2,4]{\bf Mingyang Wang}
\author[1,2]{\bf Amir Hossein Kargaran}
\author[5,$\dag$]{\\ \bf Sascha Rothe}
\author[6,$\dag$]{\bf François Yvon}
\author[1,2,$\dag$]{\bf Hinrich Sch\"utze}

\affil[]{Center for Information and Language Processing, LMU Munich \protect\\ $^{2}$Munich Center for Machine Learning (MCML) \protect\\ $^{3}$Technical University of Munich \  $^{4}$Bosch Center for Artificial Intelligence \protect\\ $^{5}$Google DeepMind, Zürich, Switzerland \ $^{6}$Sorbonne Université, CNRS, ISIR, France
 \protect\\ \texttt{\{yihong, hirlimann\}@cis.lmu.de} \ \ \ \ \ \ \ \texttt{runsheng.chen@tum.de}} 

\def\shortpar#1{\textbf{#1}}
\def\secref#1{\S\ref{sec:#1}}
\def\seclabel#1{\label{sec:#1}}

\newcommand{\RelationSpecificNeurons}{\emph{RelSpec} neurons\xspace}

\newcounter{notecounter}
\newcommand{\enotesoff}{\long\gdef\enote##1##2{}}
\newcommand{\enoteson}{\long\gdef\enote##1##2{{
\stepcounter{notecounter}
{\large\bf
\hspace{0cm}\arabic{notecounter} $<<<$ ##1: ##2
$>>>$\hspace{1cm}}}}}

\enoteson
\enotesoff

\begin{document}
\maketitle

\def\thefootnote{*}\footnotetext{Equal contribution.}\def\thefootnote{\arabic{footnote}}
\def\thefootnote{$\dag$}\footnotetext{Equal advising.}\def\thefootnote{\arabic{footnote}}

\begin{abstract}


In large language models (LLMs), certain \emph{neurons} can store distinct pieces of knowledge learned during pretraining. 
While factual knowledge typically appears as a combination of \emph{relations} and \emph{entities}, it remains unclear whether some neurons focus on a relation itself -- independent of any entity. 
We hypothesize such neurons \emph{detect} a relation in the input text and \emph{guide} generation involving such a relation.
To investigate this, we study the LLama-2 family on a chosen set of relations, with a \textit{statistics}-based method.
Our experiments demonstrate the existence of relation-specific neurons.
We measure the effect of selectively deactivating candidate neurons specific
to relation $r$ on the LLM's ability to handle (1) facts involving relation $r$ and (2) facts involving a  
different relation $r' \neq r$.
With respect to their capacity for encoding relation
information, we give evidence for the following three properties
of relation-specific neurons.
\textbf{(i) Neuron cumulativity.}
Multiple neurons jointly contribute to processing facts involving relation $r$, with no single neuron fully encoding a fact in $r$ on its own.
\textbf{(ii) Neuron versatility.}
Neurons can be shared across multiple closely related as
well as less related relations. 
In addition, some relation neurons transfer across languages.
\textbf{(iii) Neuron interference.} 
Deactivating neurons specific to one relation can improve
LLMs' factual recall performance for facts of other relations.
We make our code and data publicly available at \url{https://github.com/cisnlp/relation-specific-neurons}.
\end{abstract}

\section{Introduction}

Large text corpora like Wikipedia contain abundant
factual knowledge. LLMs, pretrained
on such corpora, can function as knowledge bases that
retrieve information and generate text involving factual
content \citep{petroni-etal-2019-language,jiang-etal-2020-know}.
Recent studies suggest that some knowledge is parameterized
by
LLMs \citep{dai-etal-2022-knowledge,geva-etal-2023-dissecting},
especially within the feed-forward layers of the Transformer
architecture
\citep{transformer2017vaswani}, which act as
key-value memory \citep{geva-etal-2021-transformer}.
Factual knowledge is often expressed as a relational fact in
triple form: \emph{subject}, \emph{relation},
and \emph{object}, e.g.,
(\texttt{NVIDIA}, \texttt{company\_ceo}, \texttt{Jensen
Huang}). 
However, it remains unclear whether each fact is
stored and processed separately through \emph{knowledge
neurons} \citep{dai-etal-2022-knowledge}, i.e., neurons that are responsible for encoding each fact individually;
or whether there
exist \emph{relation-specific neurons} (referred to as \textbf{\RelationSpecificNeurons}), i.e.,
neurons that do not represent specific facts but rather focus on the
relation
and guide generating the object once
the subject and relation of a triple have been
detected. 

In this work, we examine the existence of \RelationSpecificNeurons in decoder-only LLMs.  Our study focuses on the
LLama-2 family (7B and
13B) \citep{touvron2023llama2openfoundation} and examines
factual knowledge grouped into 12 types of relations. To
pinpoint \RelationSpecificNeurons for these relations, we adopt the neuron identification
method proposed by \citet{neuron2022Cuadros}, which
identifies the neurons that are uniquely activated in one
group of sentences (positive examples) while not in another
(negative examples). \citet{kojima-etal-2024-multilingual}
successfully applied this method to uncover
\emph{language-specific neurons}.
Following this line of work, we construct
zero-shot prompts featuring a specific relation for the
positive examples and prompts with other relations for the
negative examples. 
Neurons whose activation patterns are positively correlated with positive examples are regarded as \RelationSpecificNeurons.

To understand the impact of \RelationSpecificNeurons, we perform 
factual recall on
held-out prompts. 
These prompts
for each relation share the \textbf{same relation} as the positive
examples used for neuron identification but have \textbf{no entity overlap};
this disentangles the effects of entities
and relations.  
For each relation, we compare performance
between the original model and the model in which \RelationSpecificNeurons
for that relation are deactivated
-- \emph{intra-relation results}.  
We also study how deactivating neurons for one relation influences performance on others -- \emph{inter-relation results}. 
Our experiments reveal
several key properties of \RelationSpecificNeurons:

\textbf{Neuron cumulativity}.
\RelationSpecificNeurons present a cumulative effect -- a phenomenon where an LLM distributes relational knowledge across multiple neurons. 
\RelationSpecificNeurons jointly contribute to dealing with facts belonging to a relation, with no single neuron fully encoding a fact on its own.
This property aligns with the evidence of the existence of redundant and self-repair neurons \citep{dalvi-etal-2020-analyzing,mcgrath2023hydraeffectemergentselfrepair,he2024matterstransformersattentionneeded}.

\textbf{Neuron versatility}. 
As the total number of neurons is finite,
while the number of possible relations is vast, some \RelationSpecificNeurons strongly
associate with multiple relations.  Surprisingly, these
relations need not be closely linked -- two weakly related
relations can share a group of neurons, leading to
performance drops in both relations if those neurons are
deactivated. 
\RelationSpecificNeurons also generalize across
languages -- \RelationSpecificNeurons identified from English have a similar effect on other languages. 
This property aligns with neuron poly\-semanticity and superposition \citep{mu2020compositional,elhage2022toymodelssuperposition,scherlis2025polysemanticity}.

\textbf{Neuron interference}. 
Some \RelationSpecificNeurons appear to ``confuse'' the
model when it processes other relations. Deactivating such
neurons can yield improved performance on these other
relations. This property aligns with broader evidence
that \emph{sub-networks} or \emph{circuits} within LLMs may
serve several
different  functional roles \citep{wang2023circuit,bayazit-etal-2024-discovering,mondorf2024circuit}.



\section{Methodology}\seclabel{method}

\subsection{Dataset Manipulation}\seclabel{manipulation}

We use the factual knowledge dataset from \citet{lre2024Hernandez} for this research, which contains 25 relations. Each relation has a different number of facts. 
Each fact can be represented as a \emph{subject-relation-object} triple $(s, r_i, o)$.
We only consider relations that have more than 300 facts
to ensure the reliability of our findings.
This results in \textbf{12} relations. We refer to the set of triples for relation $r_i$ as $\mathcal{D}_{r_i}$.
We then perform the following steps for each relation $r_i$ to construct the data used to identify its corresponding \RelationSpecificNeurons.

\shortpar{Step 1: Creating Evaluation Data.} For each triple set $\mathcal{D}_{r_i}$, we randomly select \textbf{50 triples} as a held-out set for evaluation (cf.\ \secref{controlled}). 
We refer to the selected triples as
$\mathcal{D}_{r_i}^{\text{eva}}$ (for evaluation) and all
other triples as $\mathcal{D}_{r_i}^{\text{det}}$ (for detection).
To ensure disjointness, $\mathcal{D}_{r_i}^{\text{eva}}$ and $\mathcal{D}_{r_i}^{\text{det}}$ do not share any subjects.

\shortpar{Step 2: Formulating Prompts.} For each triple $(s, r_i, o)$ in $\mathcal{D}_{r_i}^{\text{det}}$, we create prompts containing the \textbf{subject} $s$ and the \textbf{relation} $r_i$ using the templates provided by \citet{lre2024Hernandez}.
 \textbf{Note that the object $o$ is not included in the prompt}.
For example, we construct a prompt ``\textit{The CEO of NVIDIA is? Answer:}'' for the triple $(\texttt{NVIDIA}, \texttt{company\_CEO}, \texttt{Jensen Huang})$ with an expected answer ``\textit{Jensen Huang}''.
We also create prompts for $\mathcal{D}_{r_i}^{\text{eva}}$ in the same way. 
We refer to the resulting prompt sets as $\mathcal{P}_{r_i}^{\text{det}}$ and $\mathcal{P}_{r_i}^{\text{eva}}$.

\shortpar{Step 3: Validating Prompts.} 
We hypothesize that the model will leverage \RelationSpecificNeurons to generate the correct answer, i.e., the object.
Therefore, such neurons should ``fire'' for those prompts for which \textbf{the model answers correctly}.
For the prompt selection,
we feed each prompt in $\mathcal{P}_{r_i}^{\text{det}}$ to
the model and set the maximum generation length to be
2.\footnote{Some prior studies evaluate correctness by only
checking the model's first predicted
token \citep{geva-etal-2023-dissecting,lre2024Hernandez}. This
evaluation can be ambiguous if the answer/object is split
into multiple tokens. Considering 2 predicted tokens
increases reliability.}
We then check if the predicted 2 tokens are a prefix of the object: 
if they are, we regard the output as being correct.
We exclude prompts that the model answers wrongly from $\mathcal{P}_{r_i}^{\text{det}}$.\footnote{
We exclude prompts that do not yield the correct answer in order to maintain high precision in identifying \RelationSpecificNeurons. 
While the exclusion seems conservative, it helps preserve the clarity and discriminative power of the method.}


\subsection{Relation-Specific Neuron Identification}\seclabel{identification}

This work's purpose is to identify \textbf{\RelationSpecificNeurons} -- neurons that solely focus on the relation rather than specific relational facts concerning the subject-relation-object triple.
Therefore, these neurons are different from \emph{knowledge
neurons} (which encode certain facts) or \emph{entity
neurons} (which encode certain subject entities).
Following \citet{neuron2022Cuadros}, we identify \RelationSpecificNeurons using statistical association measures. This method
assigns a score for each neuron, representing its level of ``expertise'' in \textbf{distinguishing a specific relation from other considered relations}.

\shortpar{Defining Neurons.} A neural network, or specifically a Transformer \citep{transformer2017vaswani}, consists of many weight matrices. 
For a given weight matrix $\boldsymbol{W}\in\mathbb{R}^{d_1\times d_2}$, we define a neuron as a column,
mapping a representation from $\mathbb{R}^{d_1}$ to $\mathbb{R}$. 
We assign a unique index $m \in M$ to each neuron and investigate its output value.
We only consider the neurons in feed-forward networks (FFNs), i.e., neurons in \texttt{up\_proj}, \texttt{gate\_proj}, and \texttt{down\_proj}, since previous studies have shown that knowledge is mostly stored there \citep{dai-etal-2022-knowledge}. 
We also investigate neurons in other modules, e.g.,
attention heads, but find they are less relation-specific
(see \secref{neuron_type}).

\shortpar{Grouping Prompts.} 
For each relation $r_i$, we collect positive and negative examples. Specifically, we regard $\mathcal{P}_{r_i}^{\text{det}}$ as positive examples and randomly sample $4 \times |\mathcal{P}_{r_i}^{\text{det}}|$ prompts from
the prompt sets of other relations as negative
examples.\footnote{
Negative samples play an important role in identifying \RelationSpecificNeurons.
We restrict negative samples to counter-relation examples (i.e., samples from other relations) to ensure a controlled and interpretable comparison. 
In theory, the negative examples can also be in natural language. 
However, this would introduce a vast and unconstrained search space, possibly making it difficult to isolate the influence of relation-specific information.}
We refer to the positive and negative examples selected for relation $r_i$  as $\mathcal{E}^{+}_{{r_i}}$ and $\mathcal{E}^{-}_{{r_i}}$.\footnote{
The sampling ratio is based on previous
research \citep{kojima-etal-2024-multilingual}. 
Too large or too small ratios are bad for computing reliable $AP$ values. 
We also sample negative examples with different seeds in our preliminary experiments. 
The identified relation neurons show little change, suggesting stability of the identification method.}
The final data used to detect \RelationSpecificNeurons for relation ${r_i}$ is then $\mathcal{E}_{r_i} = \mathcal{E}^{+}_{r_i} \cup{} \mathcal{E}^{-}_{r_i}$. 
Each example $e^{j}_{r_i}$ is associated with binary label $b^{j}_{r_i}$: 1 if $e^{j}_{r_i} \in \mathcal{E}^{+}_{r_i}$, 0 otherwise.

\shortpar{Neuron Output Values.}
Let $o^{m,j,t}_{r_i}$ be the output value of neuron $m$ for the $t$-th token in $e^{j}_{r_i}$ when feeding the example to the model. 
Following \citet{kojima-etal-2024-multilingual}, we average the outputs over tokens to form the final output value of neuron $m$ for the entire example $e^{j}_{r_i}$: $o^{m,j}_{r_i} = \frac{1}{T} \sum_{t=1}^{T} o^{m,j,t}_{r_i}$, where $T$ is the number of effective tokens in $e^{j}_{r_i}$.

\shortpar{Computing Experts.}
The level of expertise of each neuron for relation ${r_i}$ is computed by formulating a classification task. 
Specifically, we regard the output value $o^{m,j}_{r_i}$ as the prediction score with $e^{j}_{r_i}$ as input and $b^{j}_{r_i}$ as its ground-truth label.
In this way, for an individual neuron $m$, we have the following data: $\{o^{m,j}_{r_i}, b^{j}_{r_i}\}_{j=1}^{|\mathcal{E}_{r_i}|}$.
We then measure this neuron's performance by setting all output values as classification thresholds and comparing the predictions with the ground truth labels.
Average precision ($AP$) is used as the metric (the area under the precision-recall curve).
By doing this, we obtain $AP^m_{{r_i}}$ for all $m \in M$, allowing us to rank them by their level of expertise in differentiating relation ${r_i}$ from others. 
The top $k$ neurons are regarded as \RelationSpecificNeurons in descending order.

\enote{hs}{i don't think that AP and area under the
precision recall curve are the same -- i assume this is from
the original apper where they also equated the two?}
\enote{yl}{The original paper has the following claim: ``We can measure the performance of neuron $m$ for the task using its average precision ($AP_m = AP(z_m, b) \in [0, 1]$), which is the
area under the precision-recall curve with different
prediction thresholds.'', and we checked on wiki:
}

\subsection{Controlled Generation}\seclabel{controlled}

For each relation ${r_i}$, we want to investigate the impact of the identified top-$k$ \RelationSpecificNeurons. 
Therefore, we control text generation by overriding their
output values with 0 during inference, aiming to deactivate or suppress these neurons.
Specifically, we feed
$\mathcal{P}_{r_i}^{\text{eva}}$,
the prompts from the held-out evaluation prompt set for relation $r_i$, into the model. 
During inference, we simply set the output values of all top-$k$ \RelationSpecificNeurons 
to a constant 0 and set the maximum generation length to 2 (similar to the setup in
validating prompts, cf.\ \secref{method}).
The predicted 2 tokens are then compared to the object.
The prediction is regarded as correct if the predicted 2 tokens are a prefix of the object.

\section{Experimental Setup}\seclabel{exp_setup}

\begin{table}
\setlength{\belowcaptionskip}{-0.2cm}
\footnotesize
\centering
\setlength{\tabcolsep}{0.8mm}{}
\begin{tabular}{lrrr}
\toprule
Model & \#Layers & \#Neurons (FFNs) & \#Neurons (total)\\
\midrule
LLama-2-7B & 32 & 835,584 & 1,359,872 \\
LLama-2-13B & 40 & 1,310,720 & 2,129,920 \\
\bottomrule
\end{tabular}
\caption{LLama-2 model neuron statistics}\label{tab:model_info}
\end{table}

\subsection{Models}

We consider the 7B and 13B models from the \textbf{LLama-2} family \citep{touvron2023llama2openfoundation}.\footnote{We conduct a similar investigation on \textbf{Gemma-7B} \citep{gemma2024team}, as detailed in \secref{gemma}, and observe experimental results consistent with those of LLama-2.} 
As mentioned in \secref{identification}, we consider the neurons in \textbf{FFNs},
which account for more than half of neurons in both 7B and 13B models, as shown in Table \ref{tab:model_info}. We also report our preliminary results when considering neurons in other modules, i.e., attention heads, in \secref{neuron_type}. Their effectiveness tends to be unsatisfactory compared with FFNs, supporting our choice.

\subsection{Datasets}\seclabel{datasets}

We manipulate the relational knowledge datasets from \citet{lre2024Hernandez} using the procedure described in \secref{manipulation}. 
Recall that we cover 12 relations in our experiments.
Prompt sets $\mathcal{P}_{r_i}^{\text{det}}$ (for 
neuron identification) and $\mathcal{P}_{r_i}^{\text{eva}}$ (for evaluation) are constructed for each relation $r_i$,
yielding varying numbers $|\mathcal{P}_{r_i}^{\text{det}}|$ of prompts.
$\mathcal{P}_{r_i}^{\text{eva}}$ is constructed by randomly selecting 50 triples for each relation.
Since these 50 triples are not used when creating $\mathcal{P}_{r_i}^{\text{det}}$, this setup ensures \textbf{no subject entity overlap between $\mathcal{P}_{r_i}^{\text{det}}$ and $\mathcal{P}_{r_i}^{\text{eva}}$ for the same relation $r_i$}. 
The elimination of subject entity overlap allows us to disentangle the effect of entities and focus on the only shared attribute between $\mathcal{P}_{r_i}^{\text{det}}$ and $\mathcal{P}_{r_j}^{\text{det}}$ -- the relation itself.
In addition, we ensure \textbf{minimal subject entity overlap across relations} 
(mostly 0 
between $\mathcal{P}_{r_i}^{\text{det}}$ and $\mathcal{P}_{r_j}^{\text{det}}$). The only exception is between \texttt{person\_mother}
and \texttt{person\_father}, which share a lot of subject entities in $\mathcal{P}_{r_i}^{\text{det}}$; however, the two relations \textbf{share no subject entities in $\mathcal{P}_{r_i}^{\text{eva}}$}.
A detailed analysis of entity overlap 
is presented in~\secref{entity_overlap}.

\begin{figure}
    \centering
    \setlength{\belowcaptionskip}{-0.2cm}
    \includegraphics[width=0.15\textwidth]{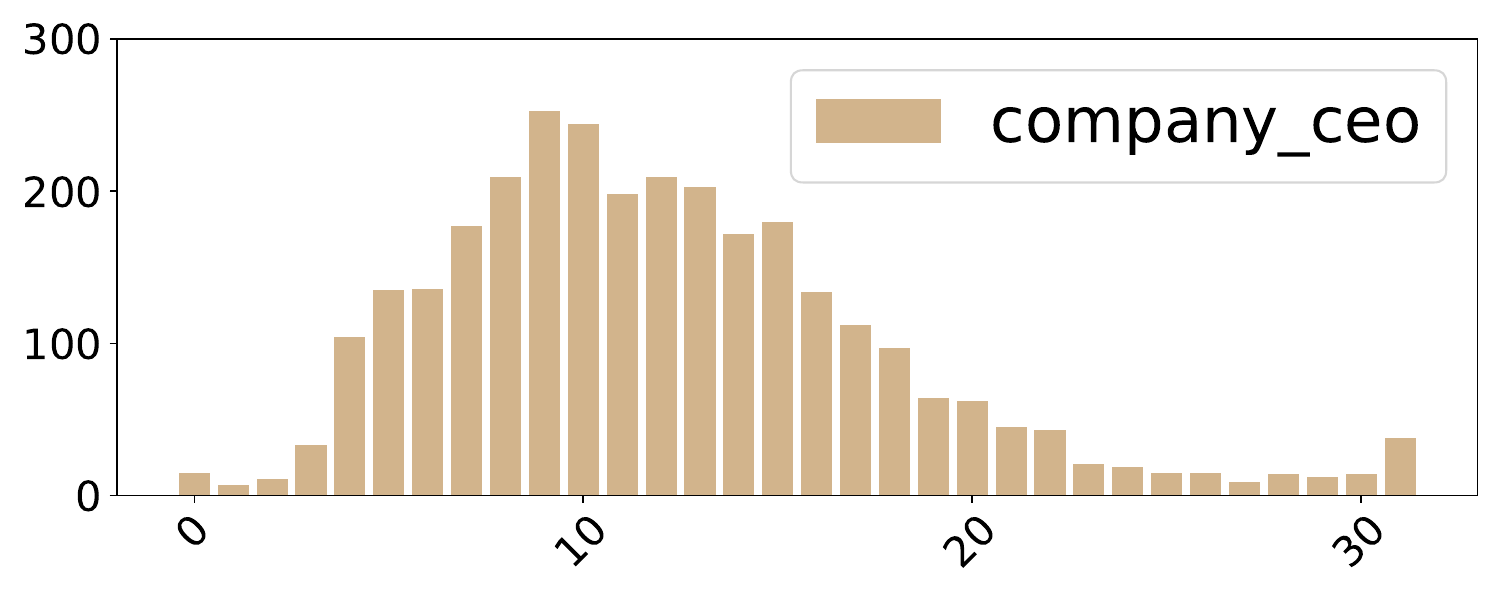}
    \includegraphics[width=0.15\textwidth]{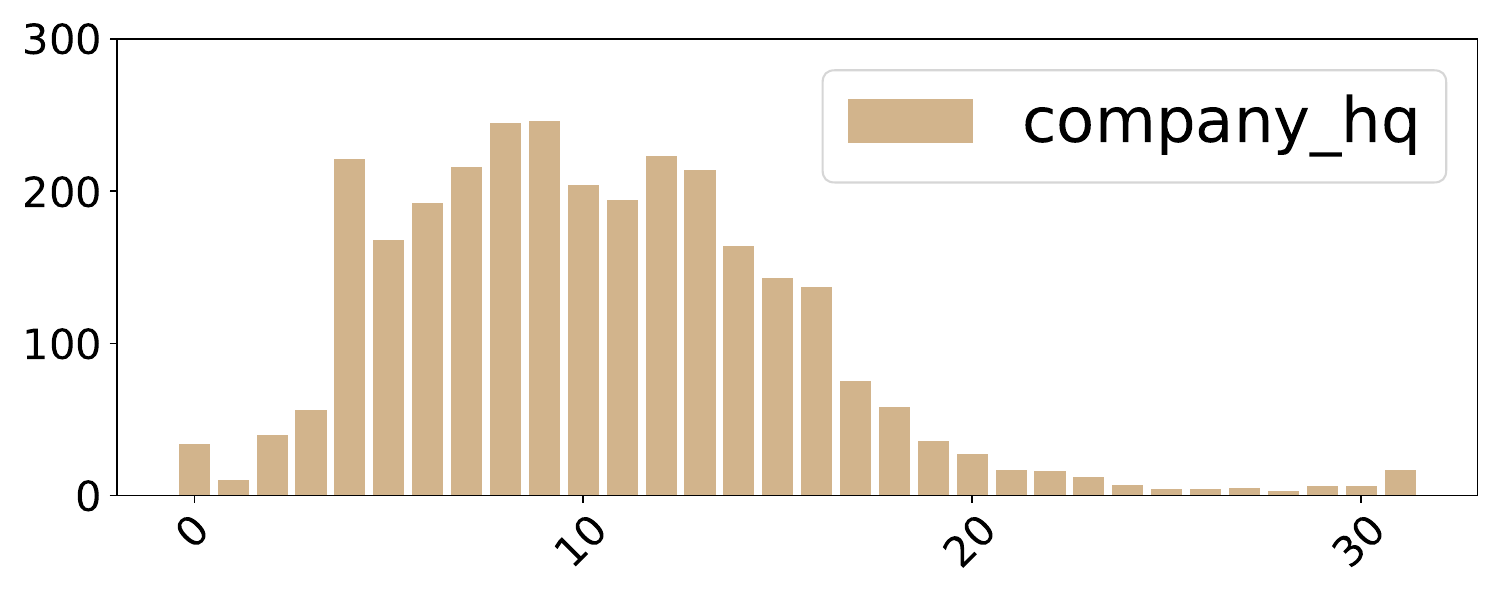}
    \includegraphics[width=0.15\textwidth]{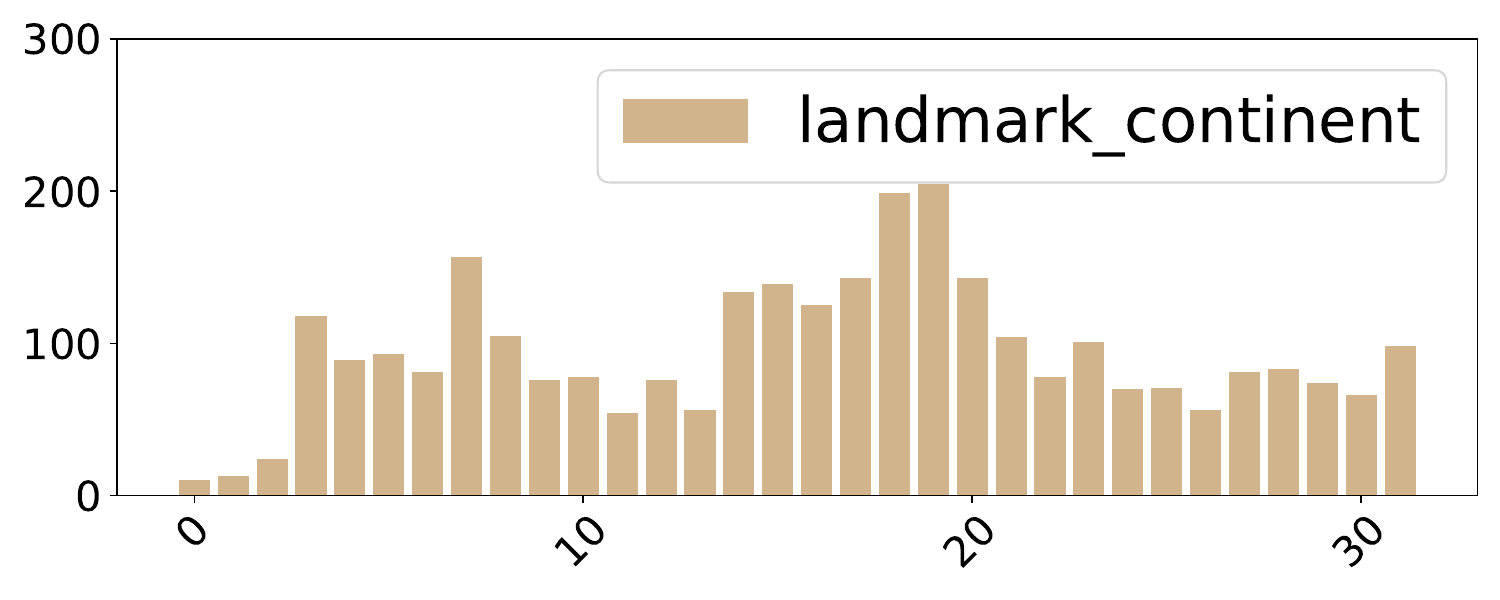}
    \includegraphics[width=0.15\textwidth]{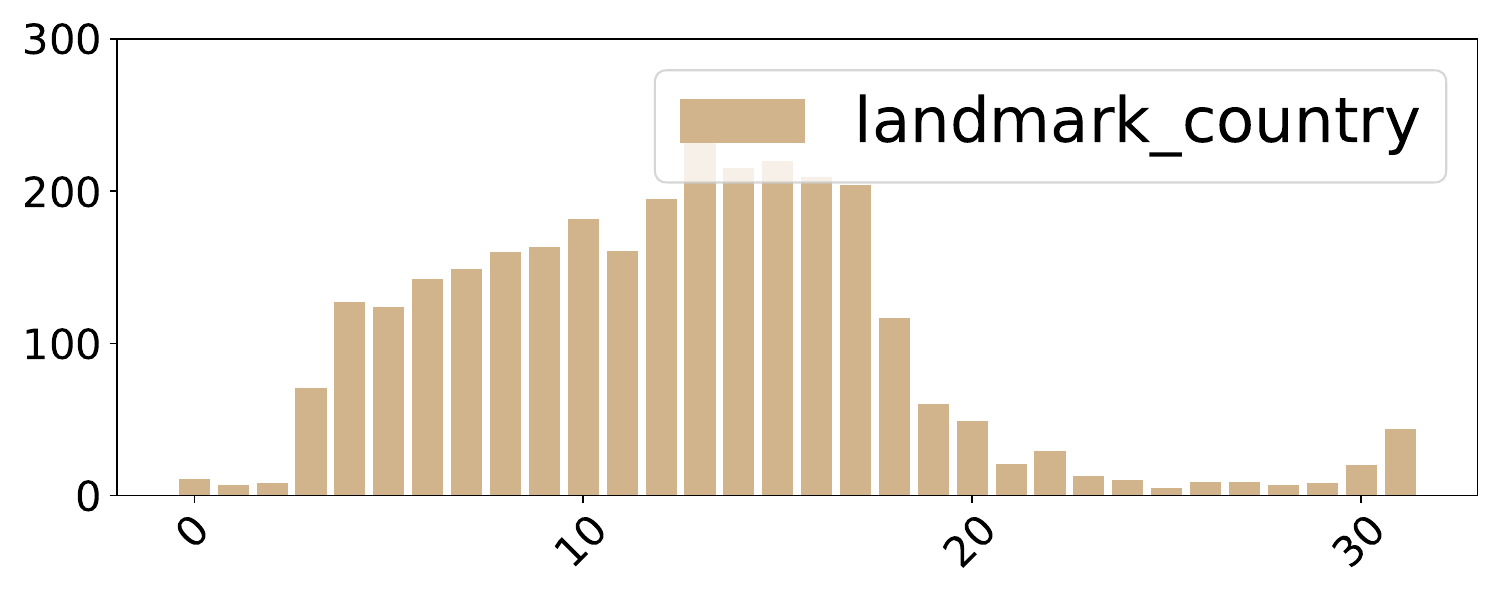}
    \includegraphics[width=0.15\textwidth]{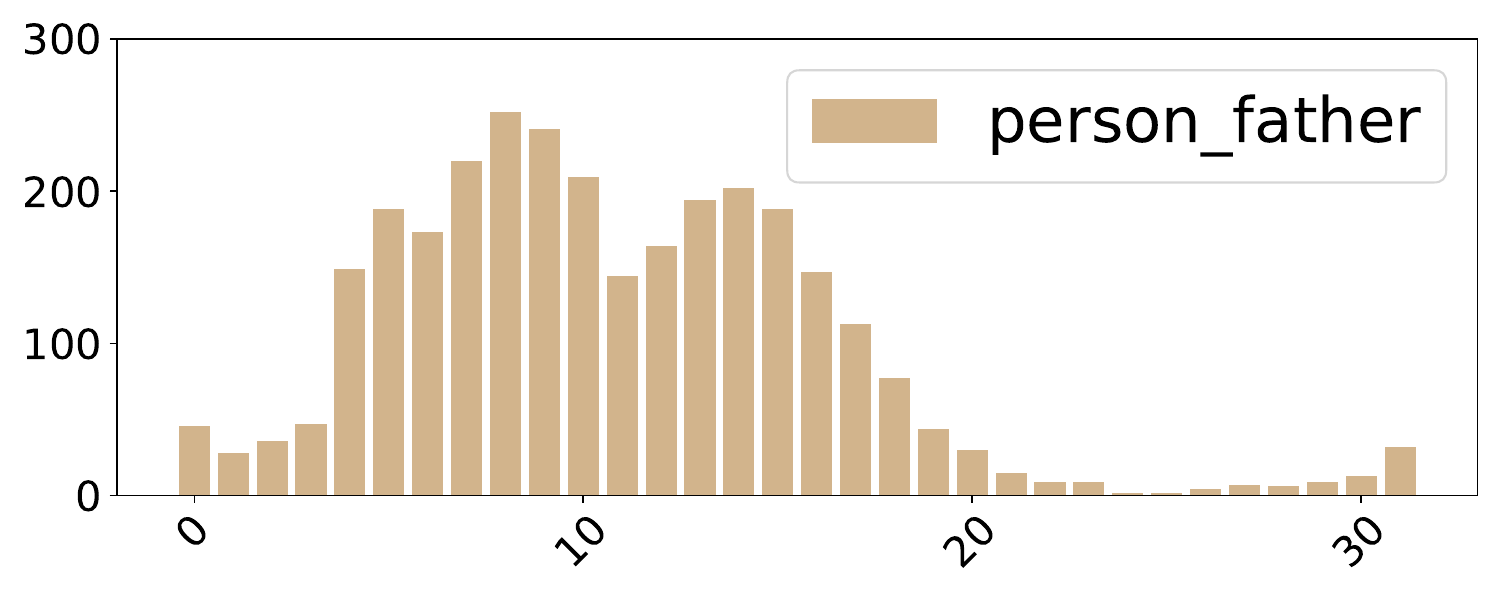}
    \includegraphics[width=0.15\textwidth]{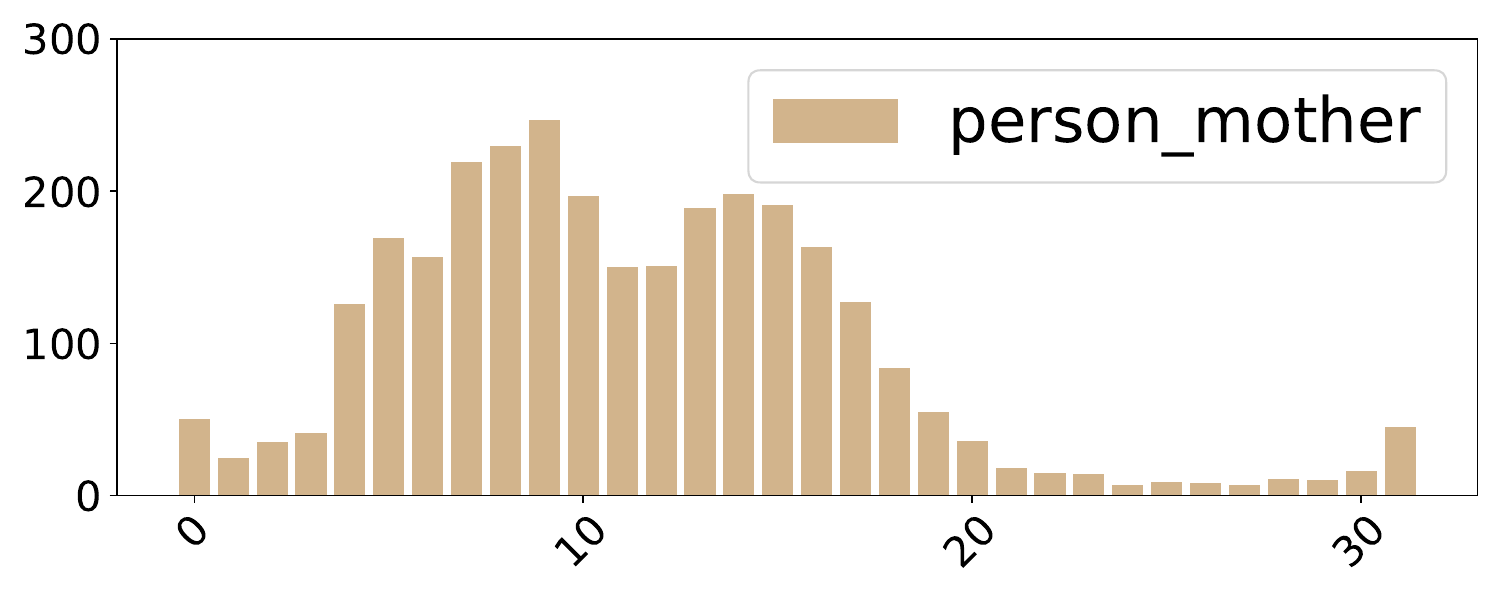}
    \includegraphics[width=0.15\textwidth]{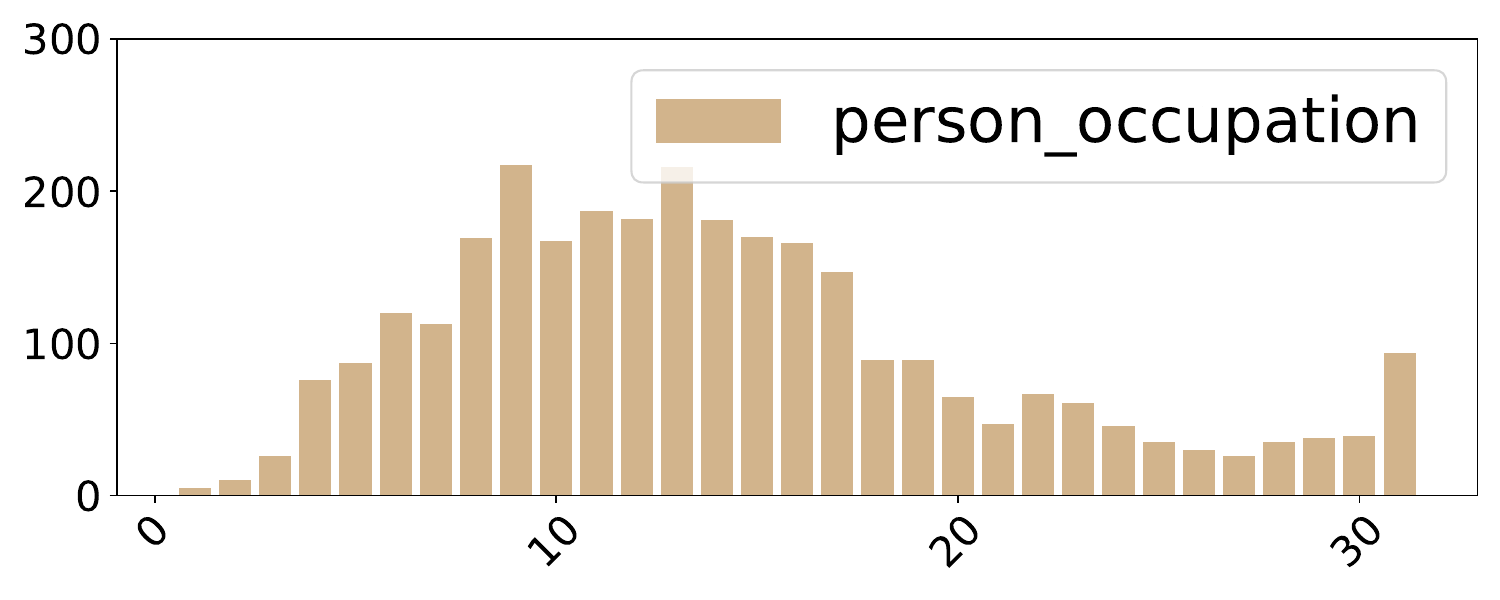}
    \includegraphics[width=0.15\textwidth]{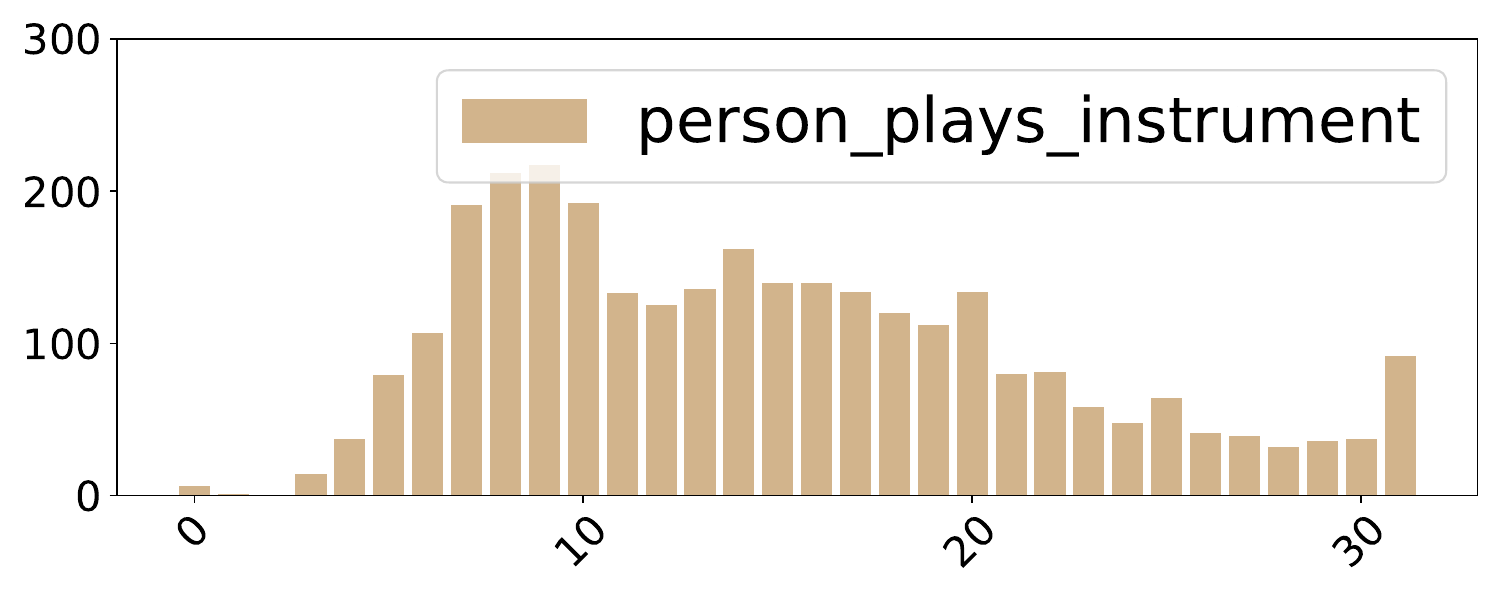}
    \includegraphics[width=0.15\textwidth]{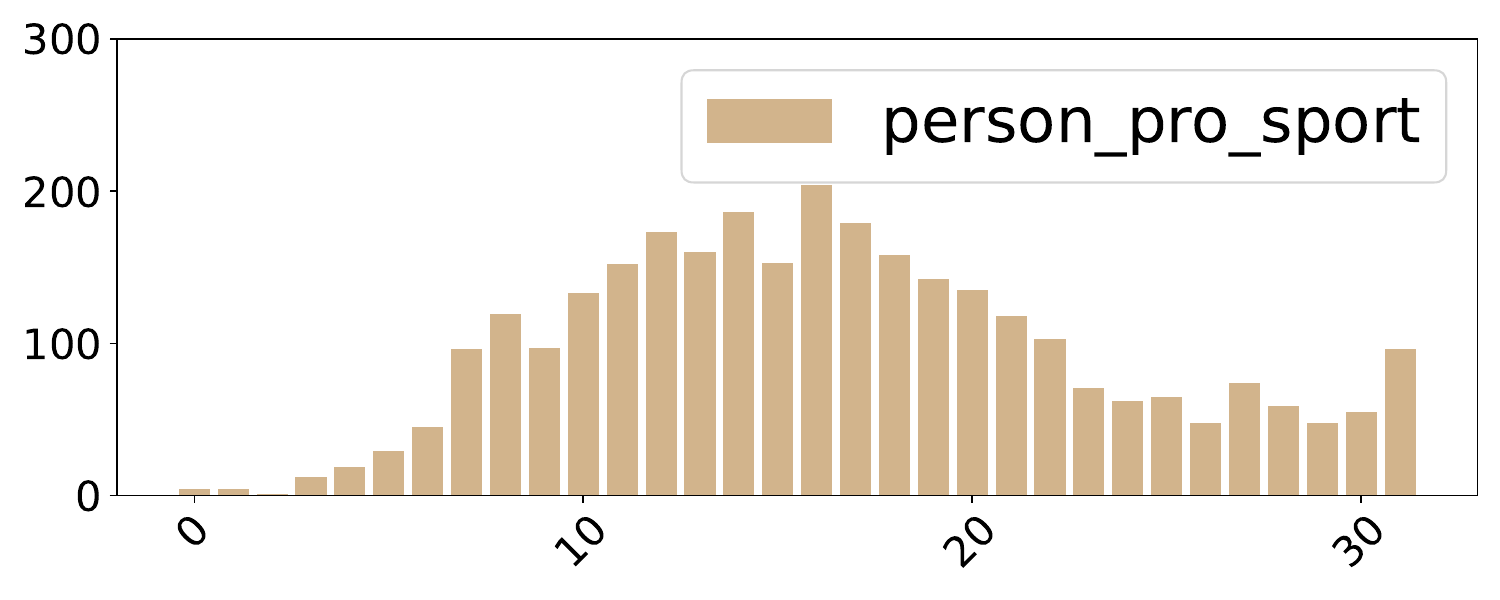}
    \includegraphics[width=0.15\textwidth]{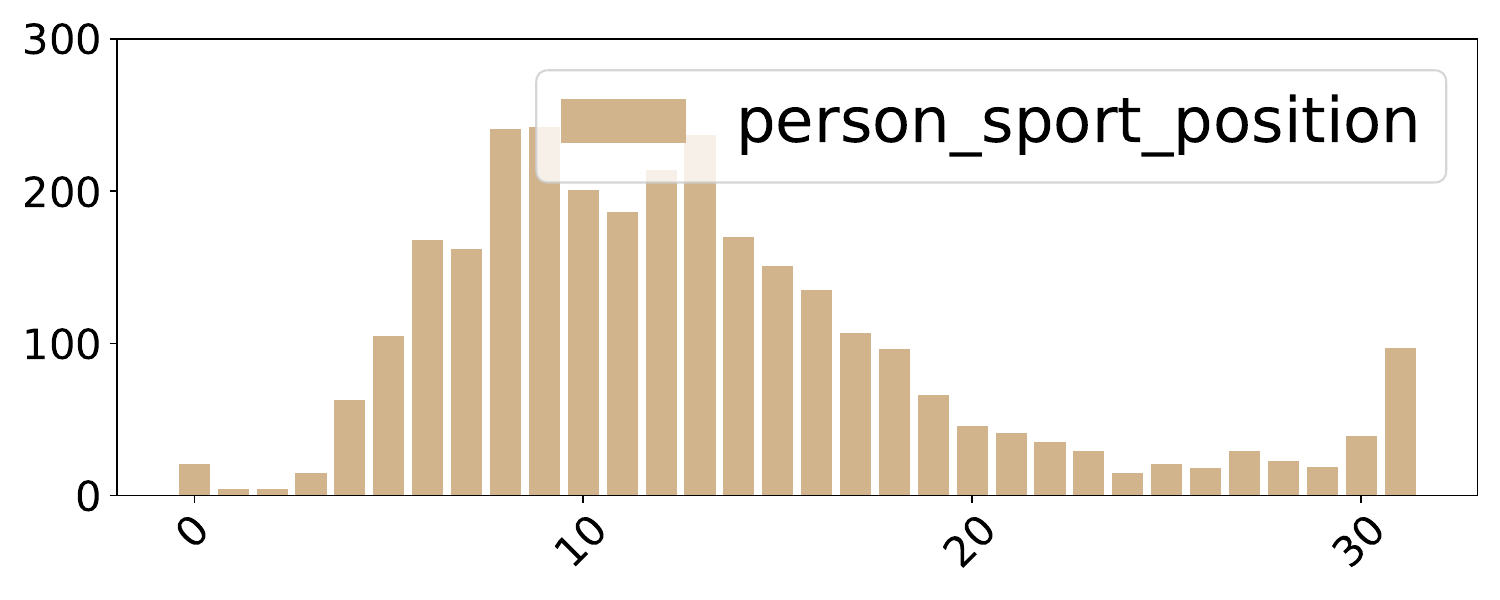}
    \includegraphics[width=0.15\textwidth]{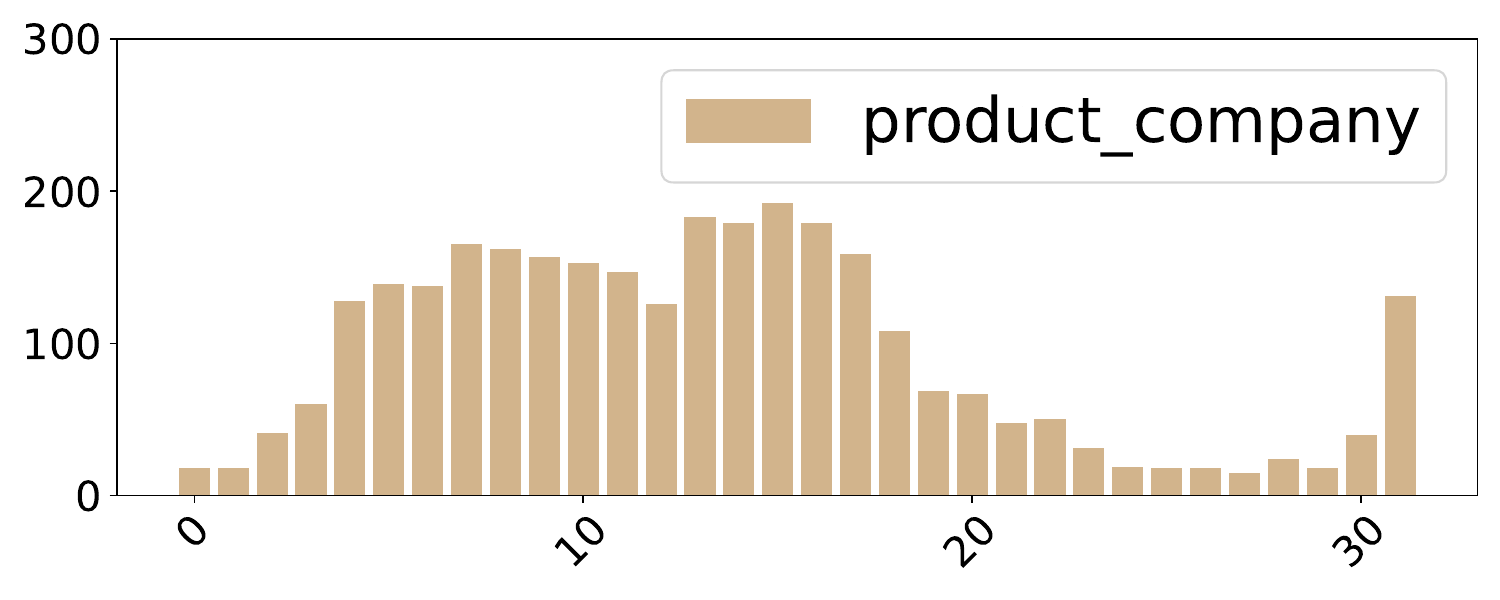}
    \includegraphics[width=0.15\textwidth]{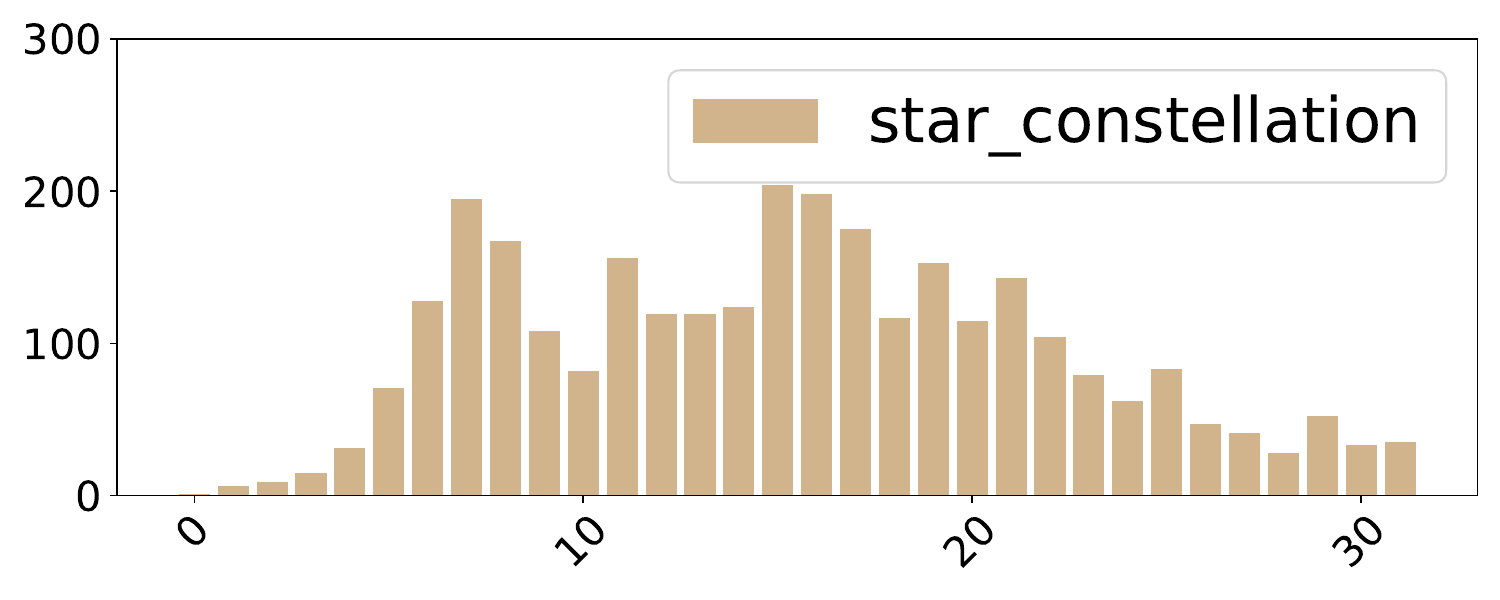}
    \caption{Distribution of \RelationSpecificNeurons
    across layers in the 7B model. Most are located in the middle layers.}
    \label{fig:layer_dist}
\end{figure}

\section{Results and Discussion}\seclabel{results}

We apply our identification method to 
both LLama-2 7B and 13B models
for all 12 relations. We regard the \textbf{top 3,000}
neurons with the highest $AP$ values as the
\RelationSpecificNeurons; for this threshold,
we achieve good coverage of relation-specific neurons with a
set of neurons that is not too large. 
We discuss the impact of this meta-parameter 
in~\secref{neuron_number}.

\subsection{Identified Relation-Specific Neurons}\seclabel{neurons}

\shortpar{Distribution Across Layers.} We display the distribution of relation-specific 
neurons across layers in the 7B model in Figure~\ref{fig:layer_dist} (see \secref{13b_analysis} for the 13B model).
Most neurons are located in the model's \textbf{middle layers}. Such a distribution differs from language-specific neurons, which are mostly located in the first and last few layers \citep{kojima-etal-2024-multilingual}.
We hypothesize that relational knowledge requires more than surface-level information that is mainly encoded and processed in the first and last few layers.
\textbf{Therefore, \RelationSpecificNeurons naturally emerge in the middle layers, where the model has integrated enough lexical and syntactic signals to model and process the relation.} 
This finding is consistent with several studies that show functional mapping vectors can be extracted from the middle layers of LLMs \citep{merullo-etal-2024-language,lre2024Hernandez,Todd2024functon}.

\begin{figure}
    \centering
    \setlength{\abovecaptionskip}{-0.1cm}
    \setlength{\belowcaptionskip}{-0.2cm}
    \includegraphics[width=0.36\textwidth]{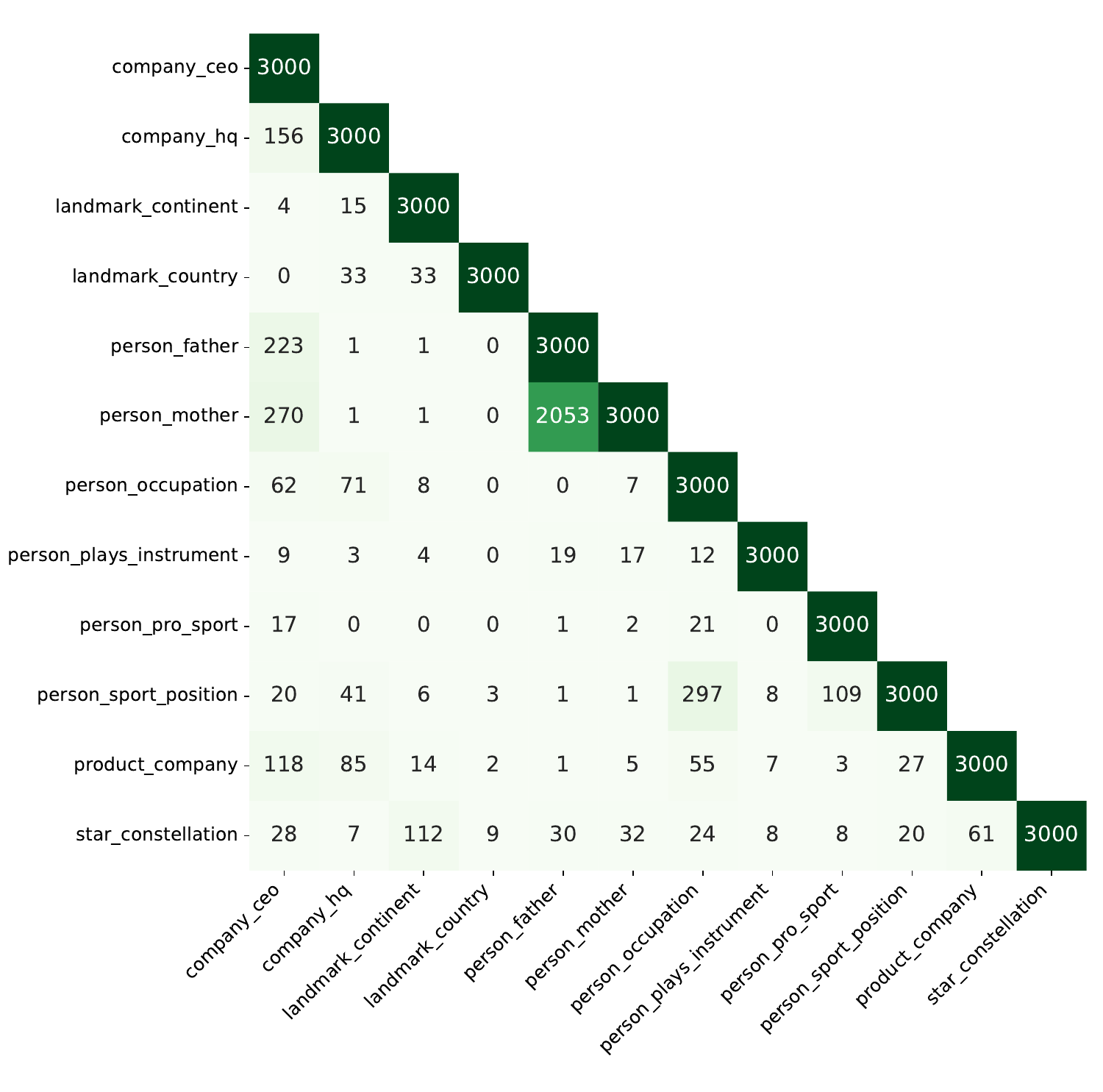}
    \caption{Neuron overlap of \RelationSpecificNeurons across 12 relations in the 7B model. For example, the number of neurons shared between the 3,000 identified neurons for \texttt{person\_father} and the 3,000 for \texttt{person\_mother} is 2053 (in green).}
    \label{fig:neuron_overlap}
\end{figure}

\shortpar{Neuron Overlap Across Relations.} 
We display the overlap of \RelationSpecificNeurons across relations for the 7B model in Figure \ref{fig:neuron_overlap} 
(13B is in \secref{13b_analysis}). 
We see that \texttt{person\_mother}
and \texttt{person\_father} share many  neurons,
possibly due to the large overlap between their subject entities, 
(see \secref{entity_overlap}).
\textbf{However, even though there is almost no subject overlap between any other relations,
many relations still share some neurons with others.} 
For instance, \texttt{person\_occupation}
and \texttt{person\_sport\_position} share $297$ neurons, possibly because they are similar relations -- a sport is
a kind of occupation.
Extensive neuron overlap can also be observed when two relations are mapping from the same type of subjects, e.g., \texttt{company\_ceo} and  \texttt{company\_hq}, or mapping to the same type of objects, e.g., \texttt{company\_ceo} 
and \texttt{person\_father}.
However, we show in \secref{inter_results} that a high neuron overlap does not necessarily imply a high level of mutual interference.

\begin{figure*}
    \centering
    \setlength{\belowcaptionskip}{-0.1cm}
    
    \subcaptionbox{Held-out evaluation prompts $\mathcal{P}_{r_i}^{\text{eva}}$\label{fig:intra-left}}{\includegraphics[width=0.48\textwidth]{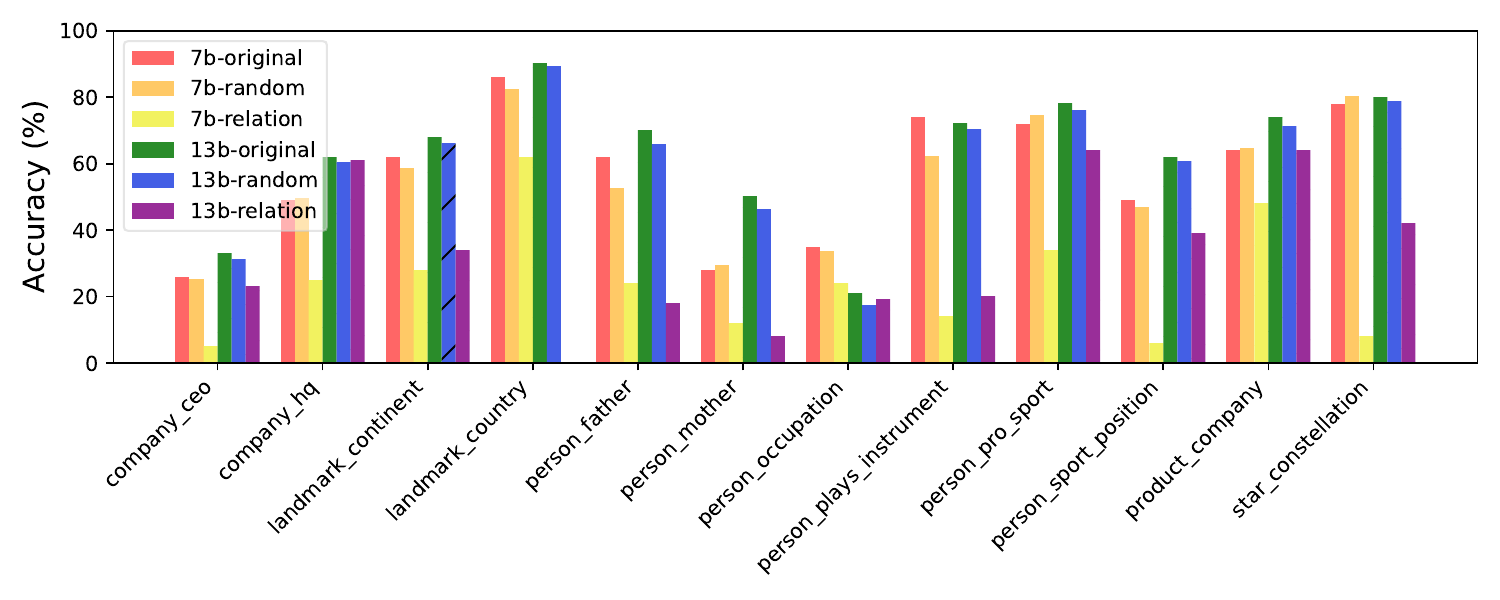}
    }
    \subcaptionbox{Identification prompts $\mathcal{P}_{r_i}^{\text{det}}$\label{fig:intra-right}}{\includegraphics[width=0.48\textwidth]{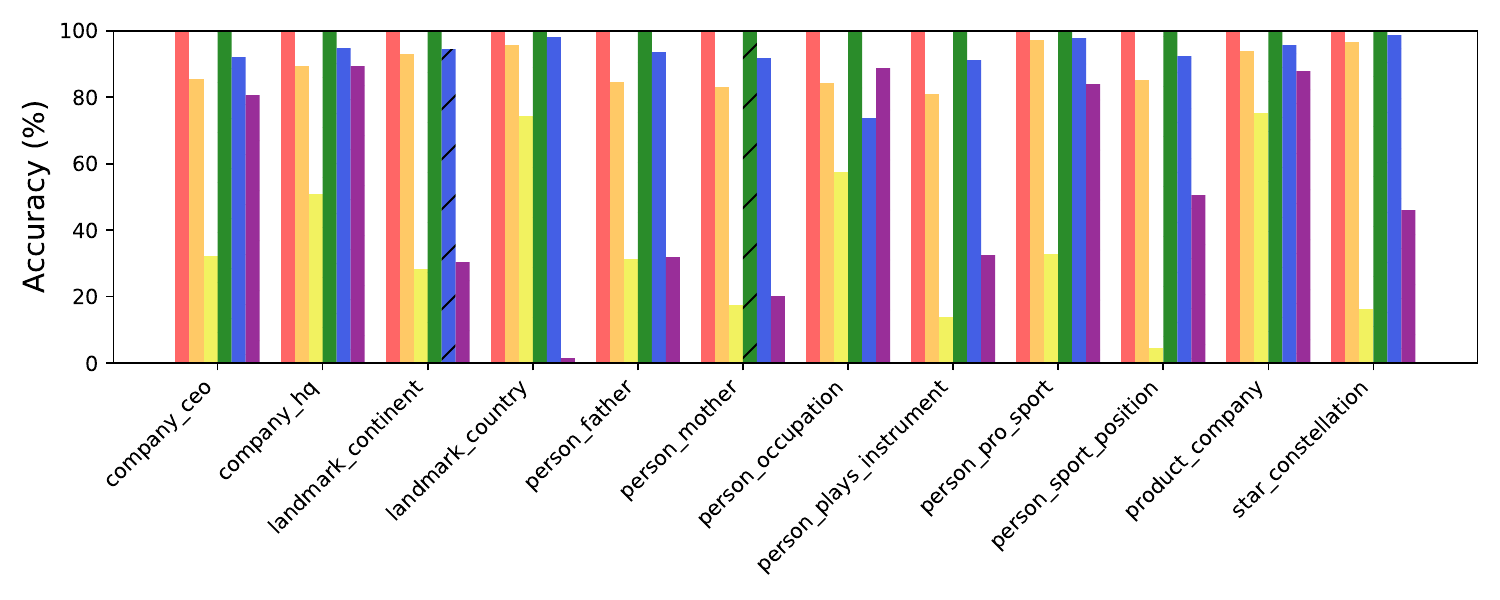}
    }
    \caption{Intra-relation results. The \textbf{left} (resp.\  \textbf{right})
    figure displays the results of held-out evaluation prompt
    set $\mathcal{P}_{r_i}^{\text{eva}}$
    (resp.\  identification prompt set
    $\mathcal{P}_{r_i}^{\text{det}}$). We report the
    performance of the original model (without any
    deactivation), e.g., \texttt{7b-original}, the model
    with 3,000 random neurons deactivated (averaged over 10 seeds), e.g., \texttt{7b-random}, and the model with \RelationSpecificNeurons deactivated, e.g., \texttt{7b-relation}.}
    \label{fig:intra-relation}
\end{figure*}

\begin{figure*}
    \centering
    \setlength{\abovecaptionskip}{-0.2cm}
    \setlength{\belowcaptionskip}{-0.1cm}
    \begin{tabular}{c}
    \includegraphics[width=0.42\textwidth]{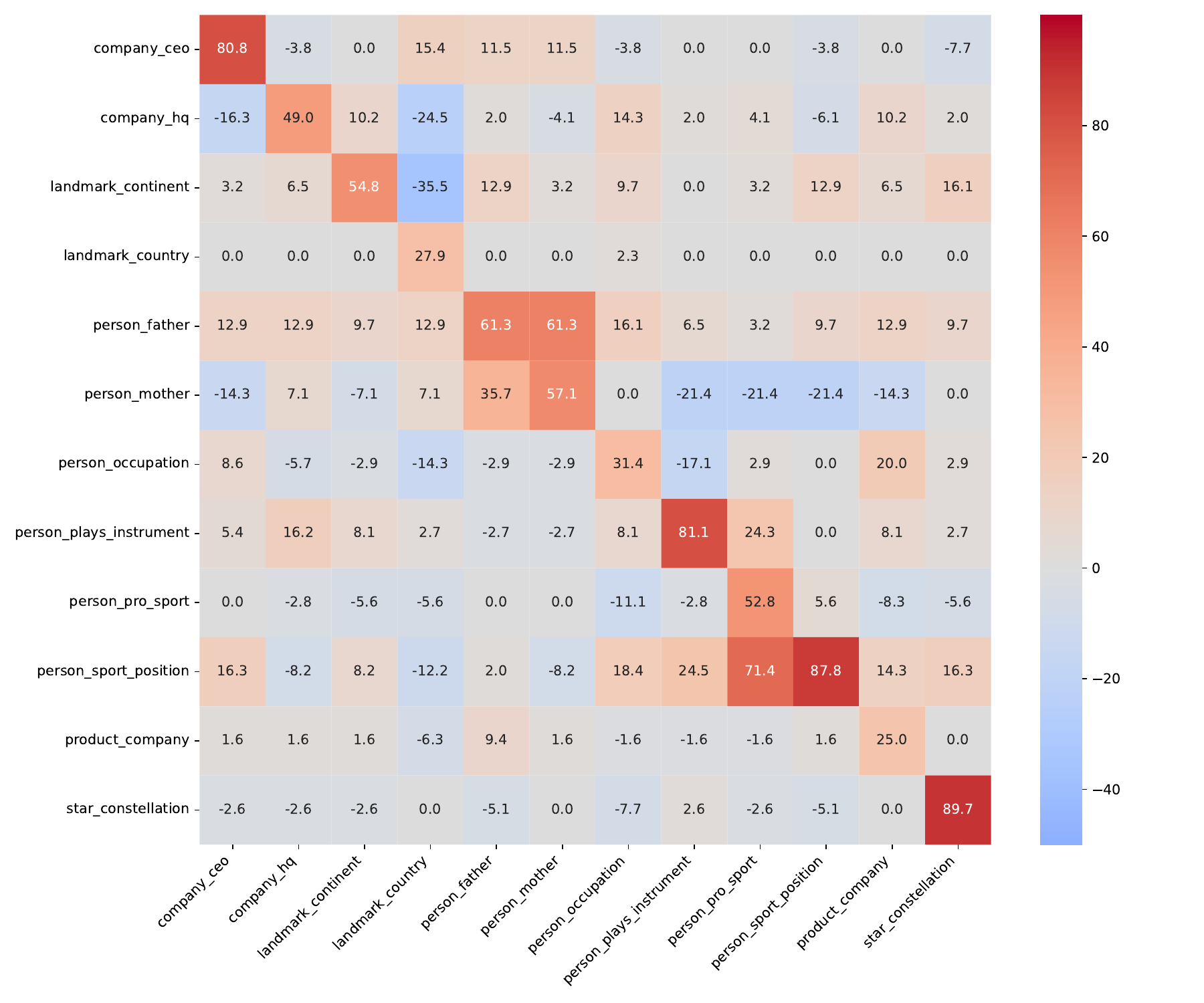}
    \includegraphics[width=0.42\textwidth]{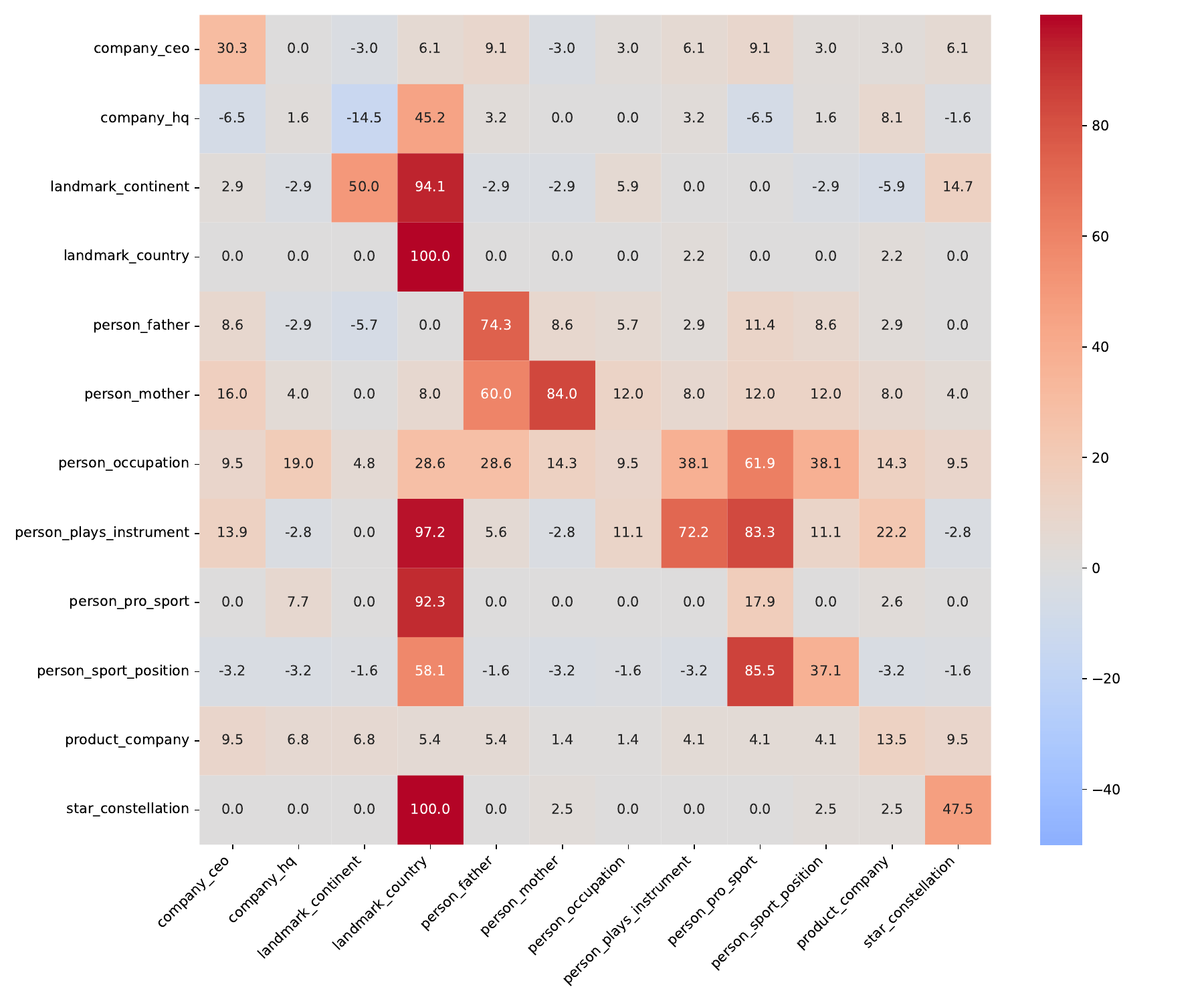}
    \end{tabular}
    \caption{Inter-relation results. Accuracy drops (in \%) for the 7B (\textbf{left})
    and the 13B model (\textbf{right}) on $\mathcal{P}_{r_i}^{\text{eva}}$. The number in cell $(r_i, r_j)$ indicates the accuracy drop of relation $r_i$ when deactivating the relation neurons of $r_j$.}
    \label{fig:inter-relation}
\end{figure*}

\subsection{Controlled Generation}\seclabel{generation}

For each relation, we set the output values of its identified 3,000 \RelationSpecificNeurons to 0, and observe how the deactivation impacts the relation itself and other relations in terms of accuracy.

\subsubsection{Intra-Relation Results}\seclabel{intra_results}
In addition to intra-relation results, i.e., deactivating the 3,000 identified \RelationSpecificNeurons for a relation and evaluating the same relation, we also create a baseline by \textbf{randomly} deactivating 3,000 neurons in the model. Results for the original models and for the two interventions are in Figure~\ref{fig:intra-relation}.

We can observe a clear performance drop on the identification prompt set $\mathcal{P}_{r_i}^{\text{det}}$ when comparing the accuracy of the original model and the model whose \RelationSpecificNeurons are deactivated.\footnote{For some relations, the drop is moderate, e.g., \texttt{product\_company}. We show in \secref{neuron_number} that the drop can become noticeable when we deactivate more than 3,000 neurons.} 
On the other hand, the model with 3,000 random deactivated neurons does not show much difference compared with the original model,
indicating the 3,000 relation neurons are closely associated with the facts included in  $\mathcal{P}_{r_i}^{\text{det}}$.
On the evaluation set $\mathcal{P}_{r_i}^{\text{eva}}$, we also observe a notable accuracy drop across models for most relations. \textbf{As $\mathcal{P}_{r_i}^{\text{eva}}$ and $\mathcal{P}_{r_i}^{\text{det}}$ do not share any subject entities, this drop can only be attributed to the fact that deactivating 3,000 neurons affects the relation itself -- the common characteristic between $\mathcal{P}_{r_i}^{\text{eva}}$ and $\mathcal{P}_{r_i}^{\text{det}}$.}\footnote{There might be another confounding variable since $\mathcal{P}_{r_i}^{\text{eva}}$ and $\mathcal{P}_{r_i}^{\text{det}}$ use the same prompt templates for each relation. But we show in \secref{effect_prompt} that even when other prompt templates are used, the effectiveness of these neurons is still preserved.} 
We thus argue that \textbf{\RelationSpecificNeurons exist in LLMs}: 
they are entity-irrelevant and focus on specific relations. 

On the other hand, the accuracy does not drop to 0 for any relation (except \texttt{landmark\_country} in the 13B model) when its identified \RelationSpecificNeurons are deactivated. 
This indicates these 3,000 neurons do not equally influence all facts that belong to a certain relation,
which highlights that \textbf{LLMs do not uniformly encode all facts belonging to a given relation, but rather distribute relational knowledge across neurons in a manner that can vary significantly from fact to fact.}
We validate this by showing that the accuracy further
drops by deactivating more neurons
in \secref{neuron_number}. 
We also show that the sensitivity of a fact to a given population of neurons may correlate with how frequently it appears in the pretraining data in \secref{frequency}.

\begin{figure*}
    \centering
    \includegraphics[width=0.23\textwidth]{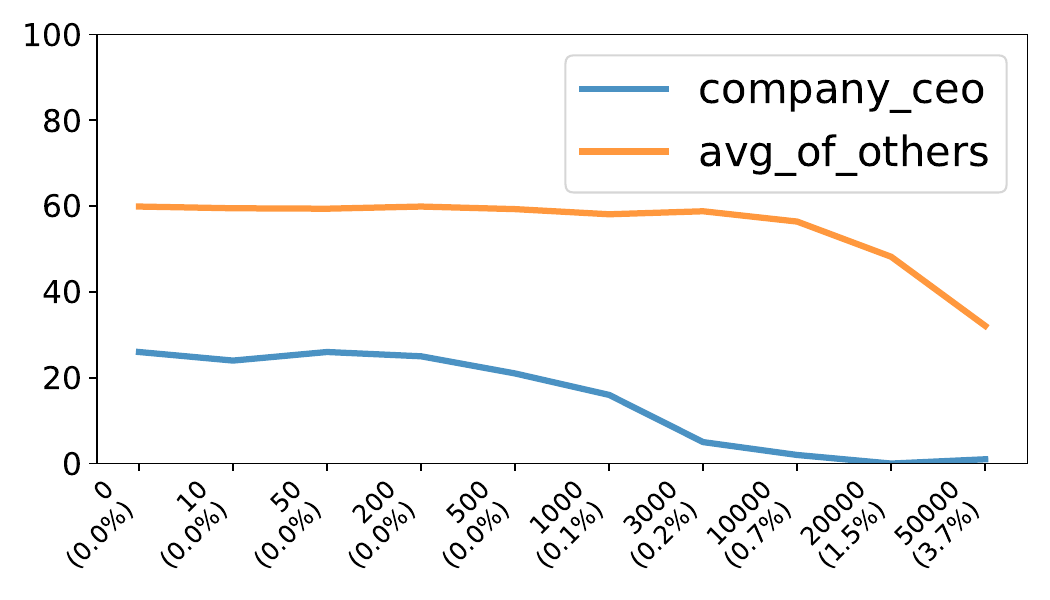}
    \includegraphics[width=0.23\textwidth]{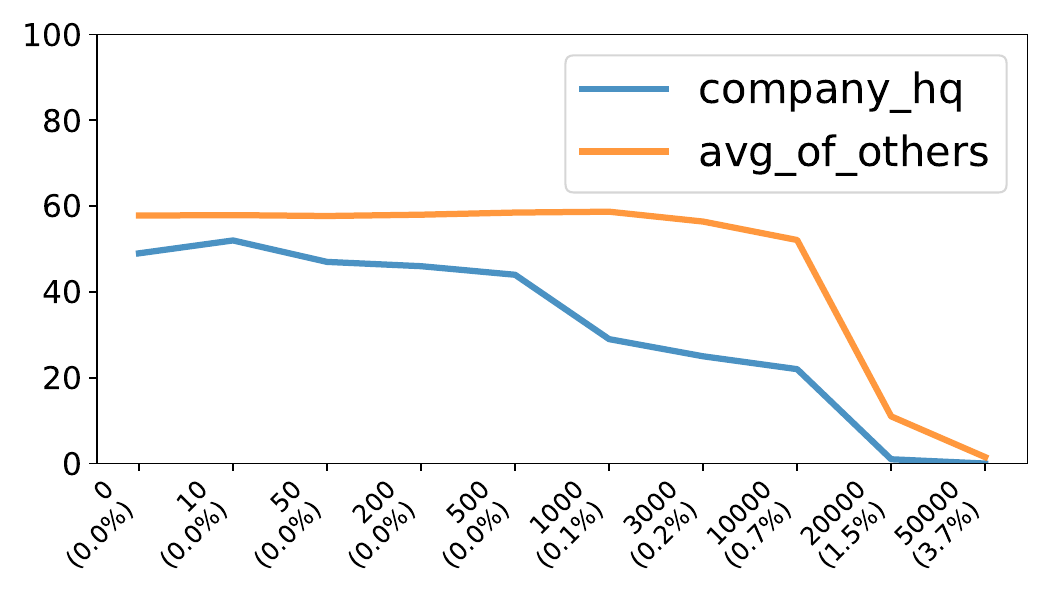}
    \includegraphics[width=0.23\textwidth]{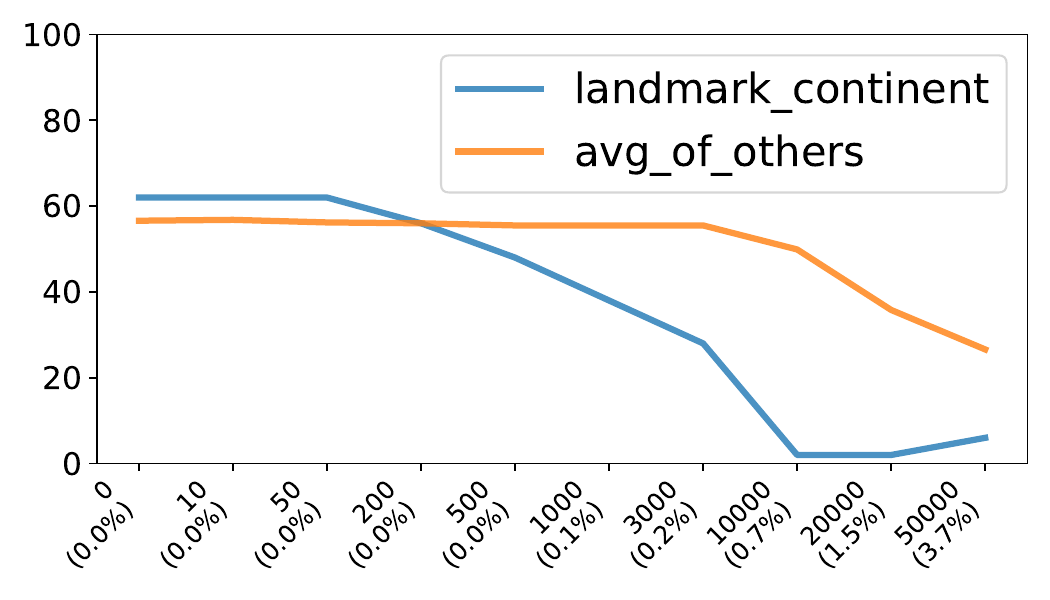}
    \includegraphics[width=0.23\textwidth]{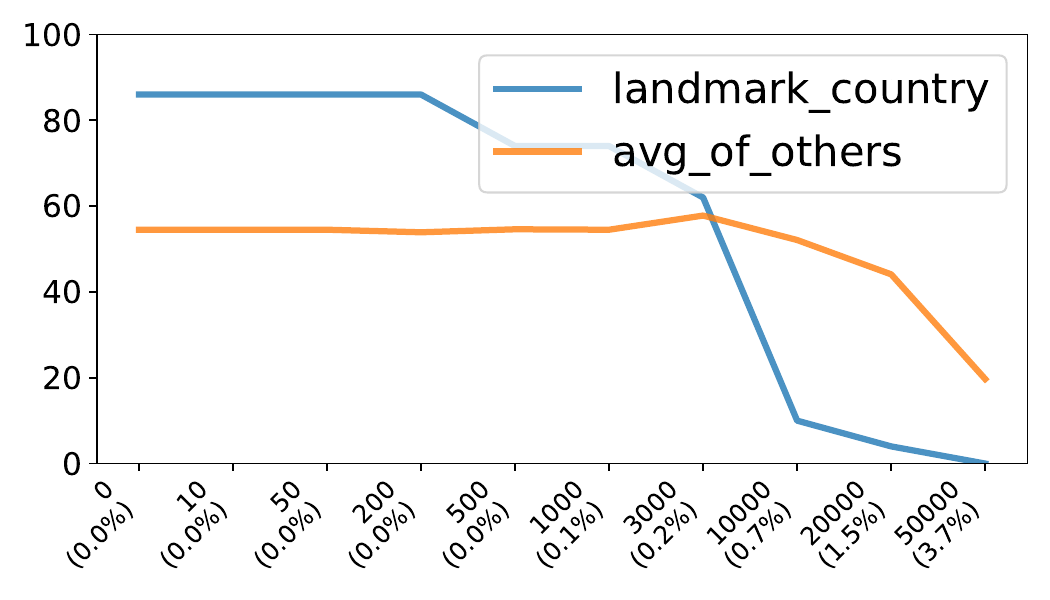}
    \includegraphics[width=0.23\textwidth]{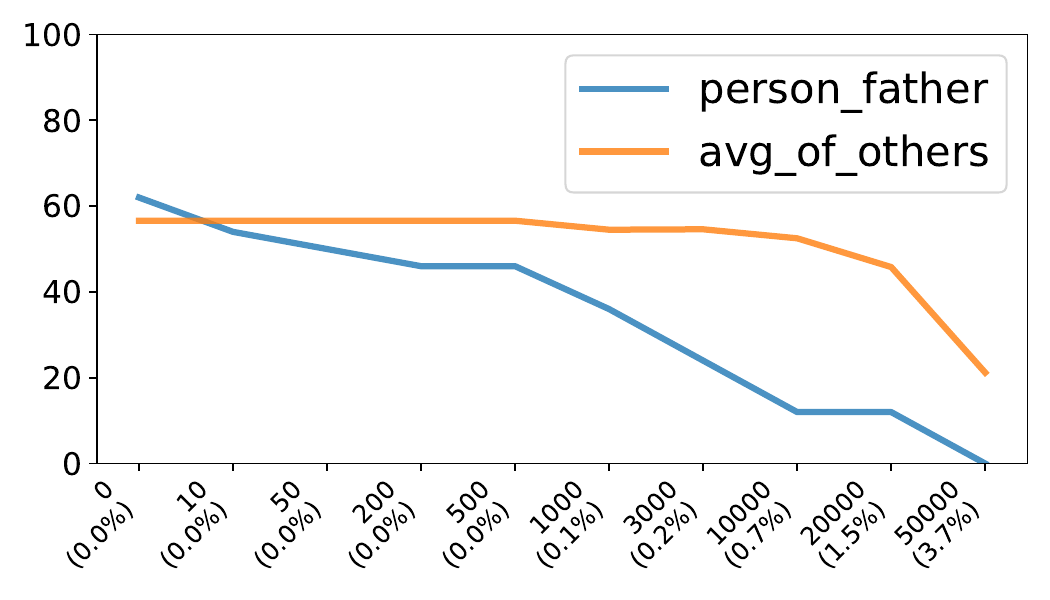}
    \includegraphics[width=0.23\textwidth]{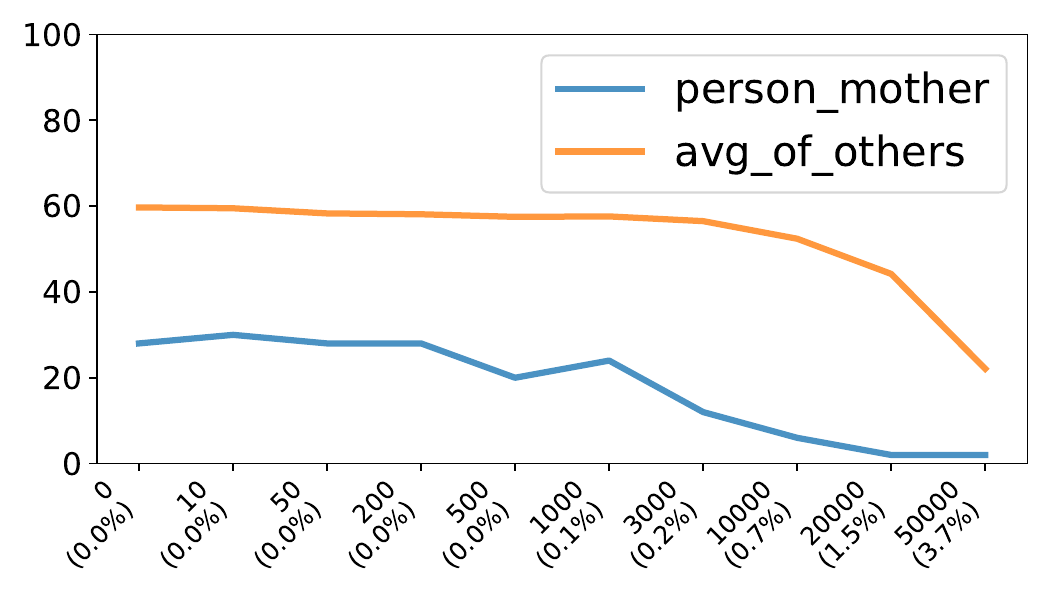}
    \includegraphics[width=0.23\textwidth]{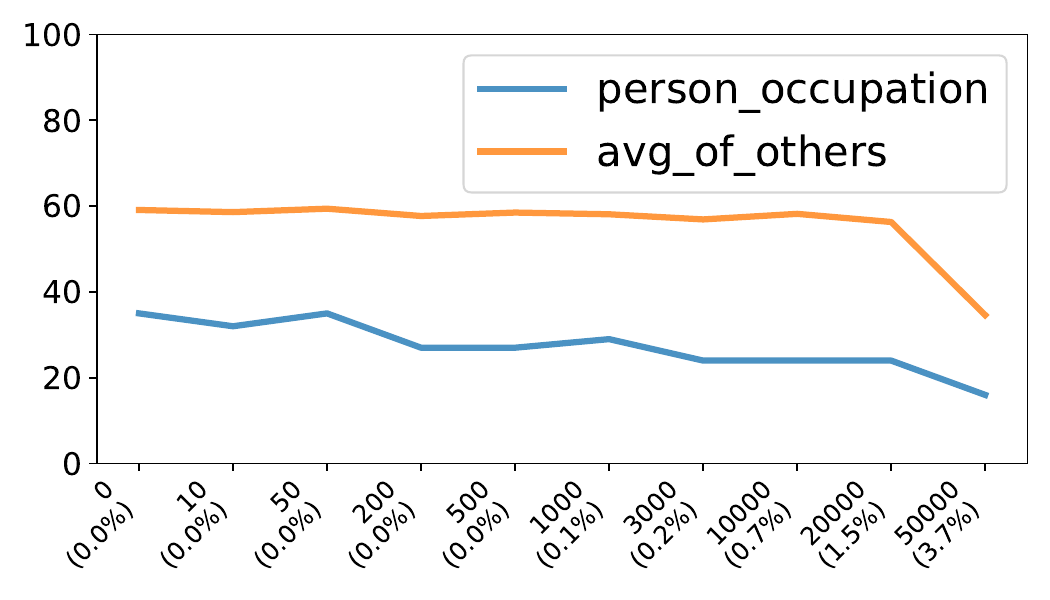}
    \includegraphics[width=0.23\textwidth]{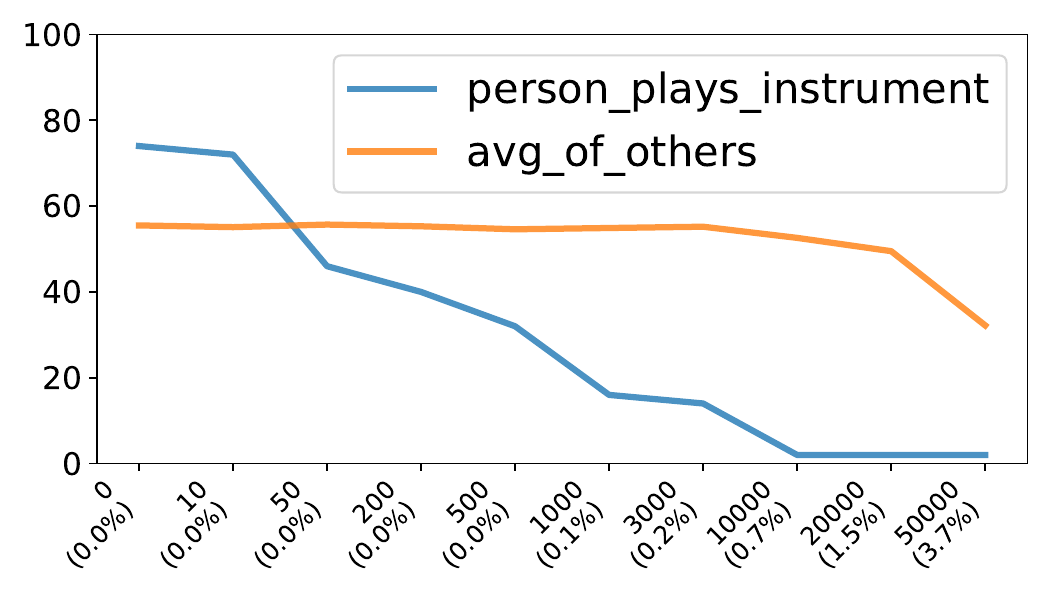}
    \includegraphics[width=0.23\textwidth]{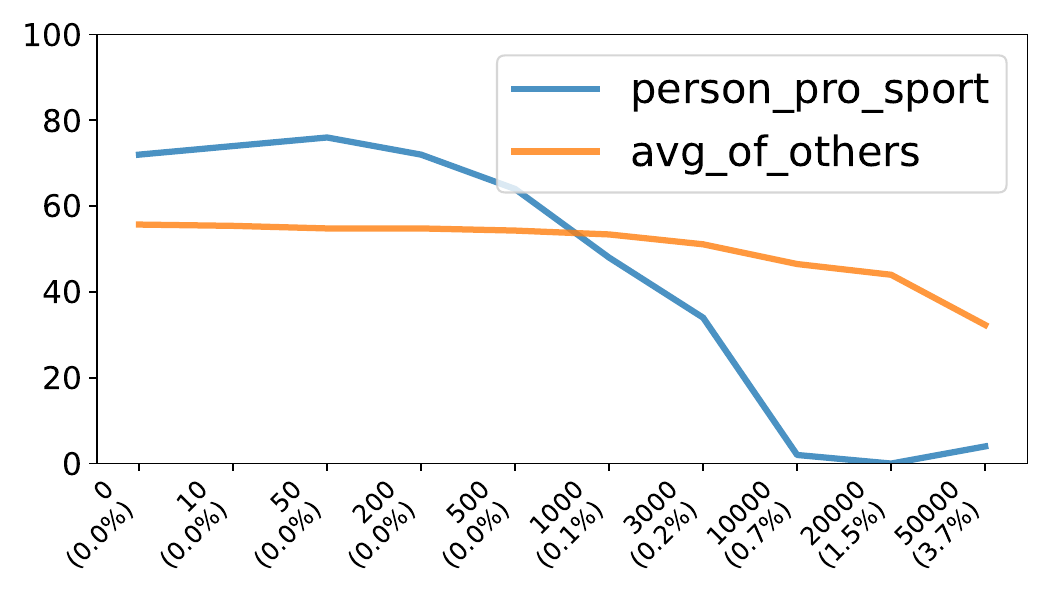}
    \includegraphics[width=0.23\textwidth]{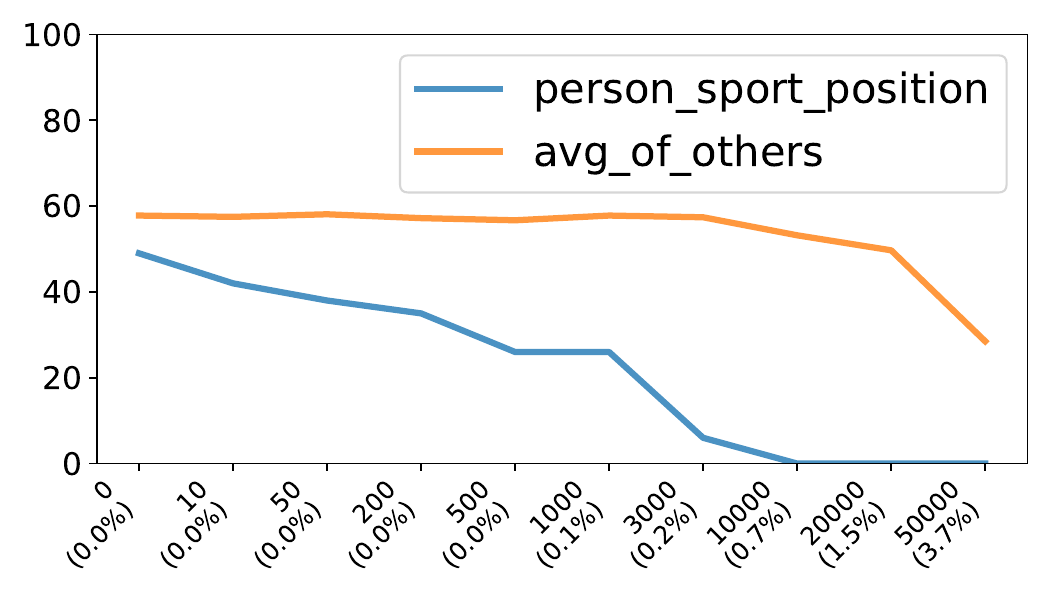}
    \includegraphics[width=0.23\textwidth]{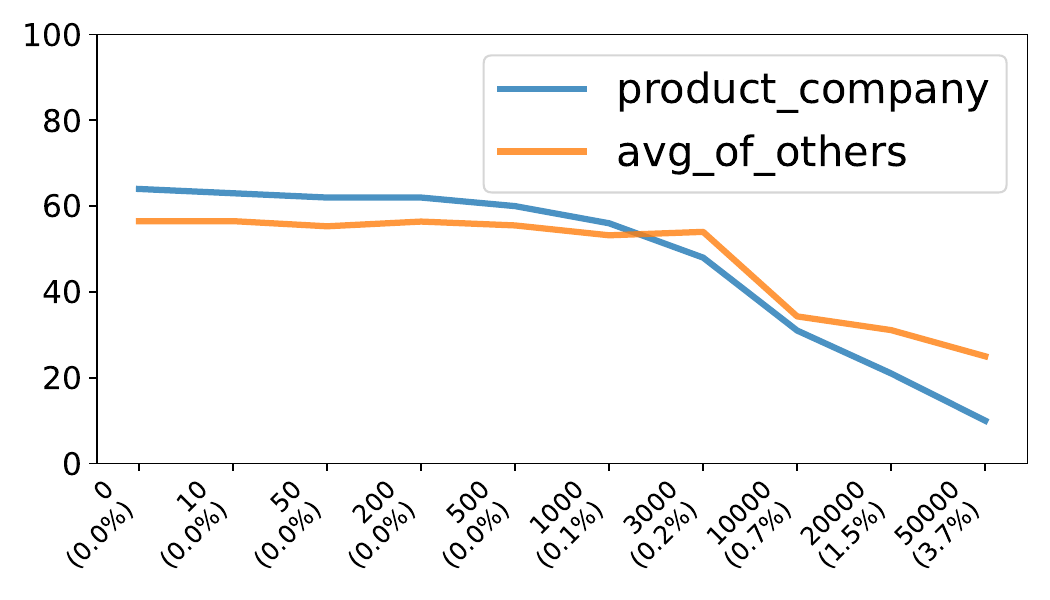}
    \includegraphics[width=0.23\textwidth]{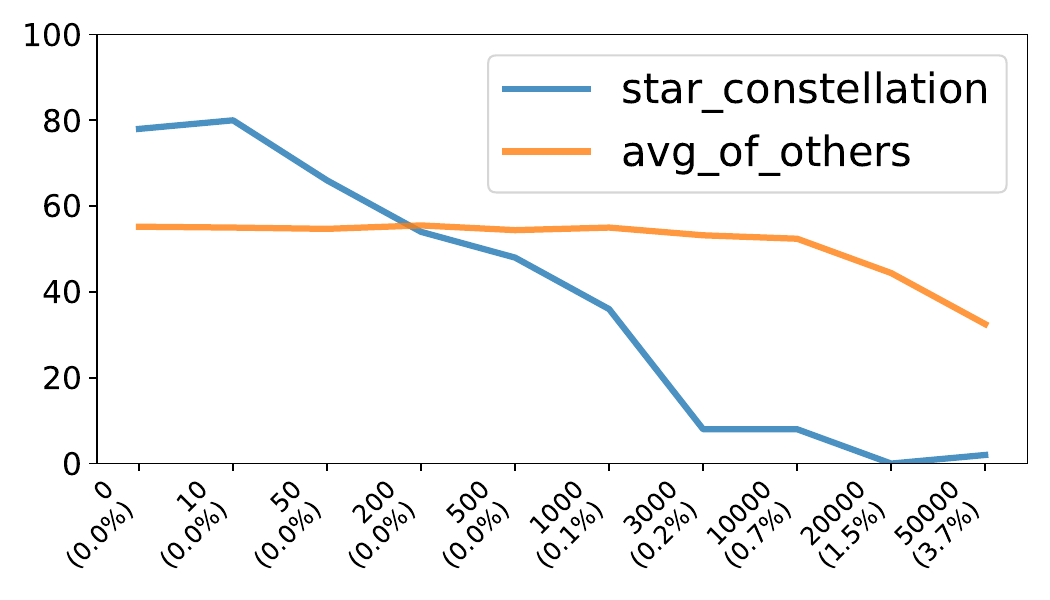}
    \caption{Influence of deactivating different numbers of
    \RelationSpecificNeurons for each relation. 
    On the $x$-axis, we report both the absolute number of deactivated neurons and the corresponding percentage of the model's total neurons. 
    We show accuracy on the relation itself and the average accuracy on other relations. 
    Increasing the number clearly affects the relation itself, while noticeable effects on other relations emerge only beyond 3,000–10,000 neurons.}
    \label{fig:neuron_num}
\end{figure*}

\subsubsection{Inter-Relation Results}\seclabel{inter_results}

To understand how \RelationSpecificNeurons{}  influence the model's ability to answer 
prompts across multiple relations, we use \textbf{accuracy drop} as a metric:
$\text{acc\_drop}_{r_i, r_j} = \frac{\text{acc}^\text{original}_{r_i} - \text{acc}^{\text{deactivated-}{r_j}}_{r_i}}{\text{acc}^\text{original}_{r_i}}$,
where $\text{acc}^\text{original}_{r_i}$ and  $\text{acc}^{\text{deactivated-}{r_j}}_{r_i}$  are the respective accuracy for $\mathcal{P}_{r_i}^\text{eva}$  of (a) the original model and (b) when the \RelationSpecificNeurons of $r_j$ are deactivated.
Results are displayed in Figure \ref{fig:inter-relation}.\footnote{Using accuracy drop -- a relative measure -- can be noisy when the initial accuracy is low.
However, we show that most relations start with relatively high baseline accuracy (cf. Figure~\ref{fig:neuron_num} and Figure~\ref{fig:neuron_num_13b}), which mitigates the problem.}

When we compare the 7B and 13B models, no consistent pattern emerges across relations.
This indicates that, though being trained on the same data, 
\textbf{differences in model size and parameter initialization appear to substantially change the functionality of neurons}. 
Particularly, most relations in the 13B model are less influenced when neurons of other relations are deactivated than in the 7B model, except in the following cases:
deactivating neurons of \texttt{landmark\_country} strongly affects several other relations concerning the notion of ``location''; \texttt{person\_mother} and \texttt{person\_occupation} are sensitive to the deactivation of neurons of other relations. 
Despite these divergences, we propose two hypotheses that hold across both models.

\shortpar{Neuron versatility.}
We observe that deactivating neurons for one relation can
strongly affect not only that relation but also others, both
closely and loosely related relations. 
E.g., disabling \texttt{person\_pro\_sport} neurons has a large effect on \texttt{person\_sport\_position} (but not vice versa) in both models, likely because a model first needs to understand ``sport'' before inferring ``position''. 
Similarly, deactivating \texttt{person\_father} neurons reduces accuracy on \texttt{person\_mother}, as both share the concept of a 
parental relationship. 
Even loosely related relations can exhibit a clear accuracy drop: deactivating \texttt{star\_constellation} neurons affects \texttt{landmark\_continent} in both models, possibly because both involve the abstract notion of ``location''.

\shortpar{Neuron interference.}
Deactivating \RelationSpecificNeurons for one relation can sometimes \textbf{improve} the accuracy for others -- a phenomenon more pronounced in the 7B model, likely because its smaller parameter space is less capable of isolating different relations. 
In the 7B model, several relations frequently benefit from this effect: for instance, \texttt{person\_mother} improves when neurons from 5 out of 11 other relations -- mostly ``less related'' ones -- are deactivated. 
This effect is also observed for closely related relations:
disabling \texttt{company\_ceo} neurons slightly boosts accuracy on \texttt{company\_hq} for both models. 
Interestingly, the 13B model shows the opposite effect for \texttt{landmark\_continent} when disabling \texttt{landmark\_country}, implying that country information can help predict a continent for the larger model. 
These findings indicate that \textbf{neuron interference happens across model sizes, but its specific patterns vary.}

\section{Complementary Analyses}\seclabel{analysis}

\subsection{Influence of the Numbers of Neurons}\seclabel{neuron_number}

In this section, we investigate the effect of varying the number of \RelationSpecificNeurons
on the 7B model
(see \secref{13b_analysis} for 13B).
Specifically, we consider \textbf{ten} values: 10, 50, 200, 500, 1,000, 3,000, 10,000, 20,000, and 50,000. 
When deactivating varying numbers of neurons for a relation,
we report \textbf{accuracy} for that relation and
the \textbf{average accuracy} for all other relations in
Figure \ref{fig:neuron_num}. Results for all
relation-relation pairs are in Figure \ref{fig:neuron_num_all}.

\shortpar{Neuron cumulativity.}
By increasing the number of neurons for deactivation, we see a consistent accuracy drop in all relations. 
This suggests neuron cumulativity: \textbf{LLMs distribute relational knowledge across multiple neurons, which jointly contribute to dealing with facts belonging to a relation.}
However, cumulativity varies across relations.
Some relations are far more sensitive to a smaller-scale deactivation than other relations, indicating a smaller set of neurons is specifically leveraged for those relations.
We hypothesize that this sensitivity may correlate with the frequency of the facts in each relation in the pretraining data:
more frequent facts may be memorized more robustly
and thus remain less sensitive to deactivation. We empirically verify this hypothesis in \secref{frequency}.

\shortpar{Deactivating \RelationSpecificNeurons has a marginal effect on other relations until certain thresholds are reached.} 
Typically, these thresholds lie between 3,000 and 10,000 as shown in Figure \ref{fig:neuron_num}, below which the 
accuracy on other relations remains stable -- \textbf{supporting the choice of 3,000 neurons in \secref{exp_setup}.} 
Once more neurons are deactivated, other relations also deteriorate, consistent with our \textbf{neuron versatility} hypothesis. 
However, even deactivating up to 50,000 neurons seldom reduces other relations to near-zero accuracy, suggesting a high degree of relation-specificity.
One exception is \texttt{company\_hq}, for which disabling 50,000 neurons causes all relations’ accuracies to approach zero -- possibly because some of these neurons underlie more general generation capabilities of the model \citep{sun2024massive,yu2024superweight}.


\begin{figure}
    \centering
    \setlength{\abovecaptionskip}{-0.01cm}
    \setlength{\belowcaptionskip}{-0.2cm}
\includegraphics[width=0.48\textwidth]{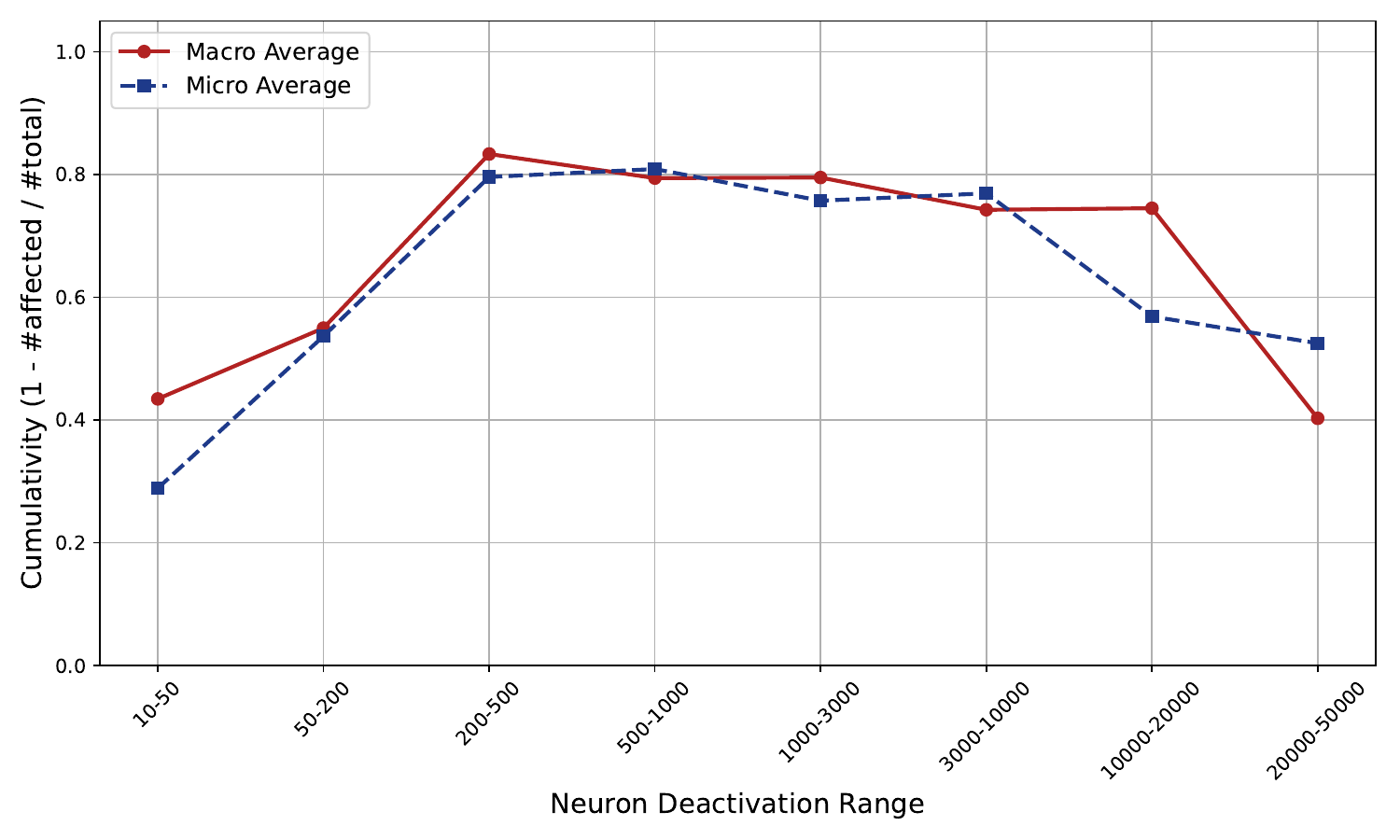}
    \caption{Macro and micro averaged neuron cumulativity for each neuron deactivation range. Cumulativity is defined as $ 1-\frac{\#\text{affected}}{\#\text{total}}$, with macro averaging across relations and micro averaging across prompts. Both trends show that cumulativity increases as the range increases.}
\label{fig:average_cumulatity}
\end{figure}

\shortpar{Validation of the cumulative effect.}
It remains unclear whether the further accuracy drop between any two thresholds in Figure \ref{fig:neuron_num} is driven by \textbf{the newly deactivated neurons} (the isolated effect of deactivated neurons) or \textbf{the cumulative effect of all deactivated neurons}. 
To further validate our neuron cumulativity hypothesis, we conduct an experiment on each consecutive pair of thresholds, e.g., 1000-3000.
Specifically, we identify prompts from $\mathcal{P}_{r_i}^{\text{eva}}$ where the model answers correctly with neurons of the smaller range being deactivated, but fails when neurons of the larger range are deactivated (\#total). 
We then deactivate only the neurons from the intermediate difference and measure the number of affected prompts -- prompts for which the model answers wrongly (\#affected).
Figure \ref{fig:average_cumulatity} shows the macro and micro averaged cumulativity, defined as $ 1-\frac{\#\text{affected}}{\#\text{total}}$.
We notice that neuron behavior becomes increasingly cumulative as the range increases, indicating that only deactivating neurons from the intermediate difference is not enough to make the model answer wrongly. 
There is a drop after the ranges 10000-20000 and 20000-50000, which can be explained by the fact that many more neurons are deactivated compared with the earlier ranges.
We also show the individual number of \#total/\#affected prompts in each relation in each range in Table \ref{tab:cumulative_effect}. 
\textbf{Thus, our results favor the cumulative effect over the isolated effect} -- multiple neurons jointly contribute to dealing with facts belonging to a relation, with no single neuron fully encoding a fact on its own.



\subsection{Are These Neurons Multilingual?}

\begin{figure}
    \centering
    \setlength{\abovecaptionskip}{-0.05cm}
    \setlength{\belowcaptionskip}{-0.2cm}
\includegraphics[width=0.48\textwidth]{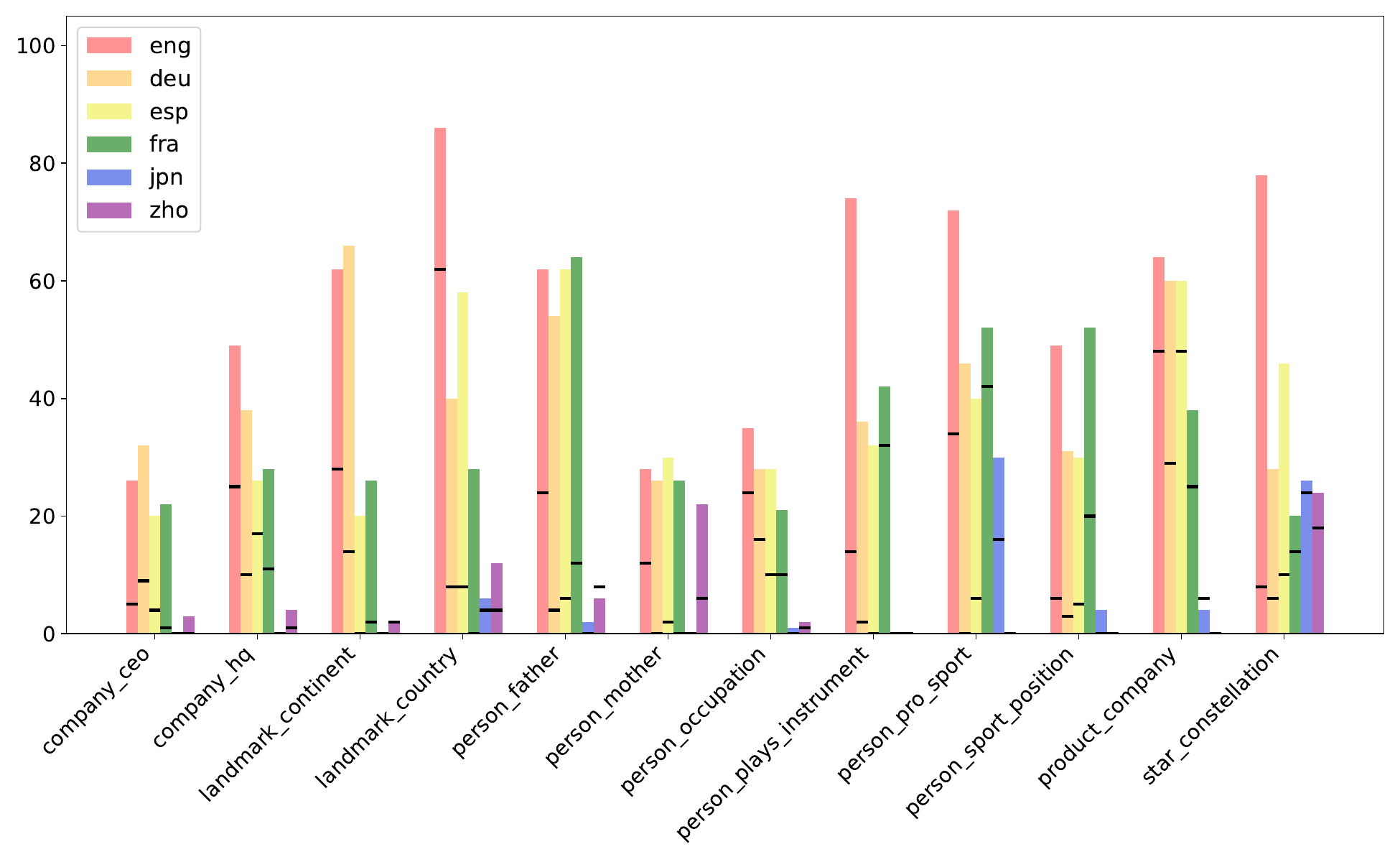}
    \caption{Accuracy on 12 relations across 6
    languages. The bars show the accuracy of the original 7B
    model. The horizontal line in each bar indicates the performance after deactivation of 3,000 \RelationSpecificNeurons.
    Even though these neurons are identified using English prompts, they usually influence other languages, indicating multilinguality of these neurons.}
    \label{fig:multilingual}
\end{figure}

Recent studies suggest that some neurons encoding factual knowledge or handling specific tasks are 
language-agnostic
\citep{stanczak-etal-2022-neurons,zhang2024multilingual,wang2024sharing}. 
A natural question is whether 
\RelationSpecificNeurons\ -- identified solely via English prompts -- also function across languages.
To explore this, we translate $\mathcal{P}_{r_i}^{\text{eva}}$ to 5 languages: German (\textbf{deu}), Spanish (\textbf{esp}), French (\textbf{fra}), Chinese (\textbf{zho}), and Japanese (\textbf{jpn}) (see \secref{translation} for details).
We then deactivate the previously identified 3,000
neurons in the 7B model and measure the effect on these languages, as shown in Figure~\ref{fig:multilingual}. 

Although the model’s accuracy is generally lower in non-English languages, it still shows good factual recall
for most relations (except for jpn and zho). 
Once the neurons for a given relation are deactivated, the accuracy drops across nearly all languages -- \textbf{supporting our neuron versatility hypothesis.} 
Our findings align with recent explanations that LLMs tend to translate the input text from any language into English for task solving in the middle layers based on a shared representation space \citep{wendler-etal-2024-llamas,dumas2024llamas,zhao2024multilingualism}.
As a result, deactivating ``English'' neurons naturally disrupts this shared space, impairing the model’s capability to generalize across languages for the affected relation.

\subsection{Effect of Prompt Templates}\seclabel{effect_prompt}

\begin{figure}
    \centering
\includegraphics[width=0.45\textwidth]{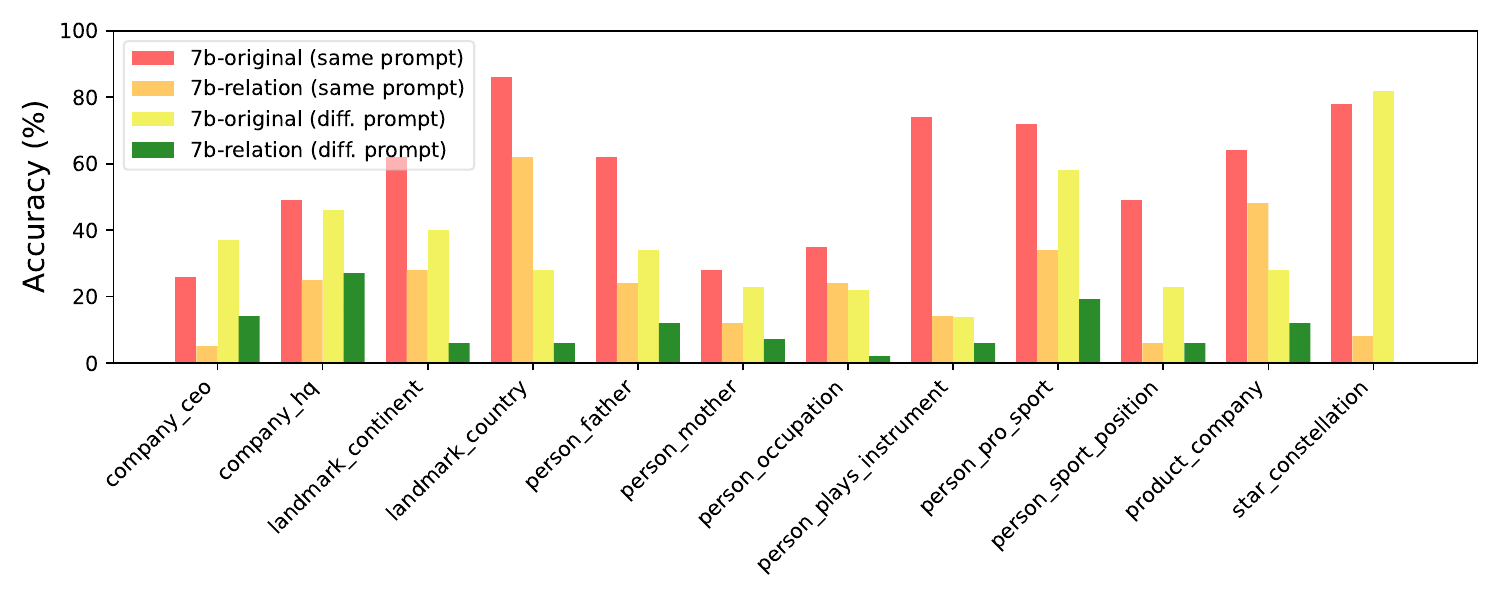}
    \caption{Intra-relation results on original prompts $\mathcal{P}_{r_i}^{\text{eva}}$ and additional prompts $\mathcal{P}_{r_i}^{\text{eva-2}}$. 
    $\mathcal{P}_{r_i}^{\text{eva-2}}$ is constructed with same triples as $\mathcal{P}_{r_i}^{\text{eva}}$ but different prompt templates are used. A consistent decrease across relations indicates that the identified neurons are not specific to prompts.}
    \label{fig:intra-relation_other_format}

\end{figure}

There is a possible confounding variable: the identified
relation-specific neurons could be associated with the
prompt templates used in $\mathcal{P}_{r_i}^{\text{det}}$.
The degradation in $\mathcal{P}_{r_i}^{\text{eva}}$ would
then be due to the identified neurons encoding syntactic structure rather than abstract relation semantics.
To exclude this confounding variable, we create an additional evaluation set $\mathcal{P}_{r_i}^{\text{eva-2}}$ where the \textbf{same triples} as $\mathcal{P}_{r_i}^{\text{eva}}$ but \textbf{different prompt templates} are used for each relation.
We then deactivate the previously identified 3,000 neurons in the 7B model and measure the effect on the new prompts. 
Figure~\ref{fig:intra-relation_other_format} presents the results. We observe that the accuracy with new prompts is a bit different from the accuracy when the original templates are used. 
This is not surprising since LLMs are sensitive to the prompt templates \citep{Sclar2024Quantifying}.
Nevertheless, we still see that the deactivation of neurons results in consistent accuracy drops for new prompts across relations.
Therefore, the neurons are not subject to the templates used to describe the relation.
Instead, \textbf{the identified neurons are only associated with the abstract relation semantics.}

\subsection{Relations vs. Concepts}

\begin{figure}[tb]
    \centering
\includegraphics[width=0.45\textwidth]{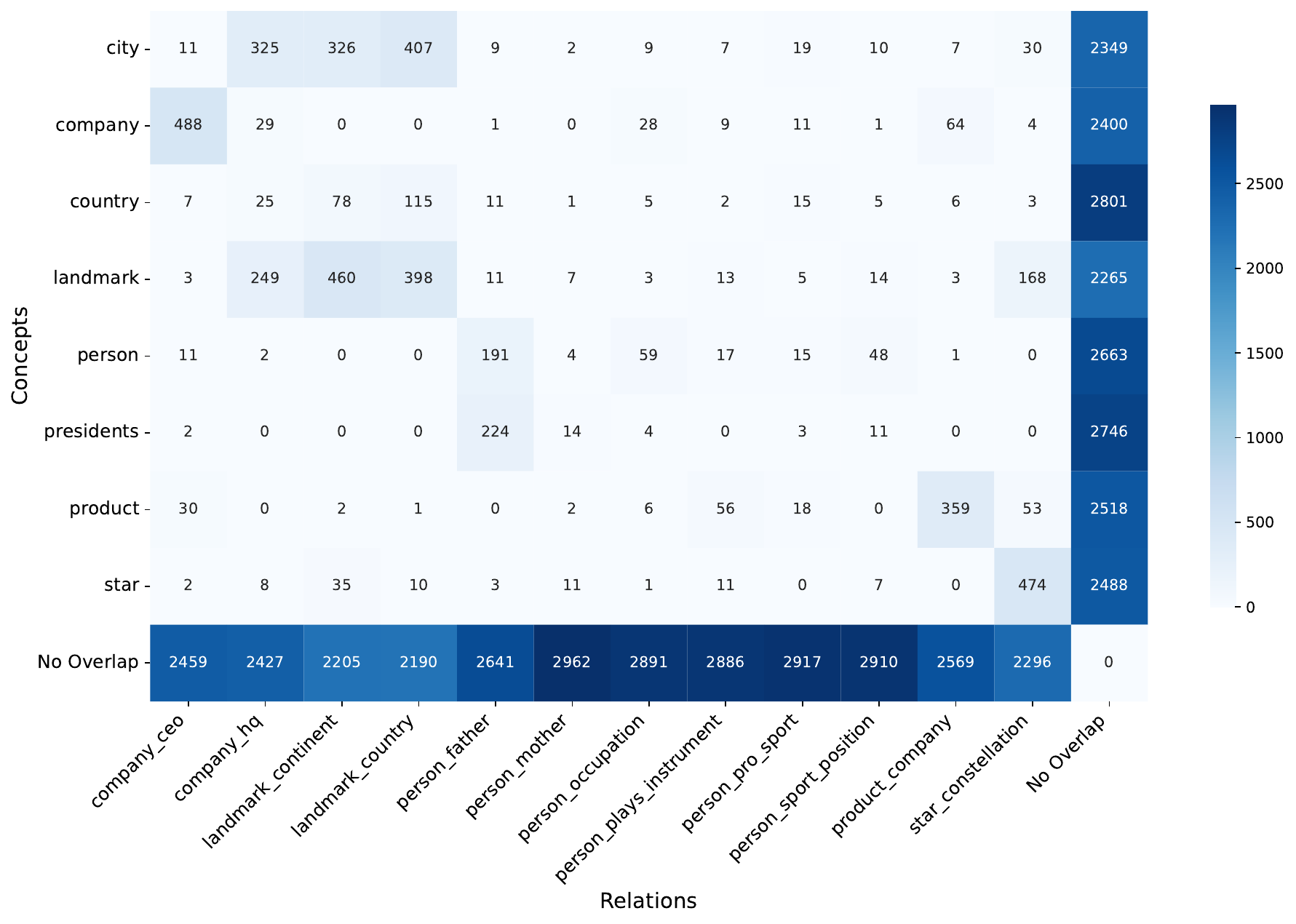}
    \caption{Overlap between the top 3000 neurons of
      relations and  concepts in the 13B model.}
    \label{fig:entity_neuron_overlap}
\end{figure}


We saw in Figure    \ref{fig:neuron_overlap} that the storage of relations is generally well separated, but there are exceptions. We can view a relation as relating two
\textbf{concepts} or \textbf{topics}, e.g.,     
\texttt{company\_ceo} relates instances of the subject concept
``company'' to instances of the object concept ``CEO''.
From this perspective, the exceptions in Figure~\ref{fig:neuron_overlap},
i.e., cases where a relation $r_1$
overlaps with a relation $r_2$, are  generally cases where the concepts of $r_1$ and $r_2$ are
the same or overlap.
To further explore this hypothesis empirically, we again use the method applied in \secref{method} to relations, but now use it for
subject concepts.\footnote{We do not consider the object concepts explicitly because the objects are not presented in the prompts for relation-specific or concept-specific neuron identification (cf. \secref{method}).} 
That is, we identify sets of \textbf{concept-specific neurons}.
We group the triples by their
subjects,
resulting in 9 different concepts. 
We then create prompts with \textbf{novel relations}
such as ``can'' and ``has a'', \textbf{balanced across
positive and negative
samples.} 
This ensures that
the model's completion for a prompt like
(``Lincoln has a'') depends on the concept instance ``Lincoln'', not on the relation.

Figure \ref{fig:entity_neuron_overlap} shows the
overlap between relation neurons and concept neurons. 
Most of the cells with large counts support our
hypothesis that the overlaps between relations we observe are rooted in these relations being representationally associated with their concepts.
Clear examples include
\texttt{company\_ceo} and its subject concept
\texttt{company}; 
\texttt{company\_hq} and its object concept 
\texttt{city} (assuming that \texttt{hq} is a subcategory of
\texttt{city});
and \texttt{landmark\_continent} and its subject concept
\texttt{landmark}. There is little overlap of
\texttt{person} with relations like \texttt{person\_mother},
potentially because \texttt{person} is a more general and
semantically unspecific concept than the others. 
However, most
identified neurons
are only concept neurons or only relation neurons,
\textbf{suggesting that
relational and conceptual representations are largely separate.}

\subsection{Effect on General Language Modeling}\seclabel{ppl}

\begin{table}[h]
\centering
\small
\resizebox{\linewidth}{!}{
\begin{tabular}{lrrr}
\toprule
\textbf{Relation} & \textbf{\# Sentences} & \textbf{PPL (Before)} & \textbf{PPL (After)} \\
\midrule
\texttt{person\_sport\_position} & 8 & 87.67 & 95.18 \\
\texttt{person\_pro\_sport} & 5 & 34.42 & 45.54 \\
\texttt{person\_occupation} & 19 & 62.71 & 84.31 \\
\texttt{company\_hq} & 50 & 63.97 & 61.27 \\
\texttt{product\_company} & 23 & 100.65 & 95.37 \\
\texttt{person\_mother} & 50 & 141.31 & 146.99 \\
\texttt{person\_father} & 50 & 90.14 & 83.54 \\
\texttt{landmark\_continent} & 4 & 90.19 & 73.82 \\
\texttt{landmark\_country} & 50 & 58.57 & 51.12 \\
\texttt{company\_ceo} & 50 & 102.32 & 110.71 \\
\texttt{person\_plays\_instrument} & 5 & 69.97 & 63.91 \\
\texttt{star\_constellation} & 26 & 47.12 & 57.48 \\
\bottomrule
\end{tabular}
}
\caption{Perplexity before and after ablation of \RelationSpecificNeurons on synthetic sentences where the object appears in a context without subject or relation.}
\label{tab:object_ppl}
\end{table}

One potential concern with ablating \RelationSpecificNeurons is the risk of inadvertently impairing general language modeling, particularly for tokens associated with the object entity in contexts unrelated to the original subject–relation pair. 
To investigate this, we design a new experiment to measure \textbf{perplexity} on synthetic sentences where the object appears in naturalistic but relation-neutral context -- without the original subject or relation. 
For each of the 12 covered relations, we construct up to 50 sentences (using distinct objects from $\mathcal{P}_{r_i}^{\text{det}}$) with fixed templates, such that the object entity appears as the final token (see \secref{prompts} for prompt templates).
We then compute the perplexity of these sentences before and after ablating the \RelationSpecificNeurons. 
The results are summarized in Table~\ref{tab:object_ppl}, which reports average perplexity across sentences for each relation.
The results show no systematic degradation in perplexity after ablation. 
For several relations (e.g., \texttt{company\_hq}, \texttt{landmark\_country}, \texttt{product\_company}), the perplexity even slightly decreases. 
This suggests that the object-token generation ability is preserved, and that the \textbf{ablation primarily targets mechanisms specific to the factual relation rather than disrupting broader lexical or contextual knowledge of the models}.

\section{Conclusion \label{ref:conclusion}}

This work highlights the existence of relation-specific neurons in LLMs -- neurons that focus on relations rather than entities. 
Our experiments show that \RelationSpecificNeurons primarily reside in the middle layers and can be shared across multiple relations. 
Through systematic deactivation, we reveal their influence on both the targeted and other relations, leading to three key hypotheses:
\textbf{neuron cumulativity} (multiple neurons jointly contribute to dealing with facts belonging
to a relation),
\textbf{neuron versatility} (neurons are shared across relations and languages), and
\textbf{neuron interference} (neurons from one relation can disrupt the processing of another).
These findings shed new light on how LLMs handle relational facts at the neuron level, contributing to the interpretability of LLMs.

\section*{Limitations}

While our findings provide valuable insights, several limitations remain and offer opportunities for future research.
First, this work focuses on factual knowledge grouped into 12 relations because the reliability of the neuron identification method requires enough facts in each relation.
Although this selection does not diminish the validity of our findings and hypotheses, it represents a relatively narrow set of relations. Future work can explore a broader range of relations and analyze how relation-specific neurons behave across a more diverse set of relations.
Second, our multilingual analysis includes only five languages. While these languages demonstrate neuron versatility, they do not fully capture linguistic diversity. Future research could investigate additional languages, particularly low-resource ones, to determine whether relation-specific neurons exhibit similar relational functionality across these languages.
Thirdly, we draw our findings from the LLama-2 family in the main content due to page limit and resource constraints. 
We also conduct the same investigation on Gemma-7B \citep{gemma2024team} (cf.\ \secref{gemma}), which shows similar trends as we observe for models from the LLama-2 family.
Future work can explore even larger models or models with post-training techniques like instruction-tuning.
Lastly, we observe that more frequent facts tend to be more robust to the deactivation of relation-specific neurons in both the 7B and 13B models (cf.\ \secref{frequency}). Fact frequency is approximated using the Dolma corpus \citep{soldaini-etal-2024-dolma} in this study. However, LLama-2 models may incorporate a larger and more diverse pretraining dataset, potentially leading to some discrepancies between these approximated fact frequencies and their actual frequencies.

\section*{Acknowledgments}

This research was supported by DFG (grant SCHU 2246/14-1). We gratefully acknowledge support from Google through a generous research grant. We appreciate suggestions and comments from other members of CIS, LMU Munich. We want to thank Lixi Liu’s suggestions for figure design.

\bibliography{custom}

\appendix

\section{Related Work}

Mechanistic interpretability (MI) is a growing subfield of interpretability that aims to understand LLMs by breaking them down into smaller components and fundamental computations. It has gained significant attention for studying how LLMs recall factual knowledge learned during pretraining \citep{meng2022locating,dai-etal-2022-knowledge,geva-etal-2023-dissecting,yu-etal-2023-characterizing,lv2024interpreting,wang-etal-2024-unveiling}. 
Following \citet{olah2020zoom, rai2024practical}, MI research can be categorized into two areas: the study of \textbf{features} and the study of \textbf{circuits}, based on the type of decomposed components. Features refer to human-interpretable properties encoded in model representations or represented by model components, such as neurons and attention heads \citep{elhage2022solu, gurnee2023finding}. Circuits are subgraphs of the model's computation graph responsible for implementing specific behaviors 
\citep{wang2022interpretability, elhage2021mathematical}.

In this work, we focus on neuron-level feature-based interpretability analysis to localize relation-specific neurons, which are responsible for encoding and recalling specific types of factual knowledge. Existing studies have utilized various approaches for neuron interpretation, each offering unique advantages and limitations \cite{sajjad-etal-2022-neuron, rai2024practical}. The \textit{visualization} method \citep{olsson2022context, elhage2022solu, lieberum2023does, bills2023language,liu-etal-2024-unraveling} involves visualizing neuron activations and manually identifying the underlying concept across input text. While being straightforward, it relies heavily on human effort and risks overgeneralization. \textit{Statistics}-based methods \citep{Bau2019Identifying,neuron2022Cuadros,kojima-etal-2024-multilingual,yu-ananiadou-2024-neuron,tang-etal-2024-language,wang-etal-2024-unveiling}, on the other hand, aggregate activation statistics across data to establish connections between neurons and concepts, identifying patterns through the co-occurrence of neuron activation values and specific input features. \textit{Probing}-based methods \citep{dalvi2019one,  durrani-etal-2020-analyzing, antverg2021pitfalls, gurnee2024universal} train diagnostic classifiers on neuron activations to identify neurons associated with predefined concepts. These methods are scalable, enabling the discovery of neuron sets across large datasets, though they depend on supervised data annotations. \textit{Causation}-based methods \citep{vig2020investigating, meng2022locating, meng2022mass, kramar2024atp,song-etal-2024-large} take a different approach by directly varying the values of specific neurons or components and analyzing changes in model behavior; significant changes indicate the importance of these neurons or components to particular functionalities. 

Building on this foundation, our work adopts the statistics-based method proposed by \citet{neuron2022Cuadros} to identify relation-specific neurons -- neurons uniquely ``fired'' for queries concerning facts sharing the same relation. This approach facilitates a scalable and targeted analysis of neuron behavior in relation to factual knowledge recall.

\section{Entity Analysis Across Relations}\seclabel{entity_overlap}

\begin{figure*}
    \centering
    \begin{tabular}{c}
    \includegraphics[width=0.48\textwidth]{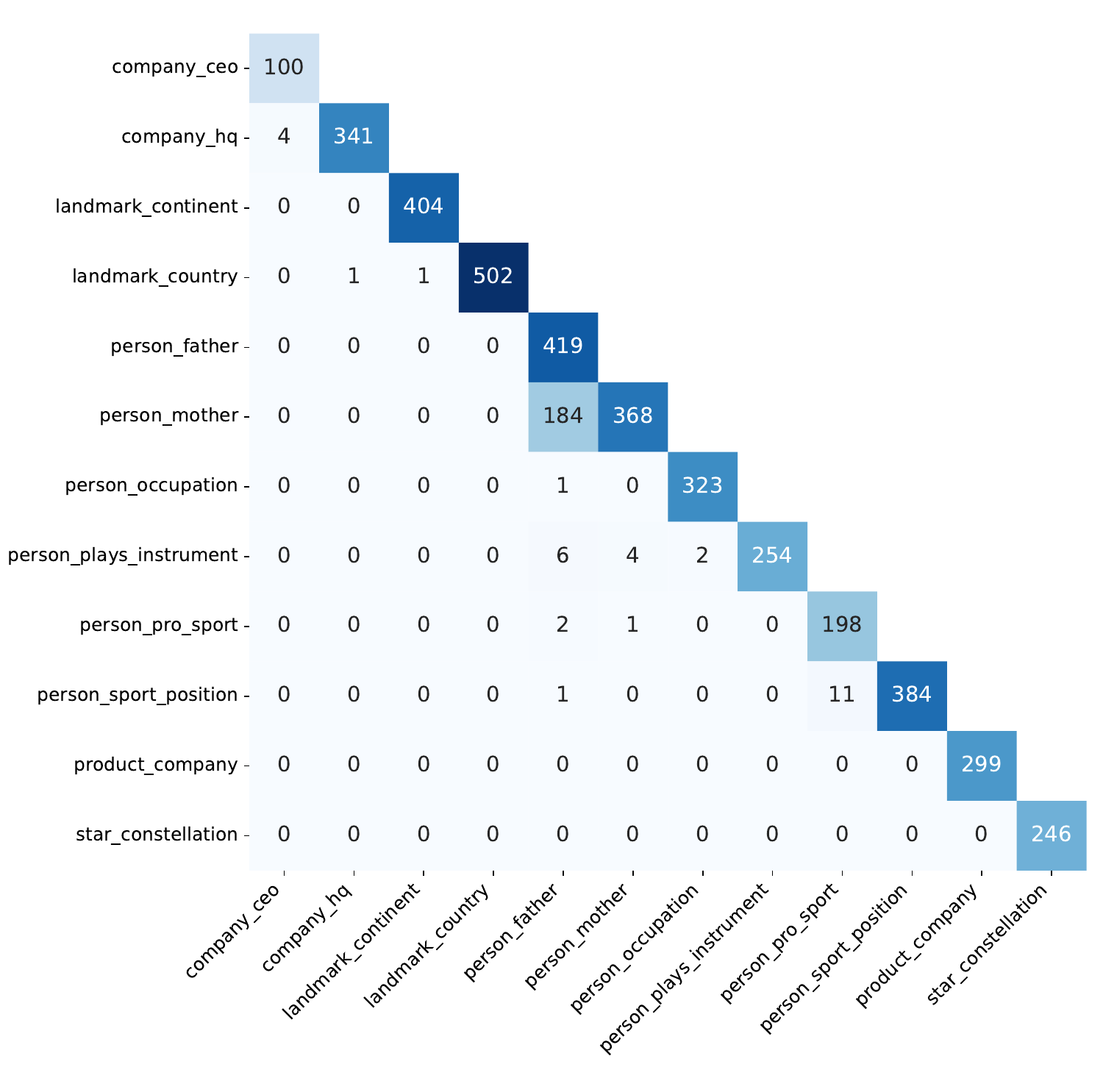}
    \includegraphics[width=0.48\textwidth]{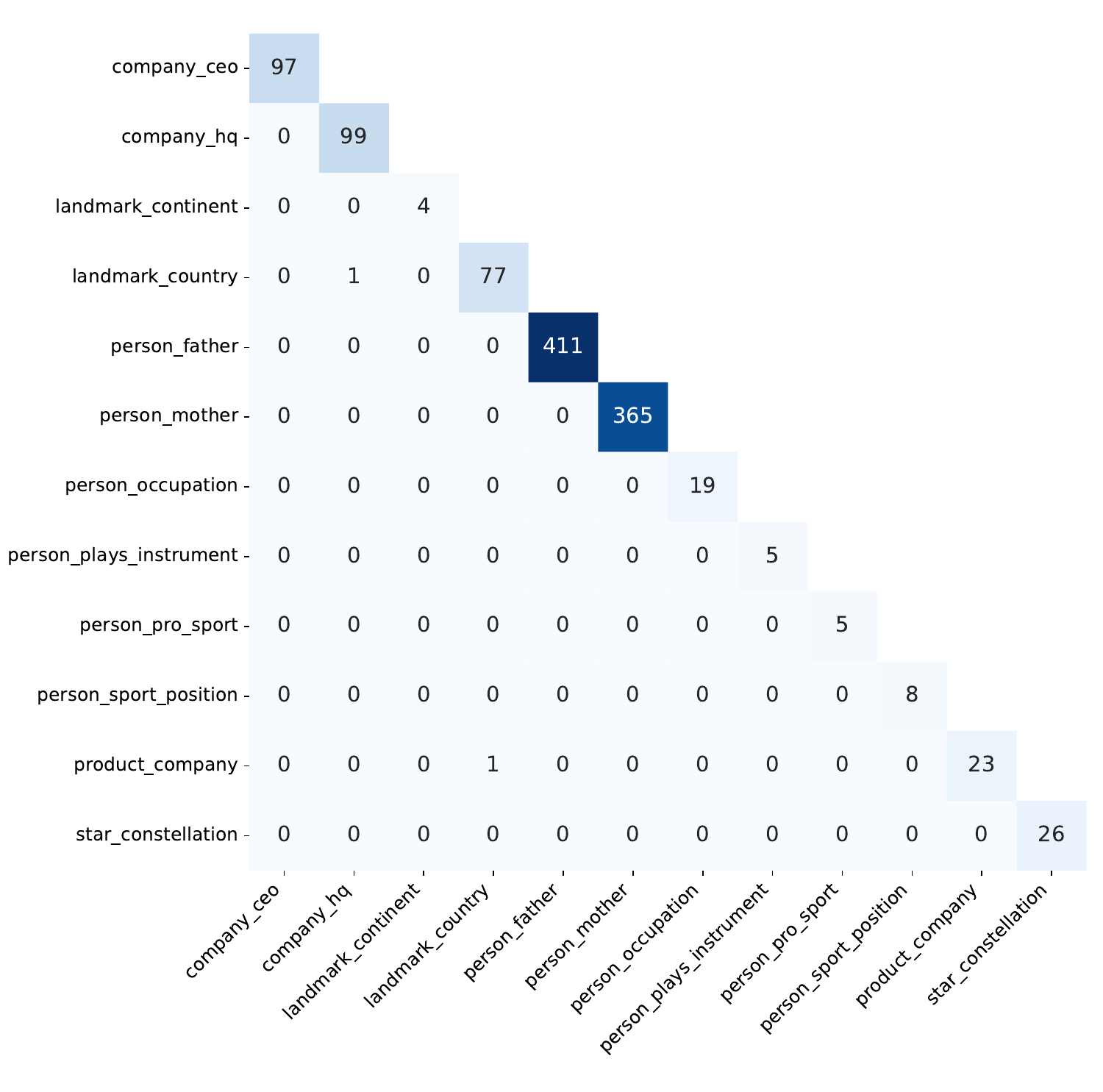}
    \end{tabular}
    \caption{Subject (left) and object (right) overlap across 12 relations obtained from the \textbf{7B} model. The diagonal in each figure shows the number of distinct subjects or objects for each relation. It can be seen that factual knowledge from different relations has almost no entity overlap except for \texttt{person\_mother} and \texttt{person\_father}, which are mostly celebrities.}
    \label{fig:subject_object_overlap_7b}
\end{figure*}

\begin{figure*}
    \centering
    \begin{tabular}{c}
    \includegraphics[width=0.48\textwidth]{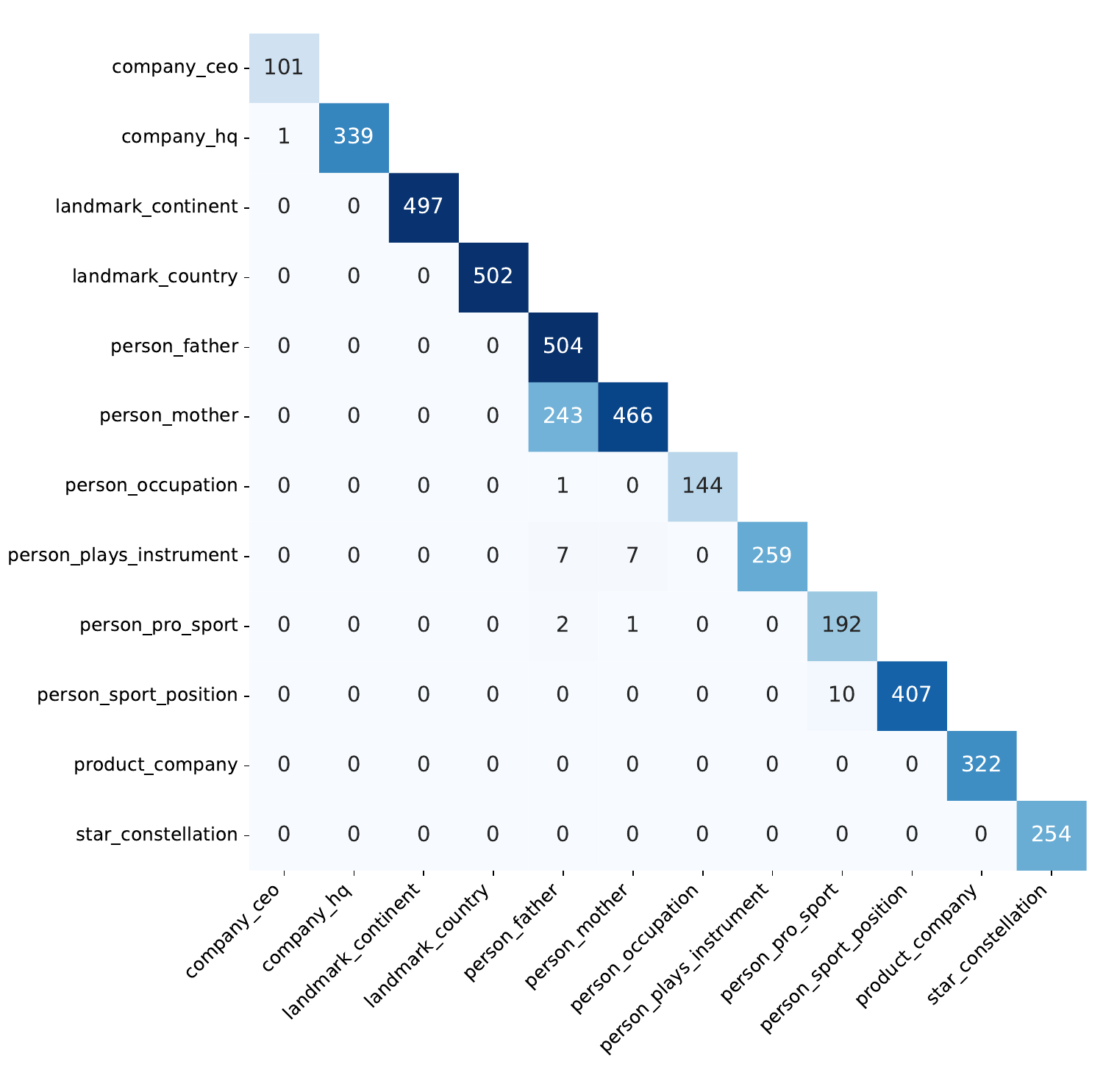}
    \includegraphics[width=0.48\textwidth]{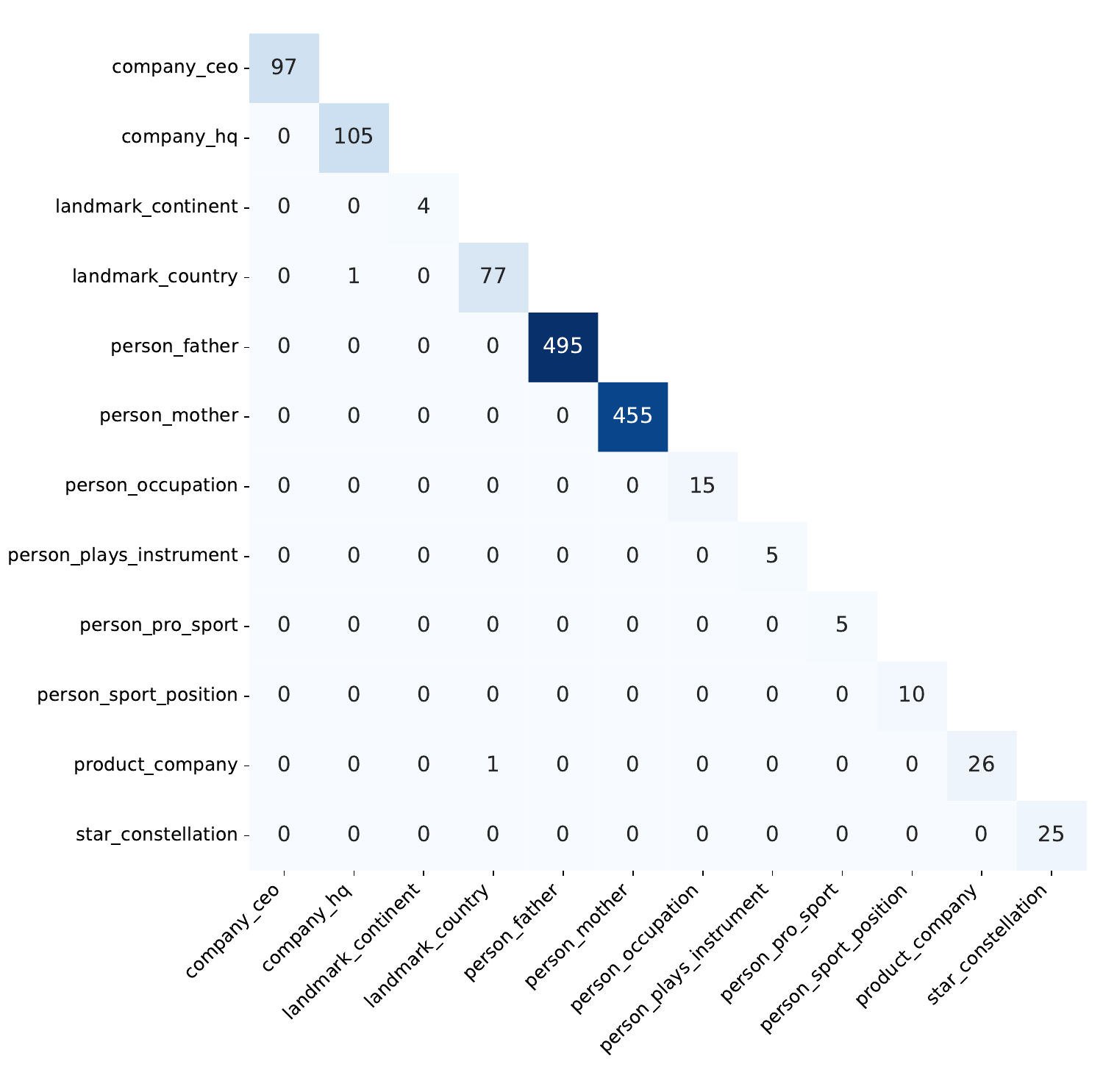}
    \end{tabular}
    \caption{Subject (left) and object (right) overlap across 12 relations obtained from the \textbf{13B} model. The trend is very similar to that in the 7B model: \texttt{person\_mother} and \texttt{person\_father} share many subjects.}
    \label{fig:subject_object_overlap_13b}
\end{figure*}

We show the number of \textbf{distinct subjects (resp. objects)} in each relation and the number of \textbf{overlapping subjects (resp. objects)} between any two relations 
in the identification prompt set $\mathcal{P}_{r_i}^{\text{det}}$
of the 7B model and the 13B model in Figure \ref{fig:subject_object_overlap_7b} and \ref{fig:subject_object_overlap_13b} respectively.
Most two relations have no common or very limited overlapping (less than 11) subjects, except for \texttt{person\_mother} and \texttt{person\_father}, which are mostly celebrities, possibly resulting in extensive neuron overlap between the two relations as we show in \secref{neurons}.
Similarly, no two relations share many objects.

\begin{figure*}
    \centering
    \begin{tabular}{c}
    \includegraphics[width=0.48\textwidth]{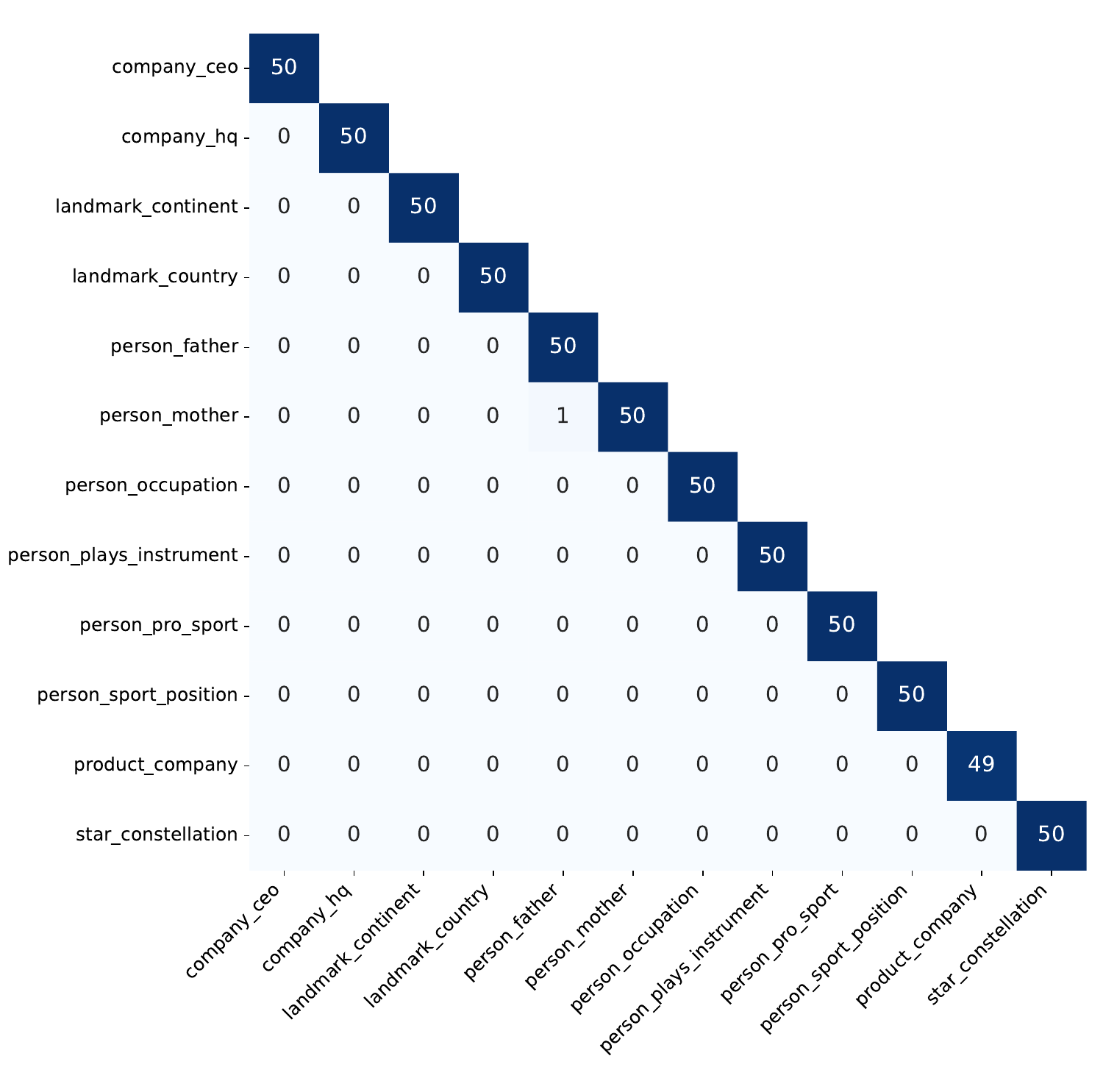}
    \includegraphics[width=0.48\textwidth]{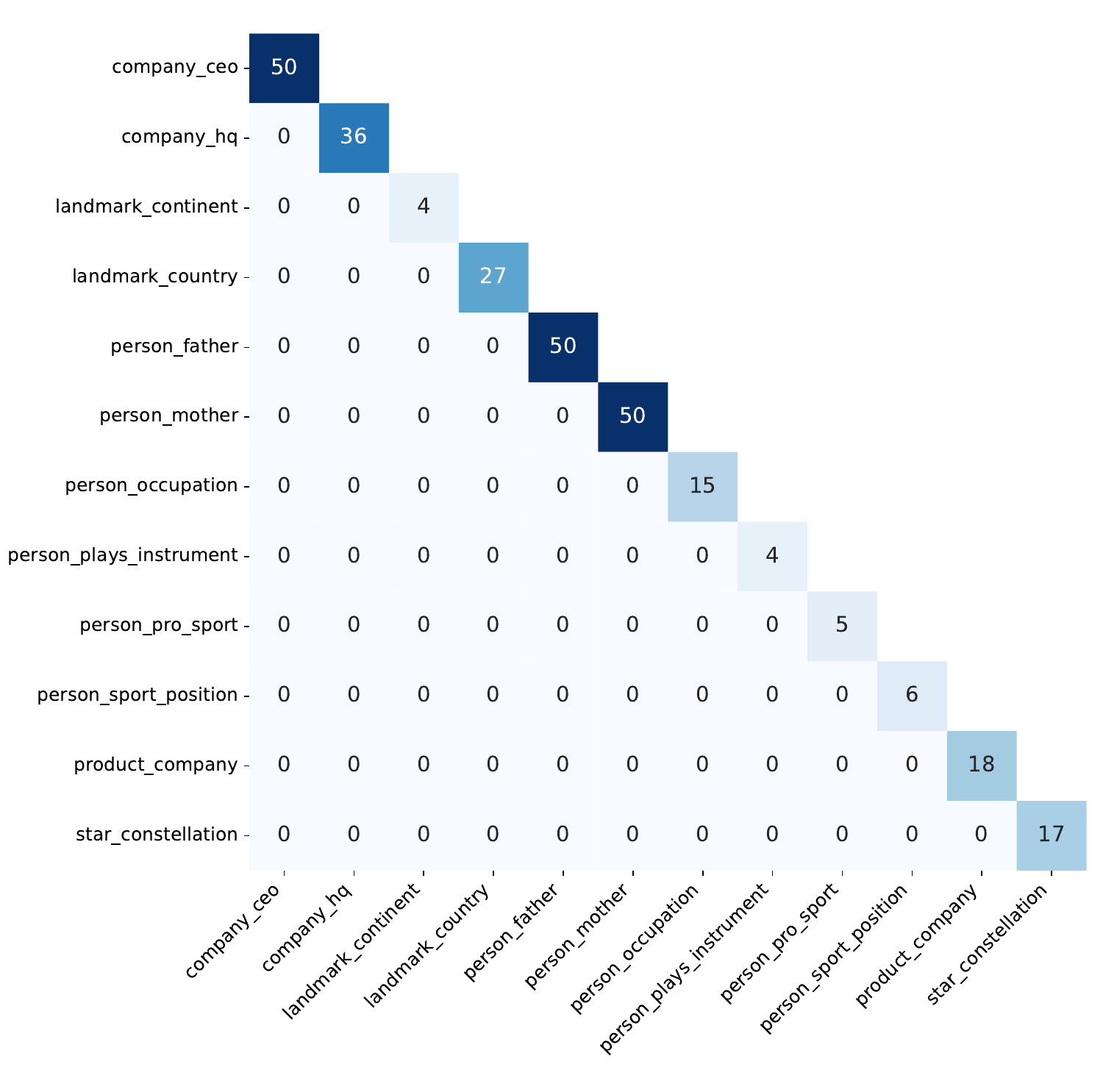}
    \end{tabular}
    \caption{Subject (left) and object (right) overlap across 12 relations in the held-out evaluation prompt set $\mathcal{P}_{r_i}^{\text{eva}}$. Almost no two relations share any subjects or objects.}
    \label{fig:subject_object_overlap_test}
\end{figure*}

Additionally, we show the number of overlapping entities in the evaluation set $\mathcal{P}_{r_i}^{\text{eva}}$ (the 7B and 13B models share the same evaluation set) in Figure \ref{fig:subject_object_overlap_test}.
The results also show almost no entity overlap across different relations: among all relations, only \texttt{person\_mother} and \texttt{person\_father} share \textbf{one} subject, and the rest of the relations do not share any subject or object overlap. 

The diagonal values of the object overlap subfigures (Figures~\ref{fig:subject_object_overlap_7b}, \ref{fig:subject_object_overlap_13b}, \ref{fig:subject_object_overlap_test}, right) reflect the number of \textbf{distinct objects}, while those of the subject overlap subfigures (left) correspond to the total number of \textbf{facts}.
This distinction reveals structural differences across relations.
For instance, in \texttt{company\_ceo}, \texttt{person\_mother}, and \texttt{person\_father}, each fact is typically associated with a unique object -- yielding an almost one-to-one mapping. 
In contrast, relations like \texttt{person\_occupation} involve a small number of frequent objects.
Furthermore, the object distribution varies: some relations (e.g., \texttt{person\_pro\_sport}) are relatively balanced, while others (e.g., \texttt{person\_occupation}) are highly skewed (with many ``actors''), largely due to biases in the original LRE dataset.

Taken together, the entity analysis suggests that entities are not a confounding factor in our experiments. 
The identified \RelationSpecificNeurons capture relation-specific behavior rather than entity-specific patterns.

\begin{figure}
    \centering
    \setlength{\belowcaptionskip}{-0.5cm}
    \includegraphics[width=0.15\textwidth]{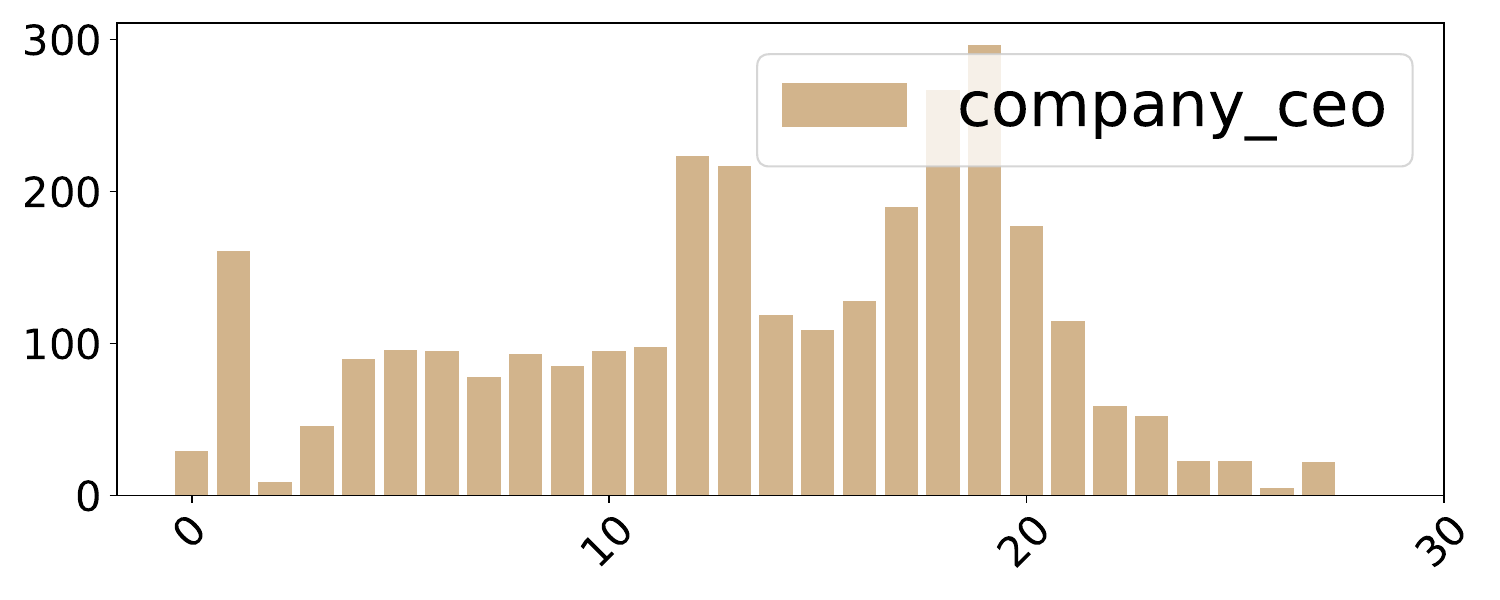}
    \includegraphics[width=0.15\textwidth]{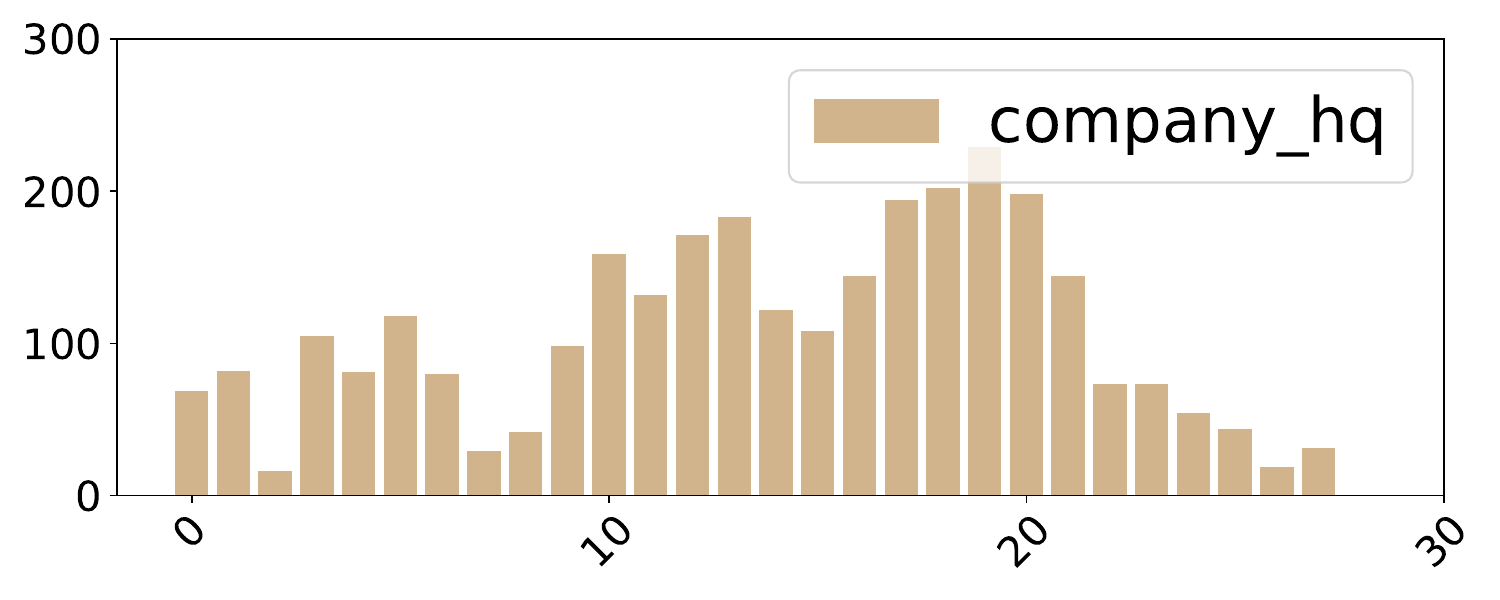}
    \includegraphics[width=0.15\textwidth]{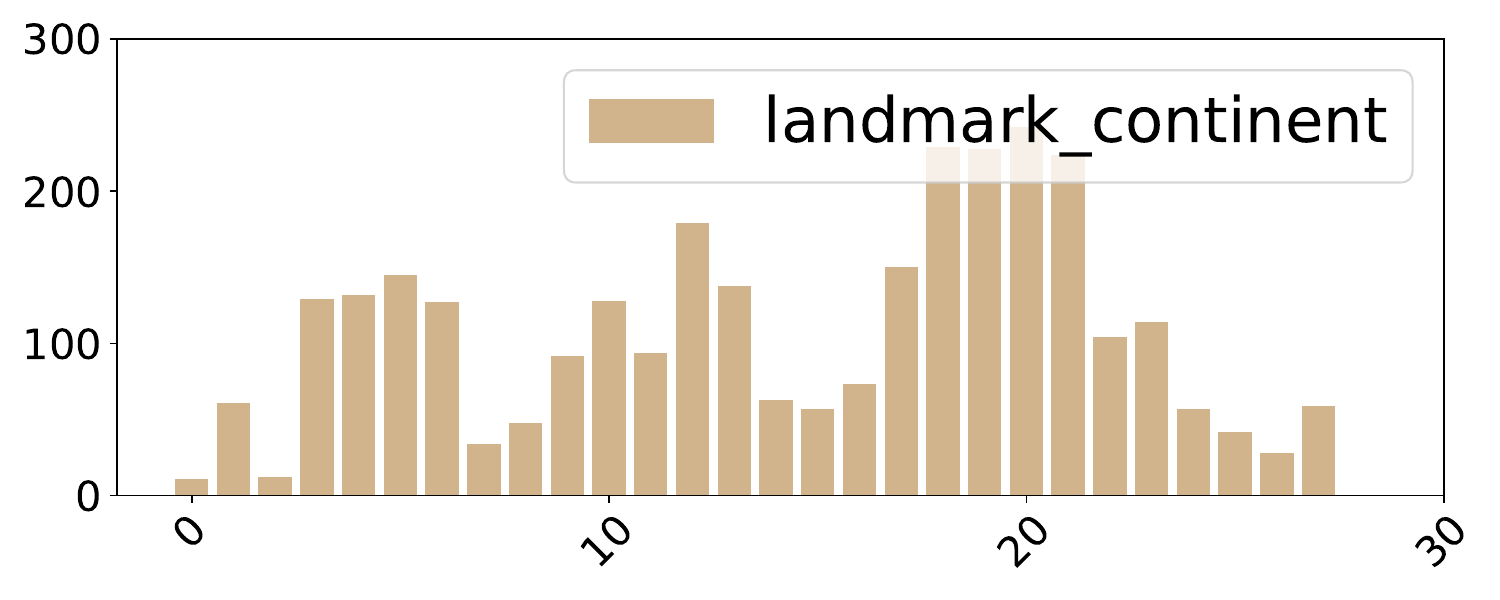}
    \includegraphics[width=0.15\textwidth]{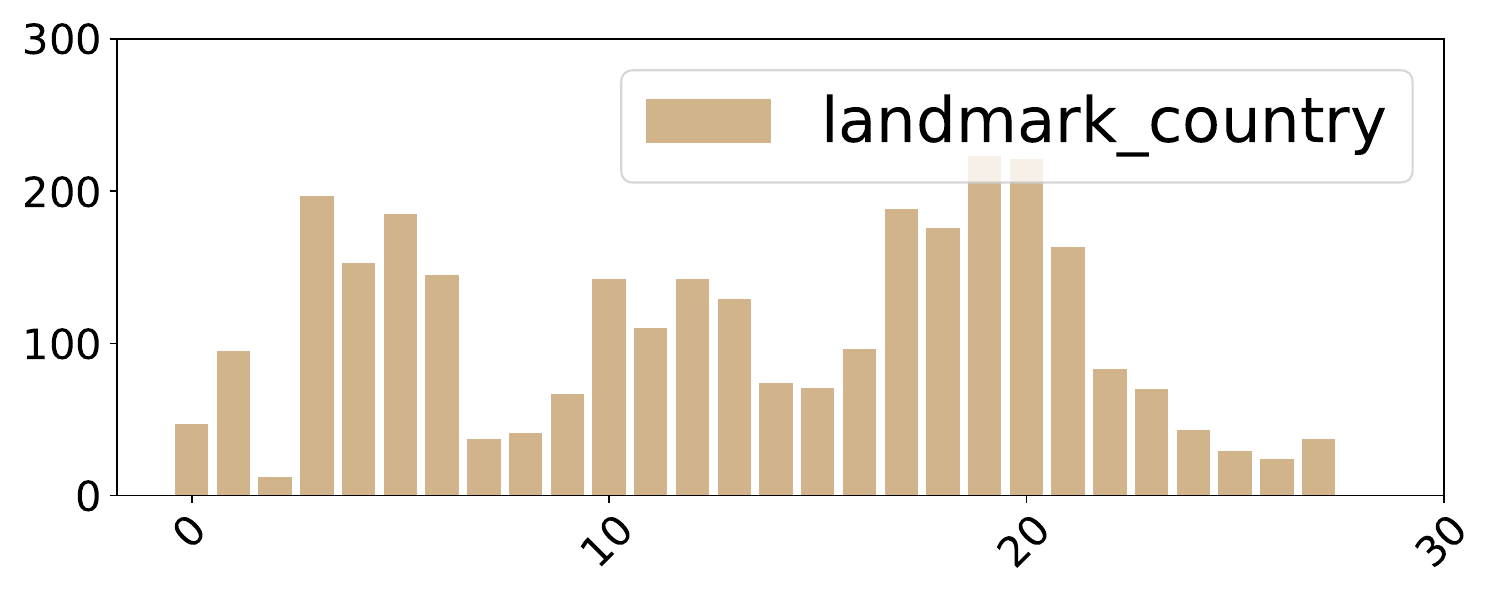}
    \includegraphics[width=0.15\textwidth]{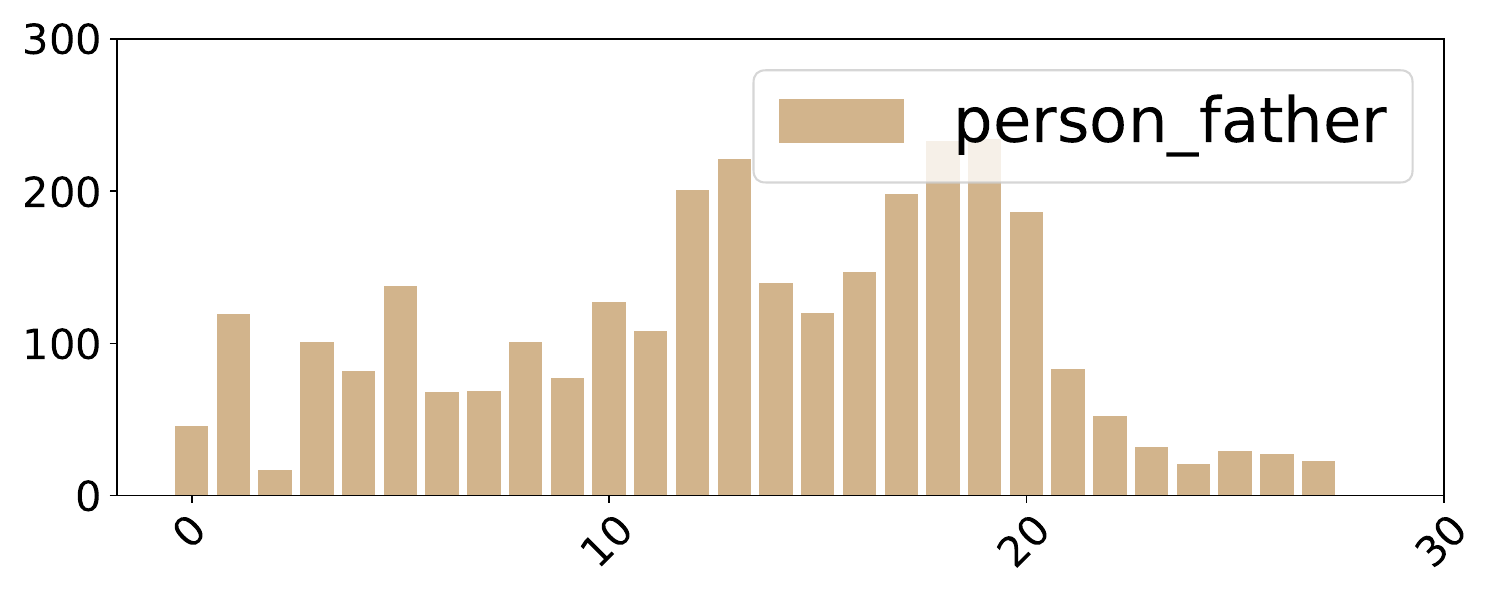}
    \includegraphics[width=0.15\textwidth]{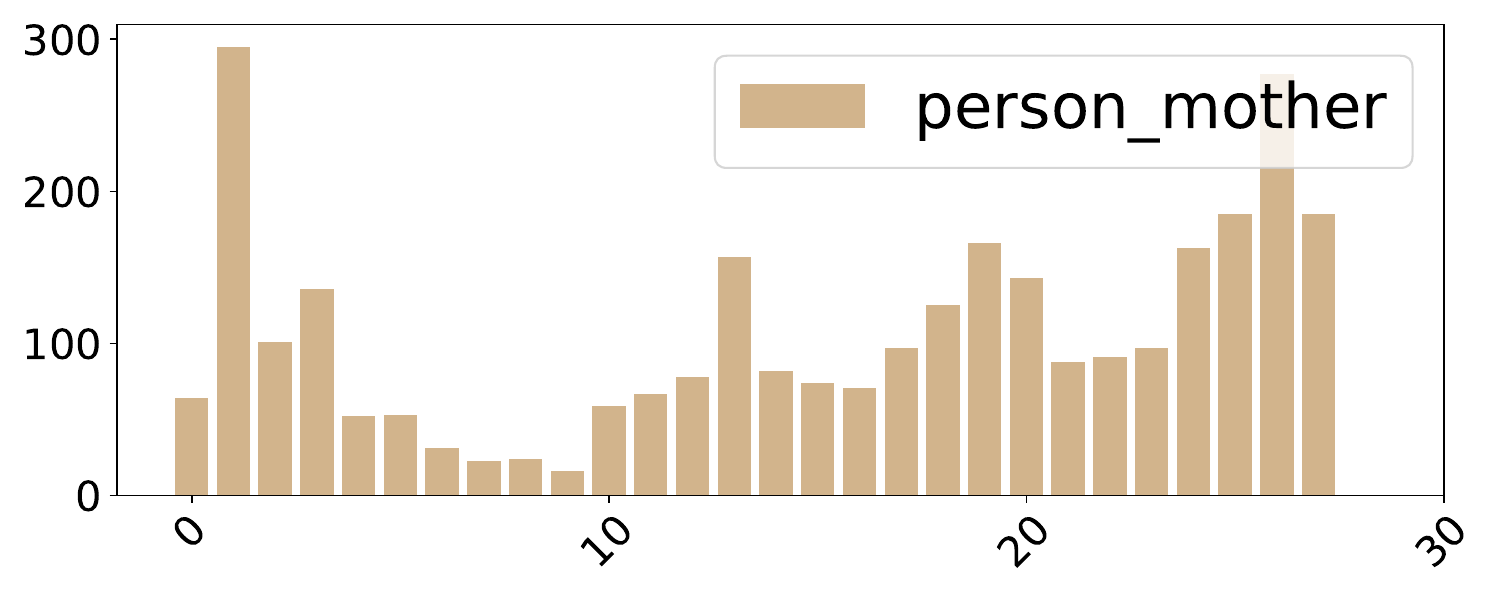}
    \includegraphics[width=0.15\textwidth]{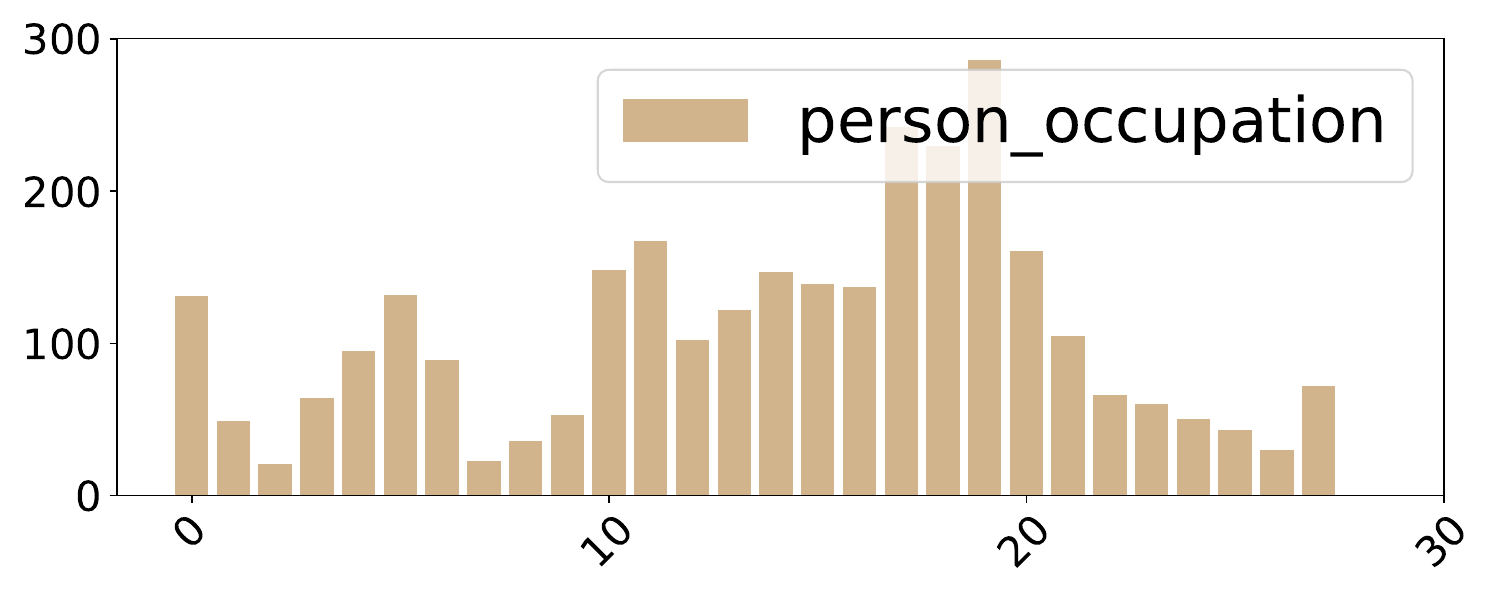}
    \includegraphics[width=0.15\textwidth]{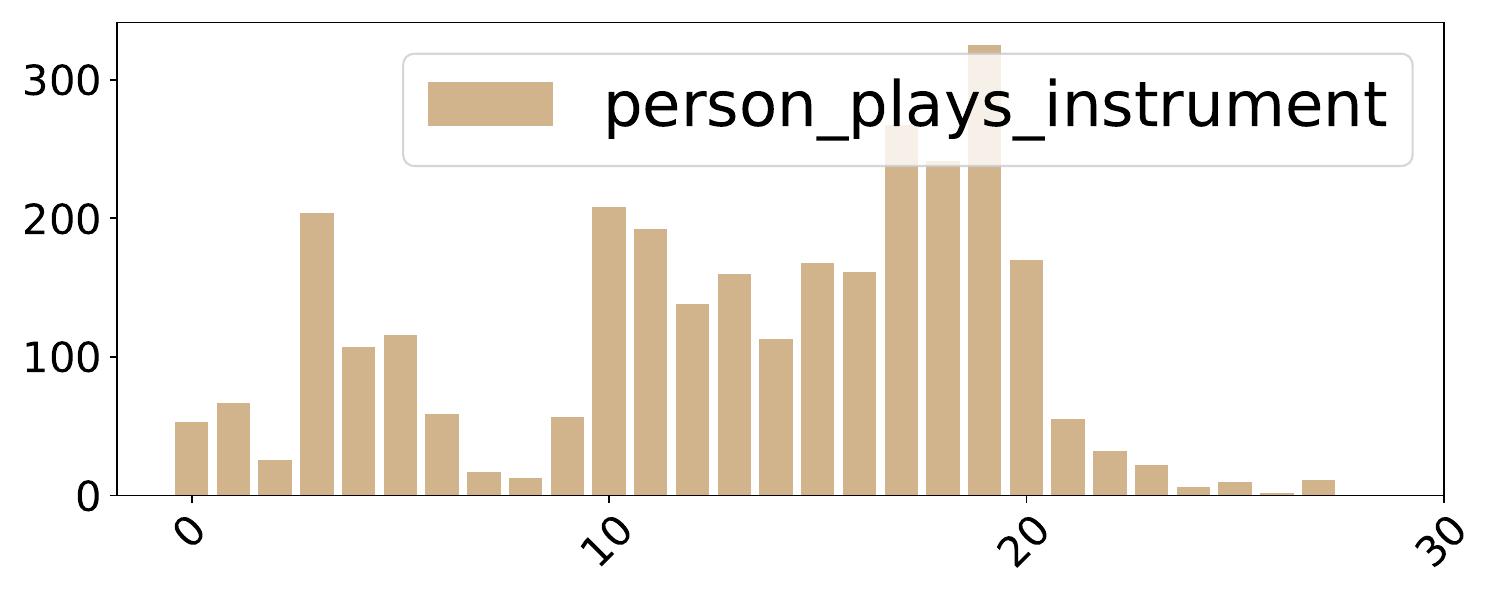}
    \includegraphics[width=0.15\textwidth]{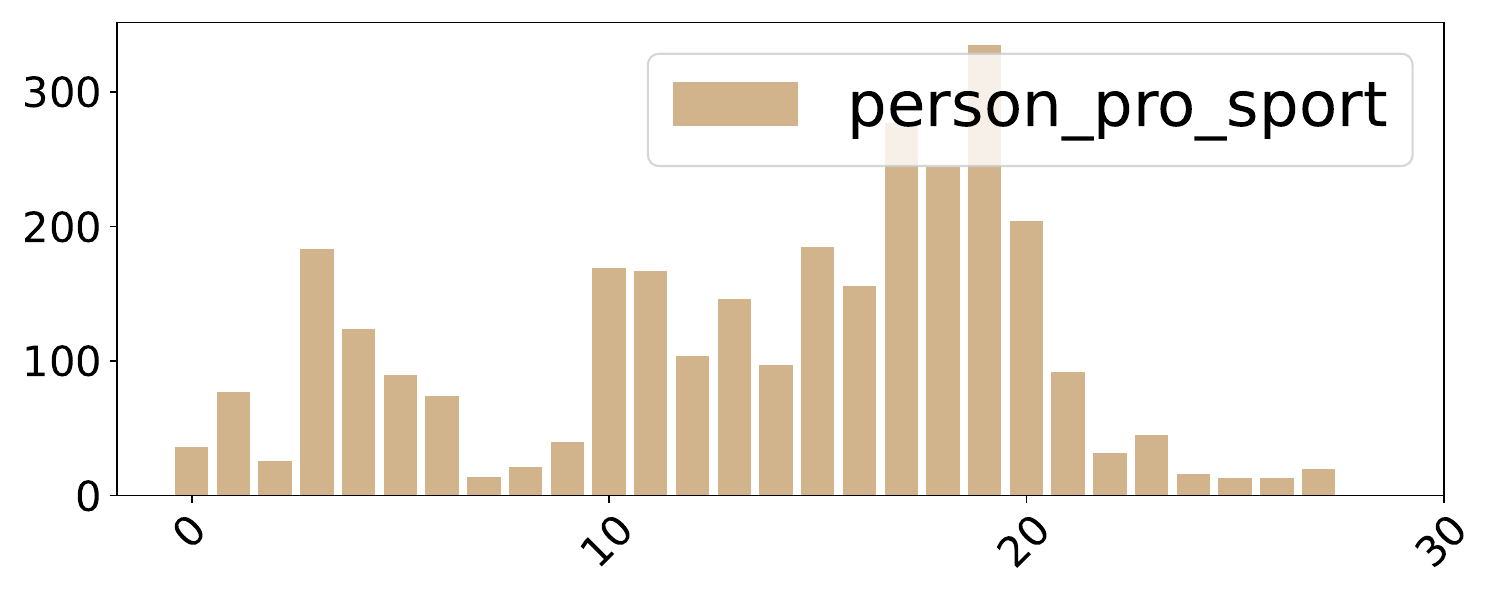}
    \includegraphics[width=0.15\textwidth]{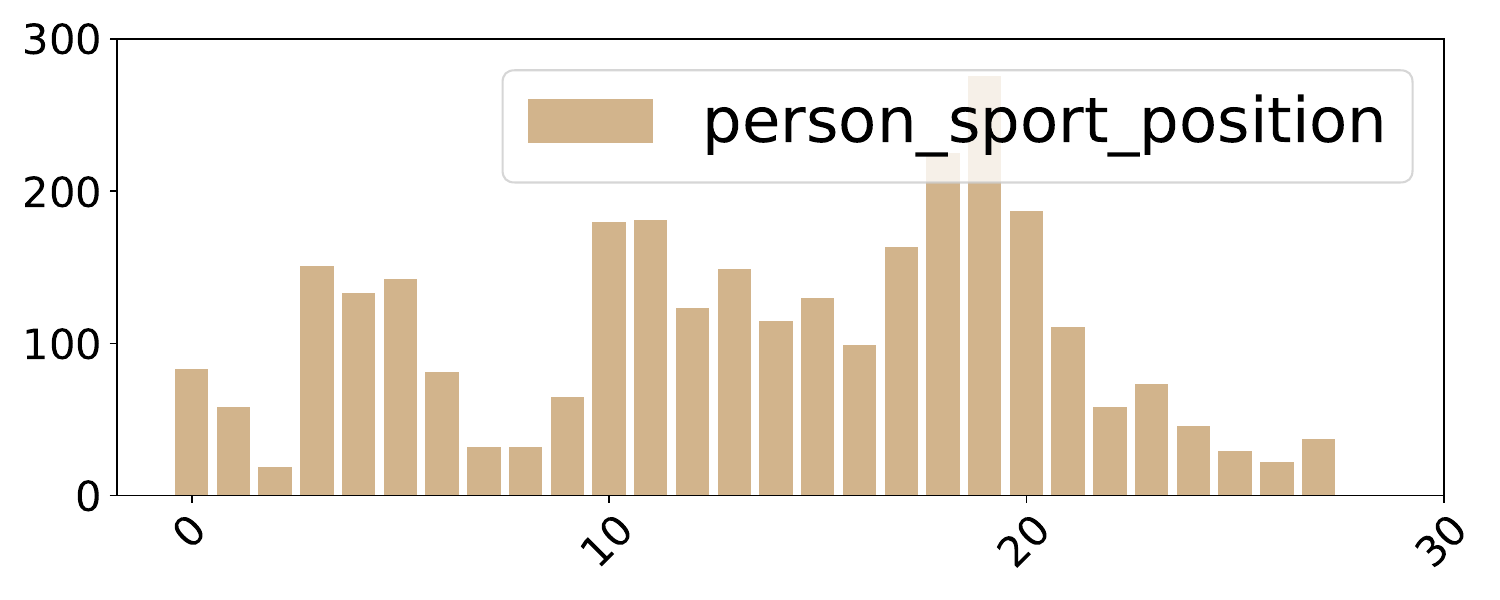}
    \includegraphics[width=0.15\textwidth]{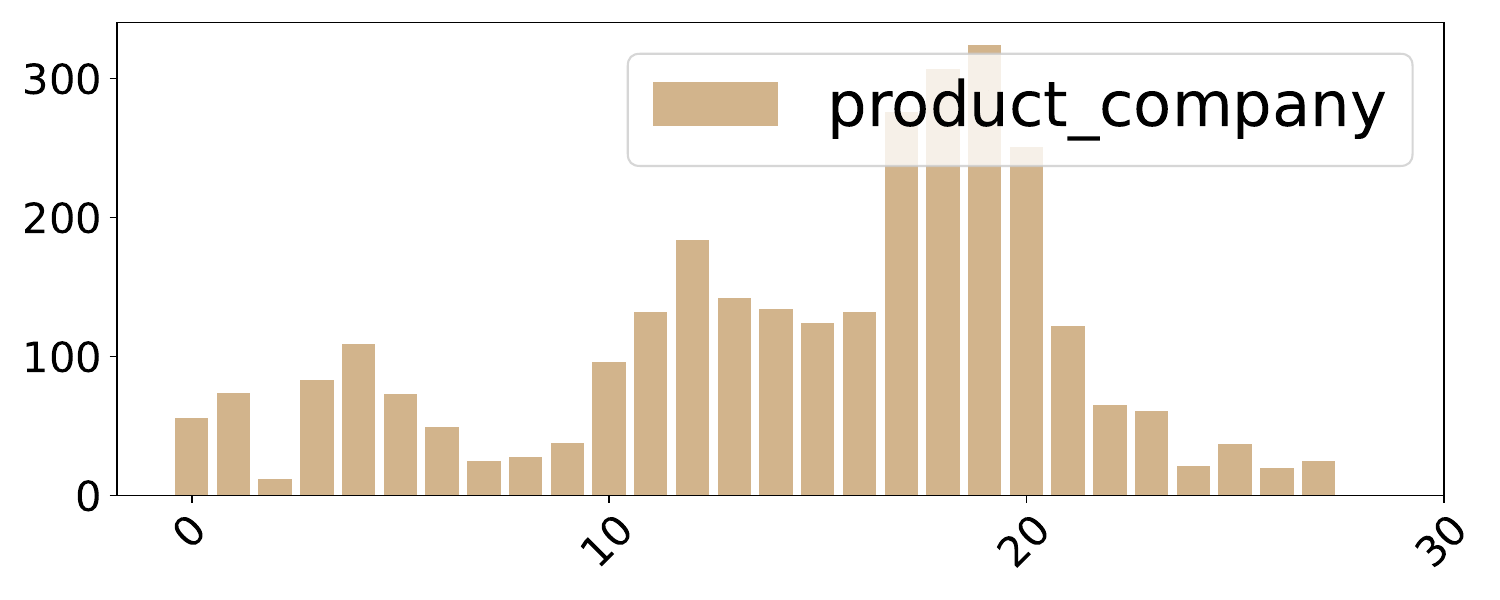}
    \includegraphics[width=0.15\textwidth]{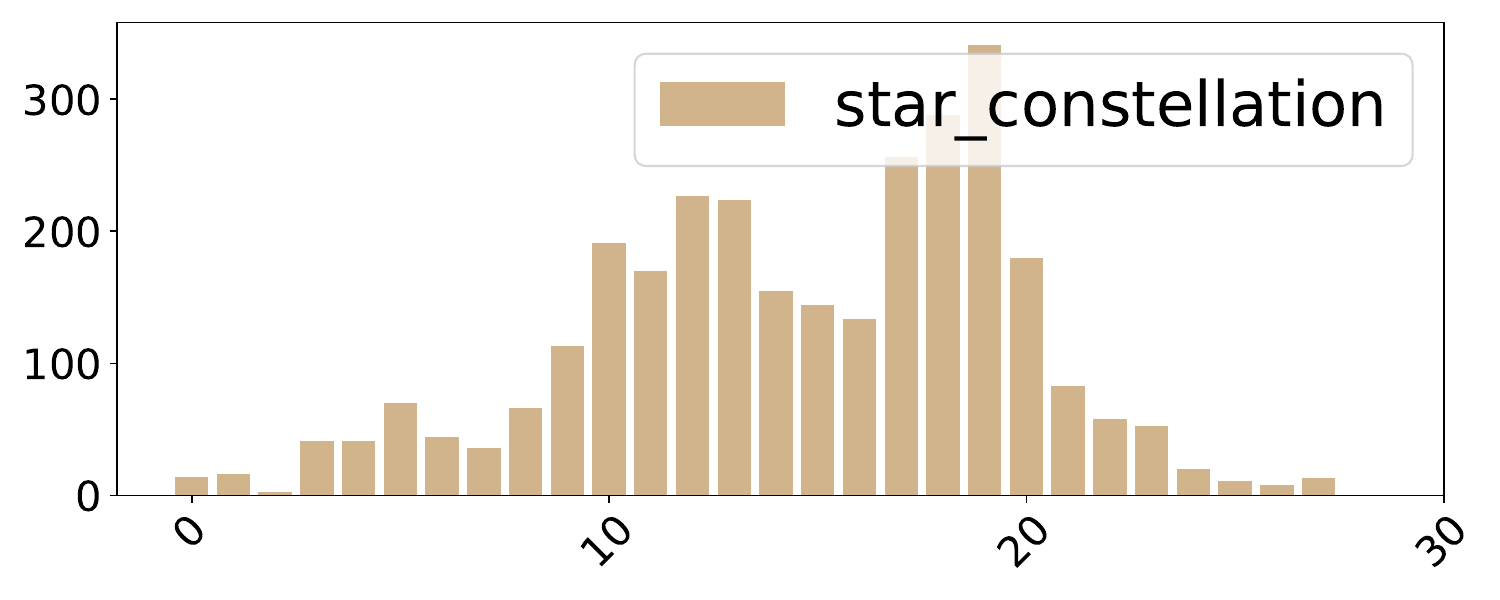}
    \caption{Distribution of \RelationSpecificNeurons
    across layers for the \textbf{Gemma-7B} model. Compared to the LLama-7B model in Figure \ref{fig:layer_dist}, identified \RelationSpecificNeurons are more evenly distributed across layers. However, the majority of the population is still located in the middle layers.}
    \label{fig:layer_dist_gemma_7b}
\end{figure}

\begin{figure*}
    \centering
    \includegraphics[width=0.23\textwidth]{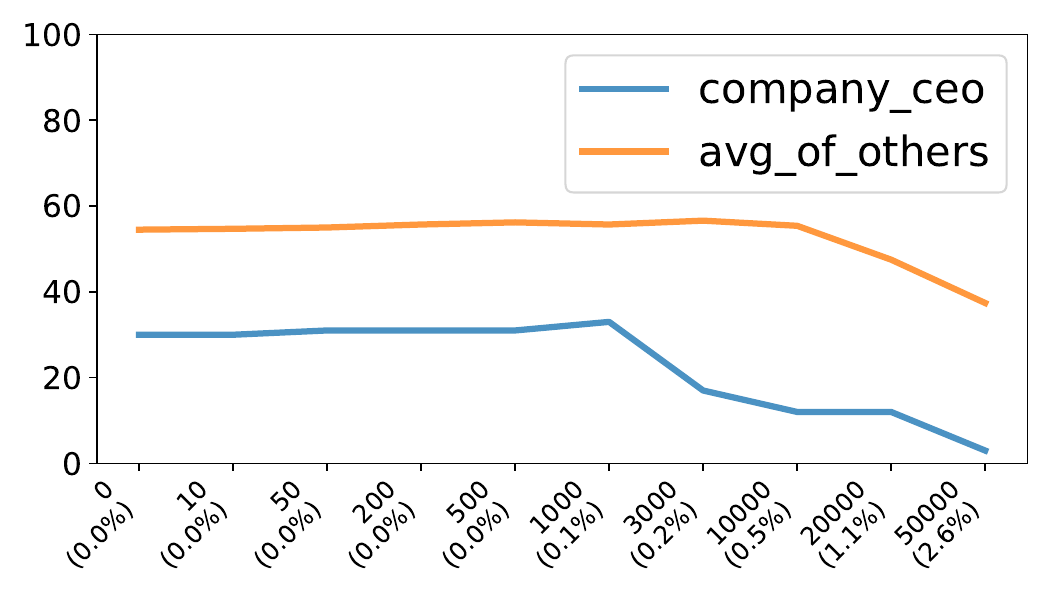}
    \includegraphics[width=0.23\textwidth]{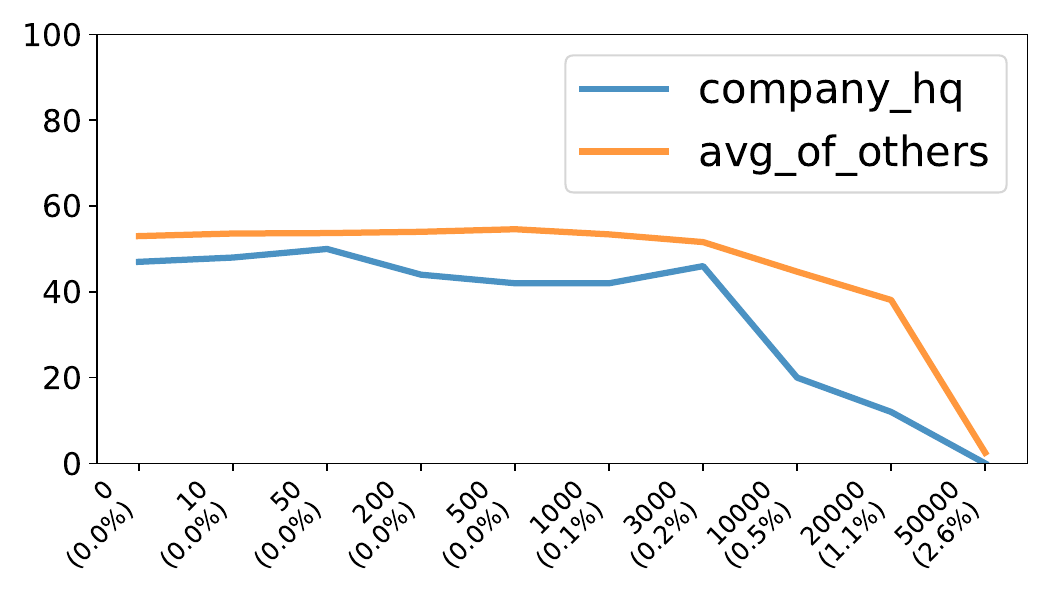}
    \includegraphics[width=0.23\textwidth]{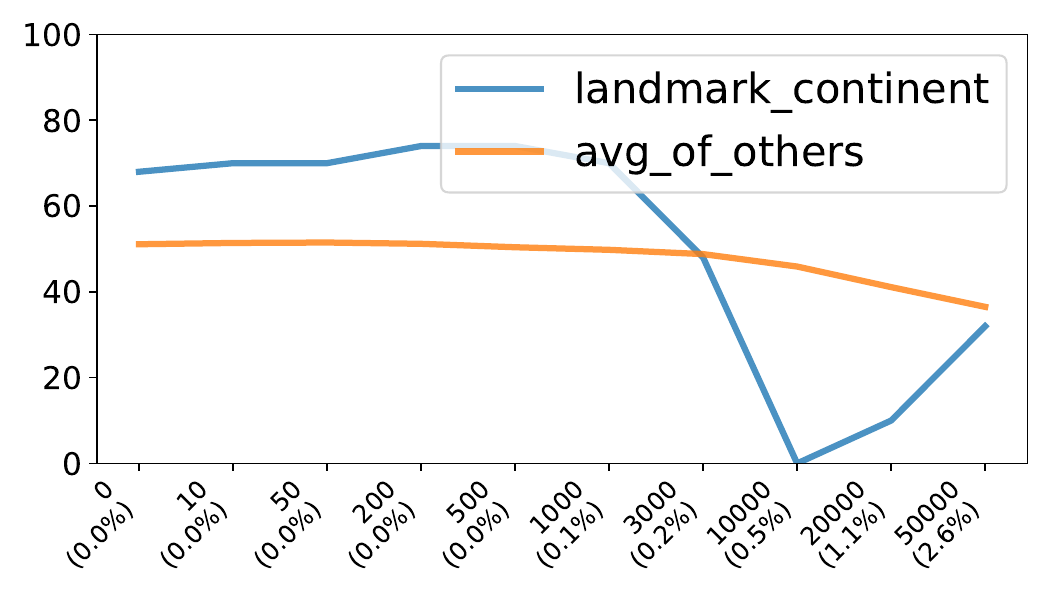}
    \includegraphics[width=0.23\textwidth]{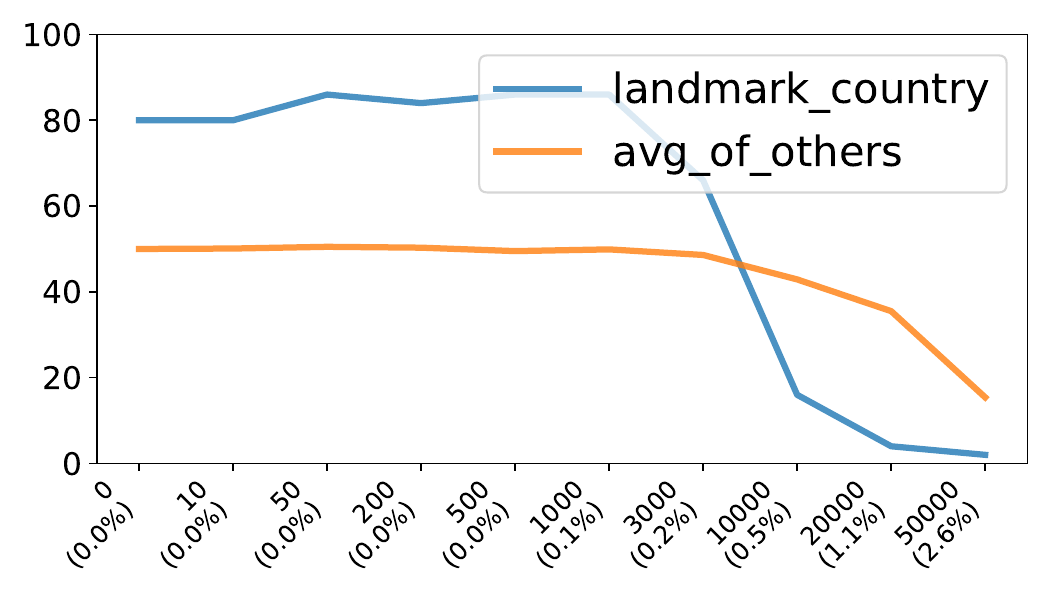}
    \includegraphics[width=0.23\textwidth]{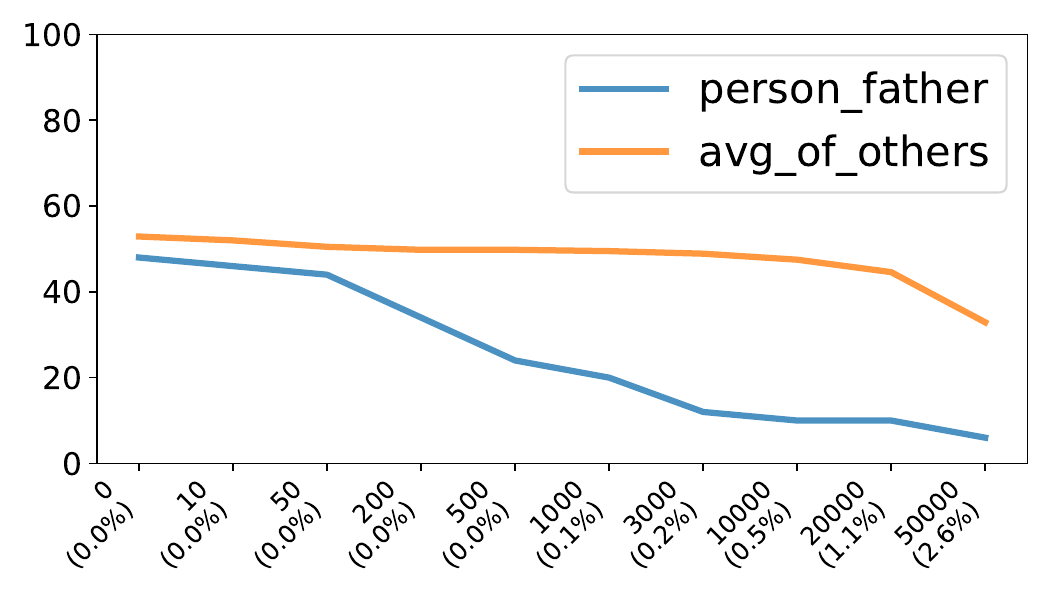}
    \includegraphics[width=0.23\textwidth]{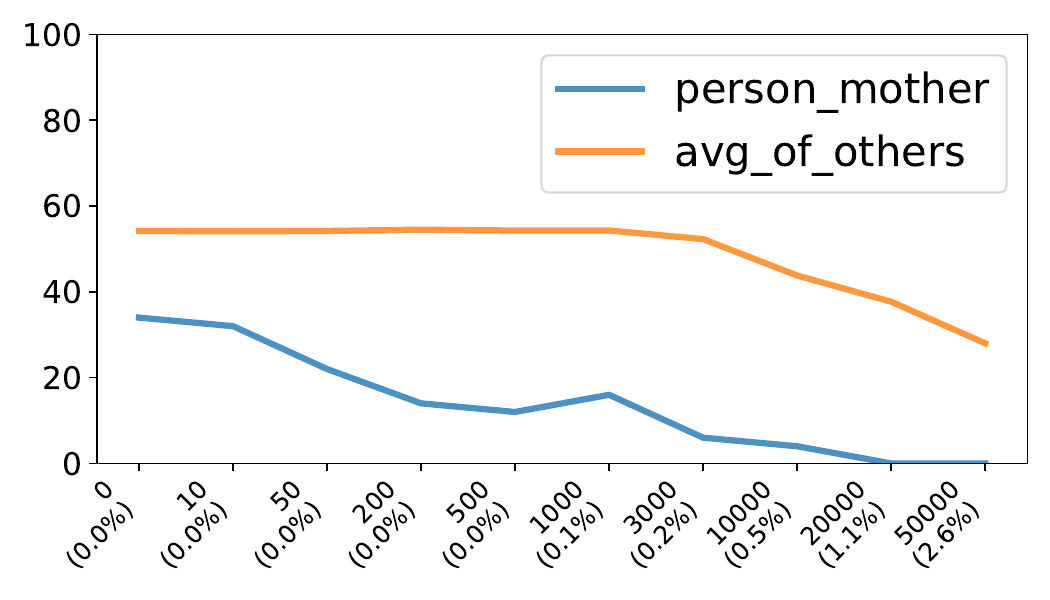}
    \includegraphics[width=0.23\textwidth]{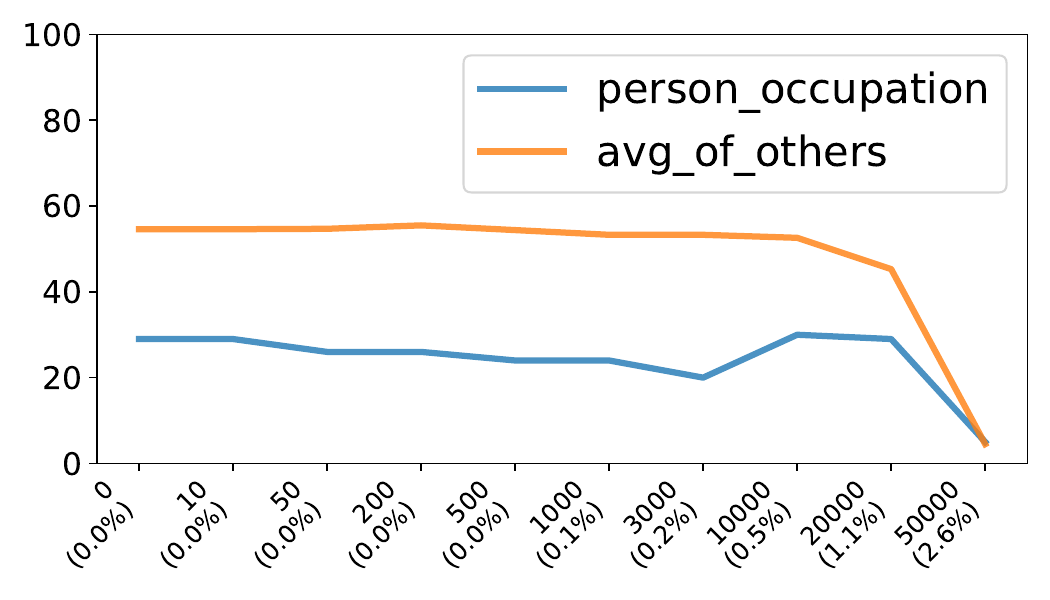}
    \includegraphics[width=0.23\textwidth]{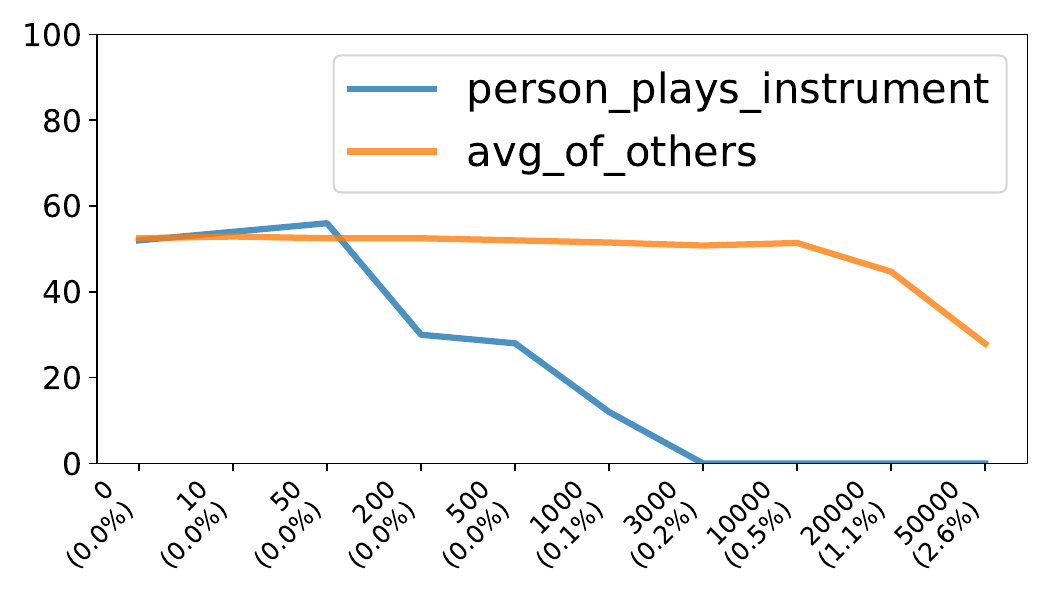}
    \includegraphics[width=0.23\textwidth]{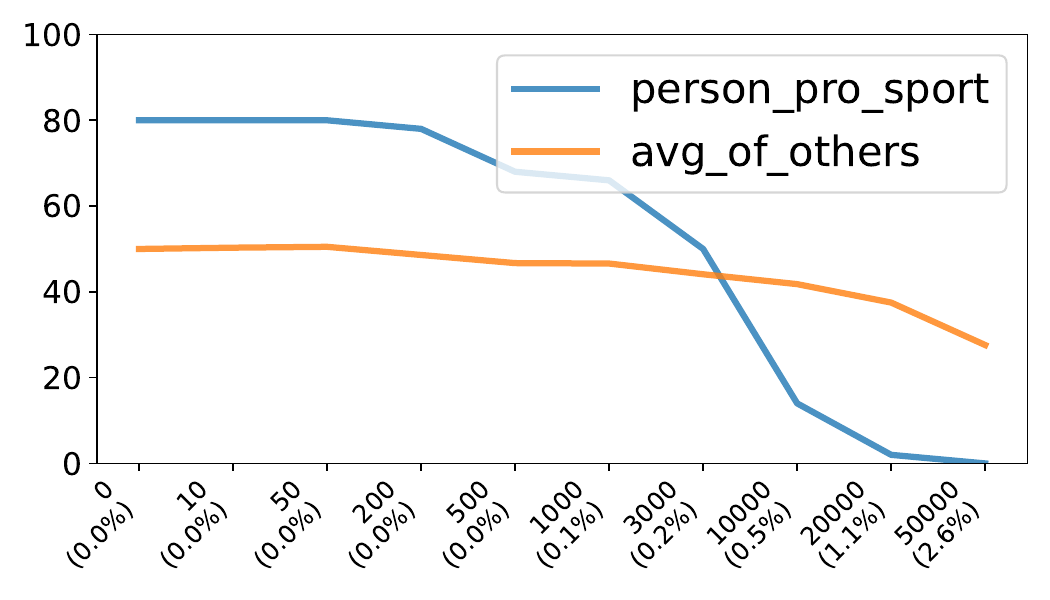}
    \includegraphics[width=0.23\textwidth]{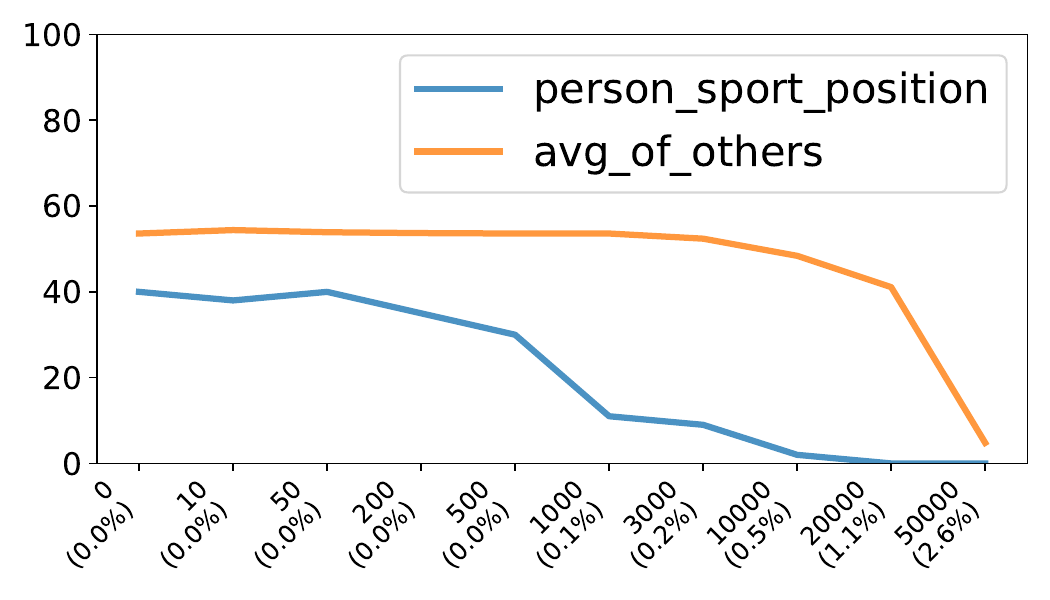}
    \includegraphics[width=0.23\textwidth]{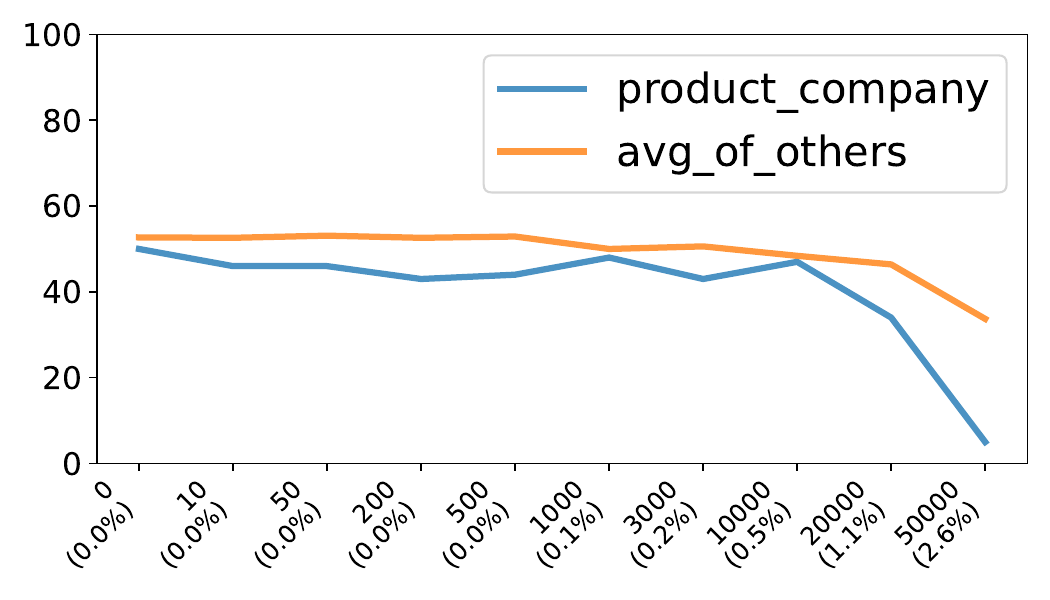}
    \includegraphics[width=0.23\textwidth]{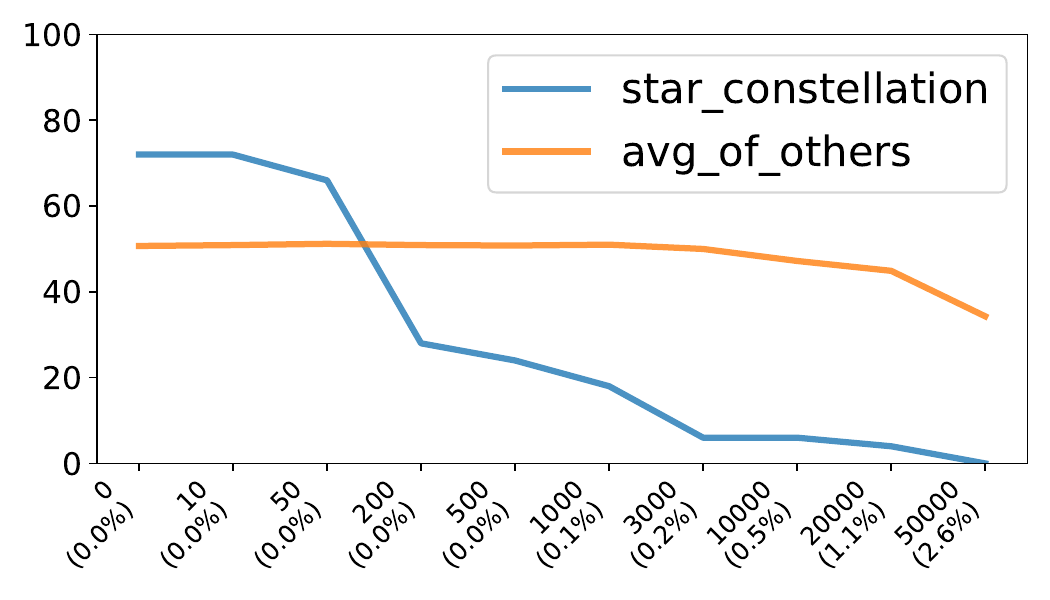}
    \caption{Influence of deactivating different numbers of \RelationSpecificNeurons for each relation (\textbf{Gemma-7B}). The variation of accuracy on the relation itself and the average accuracy on other relations is shown.}
    \label{fig:neuron_num_gemma_7b}
\end{figure*}

\begin{figure*}
    \centering
    \includegraphics[width=0.32\textwidth]{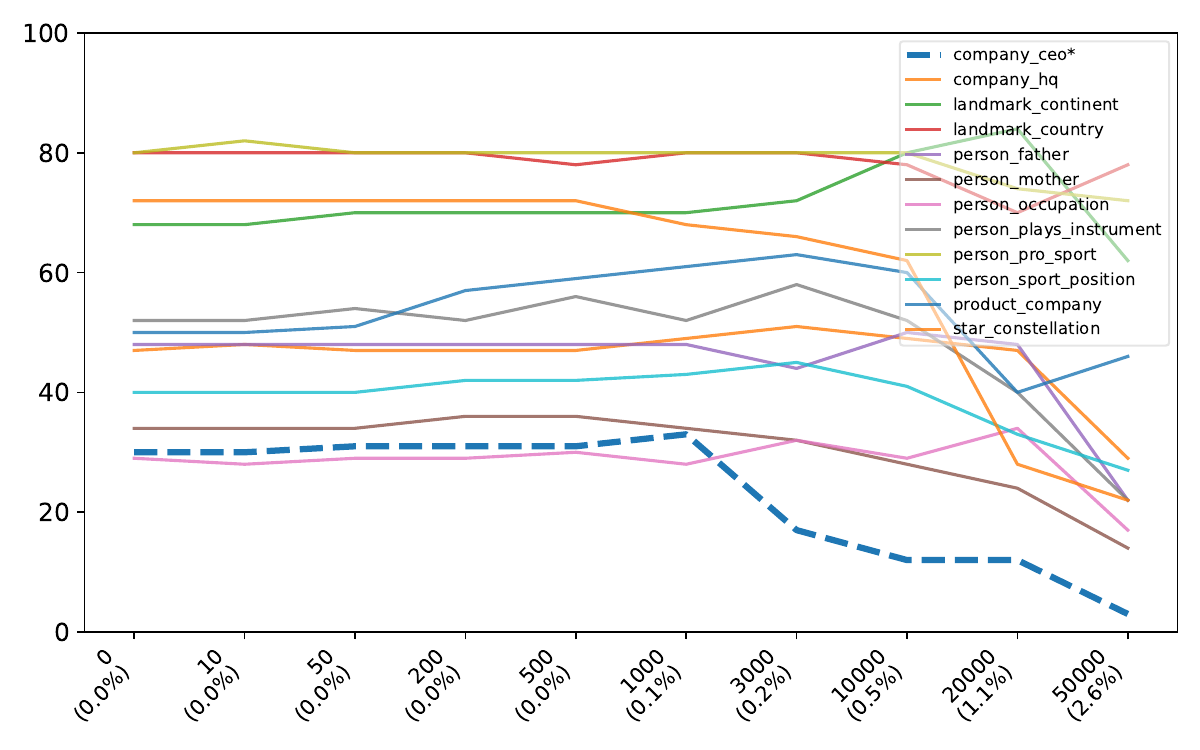}
    \includegraphics[width=0.32\textwidth]{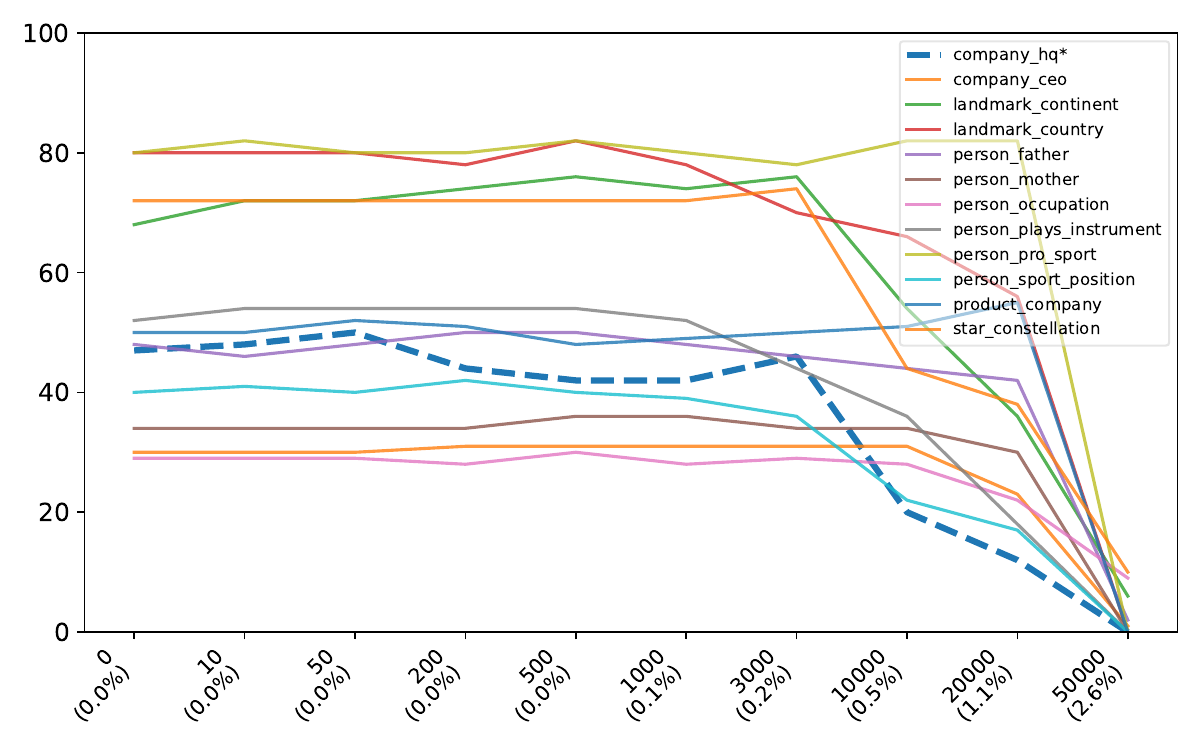}
    \includegraphics[width=0.32\textwidth]{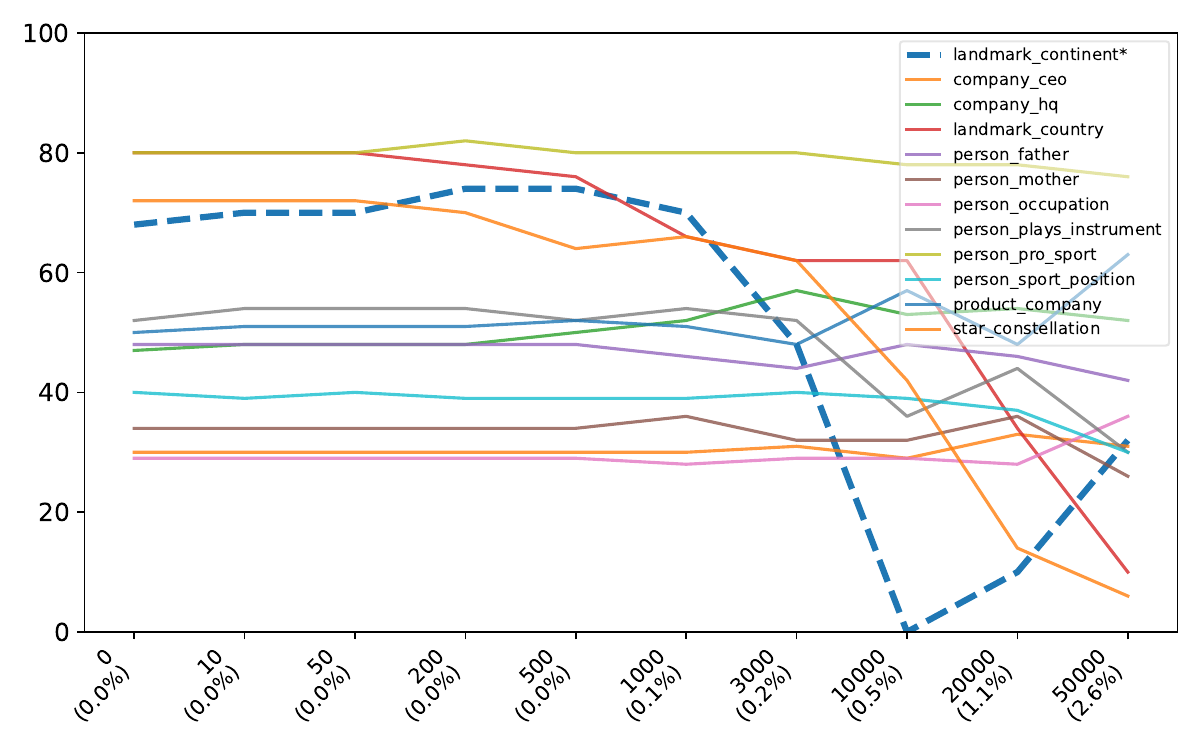}
    \includegraphics[width=0.32\textwidth]{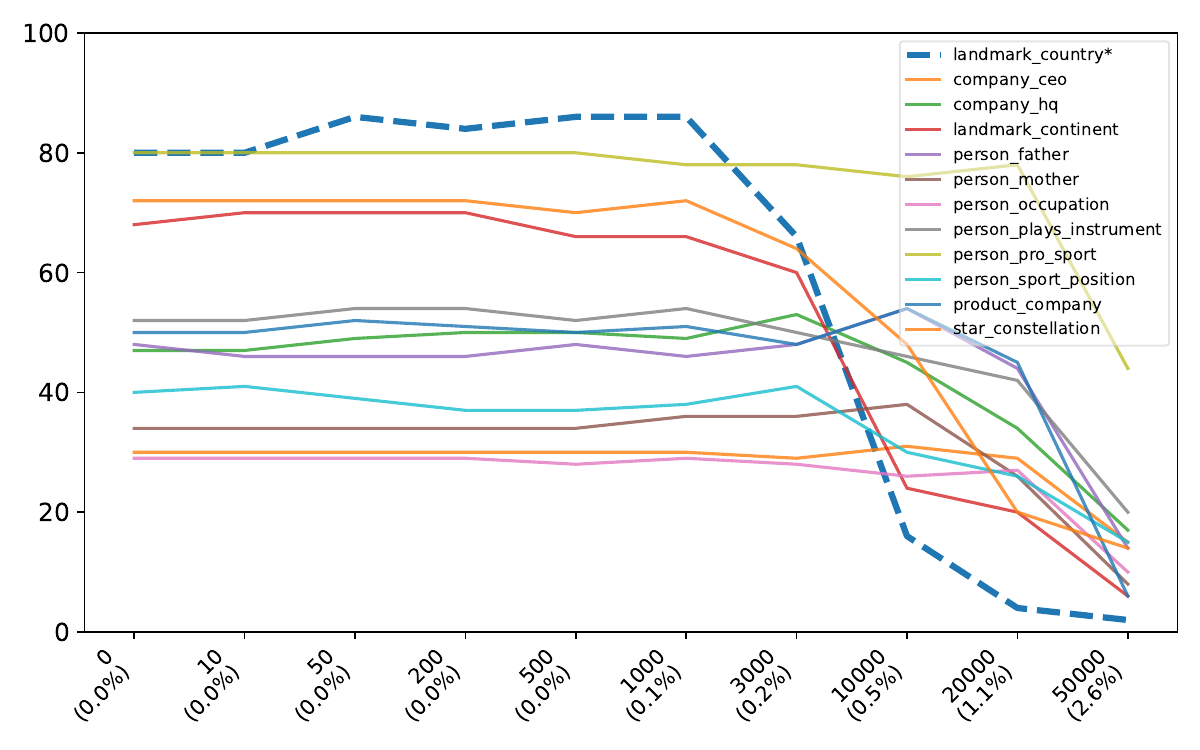}
    \includegraphics[width=0.32\textwidth]{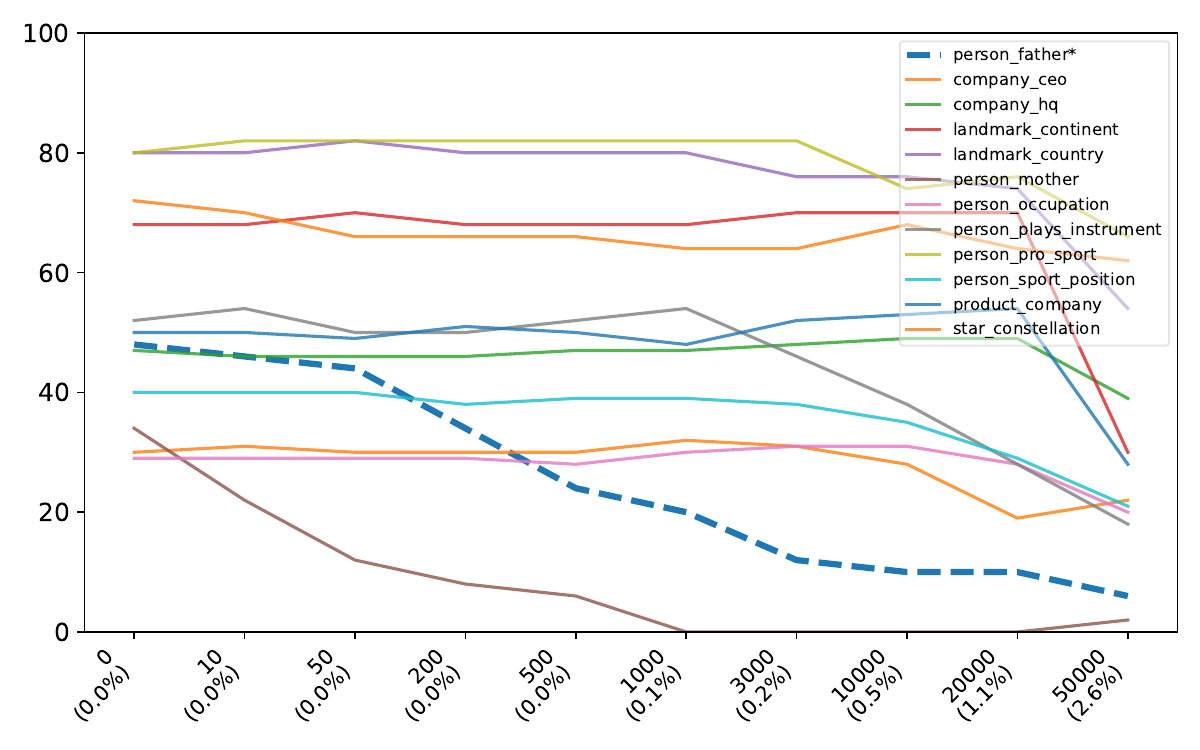}
    \includegraphics[width=0.32\textwidth]{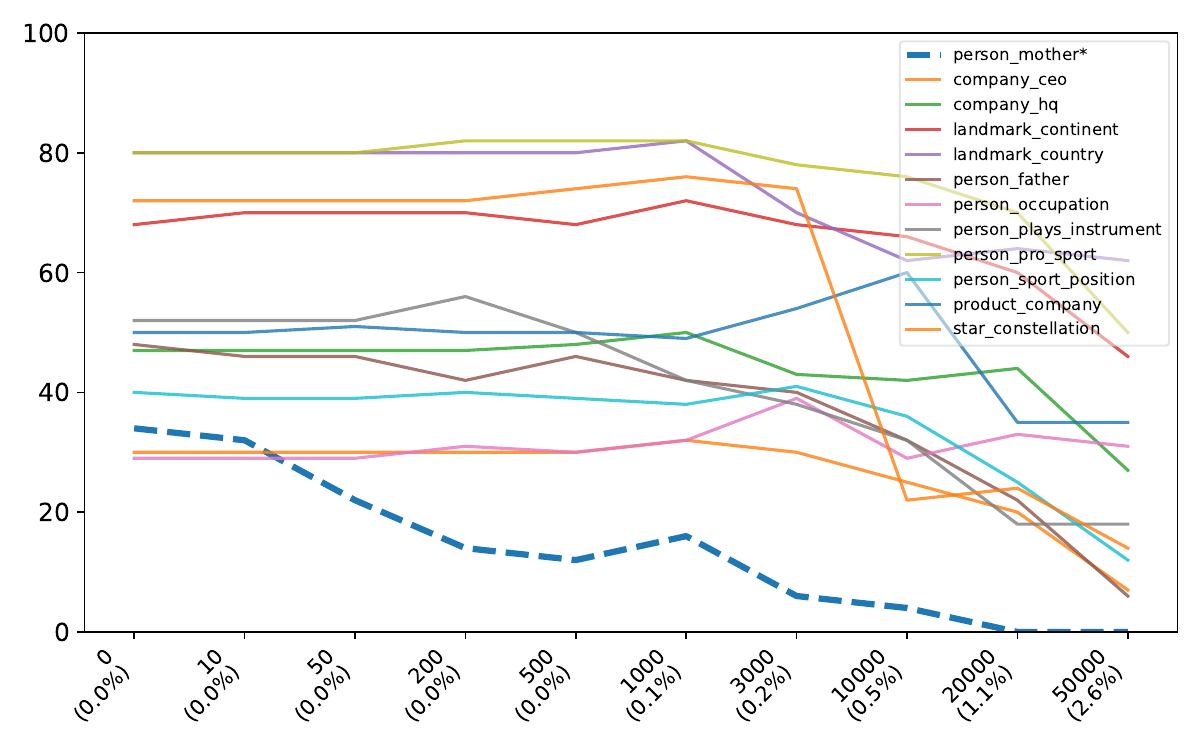}
    \includegraphics[width=0.32\textwidth]{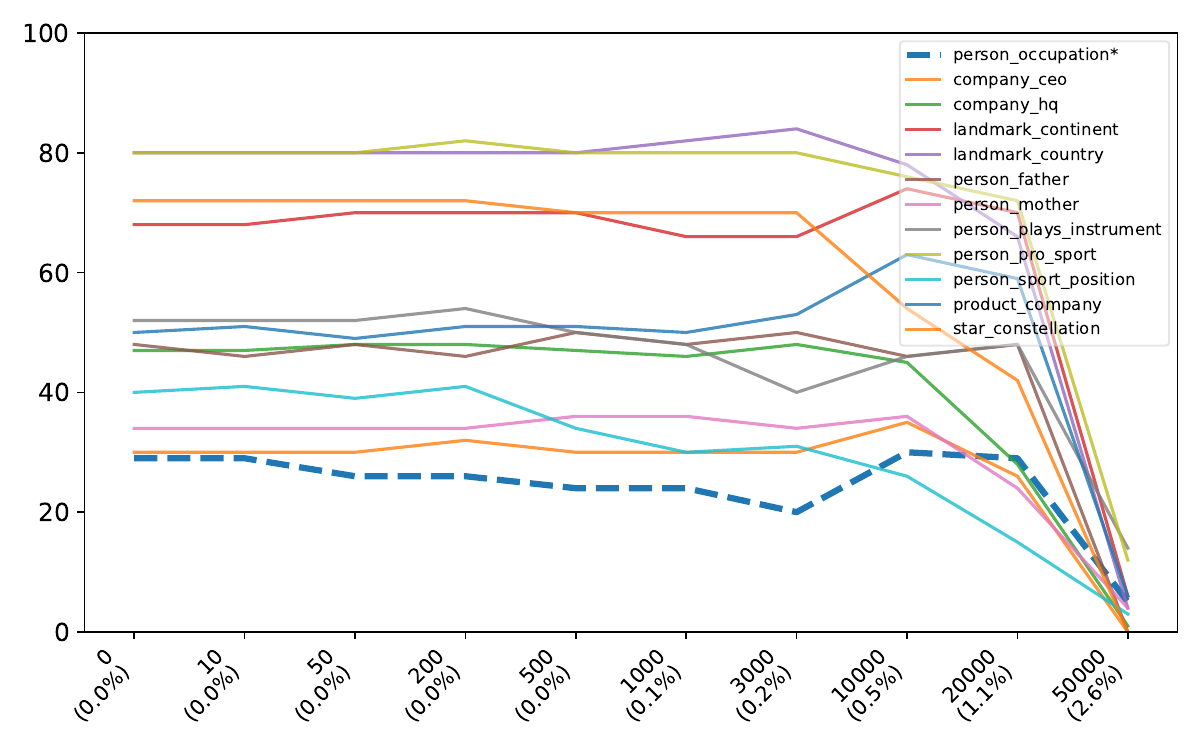}
    \includegraphics[width=0.32\textwidth]{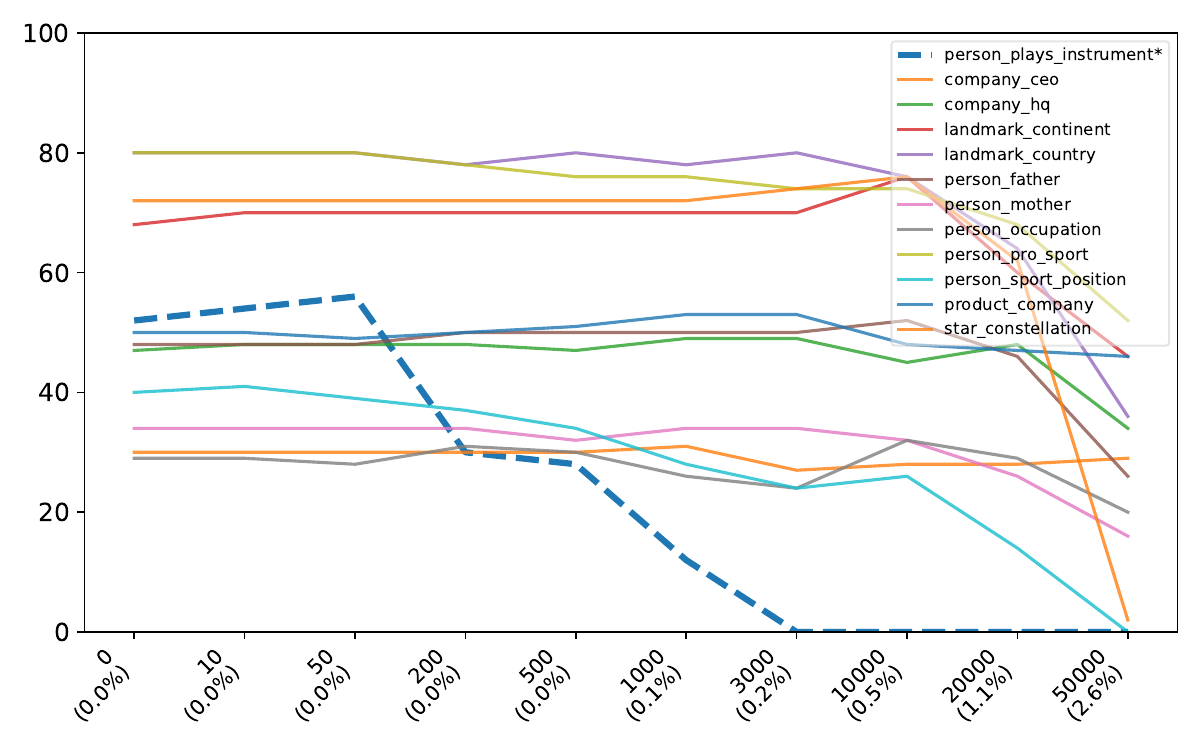}
    \includegraphics[width=0.32\textwidth]{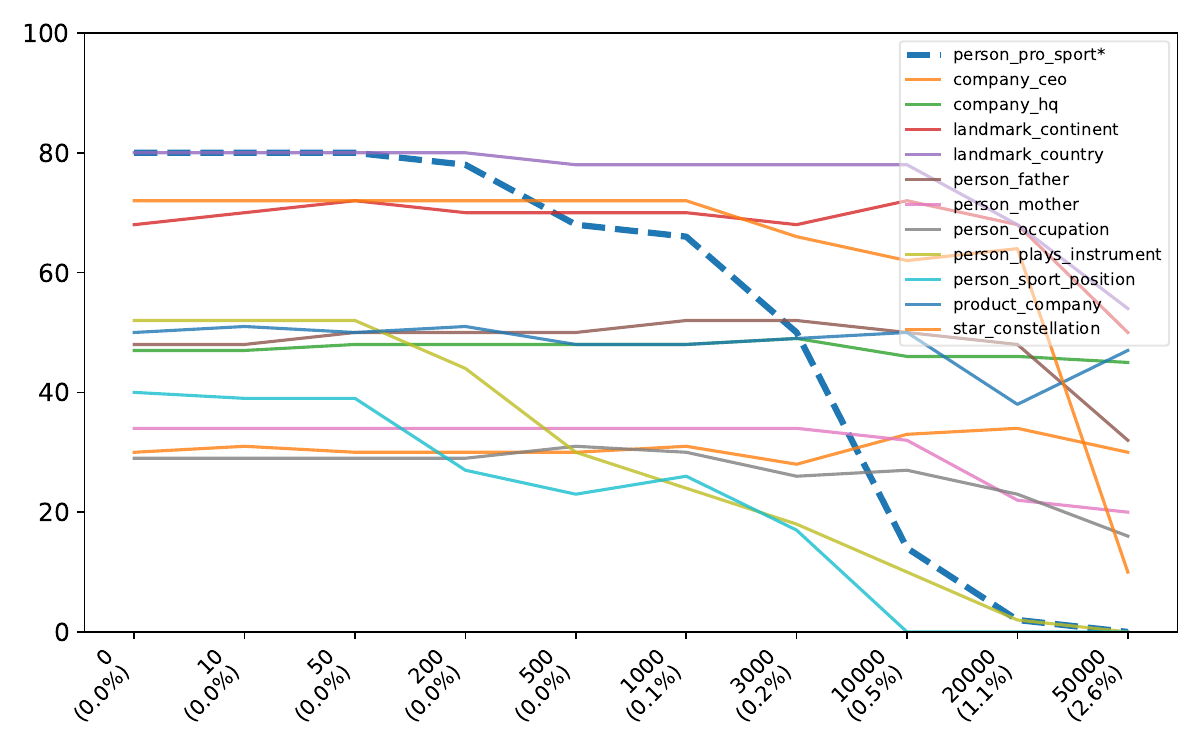}
    \includegraphics[width=0.32\textwidth]{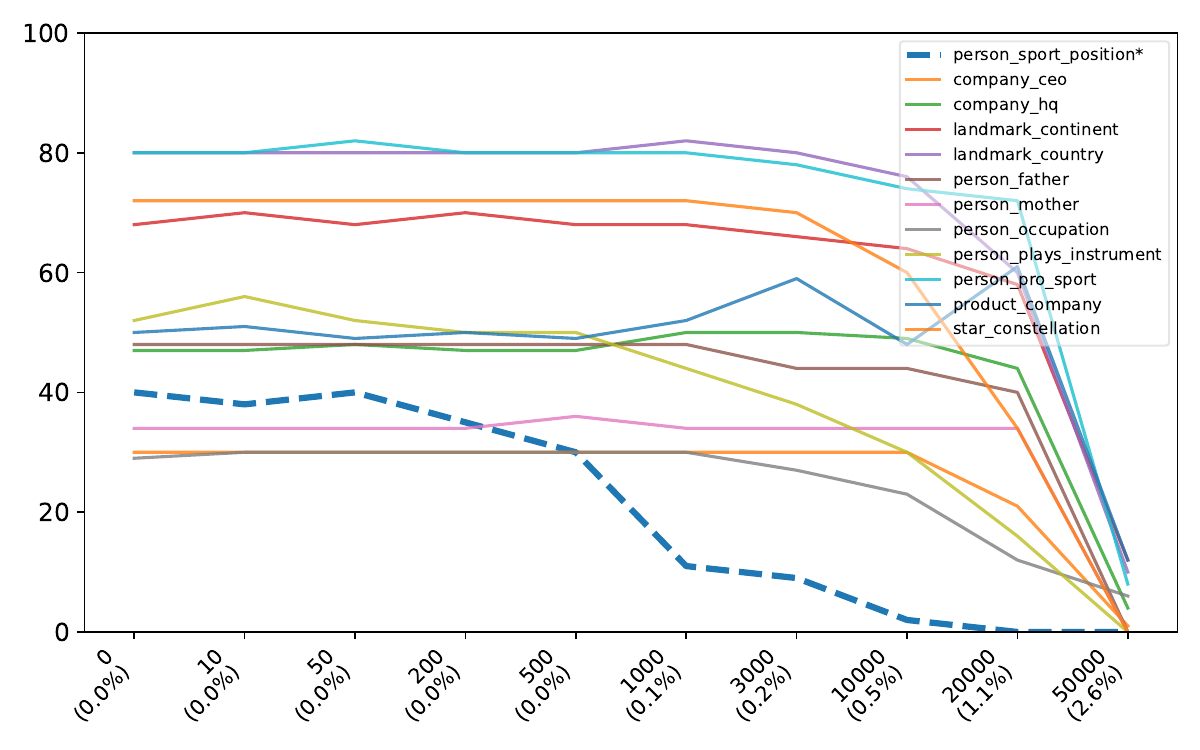}
    \includegraphics[width=0.32\textwidth]{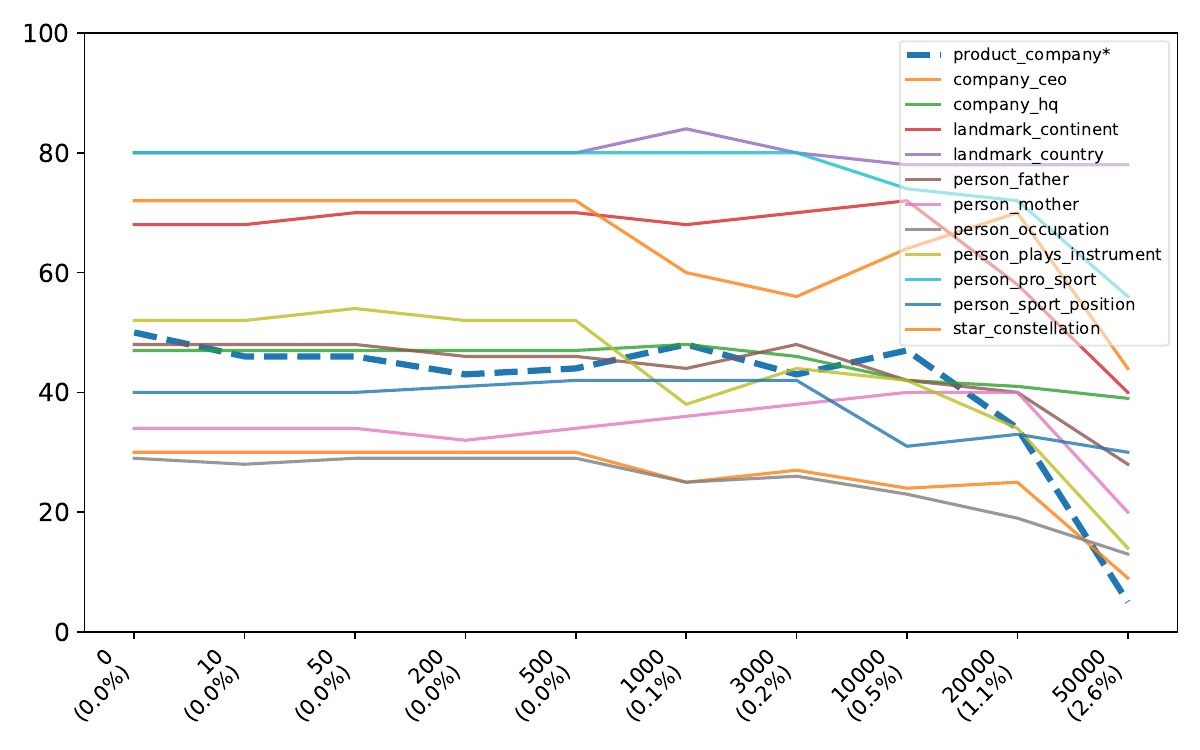}
    \includegraphics[width=0.32\textwidth]{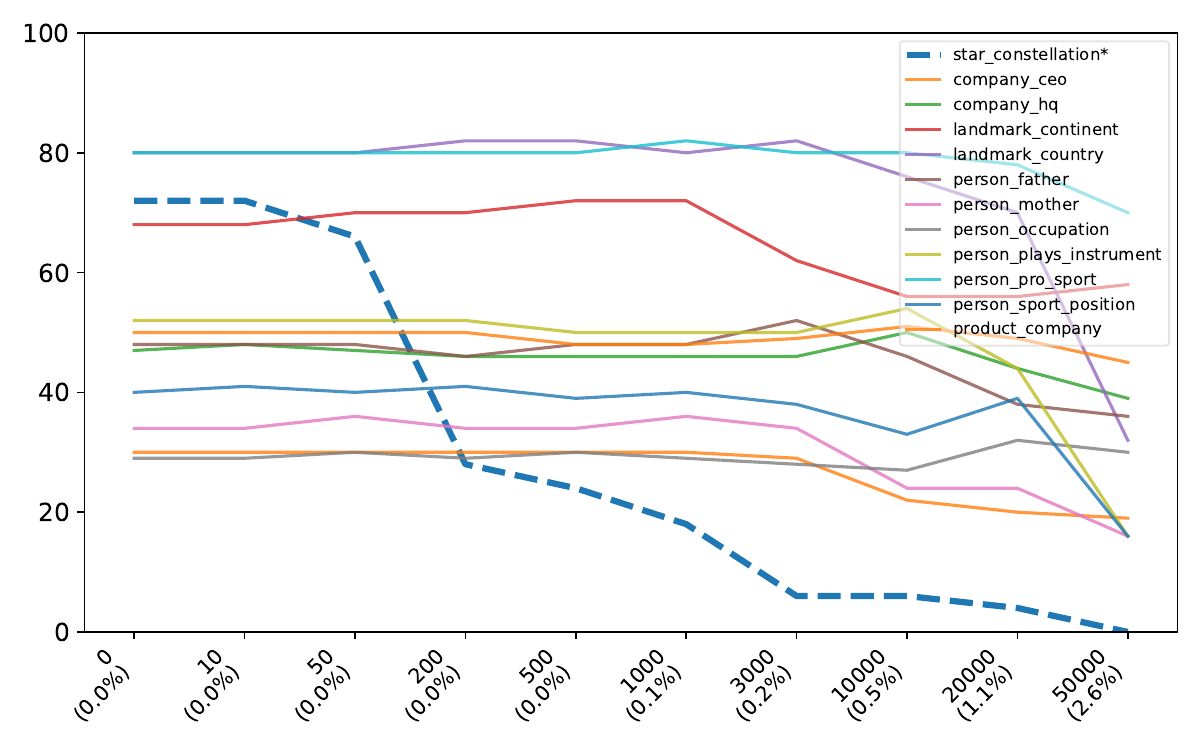}
    \caption{Influence of deactivating different numbers of
    \RelationSpecificNeurons in the \textbf{Gemma-7B} model for each
    relation. The variation of accuracy on the relation
    itself (noted with ``*'' and a dashed line style) and
    the accuracy on all other relations is shown in each
    figure.}
    \label{fig:neuron_num_all_gemma_7b}
\end{figure*}

\begin{figure*}
    \centering
    \begin{tabular}{c}
    \includegraphics[width=0.45\textwidth]{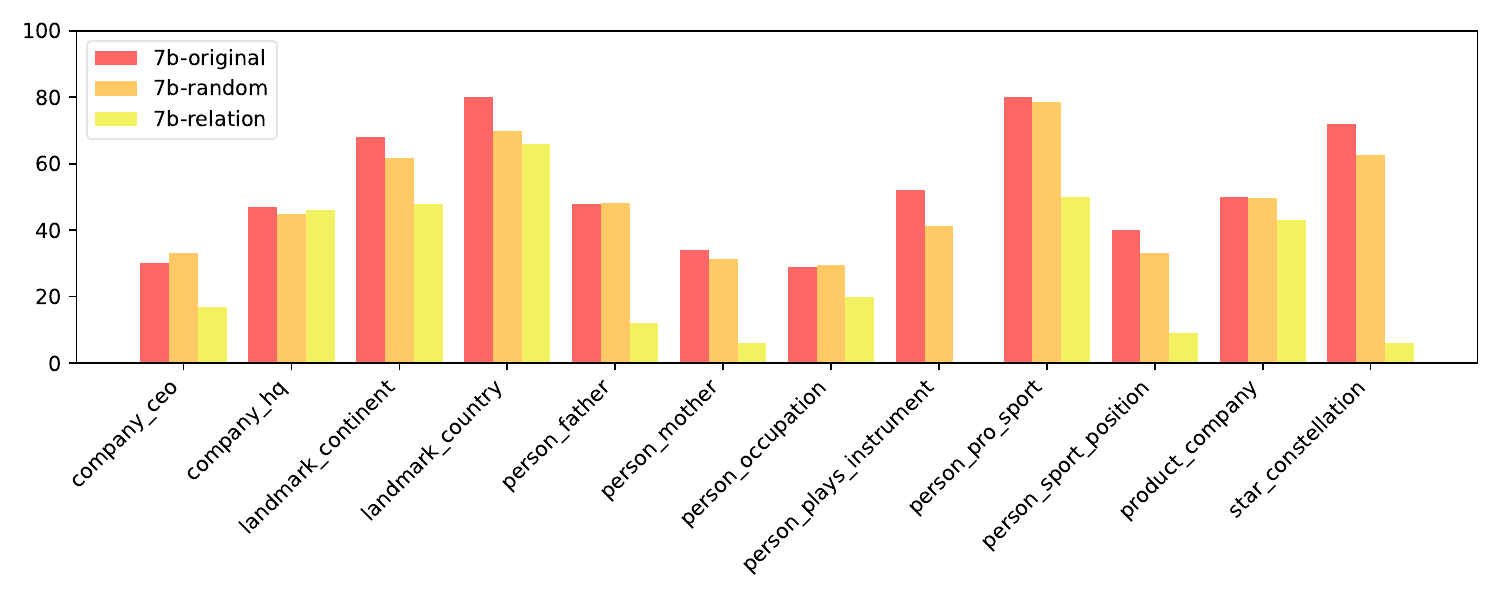}
    \includegraphics[width=0.45\textwidth]{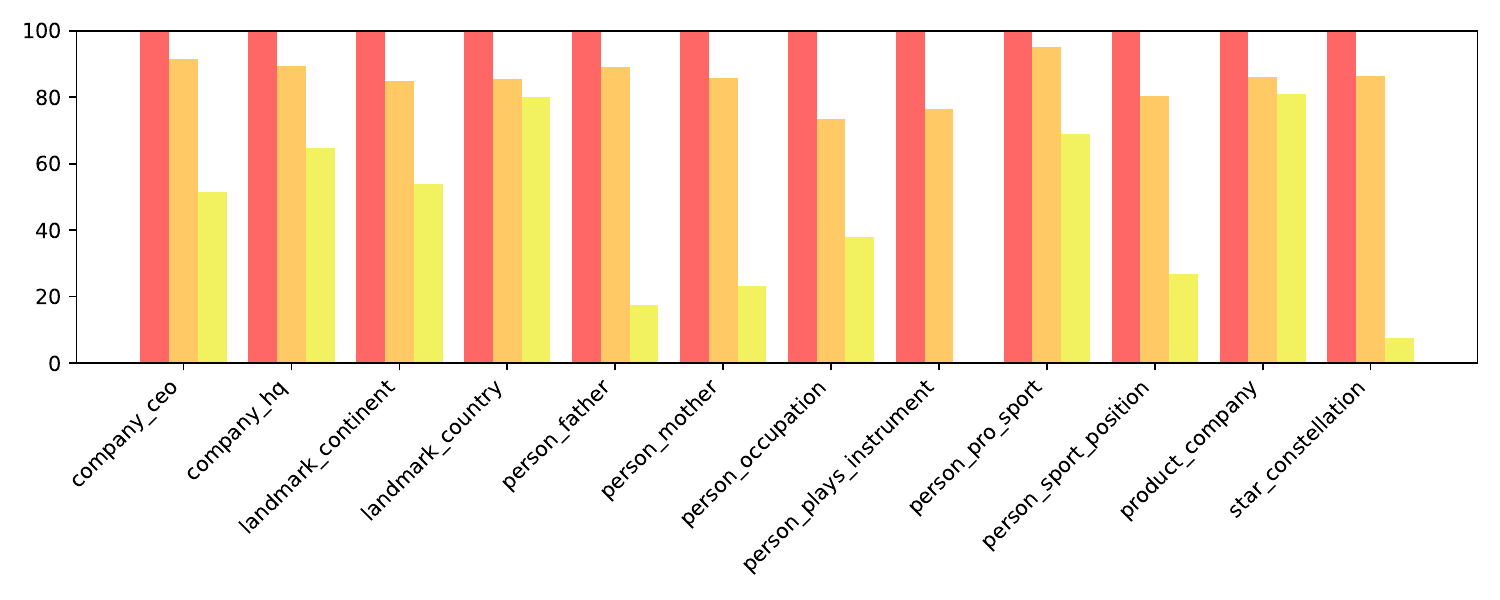}
    \end{tabular}
    \caption{Intra-relation results on \textbf{Gemma-7B}. The left (resp.\  right)
    figure displays the results of held-out evaluation prompt
    set $\mathcal{P}_{r_i}^{\text{eva}}$
    (resp.\  identification prompt set
    $\mathcal{P}_{r_i}^{\text{det}}$). We report the
    performance of the original model (without any
    deactivation), the model
    with 3,000 random neurons deactivated, and the model with relation neurons deactivated.}
    \label{fig:intra-relation_gemma-7b}
\end{figure*}

\section{Analysis On Gemma-7B}\seclabel{gemma}

We perform a similar analysis on the Gemma-7B model \citep{gemma2024team} as we do for the LLama-7B model.  
We first show how the identified 3,000 \RelationSpecificNeurons are distributed across layers for each relation in Figure \ref{fig:layer_dist_gemma_7b}. 
The trend is similar to what we observe in the 7B model (cf. Figure \ref{fig:layer_dist}): the most of these neurons are located in the middle layers, but it is more evenly distributed across layers compared to the LLama families.

We show the intra-relation results in Figure \ref{fig:intra-relation_gemma-7b}. The results indicate that the identified \RelationSpecificNeurons are also effective in the Gemma-7B model: not only for the identification prompt set $\mathcal{P}_{r_i}^{\text{det}}$ but also for the held-out evaluation prompt set $\mathcal{P}_{r_i}^{\text{eva}}$, the deactivation of the neurons result in obvious accuracy drops, especially compared with the randomly deactivated neurons, indicating the existence of \RelationSpecificNeurons are held across model families.

We then demonstrate the effect of varying numbers of \RelationSpecificNeurons using the same numbers: 10, 50, 200, 500, 1,000, 3,000, 10,000, 20,000, and 50,000. Figure \ref{fig:neuron_num_gemma_7b} and \ref{fig:neuron_num_all_gemma_7b} present the results. The global trend is similar to what we observe for the LLama-7B model: the accuracy for a relation further drops when more of its \RelationSpecificNeurons are deactivated; until 3,000 or 10,0000 neurons, the effect is almost only obvious for the concerned relation itself; after 10,000, deactivating more neurons results in a further drop in accuracy across all relations. This indicates the \textbf{neuron cumulativity} and \textbf{neuron versatility} can be observed across model families.
\newpage

\section{Analysis On the 13B Model}\seclabel{13b_analysis}

\begin{figure}
    \centering
    \setlength{\belowcaptionskip}{-0.5cm}
    \includegraphics[width=0.15\textwidth]{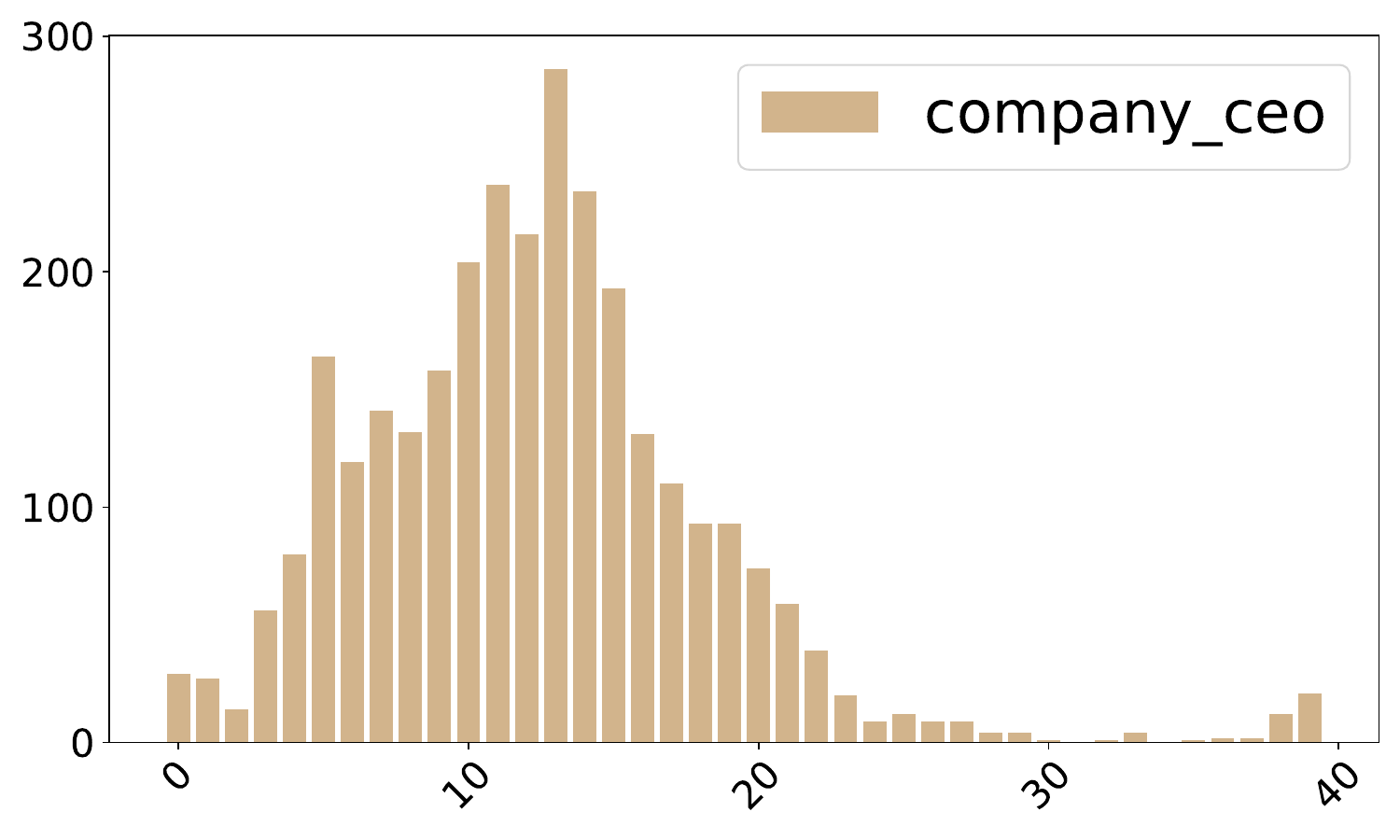}
    \includegraphics[width=0.15\textwidth]{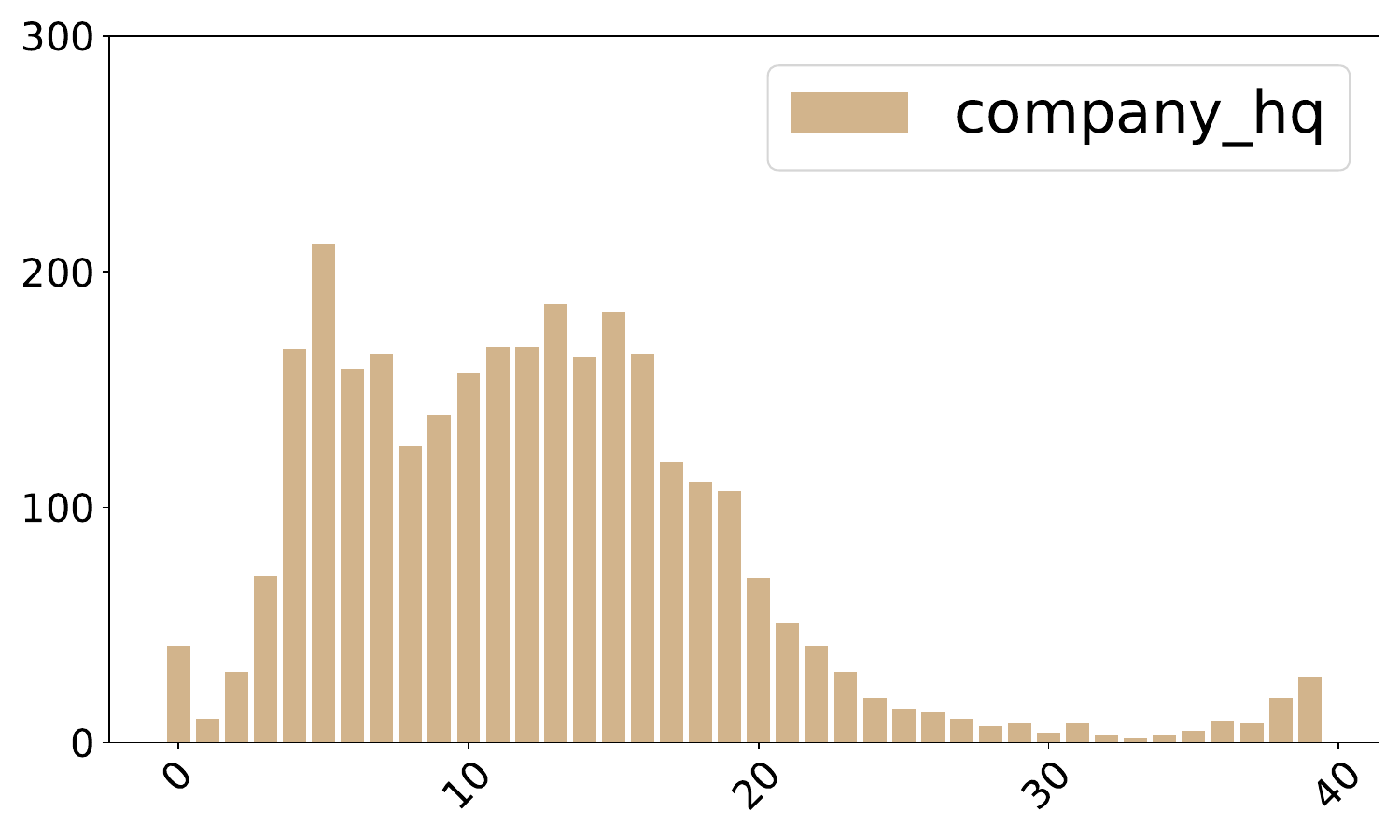}
    \includegraphics[width=0.15\textwidth]{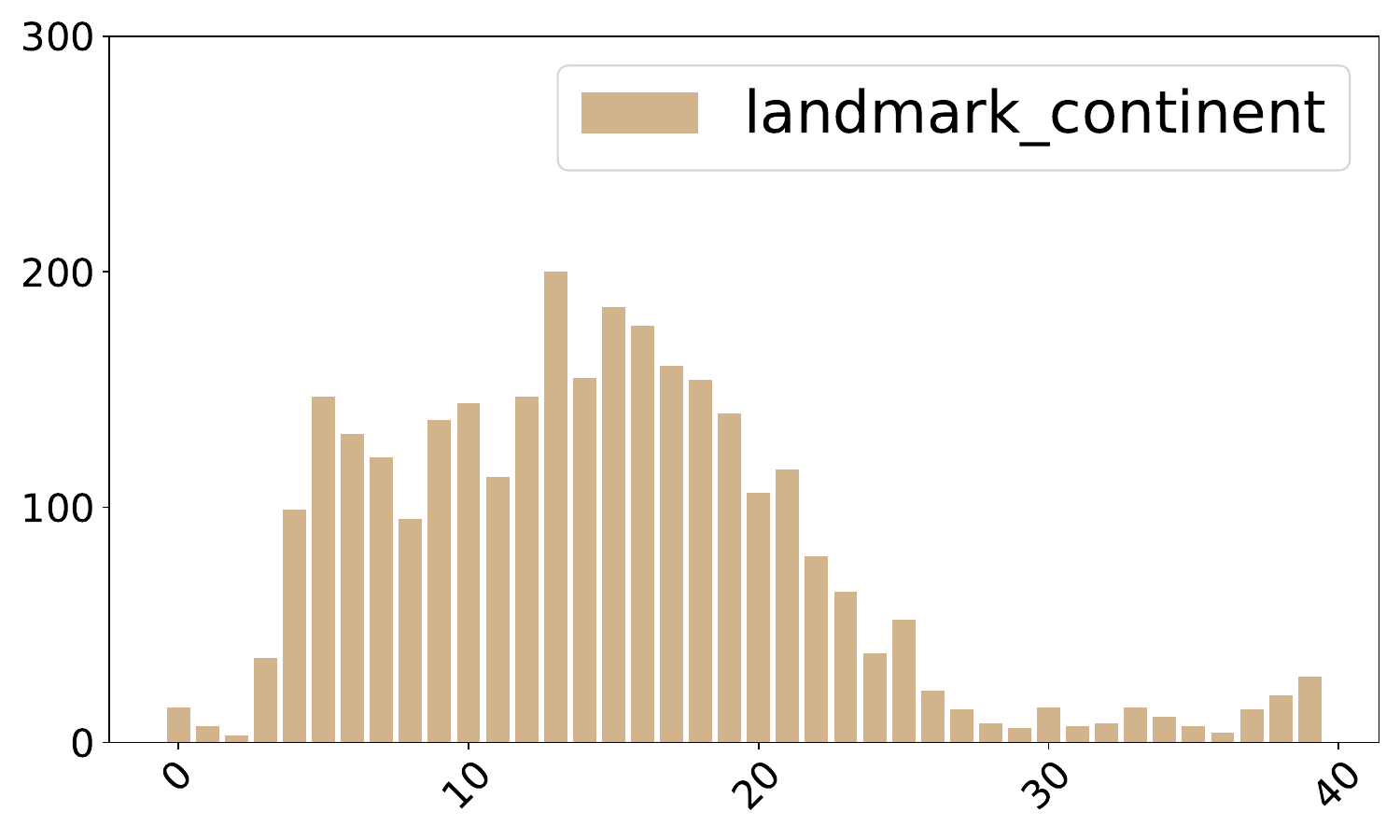}
    \includegraphics[width=0.15\textwidth]{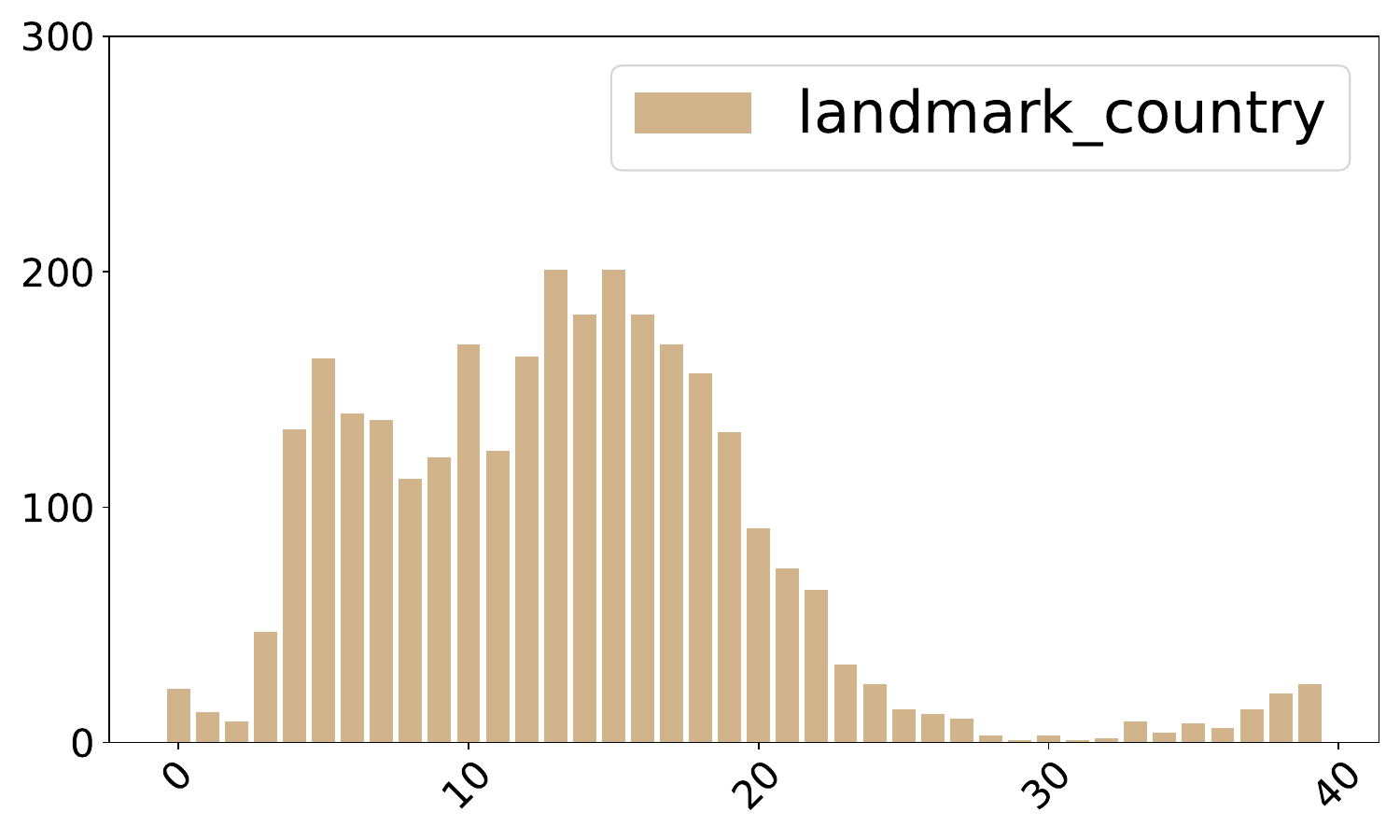}
    \includegraphics[width=0.15\textwidth]{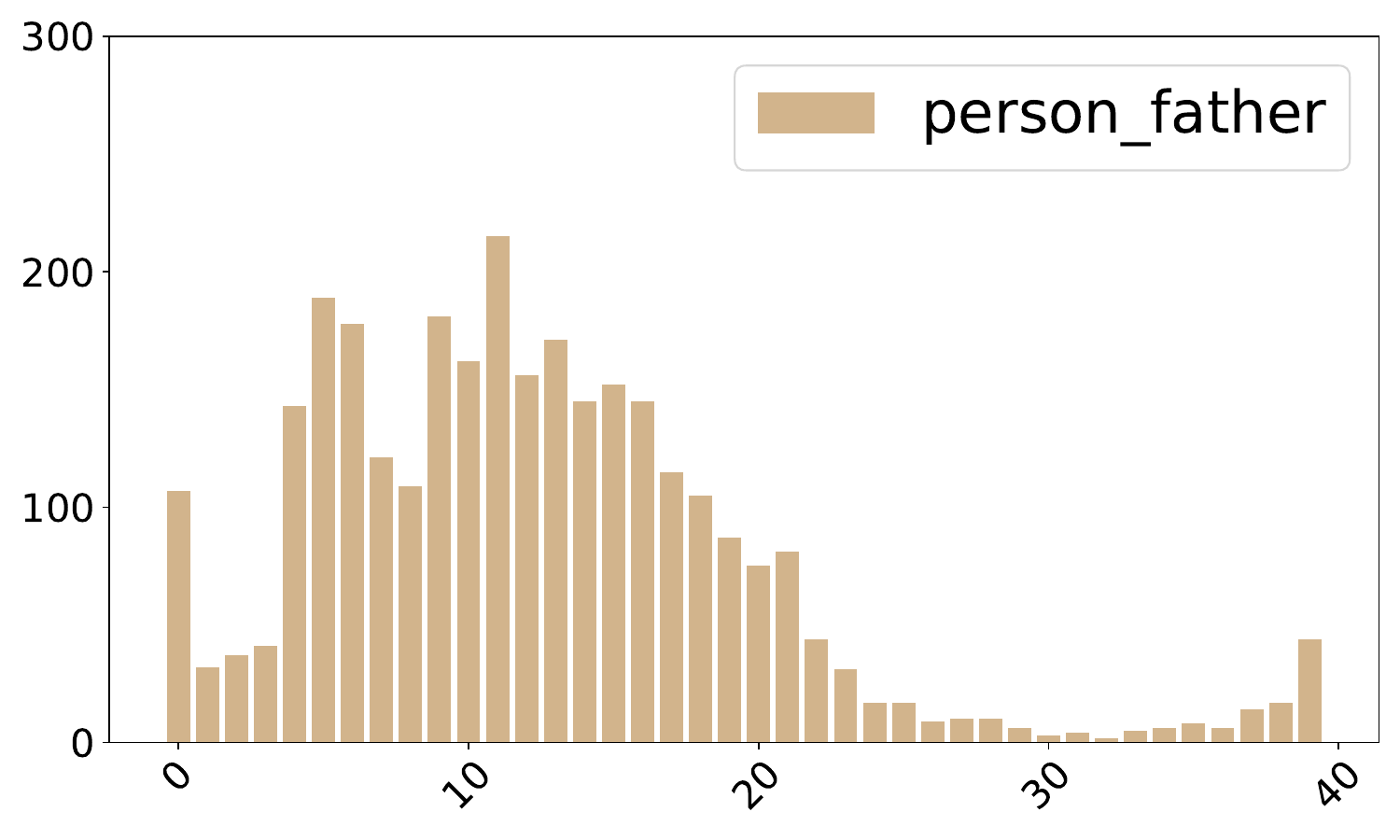}
    \includegraphics[width=0.15\textwidth]{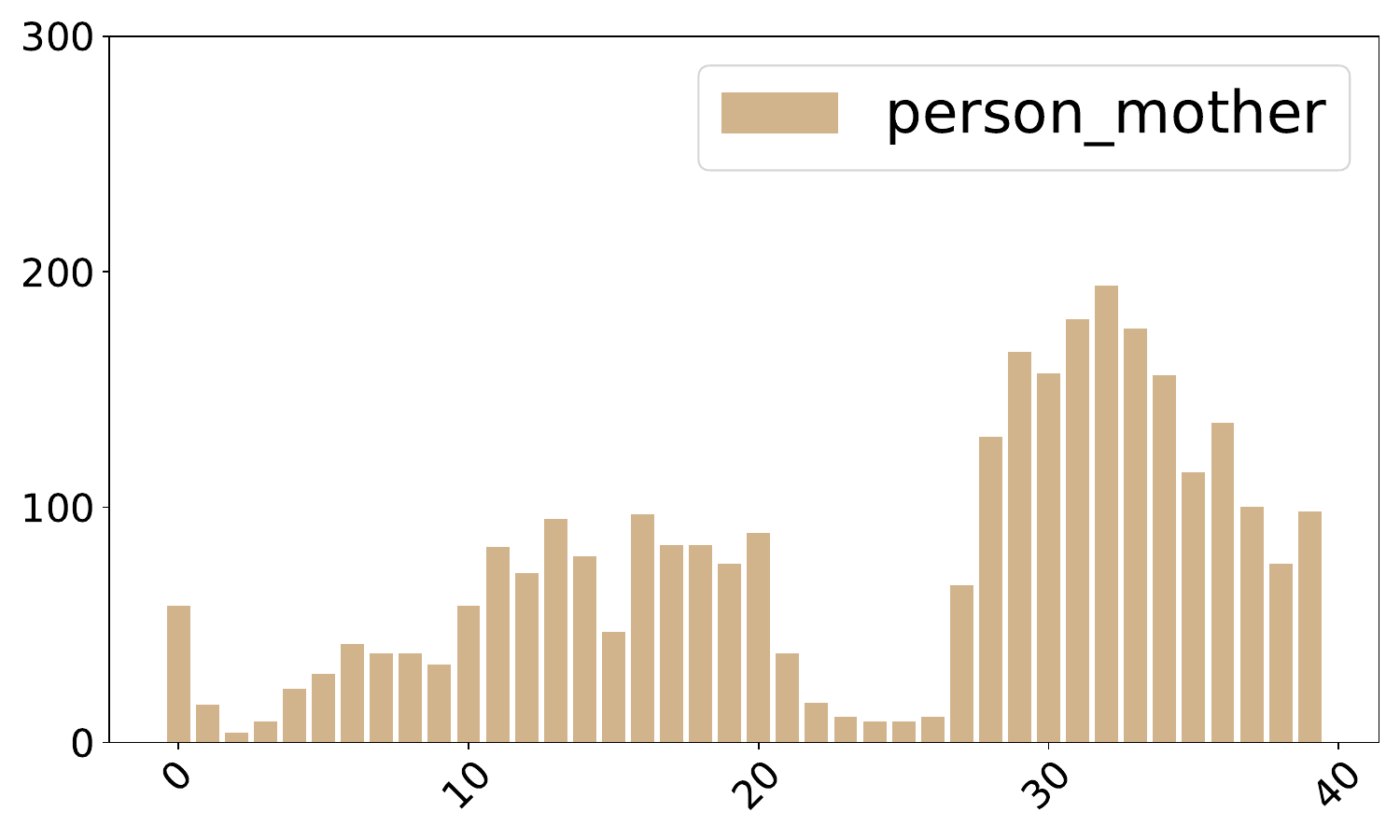}
    \includegraphics[width=0.15\textwidth]{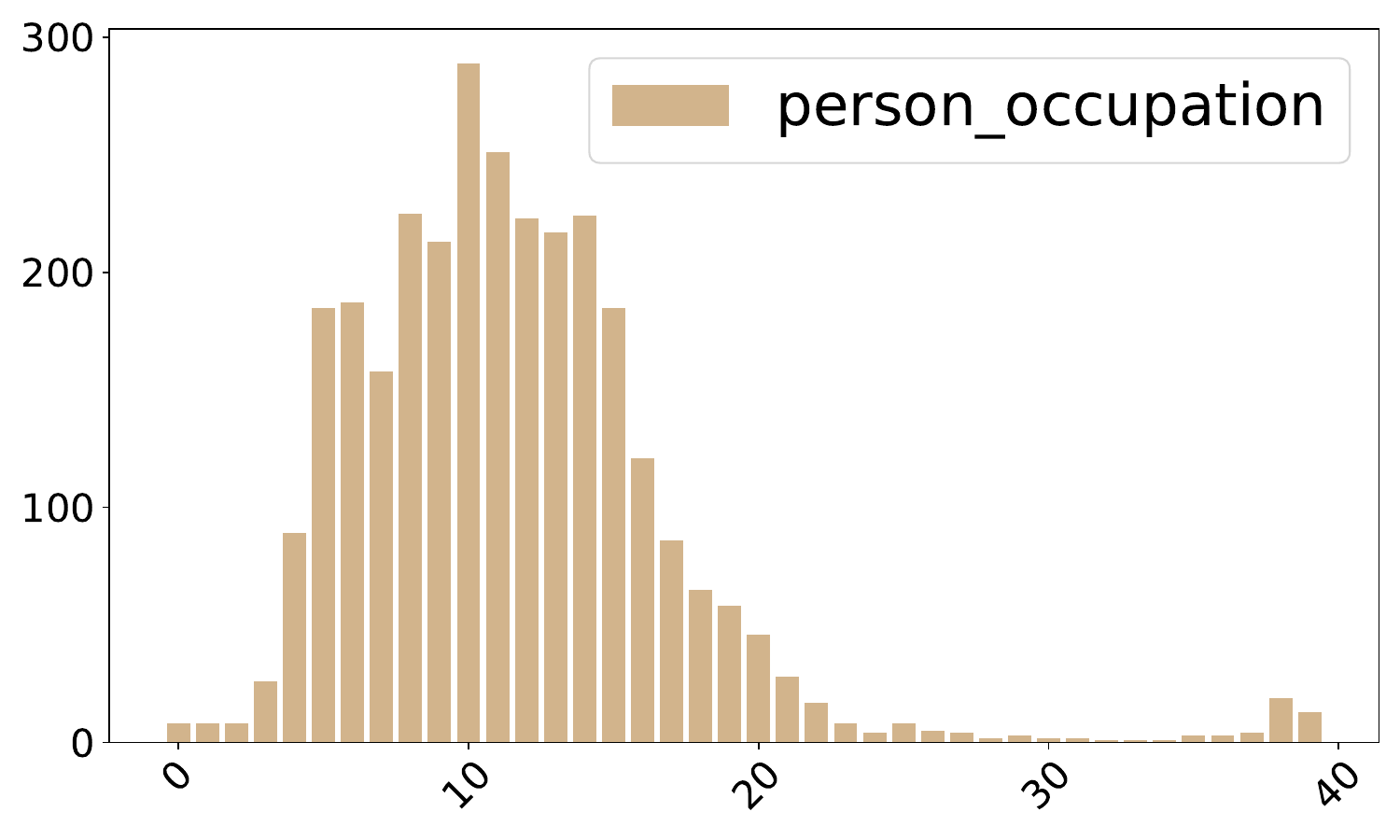}
    \includegraphics[width=0.15\textwidth]{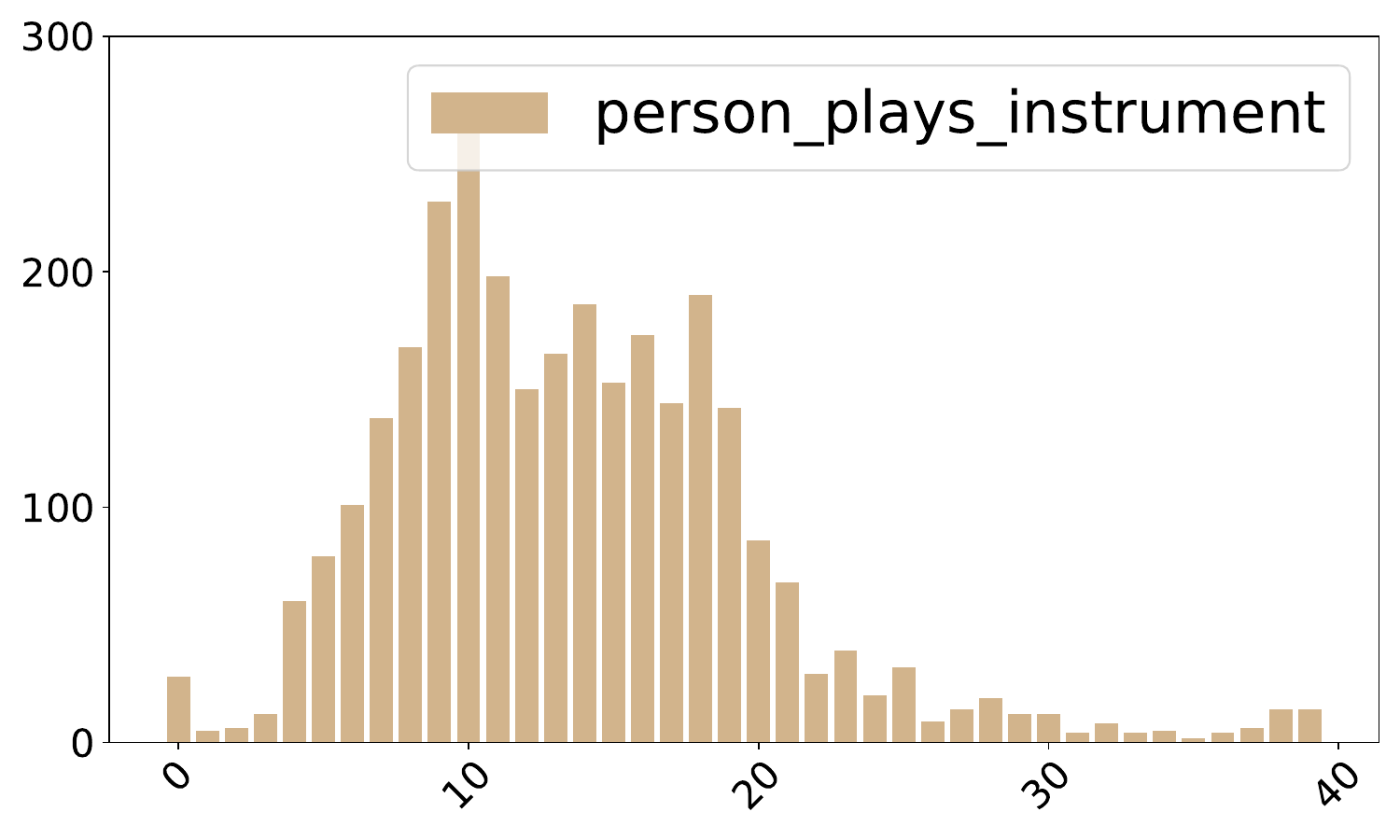}
    \includegraphics[width=0.15\textwidth]{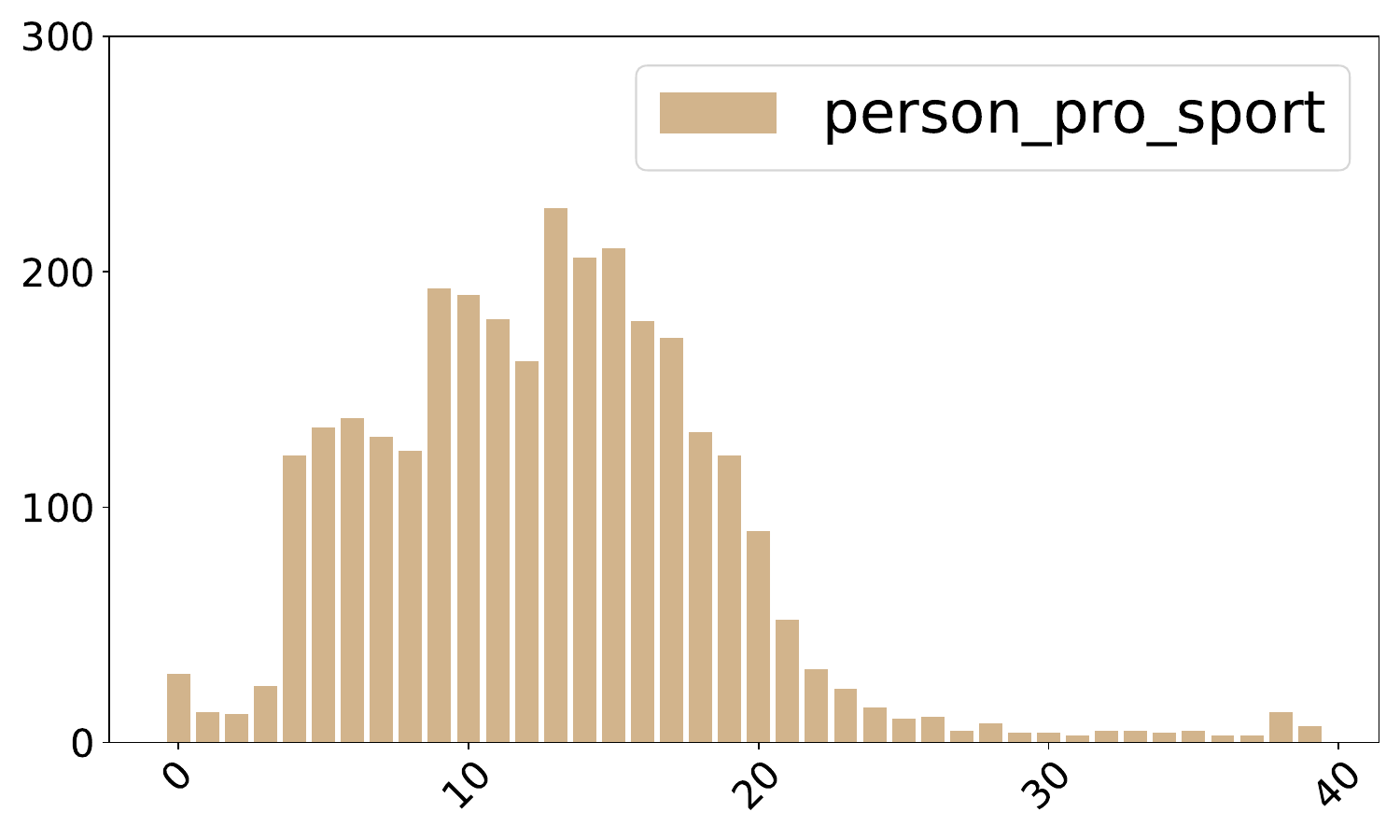}
    \includegraphics[width=0.15\textwidth]{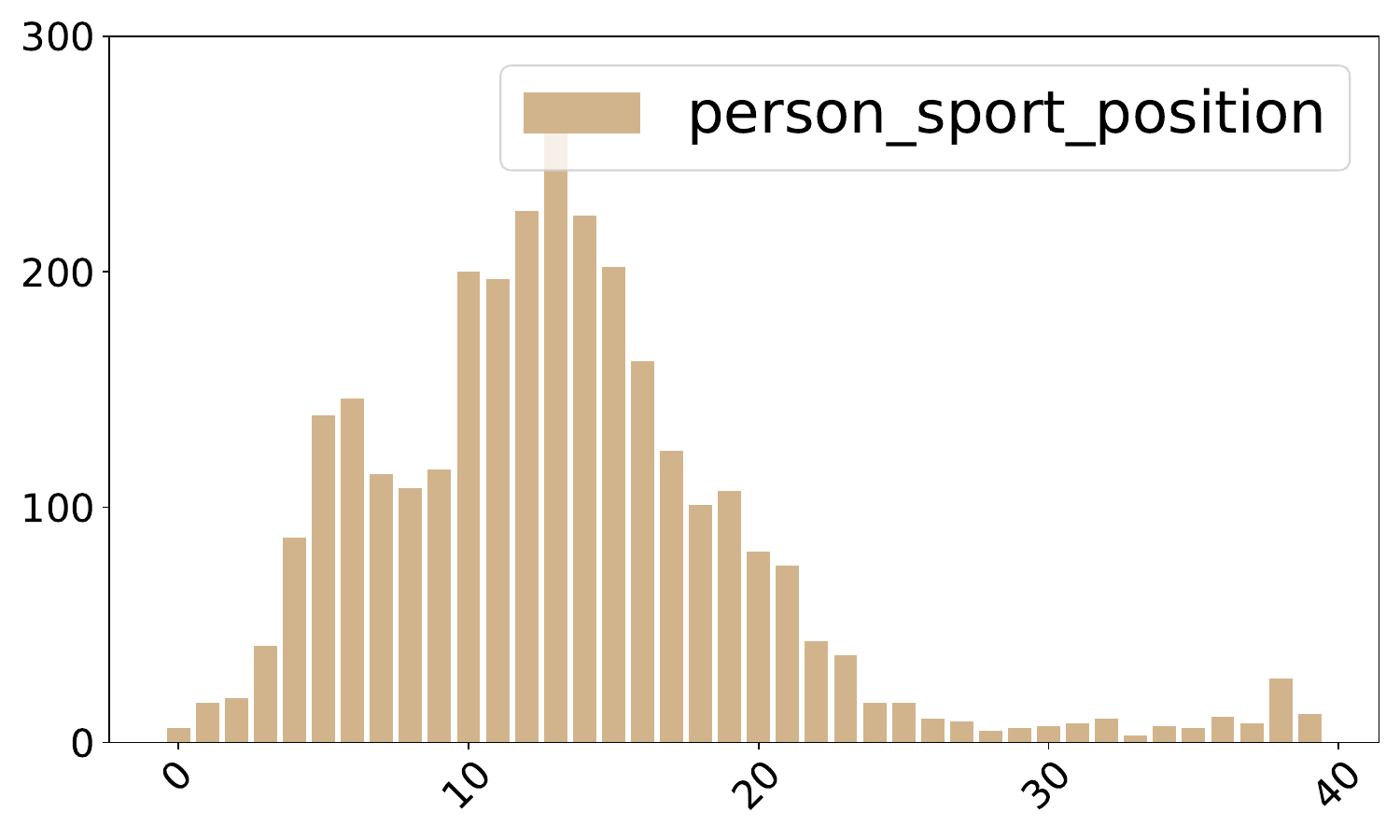}
    \includegraphics[width=0.15\textwidth]{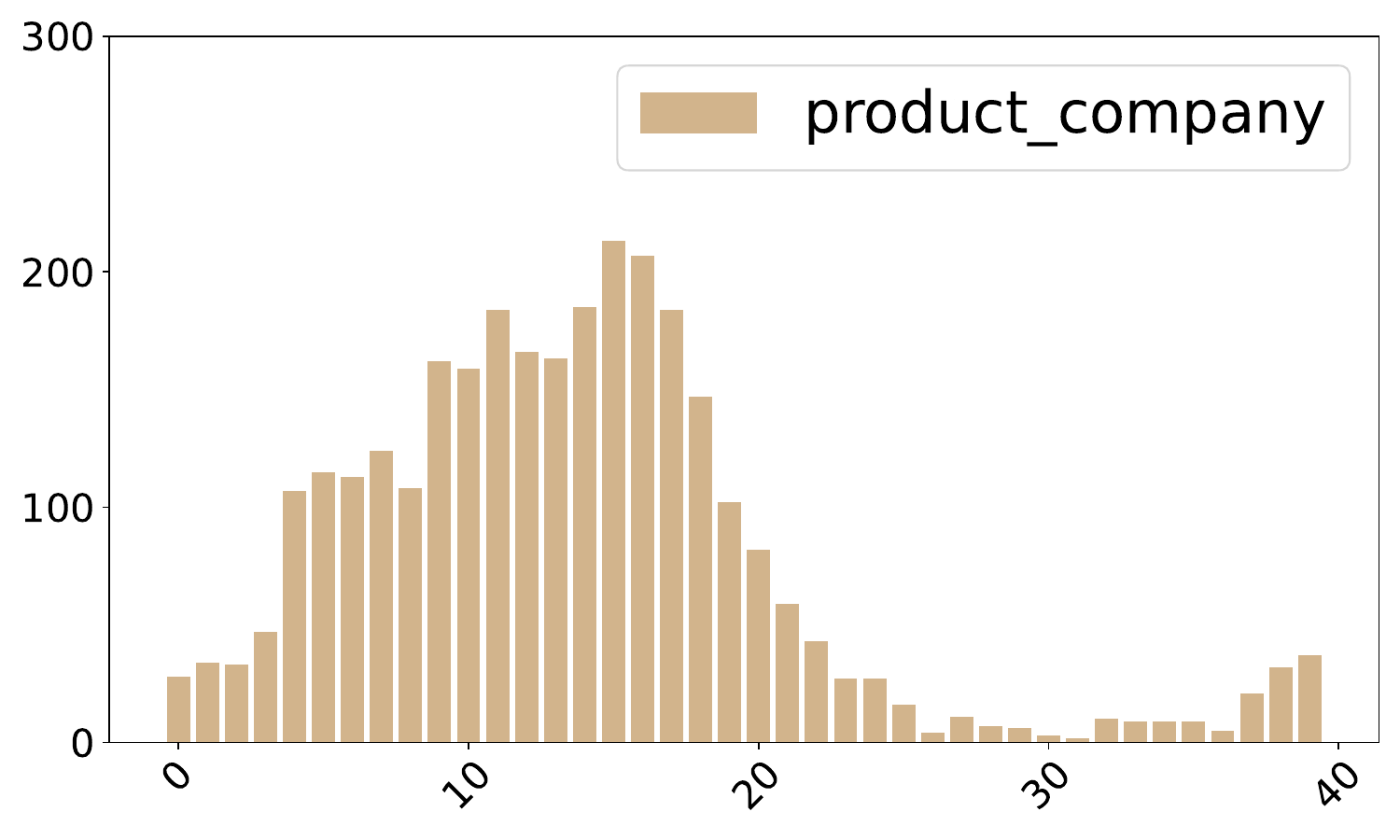}
    \includegraphics[width=0.15\textwidth]{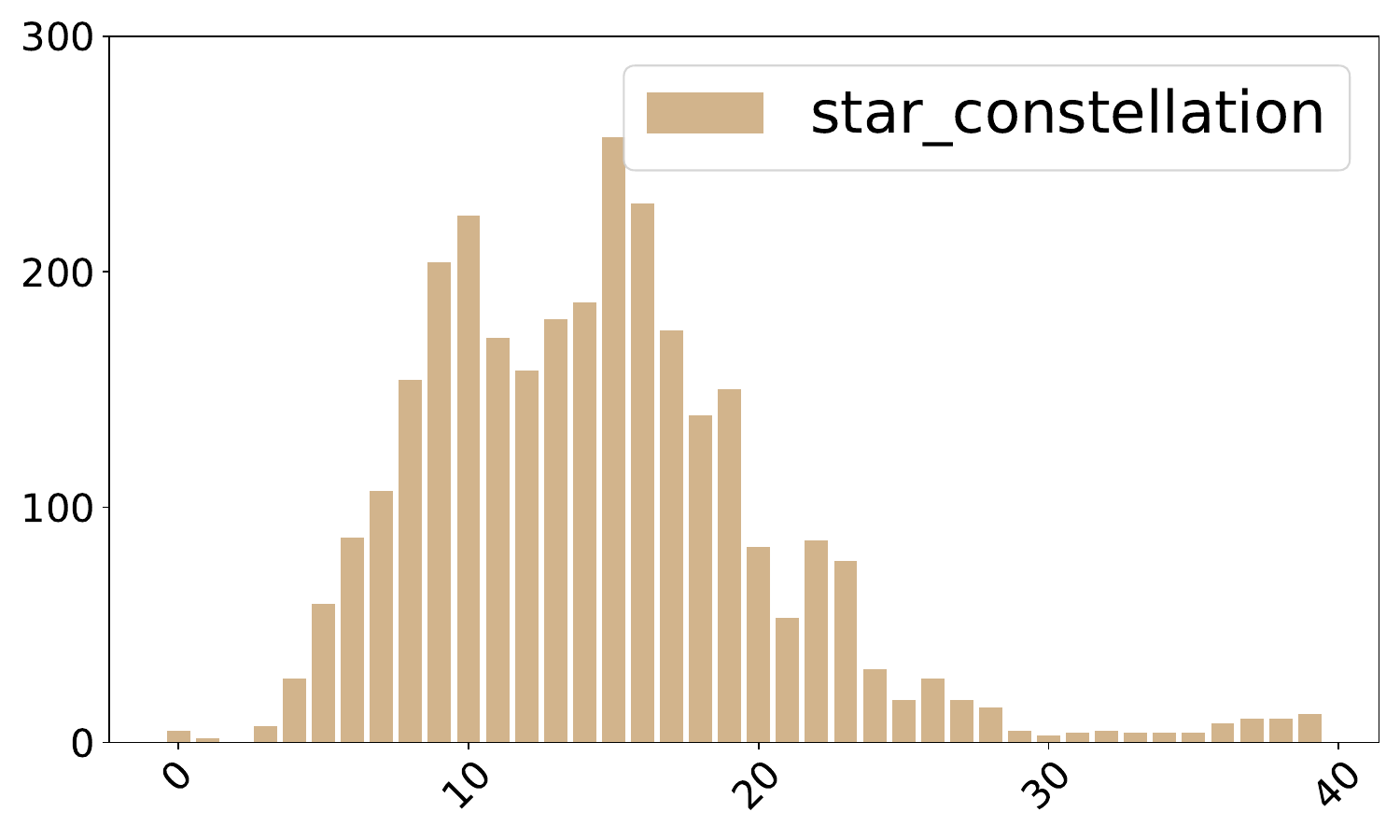}
    \caption{Distribution of \RelationSpecificNeurons
    across layers for the \textbf{13B} model. Similar to Figure \ref{fig:layer_dist}, identified \RelationSpecificNeurons are mostly located in the middle layers, except for \texttt{person\_mother}.}
    \label{fig:layer_dist_13b}
\end{figure}

\begin{figure}
    \centering
    \setlength{\abovecaptionskip}{-0.1cm}
    \setlength{\belowcaptionskip}{-0.4cm}
    \includegraphics[width=0.45\textwidth]{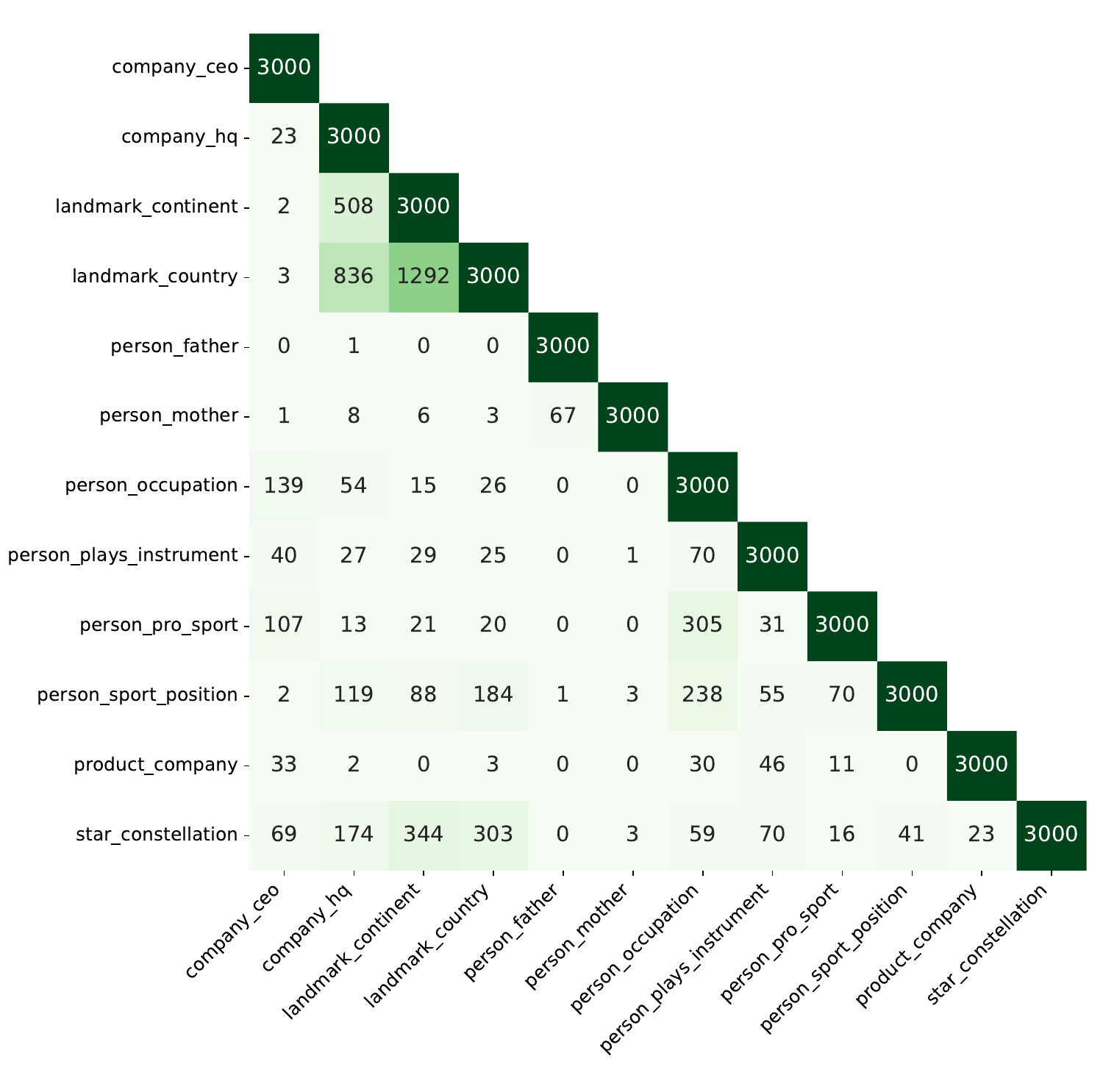}
    \caption{Neuron overlap of \RelationSpecificNeurons across 12 relations in the \textbf{13B} model. The overlap distribution is not similar to what we observe for the 7B model shown in Figure \ref{fig:neuron_overlap}, explaining the difference in inter-relation results (cf. Table \ref{fig:inter-relation}).}
    \label{fig:neuron_overlap_13b}
\end{figure}

We perform a similar analysis on the 13B model as we do for the 7B model.  We first show how the identified 3,000 \RelationSpecificNeurons are distributed across layers for each relation in Figure \ref{fig:layer_dist_13b}. The trend is similar to what we observe in the 7B model (cf. Figure \ref{fig:layer_dist}). Most of the \RelationSpecificNeurons are distributed in the middle layers. 
Then we show the overlap of \RelationSpecificNeurons across relations in Figure \ref{fig:neuron_overlap_13b}. Surprisingly, the overlap pattern is very different from what we observe in the 7B model. First, it seems that many relations that share a concept of ``location'' share extensive neurons, e.g., \texttt{company\_hq}, \texttt{landmark\_country}, \texttt{landmark\_country} and \texttt{star\_constellation}. This explains the difference in inter-relation results between the models (cf. Figure \ref{fig:inter-relation}) where we see deactivating neurons of \texttt{landmark\_country} significantly influence other relations also concerning location for the 13B model but not for the 7B model.

\begin{figure}
    \centering
\includegraphics[width=0.48\textwidth]{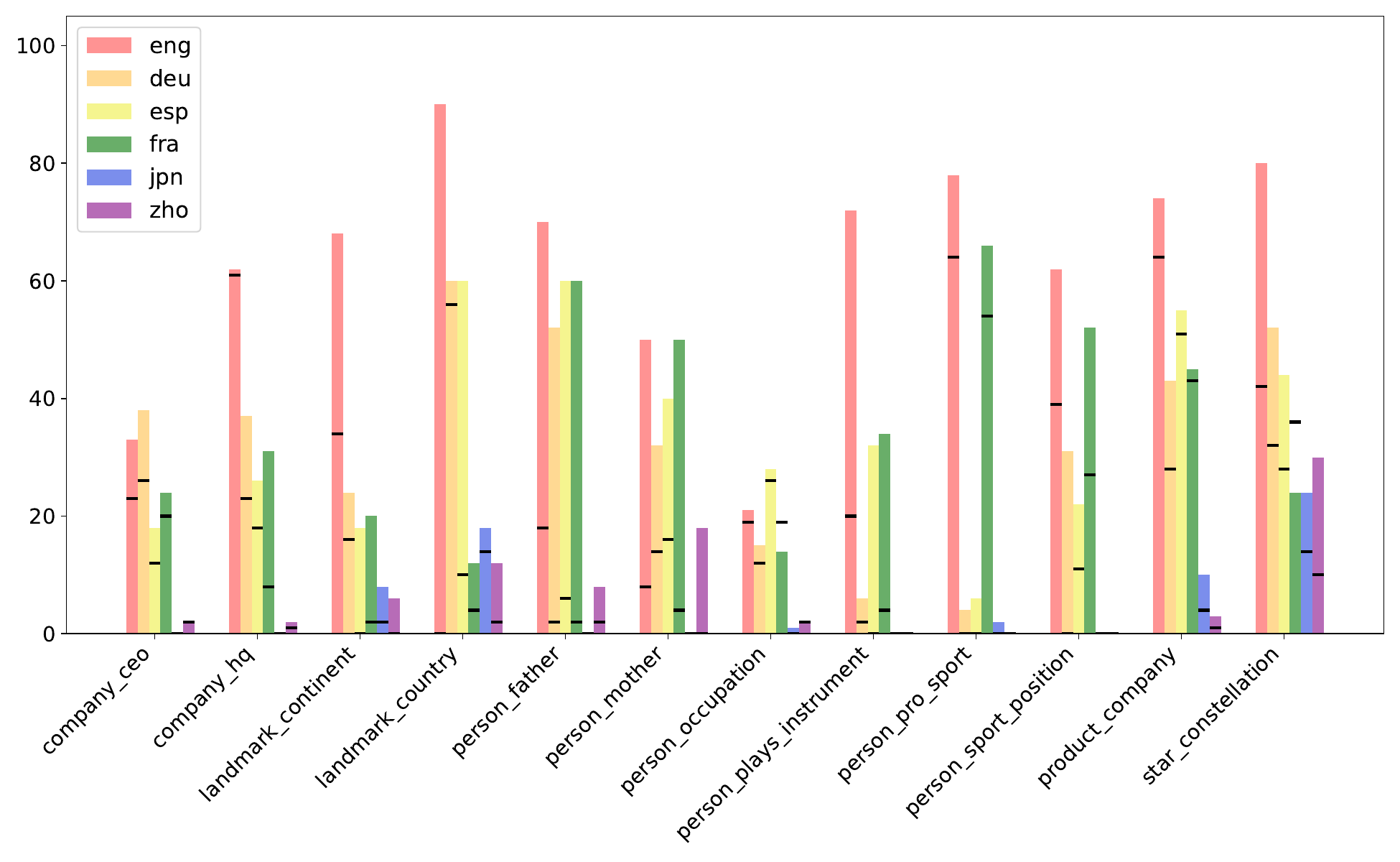}
    \caption{Accuracy on 12 relations across 6 languages from the \textbf{13B} model.
    The bars show the accuracy of the original model, with a horizontal line in each bar that indicates the performance after the deactivation of 3,000 \RelationSpecificNeurons.}
    \label{fig:multilingual_13b}
\end{figure}

\begin{figure*}
    \centering
    \setlength{\belowcaptionskip}{-0.5cm}
    \includegraphics[width=0.23\textwidth]{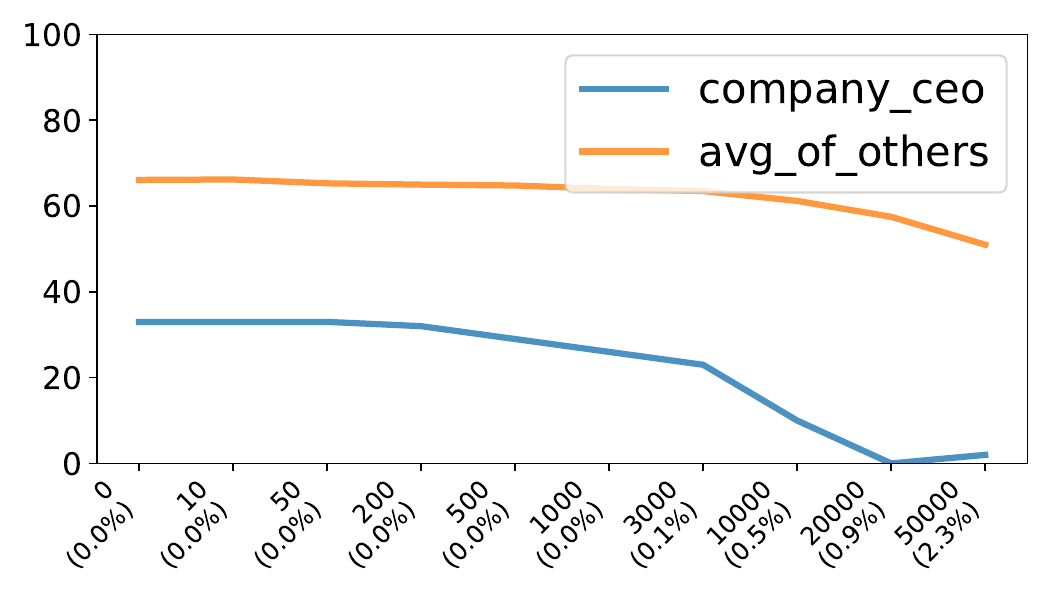}
    \includegraphics[width=0.23\textwidth]{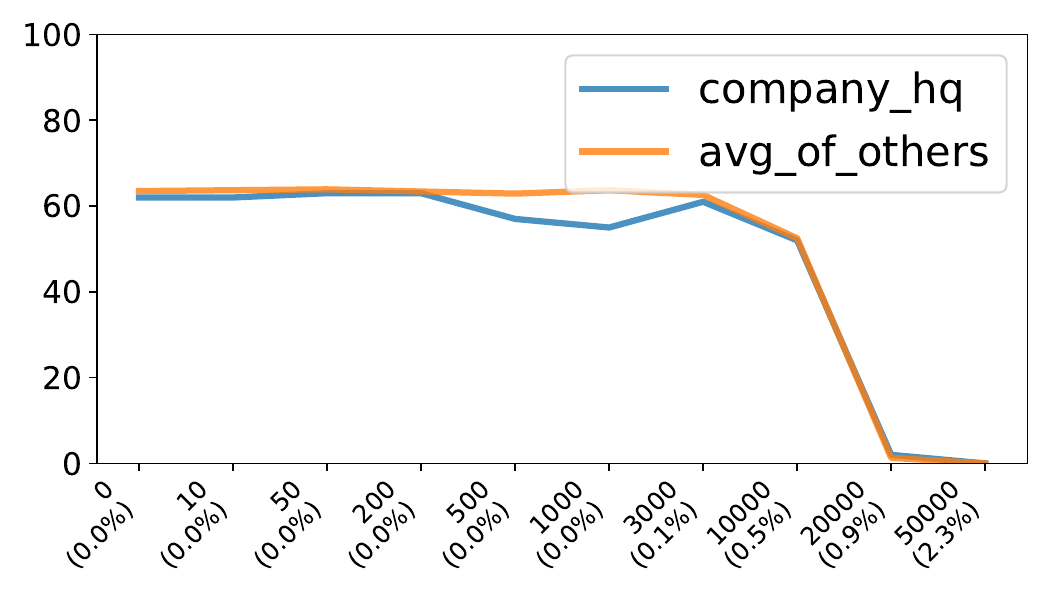}
    \includegraphics[width=0.23\textwidth]{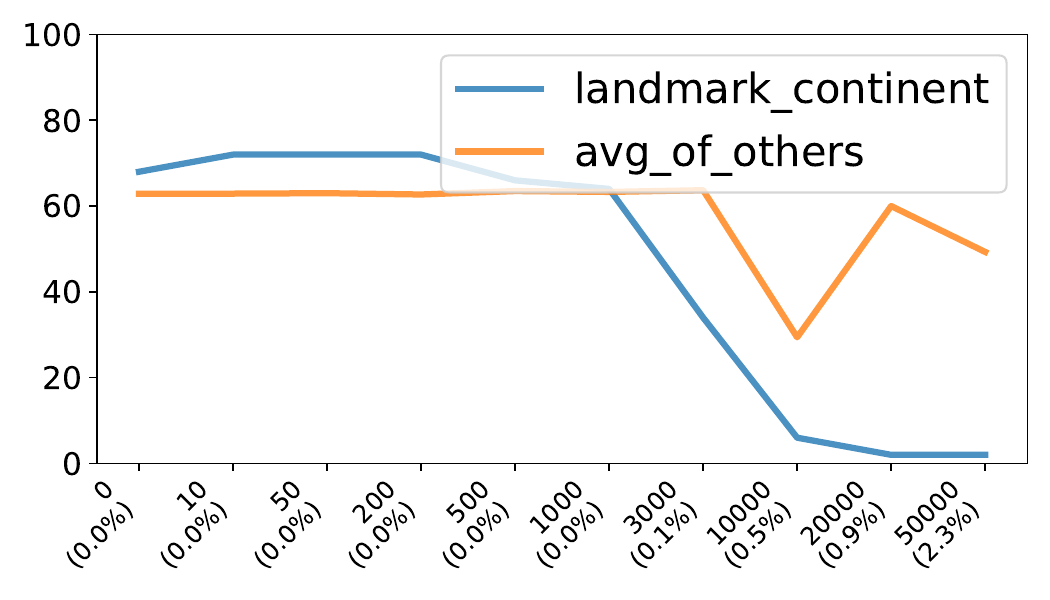}
    \includegraphics[width=0.23\textwidth]{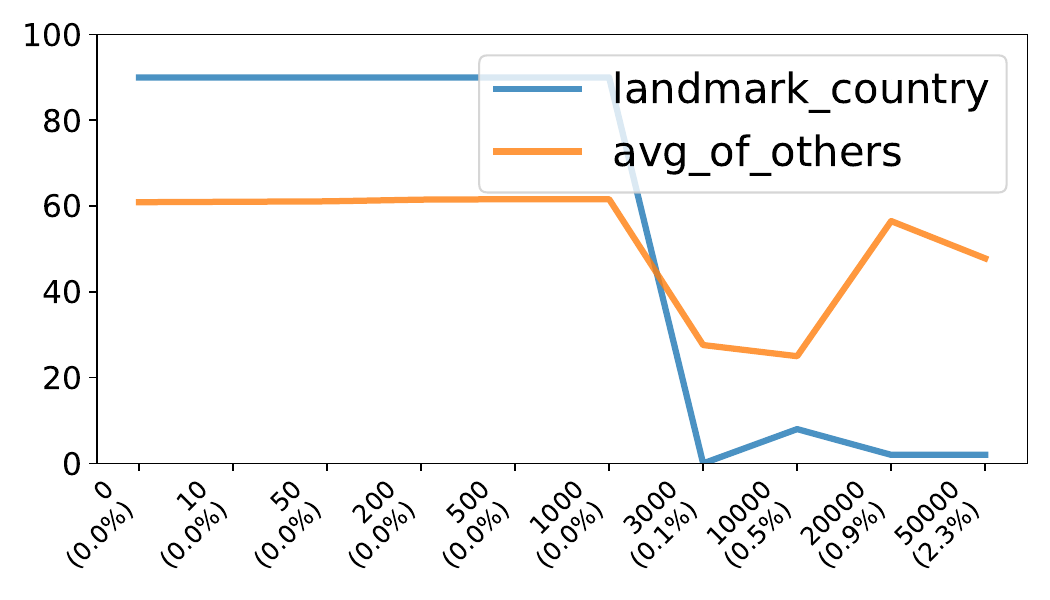}
    \includegraphics[width=0.23\textwidth]{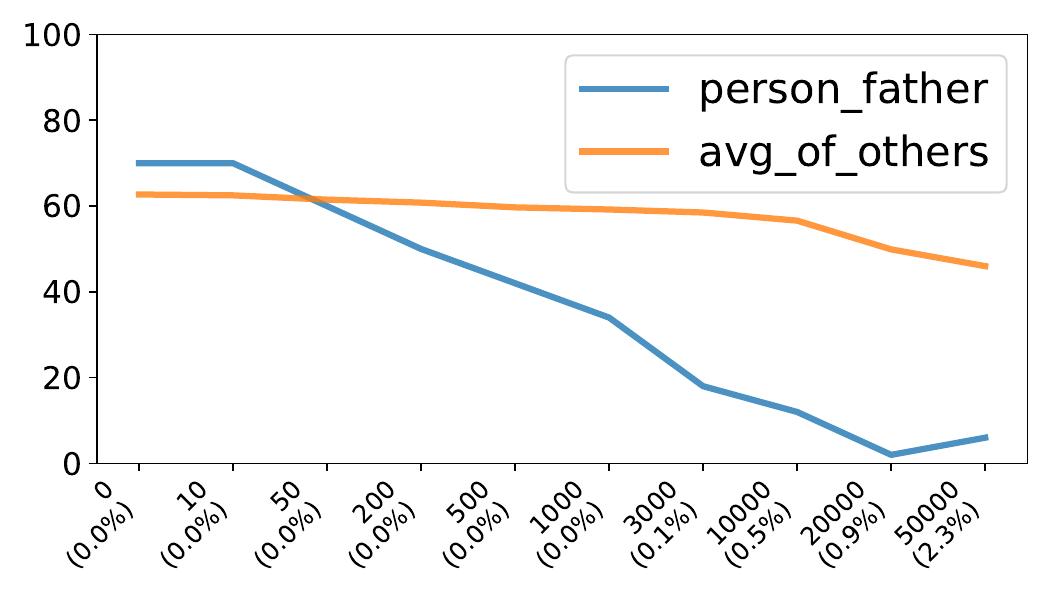}
    \includegraphics[width=0.23\textwidth]{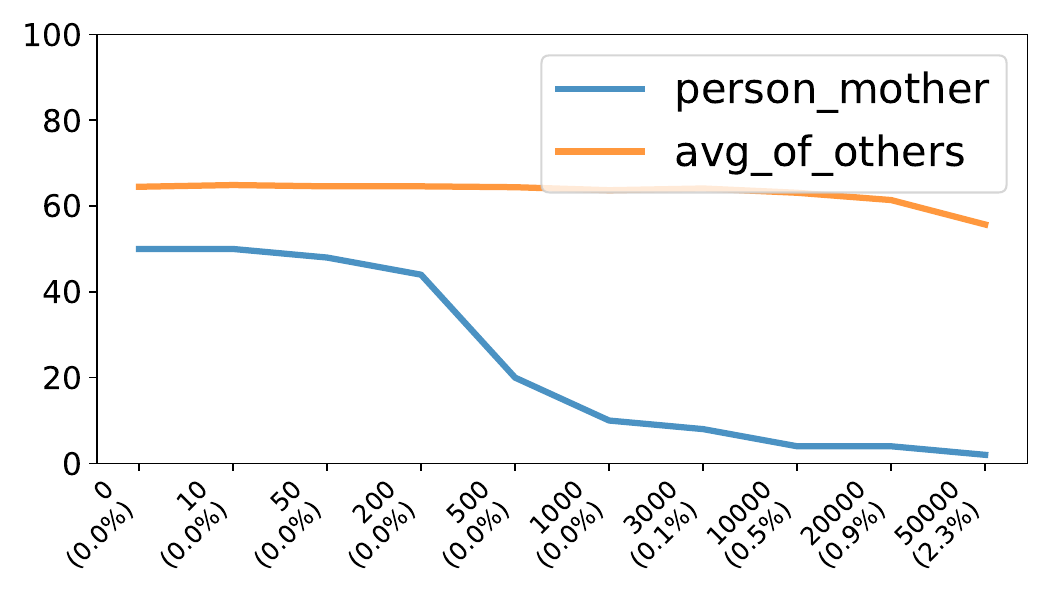}
    \includegraphics[width=0.23\textwidth]{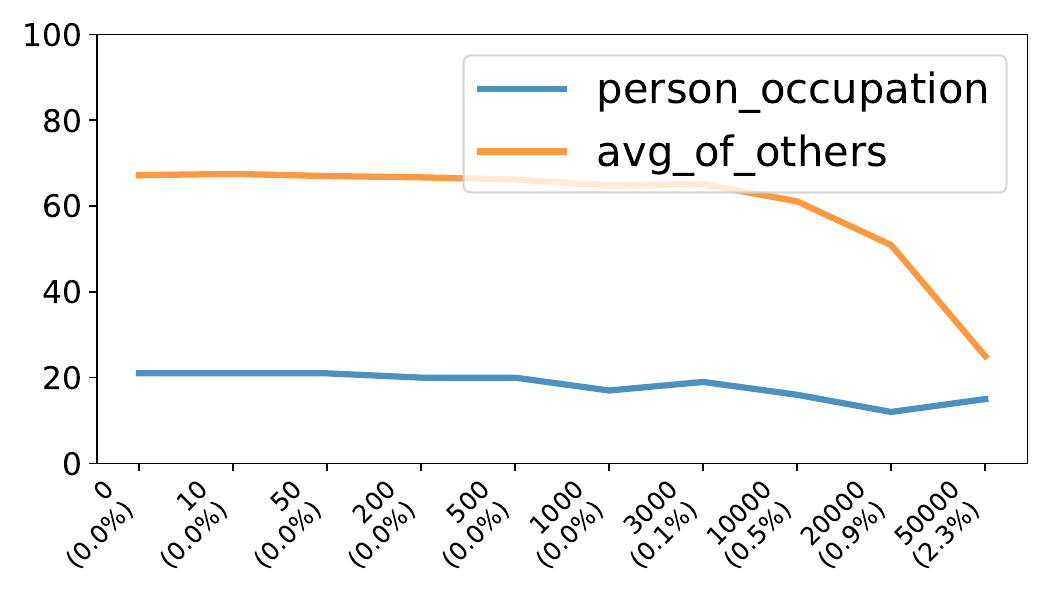}
    \includegraphics[width=0.23\textwidth]{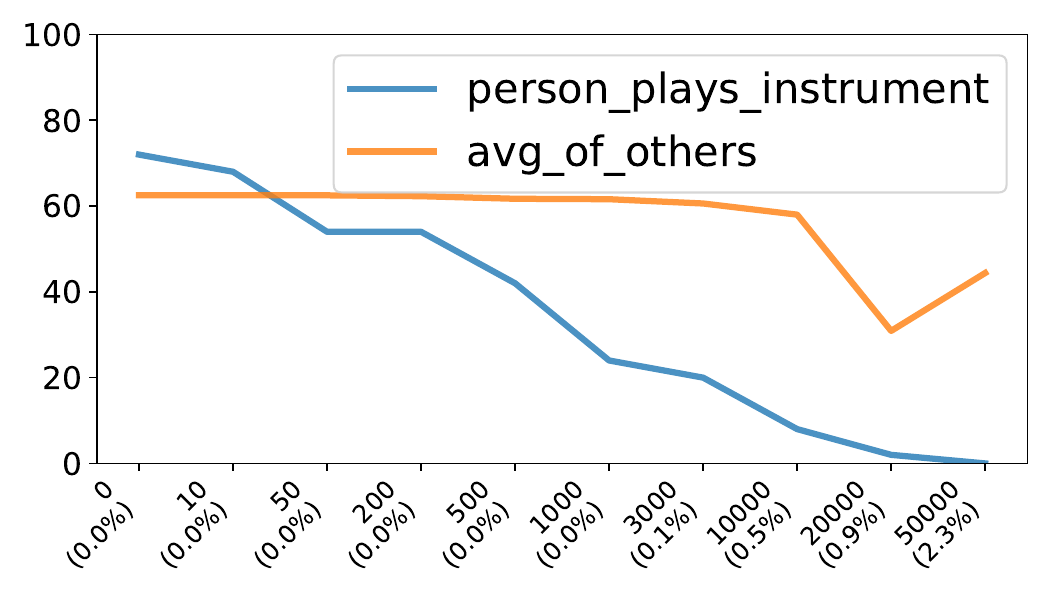}
    \includegraphics[width=0.23\textwidth]{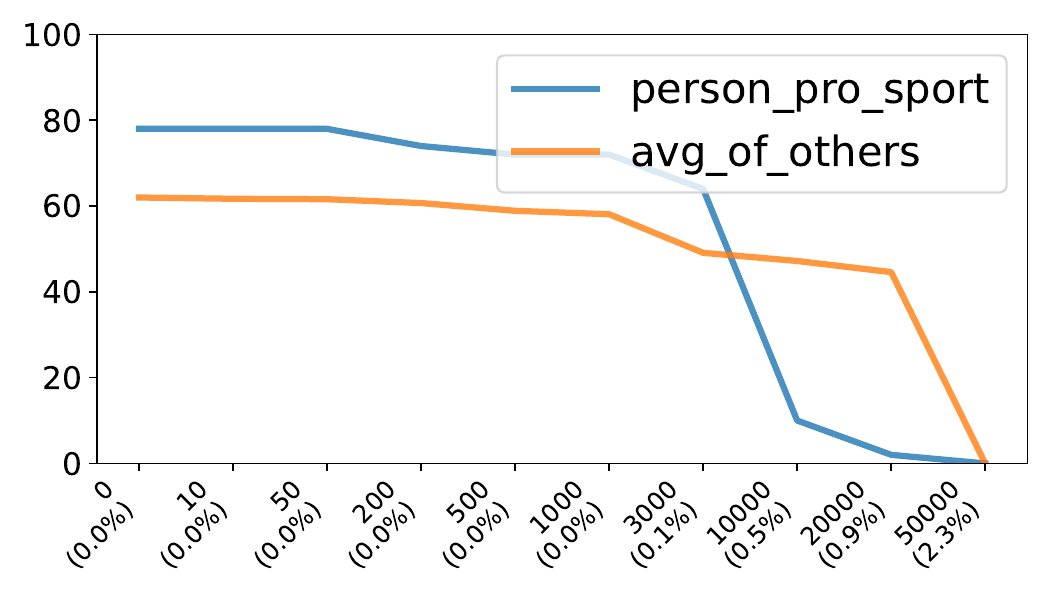}
    \includegraphics[width=0.23\textwidth]{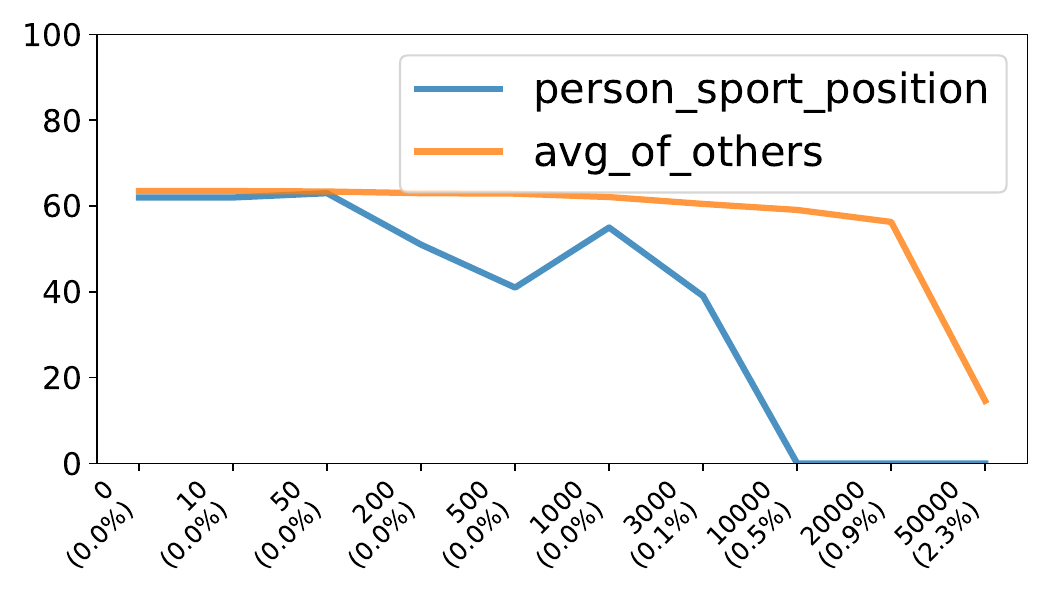}
    \includegraphics[width=0.23\textwidth]{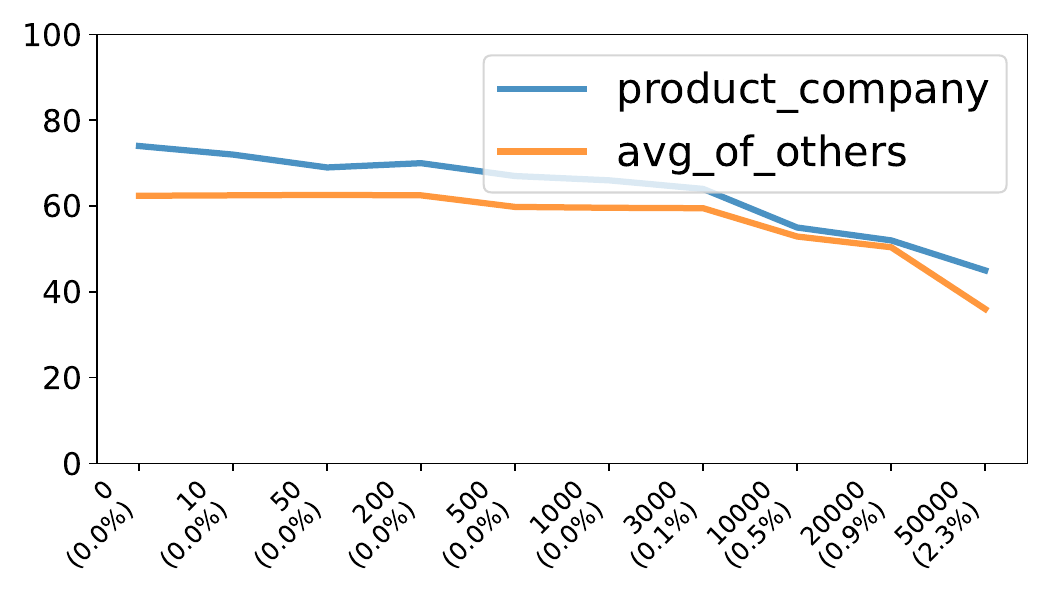}
    \includegraphics[width=0.23\textwidth]{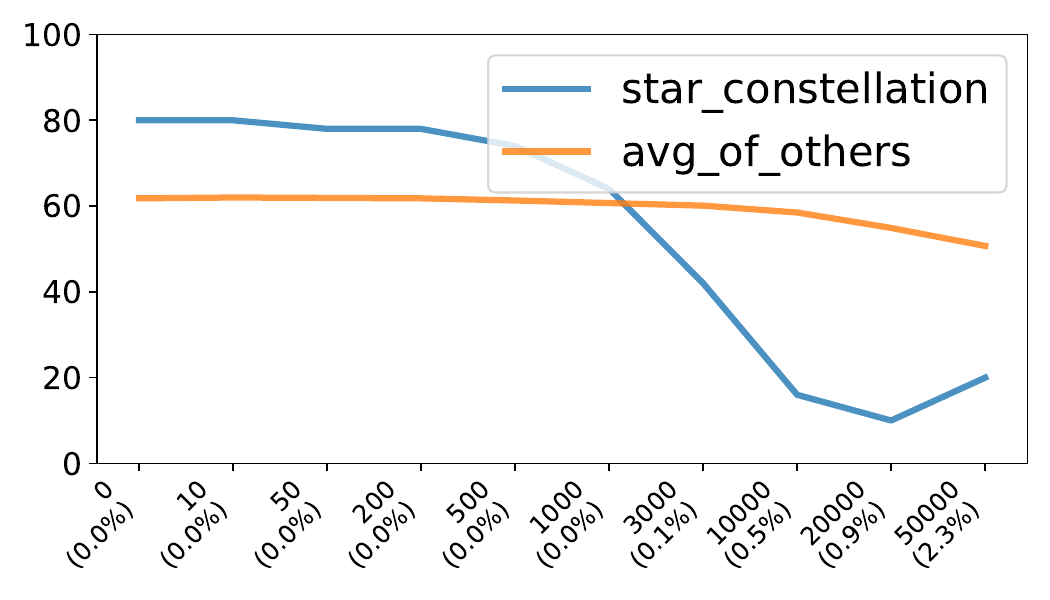}
    \caption{Influence of deactivating different numbers of \RelationSpecificNeurons for each relation (the \textbf{13B} model). The variation of accuracy on the relation itself and the average accuracy on other relations is shown.}
    \label{fig:neuron_num_13b}
\end{figure*}

We then demonstrate the effect of varying numbers of \RelationSpecificNeurons using the same numbers: 10, 50, 200, 500, 1,000, 3,000, 10,000, 20,000, and 50,000. Figure \ref{fig:neuron_num_13b} presents the results. The global trend is similar to what we observe for the 7B model: deactivating more neurons results in a further drop in accuracy across all relations. This indicates the \textbf{neuron cumulativity} is universal across models. 
\RelationSpecificNeurons for most relations present a similar cumulative effect to the 13B model. The original two outliers in the 7B model (\texttt{person\_occupation} and \texttt{person\_company} where the accuracy does not drop to 0 in the 7B model) even show a plateau, i.e., the accuracy remains almost unchanged or only slightly decreases. This might suggest that facts belonging to these two relations might be well-memorized by the models and are less sensitive to the deactivation of \RelationSpecificNeurons.

Lastly, we show whether the identified \RelationSpecificNeurons from the 13B model are also multilingual. We use the same translated prompt sets as we use for the 7B model. We deactivate the 3,000 neurons identified using English and see how this affects the performance in other languages: German (\textbf{deu}), Spanish (\textbf{esp}), French (\textbf{fra}), Chinese (\textbf{zho}), and Japanese (\textbf{jpn}). The results are presented in Figure \ref{fig:multilingual_13b}. We observe similar results as from the 7B model: when we deactivate \RelationSpecificNeurons identified using English prompts, many relations are influenced across languages, suggesting models with different sizes also have multilingual relational neurons. We also see some interesting counterexamples: deactivating \texttt{landmark\_country} neurons completely deteriorates the relation in English but not in German. This indicates while some neurons have multilingual relational functionalities, there are still some relations dealt with in a language-specific manner.

\begin{figure*}
    \centering
    \includegraphics[width=0.32\textwidth]{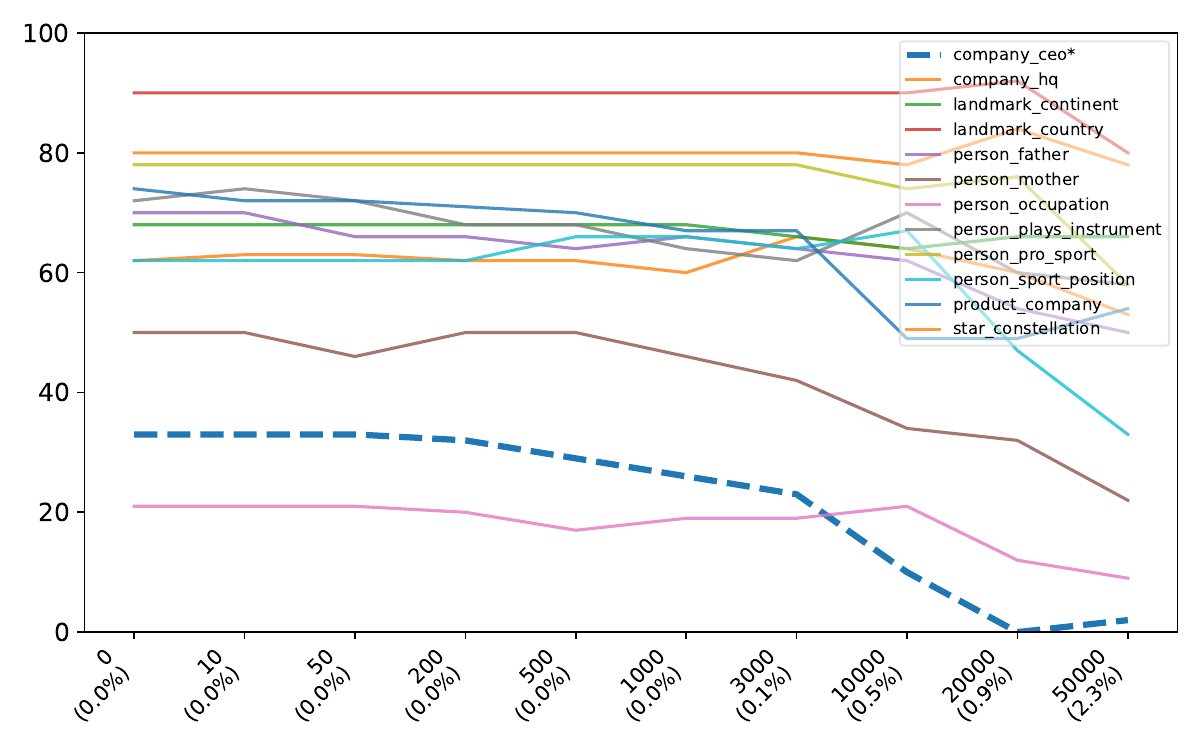}
    \includegraphics[width=0.32\textwidth]{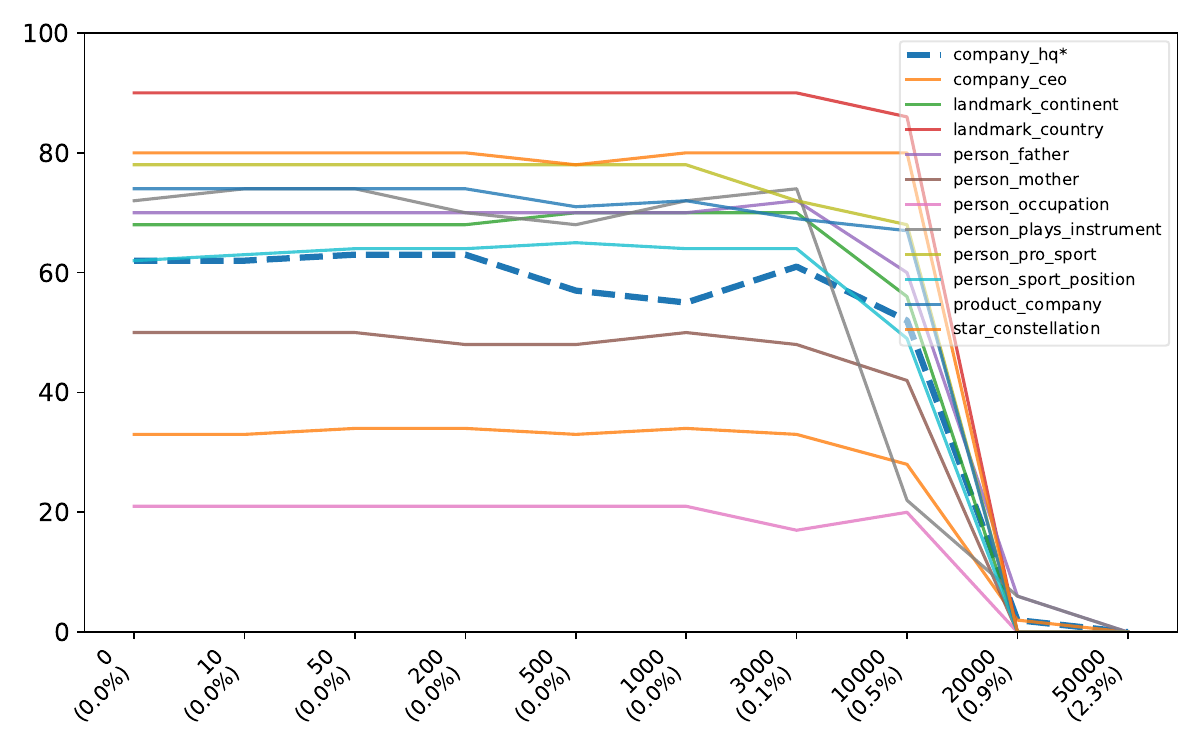}
    \includegraphics[width=0.32\textwidth]{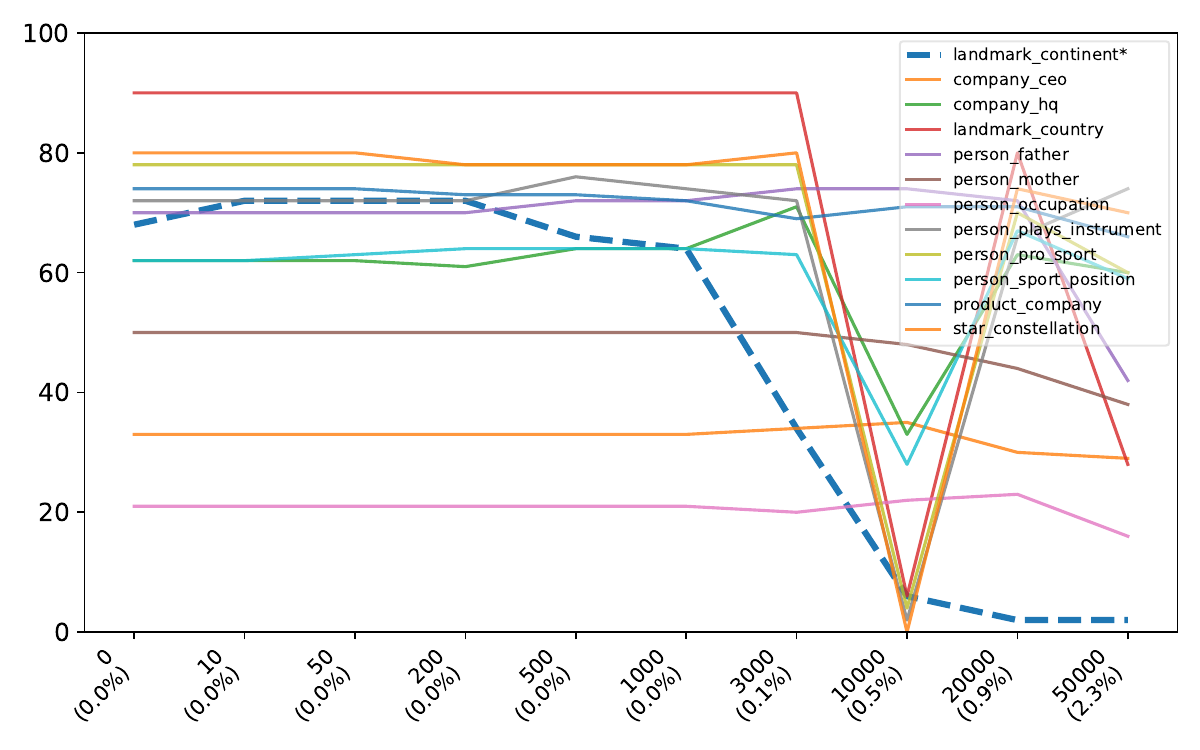}
    \includegraphics[width=0.32\textwidth]{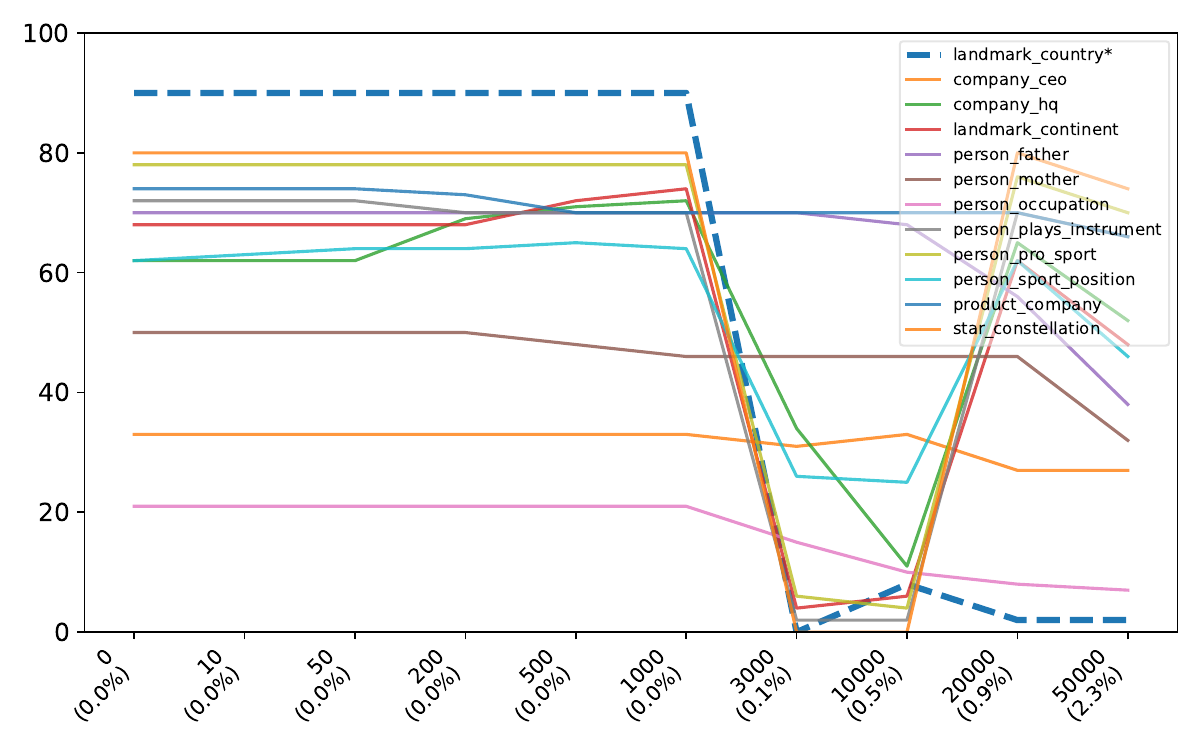}
    \includegraphics[width=0.32\textwidth]{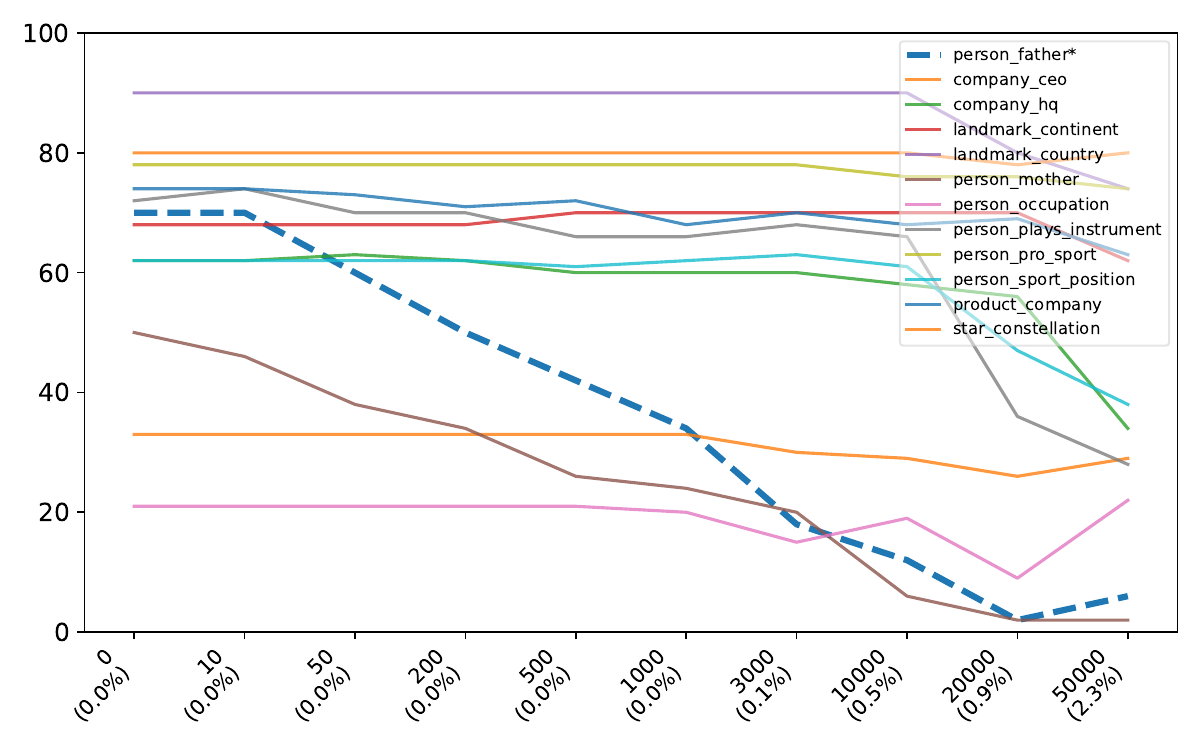}
    \includegraphics[width=0.32\textwidth]{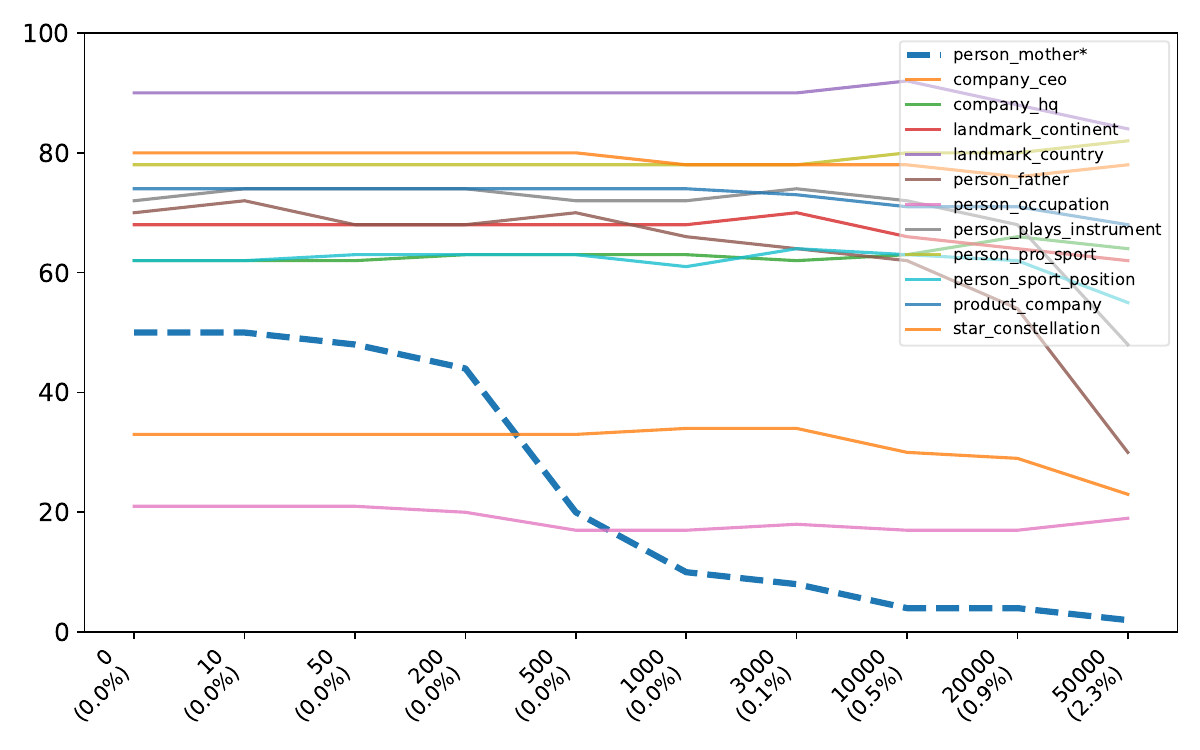}
    \includegraphics[width=0.32\textwidth]{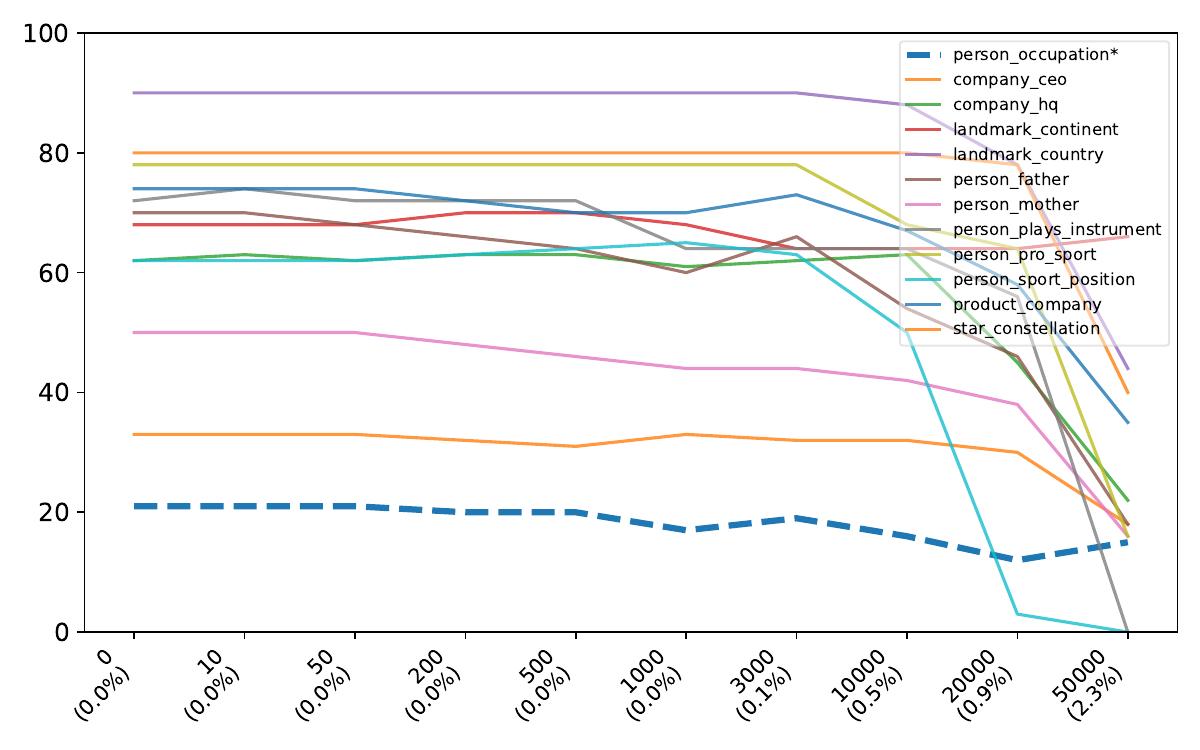}
    \includegraphics[width=0.32\textwidth]{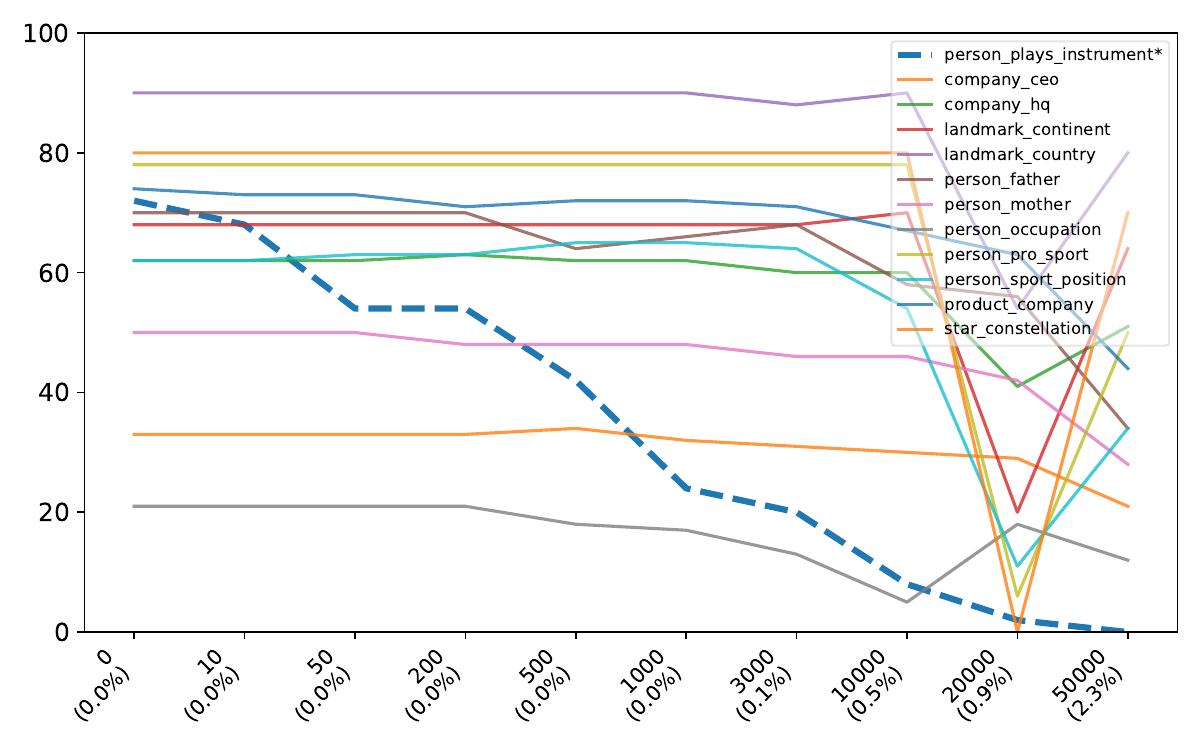}
    \includegraphics[width=0.32\textwidth]{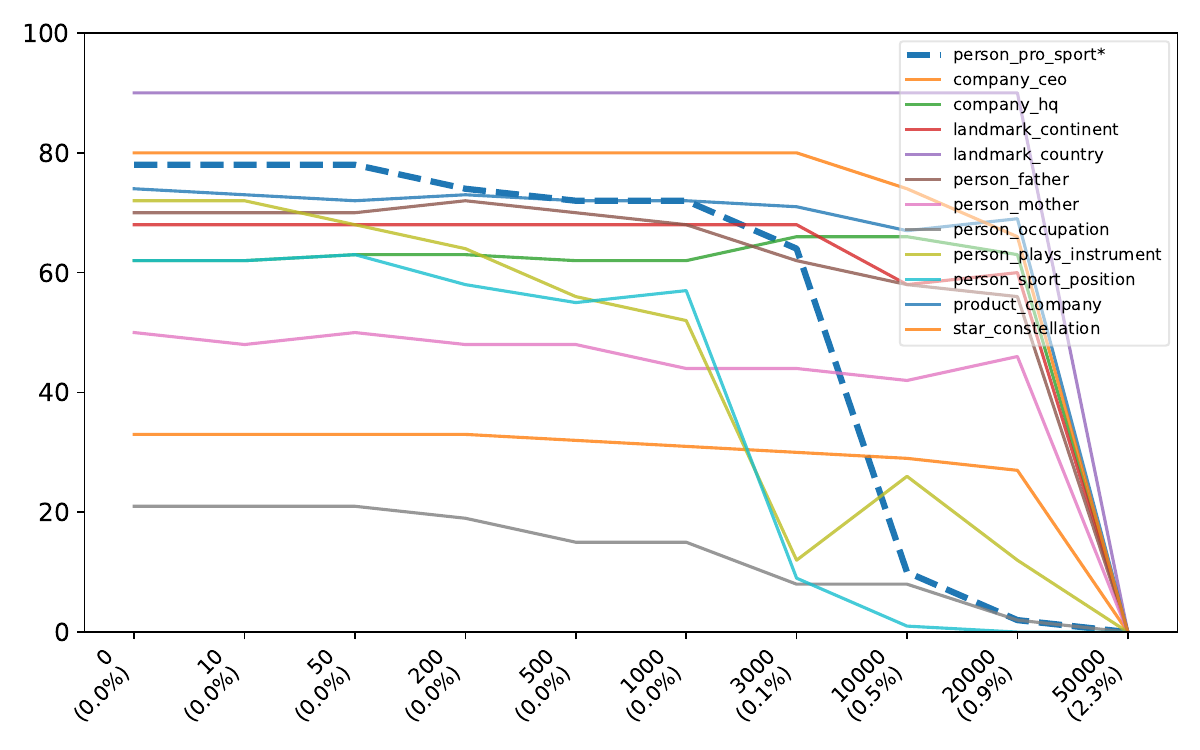}
    \includegraphics[width=0.32\textwidth]{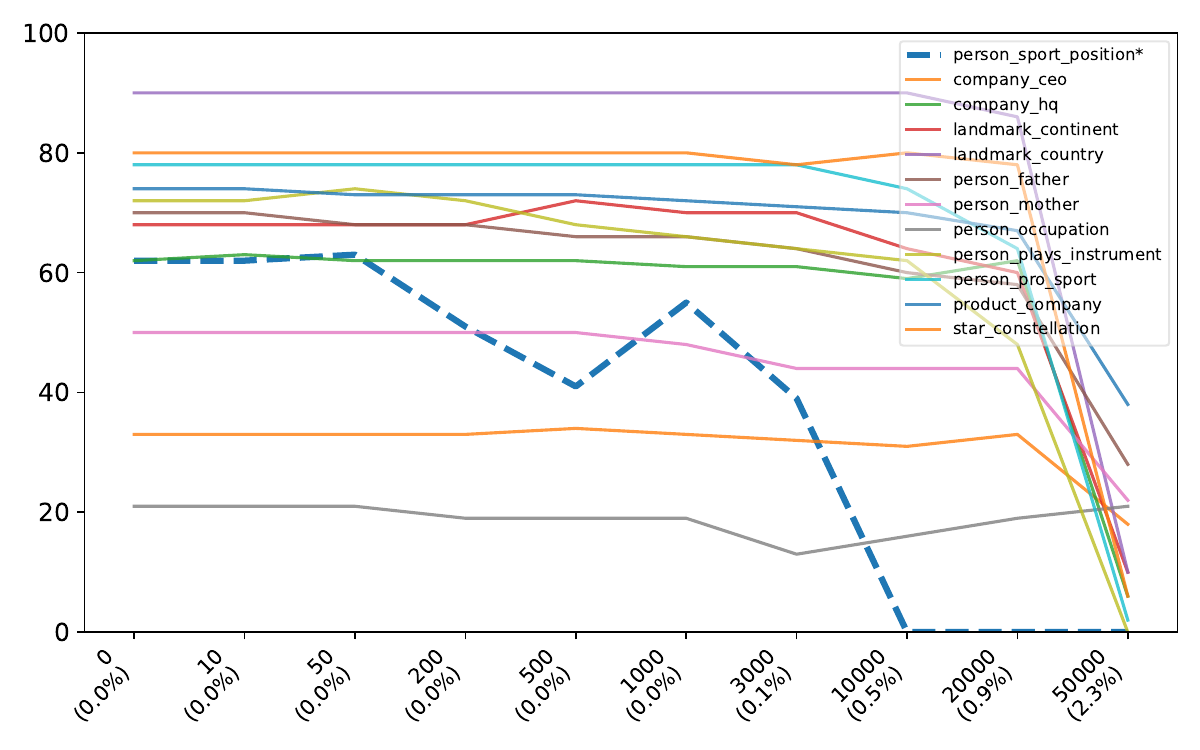}
    \includegraphics[width=0.32\textwidth]{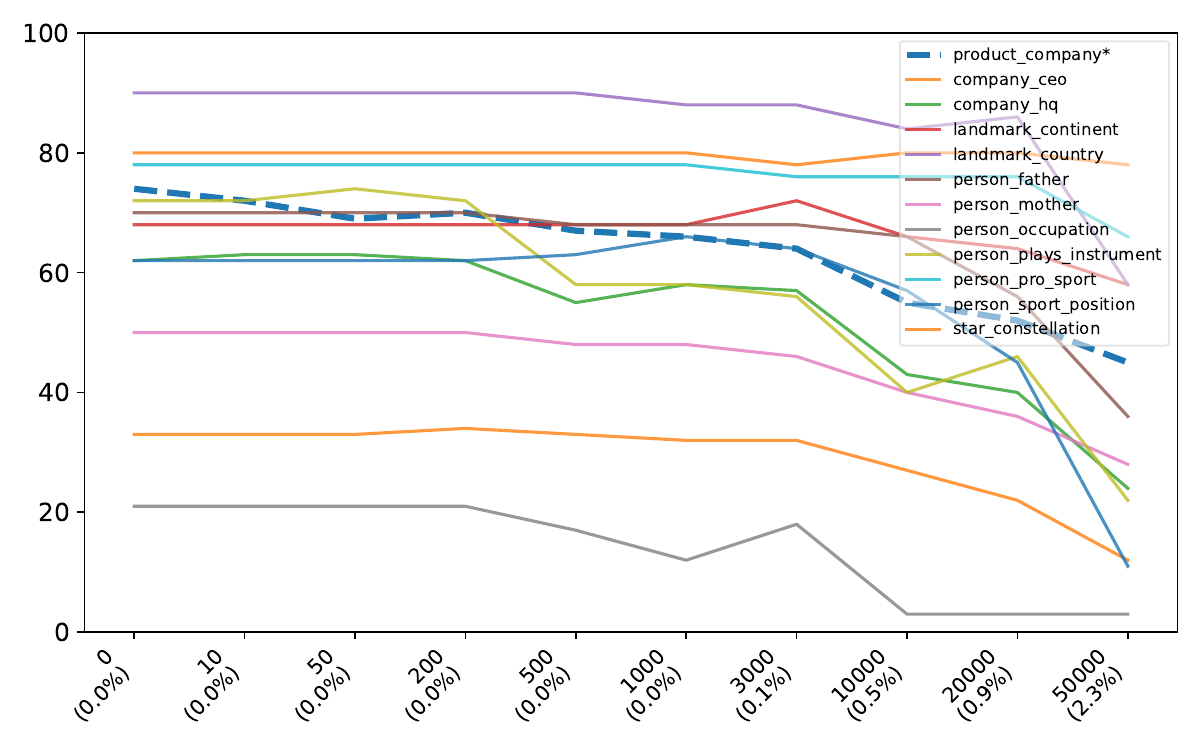}
    \includegraphics[width=0.32\textwidth]{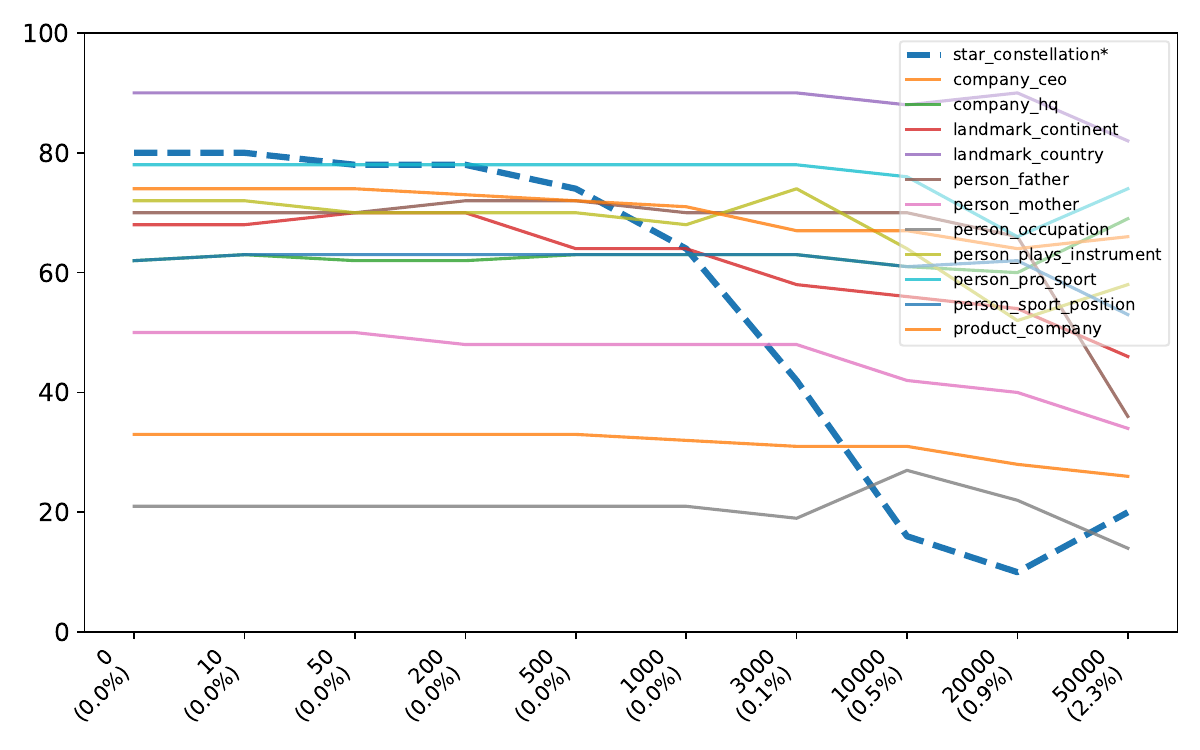}
    \caption{Influence of deactivating different numbers of
    \RelationSpecificNeurons in the \textbf{13B} model for each
    relation. The variation of accuracy on the relation
    itself (noted with ``*'' and a dashed line style) and
    the accuracy on all other relations is shown in each
    figure.}
    \label{fig:neuron_num_all_13b}
\end{figure*}
\newpage

\begin{figure*}
    \centering
    \includegraphics[width=0.32\textwidth]{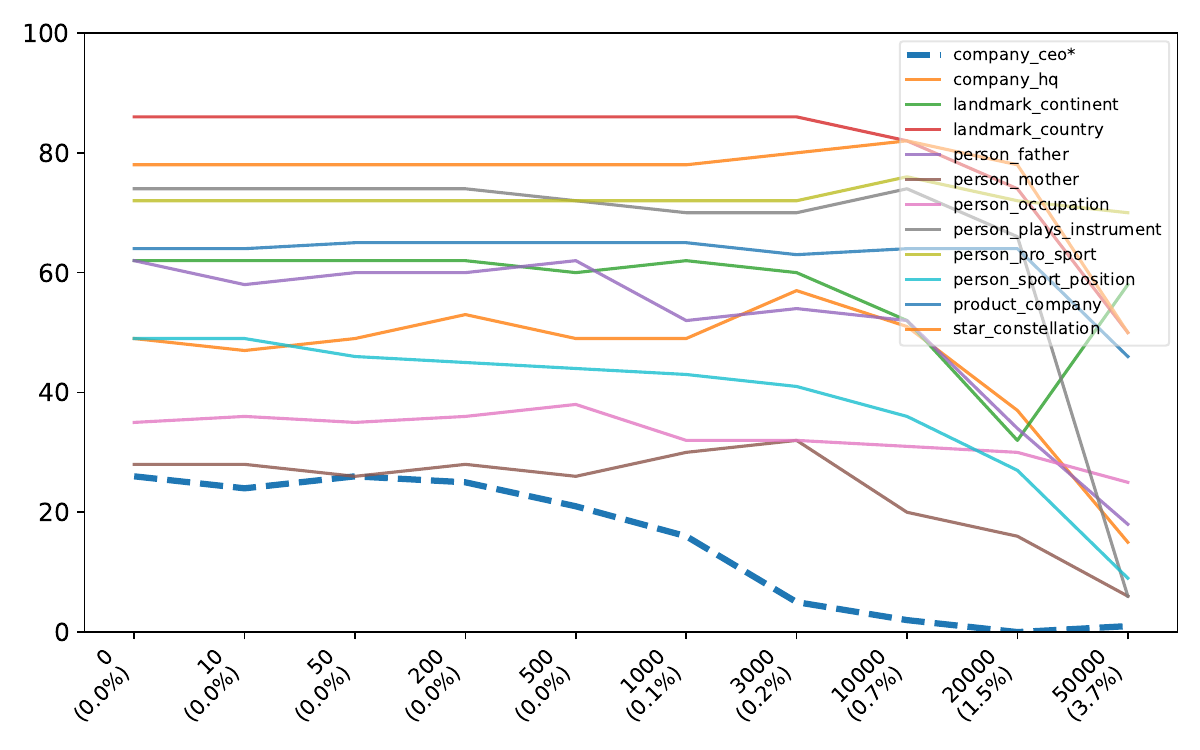}
    \includegraphics[width=0.32\textwidth]{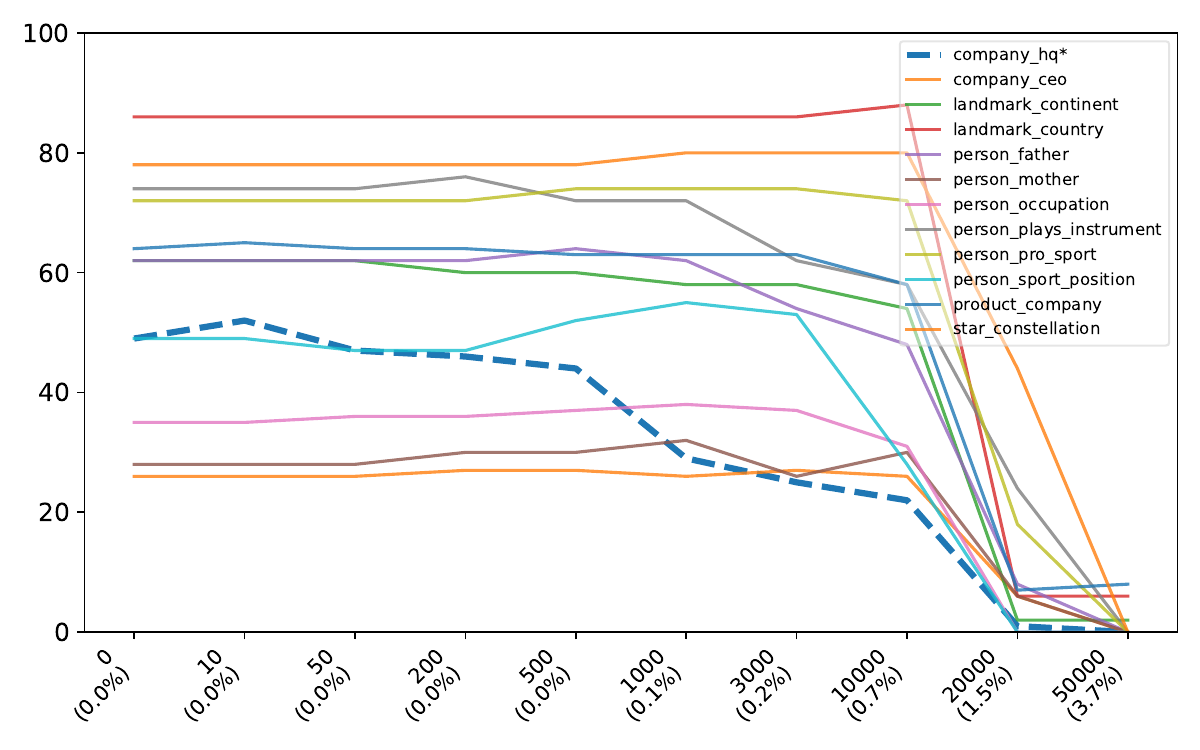}
    \includegraphics[width=0.32\textwidth]{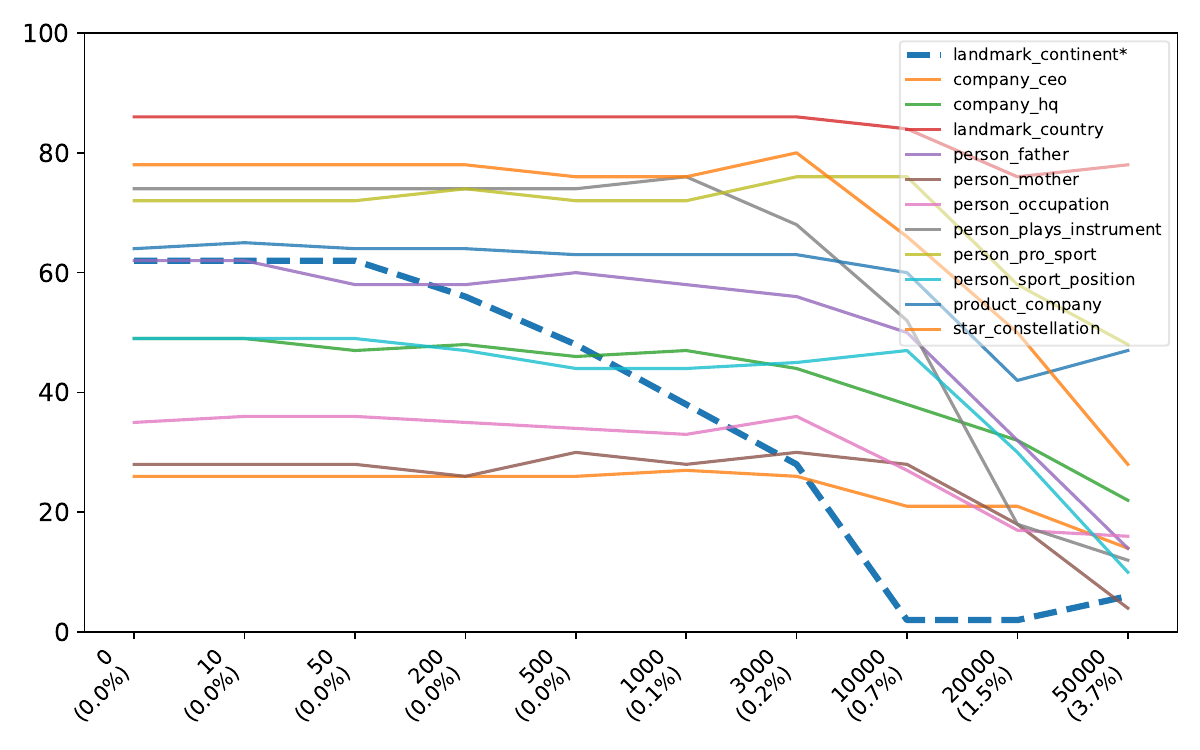}
    \includegraphics[width=0.32\textwidth]{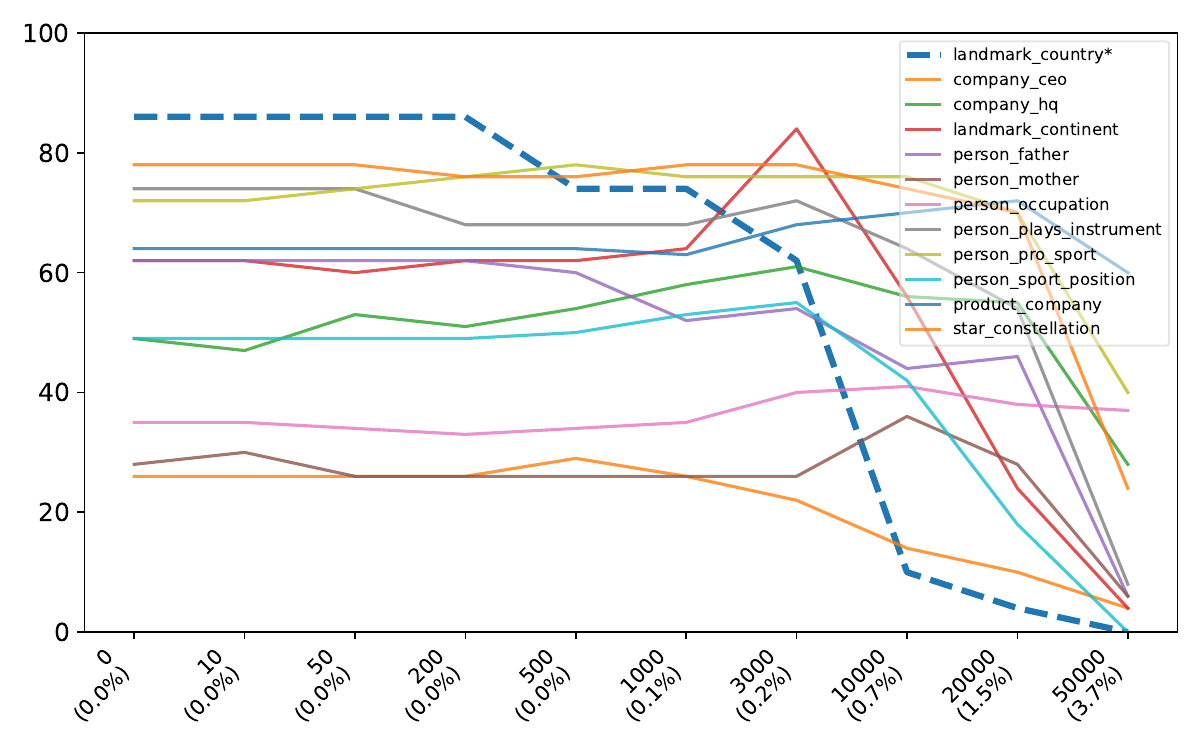}
    \includegraphics[width=0.32\textwidth]{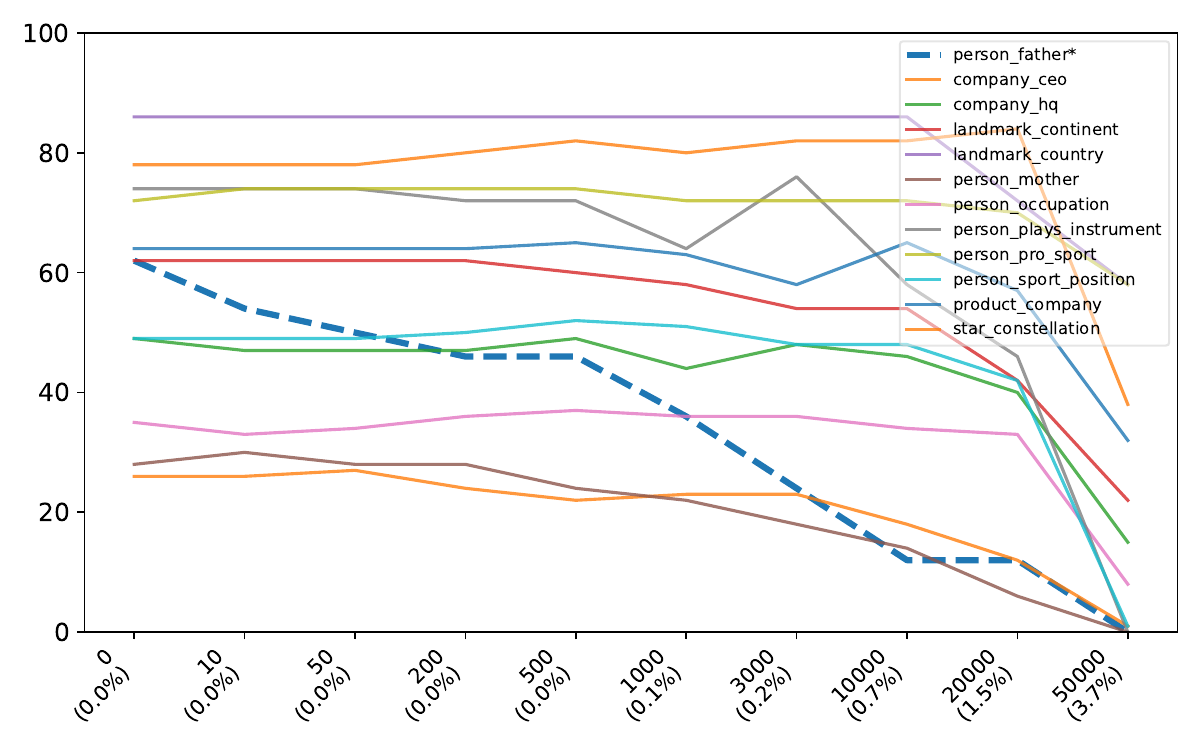}
    \includegraphics[width=0.32\textwidth]{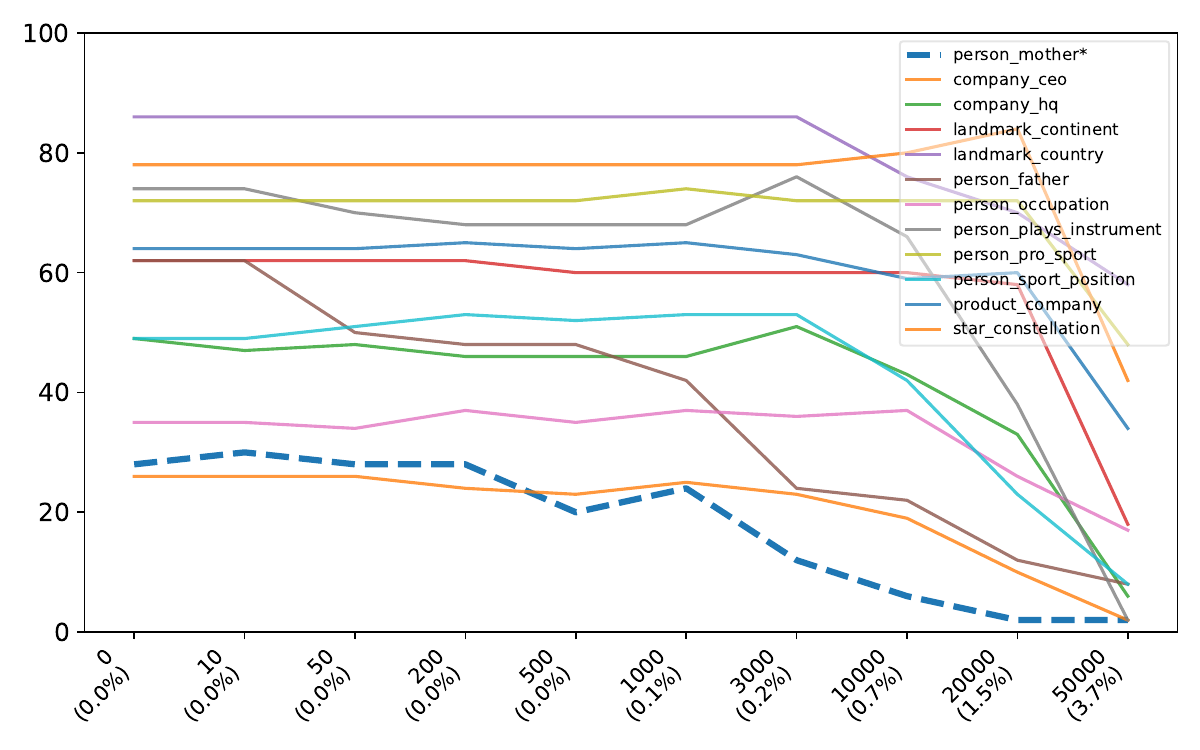}
    \includegraphics[width=0.32\textwidth]{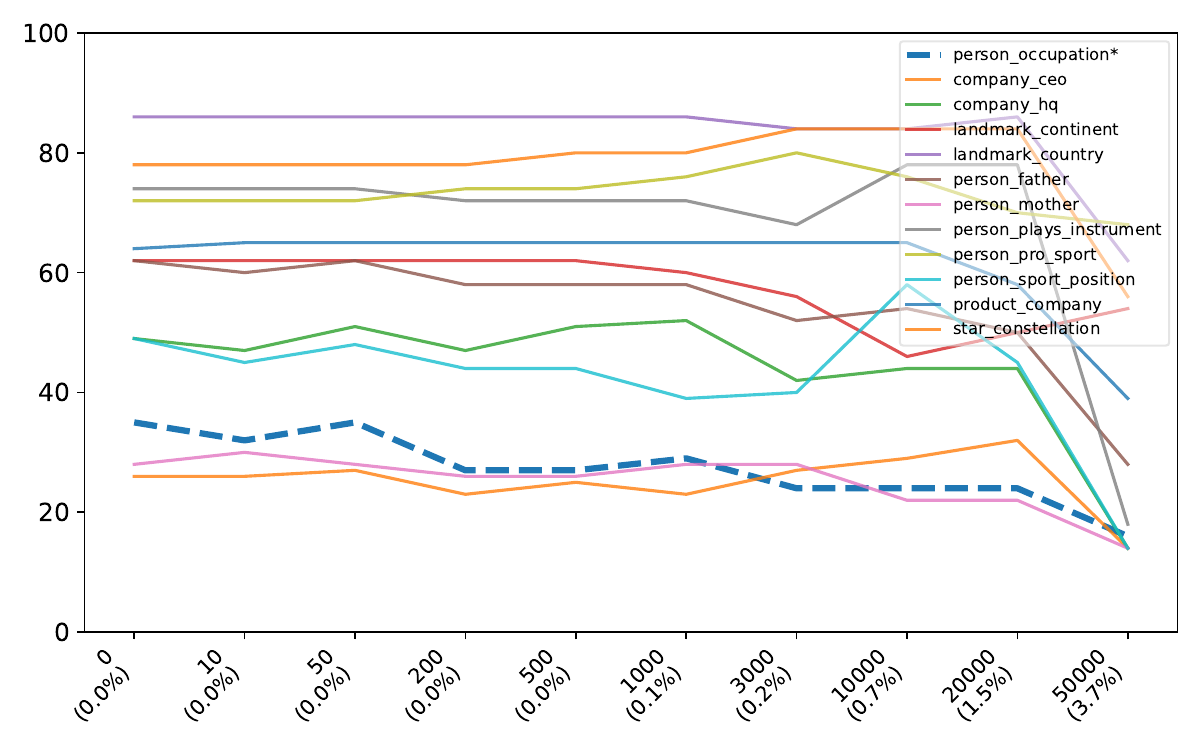}
    \includegraphics[width=0.32\textwidth]{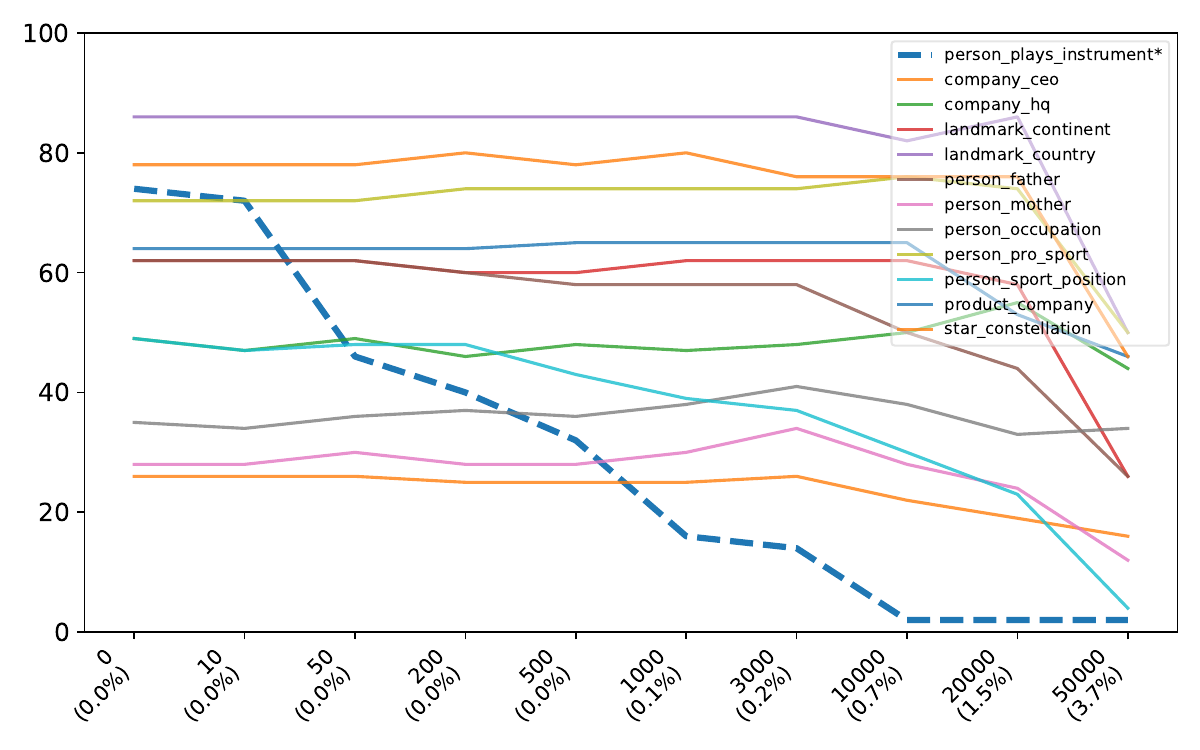}
    \includegraphics[width=0.32\textwidth]{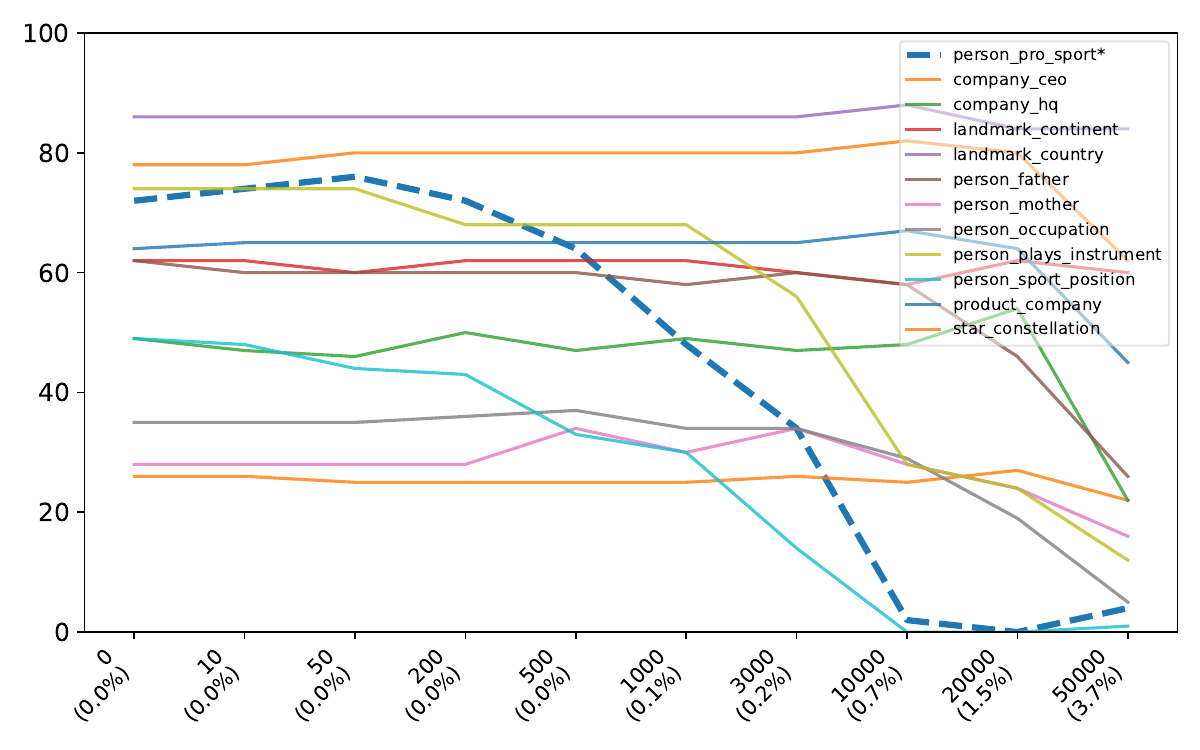}
    \includegraphics[width=0.32\textwidth]{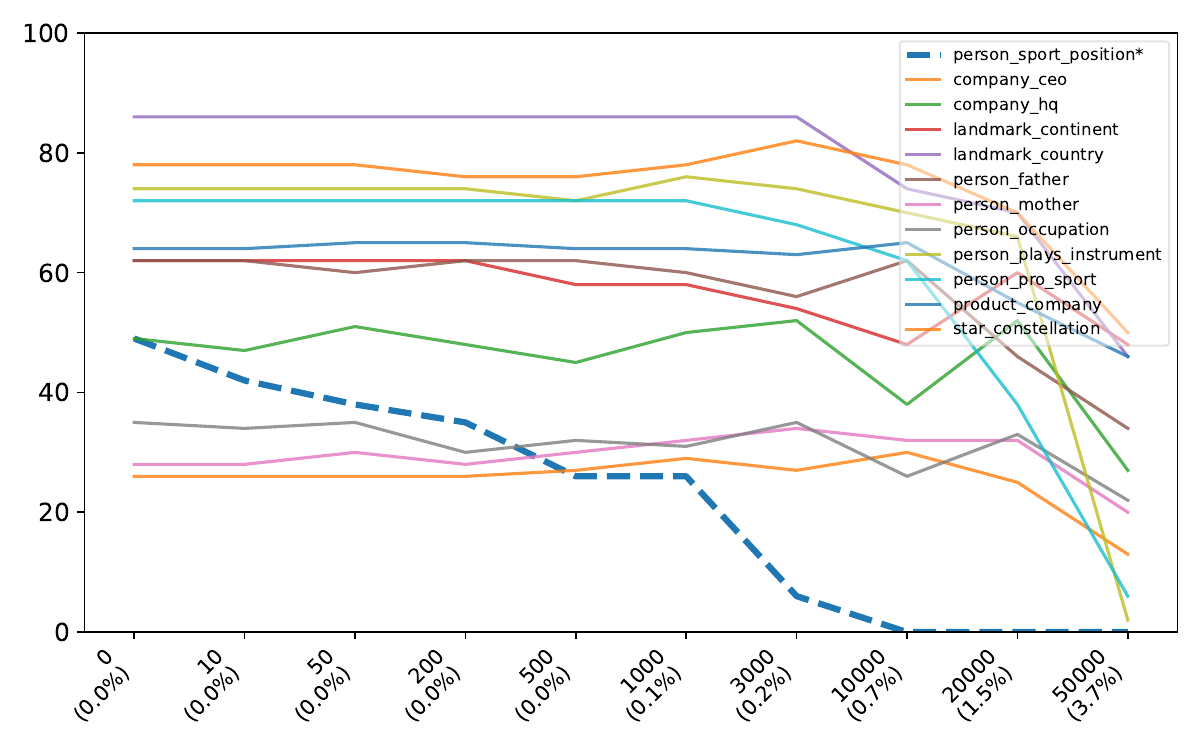}
    \includegraphics[width=0.32\textwidth]{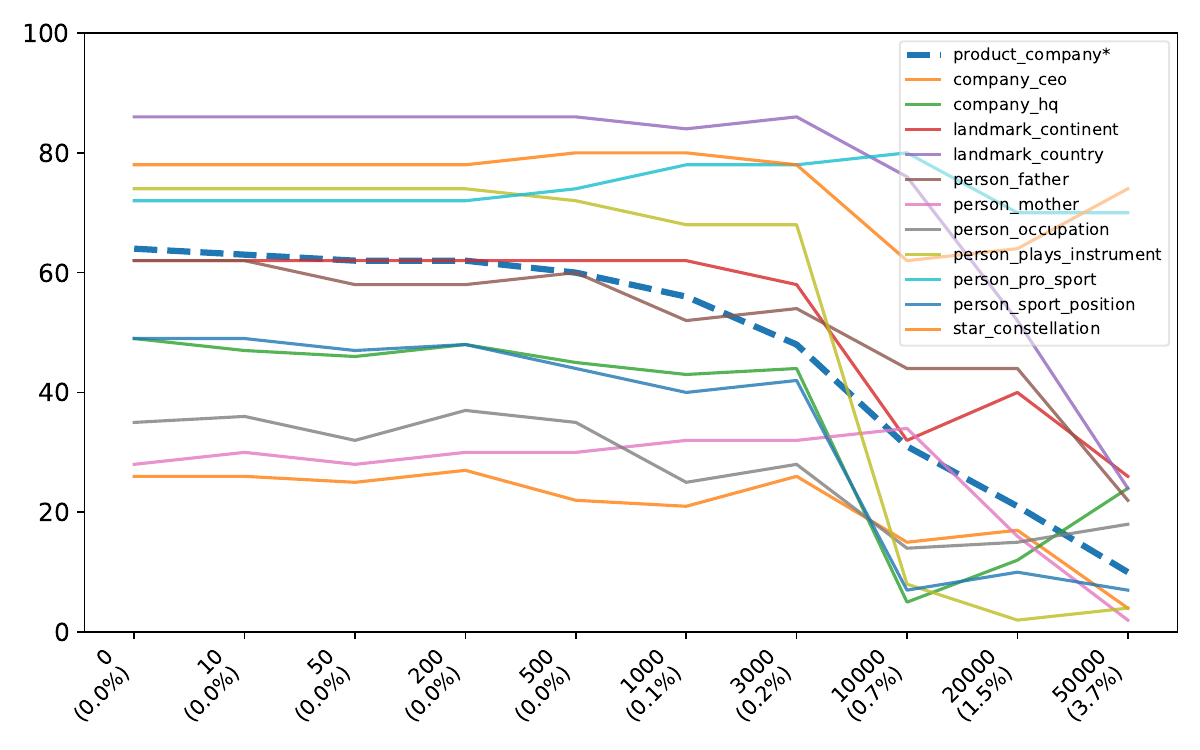}
    \includegraphics[width=0.32\textwidth]{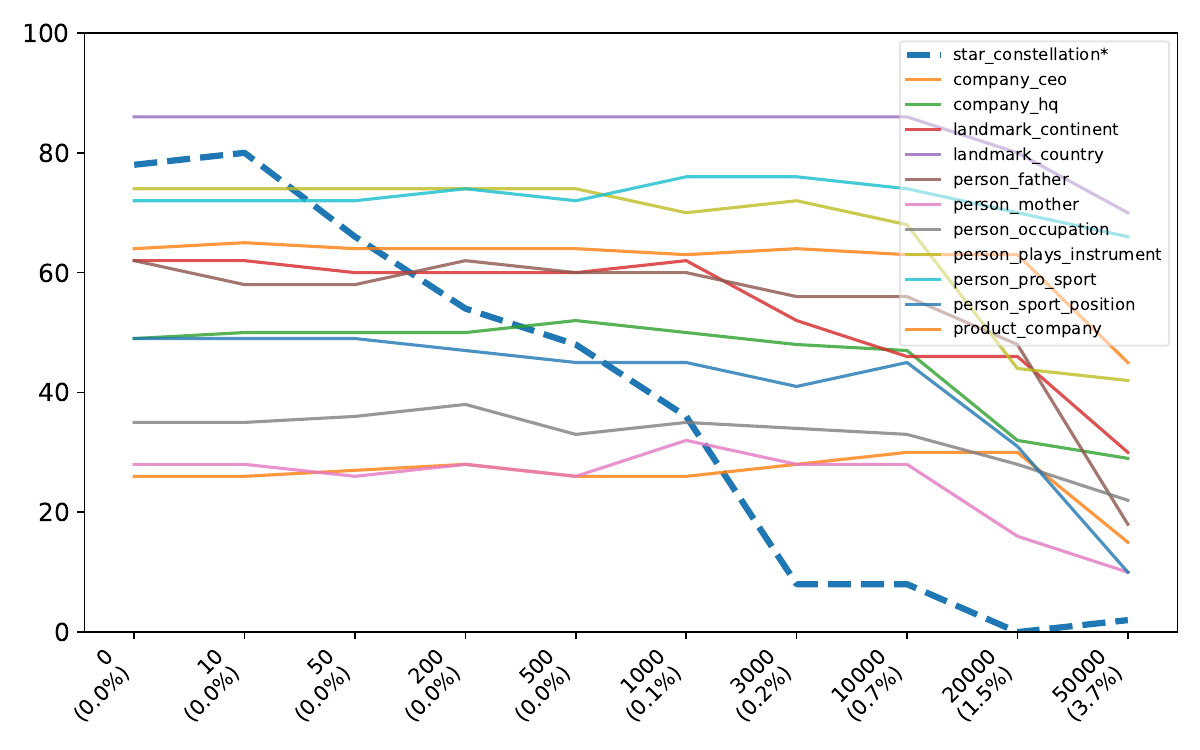}
    \caption{Influence of deactivating different numbers of
    \RelationSpecificNeurons in the \textbf{7B} model for each
    relation. The variation of accuracy on the relation
    itself (noted with ``*'' and a dashed line style) and
    the accuracy on all other relations is shown in each
    figure. Similar to Figure \ref{fig:neuron_num},
    increasing the number of neurons clearly affects the
    relation itself, but the effect on other individual
    relations does not become clearly noticeable until 3,000--10,000 neurons.}
    \label{fig:neuron_num_all}
\end{figure*}

\begin{table*}
\setlength{\belowcaptionskip}{-0.5cm}
\tiny
\centering
\setlength{\tabcolsep}{1.0mm}
\begin{tabular}{l|rr|rr|rr|rr|rr|rr|rr|rr}
\toprule
\textbf{Relation} & \multicolumn{2}{c|}{\textbf{10--50}} & \multicolumn{2}{c|}{\textbf{50--200}} & \multicolumn{2}{c|}{\textbf{200--500}} & \multicolumn{2}{c|}{\textbf{500--1000}} & \multicolumn{2}{c|}{\textbf{1000--3000}} & \multicolumn{2}{c|}{\textbf{3000--10000}} & \multicolumn{2}{c|}{\textbf{10000--20000}} & \multicolumn{2}{c}{\textbf{20000--50000}} \\
         & \#total & \#affected & \#total & \#affected & \#total & \#affected & \#total & \#affected & \#total & \#affected & \#total & \#affected & \#total & \#affected & \#total & \#affected \\
\midrule
\texttt{company\_ceo} & 1 & 0 & 3 & 2 & 5 & 0 & 7 & 2 & 11 & 2 & 3 & 3 & 2 & 0 & 0 & 0 \\
\texttt{company\_hq} & 5 & 5 & 2 & 1 & 2 & 0 & 16 & 5 & 5 & 0 & 9 & 2 & 21 & 16 & 1 & 1 \\
\texttt{landmark\_continent} & 0 & 0 & 4 & 4 & 4 & 2 & 5 & 0 & 6 & 2 & 13 & 6 & 0 & 0 & 0 & 0 \\
\texttt{landmark\_country} & 0 & 0 & 1 & 1 & 6 & 0 & 2 & 0 & 6 & 0 & 26 & 5 & 3 & 0 & 2 & 0 \\
\texttt{person\_father} & 3 & 1 & 2 & 0 & 0 & 0 & 5 & 0 & 6 & 2 & 6 & 0 & 2 & 1 & 6 & 4 \\
\texttt{person\_mother} & 4 & 3 & 1 & 0 & 4 & 3 & 0 & 0 & 7 & 5 & 4 & 1 & 2 & 1 & 1 & 1 \\
\texttt{person\_occupation} & 3 & 3 & 9 & 6 & 2 & 0 & 2 & 1 & 8 & 1 & 7 & 5 & 6 & 2 & 18 & 6 \\
\texttt{person\_plays\_instrument} & 13 & 11 & 7 & 2 & 5 & 0 & 8 & 0 & 3 & 0 & 6 & 0 & 0 & 0 & 0 & 0 \\
\texttt{person\_pro\_sport} & 0 & 0 & 2 & 1 & 4 & 1 & 9 & 0 & 8 & 0 & 16 & 0 & 1 & 0 & 0 & 0 \\
\texttt{person\_sport\_position} & 7 & 2 & 4 & 0 & 12 & 4 & 4 & 2 & 20 & 11 & 6 & 0 & 0 & 0 & 0 & 0 \\
\texttt{product\_company} & 1 & 0 & 0 & 0 & 2 & 0 & 4 & 2 & 9 & 2 & 20 & 5 & 10 & 2 & 12 & 7 \\
\texttt{star\_constellation} & 8 & 7 & 6 & 2 & 3 & 0 & 6 & 1 & 14 & 0 & 1 & 0 & 4 & 0 & 0 & 0 \\

\bottomrule
\end{tabular}
\caption{Cumulative effect validation. For each neuron
deactivation range, e.g., 1000-3000, the number of prompts
where the model answers correctly in the smaller (1000) but
not the larger range (3000) is denoted as column \#total,
and the number of prompts out of \#total that are also
affected, i.e., being answered wrongly, when deactivating
the intermediate difference (2000 = 3000 - 1000) is denoted
as \#affected. \#affected is usually much smaller
than \#total, indicating
that neurons mostly act in a cumulative way and have no
strong effect in isolation.}
\label{tab:cumulative_effect}
\end{table*}

\section{Fact Frequencies vs. Neuron Cumulativity}\seclabel{frequency}

\begin{figure}
    \centering
    \setlength{\abovecaptionskip}{-0.1cm}
    \setlength{\belowcaptionskip}{-0.5cm}
\includegraphics[width=0.46\textwidth]{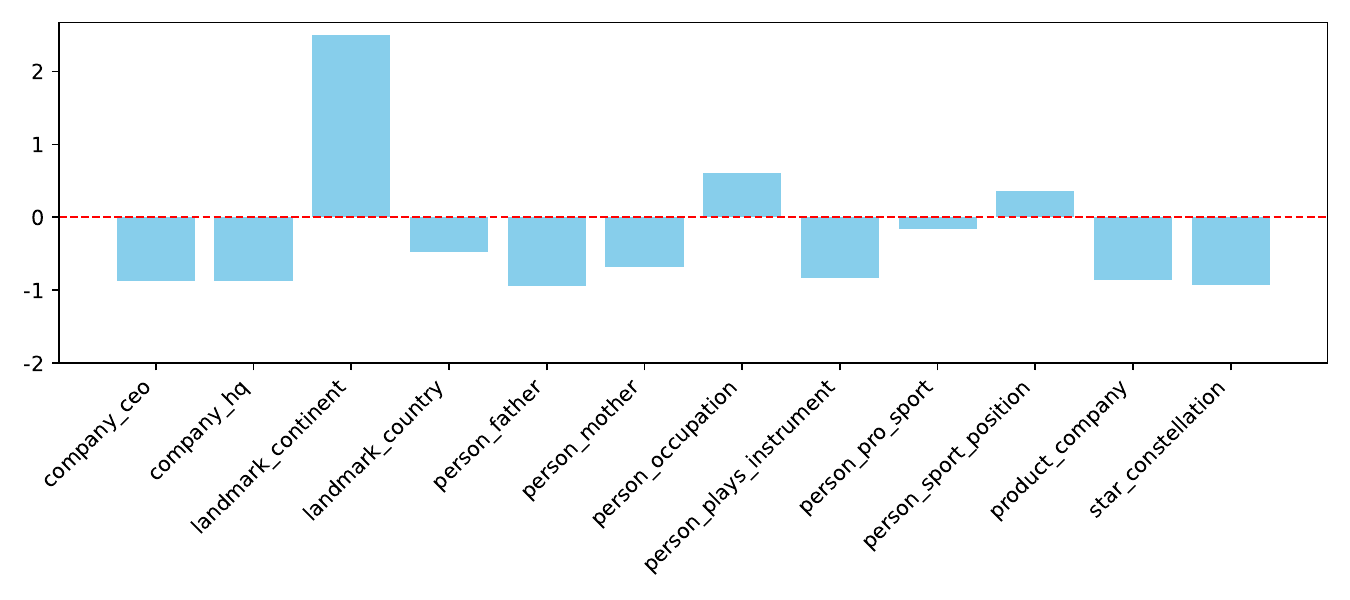}
\includegraphics[width=0.46\textwidth]{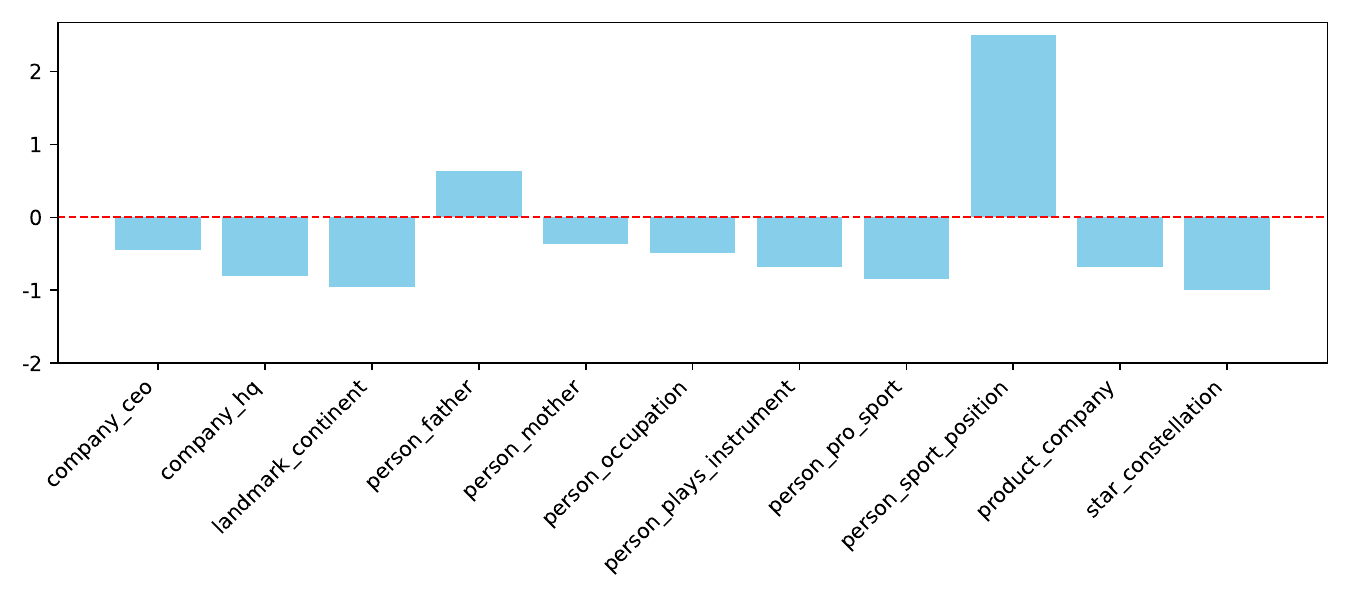}
    \caption{Relative difference between the average fact
    frequencies of the group (a) \emph{resilient facts} and (b) \emph{sensitive facts} for each relation
    in 7B (top) and 13B (bottom) models. Resilient facts generally appear more often than sensitive facts in most relations in the pertaining data.}
    \label{fig:relative_diff}
\end{figure}

We now examine our \textbf{neuron cumulativity} hypothesis by asking: \emph{why do some facts show higher sensitivity to a given set of relation neurons than others?}
We hypothesize that the frequency of a fact in the pretraining data can be a key factor, as more frequent facts may be memorized more robustly and thus remain less sensitive to deactivation.

Because the pretraining data for Llama 2 is not publicly available, we approximate it using Dolma \citep{soldaini-etal-2024-dolma}, a 3 trillion-token open-source corpus.
For each relation, we split the facts into two groups: (\textbf{a}) \emph{resilient facts}, for which the 7B (or 13B) model correctly predicts the object \textbf{both before and after} deactivating 3,000 \RelationSpecificNeurons. (\textbf{b}) \emph{sensitive facts}, for which the model is correct \textbf{before but not after} these neurons are deactivated.\footnote{We do not consider other numbers 
of 
\RelationSpecificNeurons 
because (1) if \#neurons < 3,000, there are not enough facts whose predictions change, and (2) if \#neurons > 3,000, facts belonging to other relations will also be influenced a lot.}
We then count how many documents in Dolma contain \textbf{both the subject and object} of each fact, calling this the \emph{fact frequency}.\footnote{We use ElasticSearch API from WIMBD \citep{elazar2024wimbd}
that allows for 
counting 
and searching 
in large 
corpora. 
}
Finally, we compute the average frequency for resilient and sensitive facts in each relation $r_i$, denoted respectively as  $\text{group}^{(\text{a})}_{r_i}$ and $\text{group}^{(\text{b})}_{r_i}$.

Relative difference: $\text{diff}_{r_i} = \frac{\text{group}^{(\text{b})}_{r_i} - \text{group}^{(\text{a})}_{r_i}}{\text{group}^{(\text{b})}_{r_i}}$ for each relation $r_i$ is reported in Figure \ref{fig:relative_diff}.
We find that resilient facts generally appear more often in Dolma than sensitive facts, with only 3 exceptions in the 7B model and 2 exceptions in the 13B model (note that \texttt{landmark\_country} is omitted for the 13B model because no facts fall into group~(\textbf{a})).
We evaluate this difference with the Wilcoxon Signed-Rank Test \citep{woolson2005wilcoxon} and obtain $p$-values of respectively 0.11 and 0.03 for the 7B and the 13B models.\footnote{We use a nonparametric test because the difference 
across relations does not follow a Gaussian distribution.} These results show that there is a difference (statistically significant in the 13B model at the 5\% level) between the two groups, supporting our hypothesis that
\textbf{more frequent facts are generally less sensitive to the deactivation of a given set of \RelationSpecificNeurons}.
\newpage

\section{Translation Process}\seclabel{translation}

We take a \textbf{two-step} approach to ensure the translation quality of individual prompts from English into the target languages across relations. 

\paragraph{Translating subject-object pairs.} 
The first step concerns mapping entities, i.e., subject and object pairs, into the target language. The default way of doing this is by identifying if the entity is available in Wikidata and the target language using the Wikidata API.\footnote{\url{https://www.wikidata.org/w/api.php}}
If the entity of interest is available in the target language, we directly take the entity name in that language.
If the entity is not available, we then resort to Google Translate to translate the entity from English to the target language.\footnote{\url{https://translation.googleapis.com/language/translate/v2}}. By performing this step, we obtain the subject-object pairs in all target languages and all relations.

\paragraph{Translating prompt templates.} 
We take the prompt templates of different relations written in English and use Google Translate to translate them into target languages. We then investigate how the LLama-2 7B model performs on these prompts using $\mathcal{P}_{r_i}^{\text{eva}}$ in the target languages. If the model performs suboptimally (<30\% accuracy) for a relation in a specific language, then we manually check the prompt template in that language and update the template accordingly until satisfactory accuracy (>30\%) is achieved. For Chinese and Japanese, we do not ensure more than 30\% accuracy because the models perform very badly for some relations, even if we have tried many prompt templates.


\section{Influence of Neuron Type}\seclabel{neuron_type}

\begin{figure}
    \centering
    \includegraphics[width=0.46\textwidth]{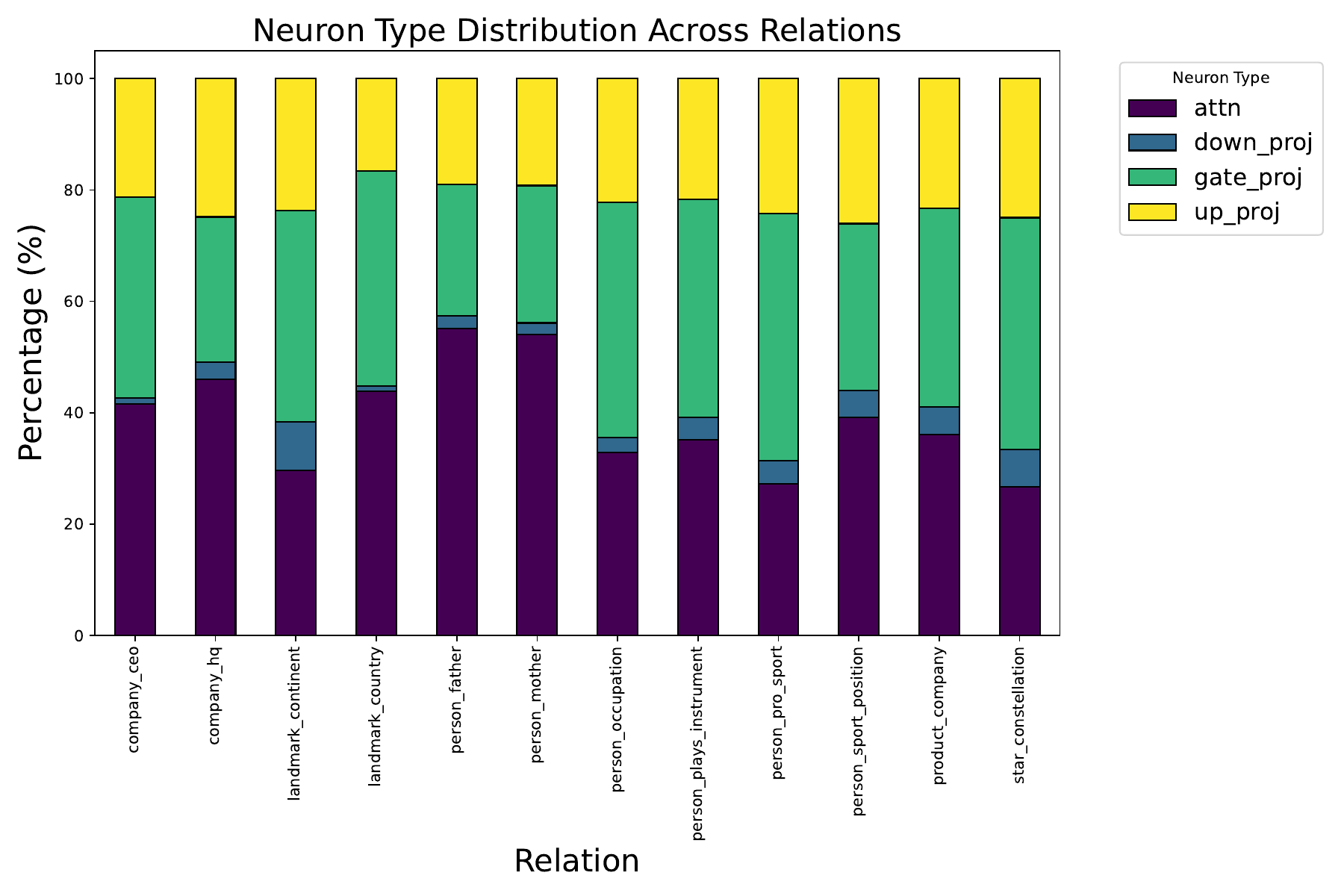}
    \caption{The distribution of the neuron types in the identified 3,000 neurons for the variety \textbf{\texttt{all}} across all relations.}
\label{fig:neuron_type_dist}
\end{figure}

\begin{figure}
    \centering
    \setlength{\abovecaptionskip}{-0.1cm}
    \includegraphics[width=0.48\textwidth]{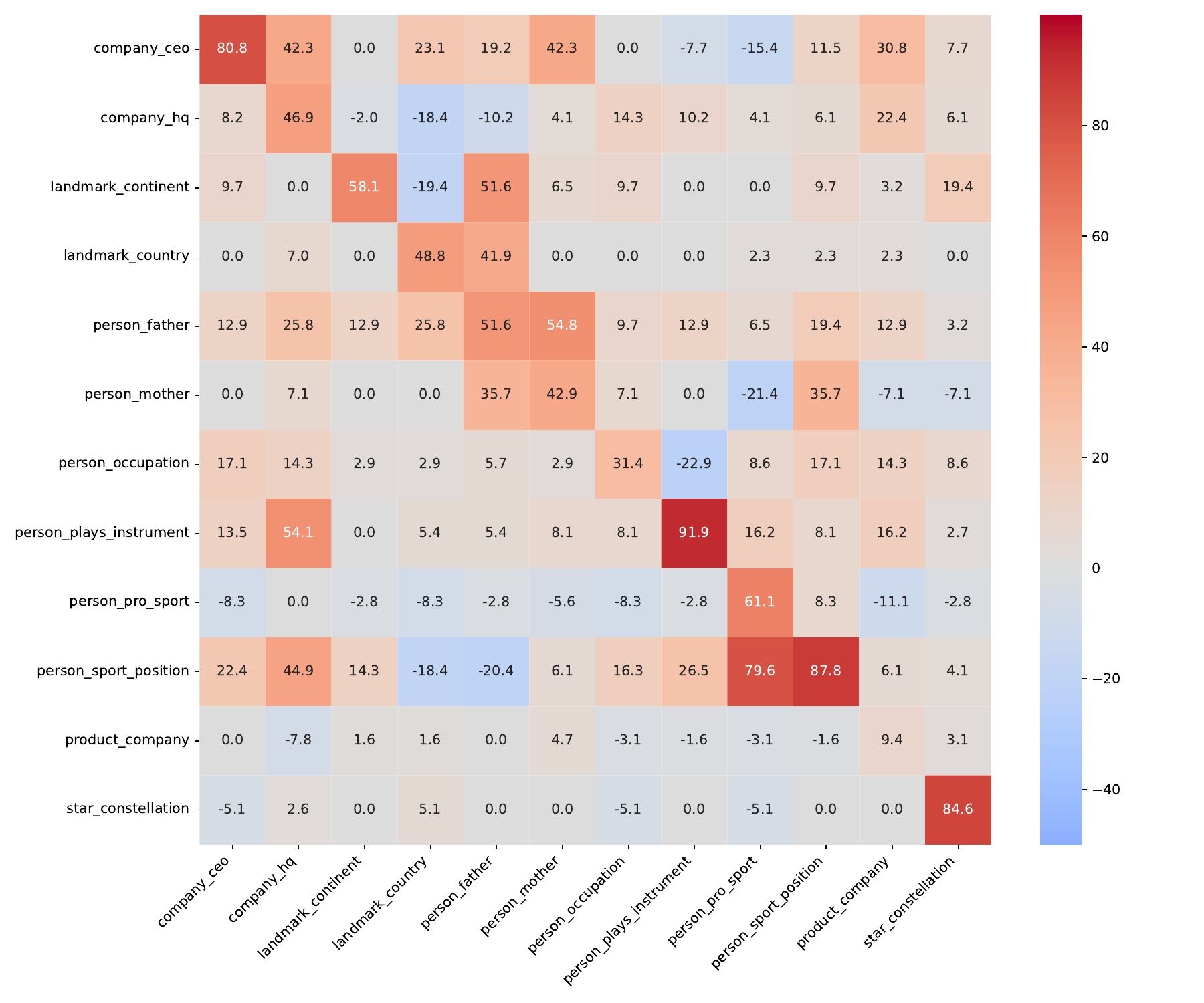}
    \caption{Inter-relation results of the 7B model when considering the neuron type variety as \textbf{\texttt{all}}.}
    \label{fig:neuron_type_all}
\end{figure}

\begin{figure}
    \centering
    \setlength{\abovecaptionskip}{-0.1cm}
    \includegraphics[width=0.48\textwidth]{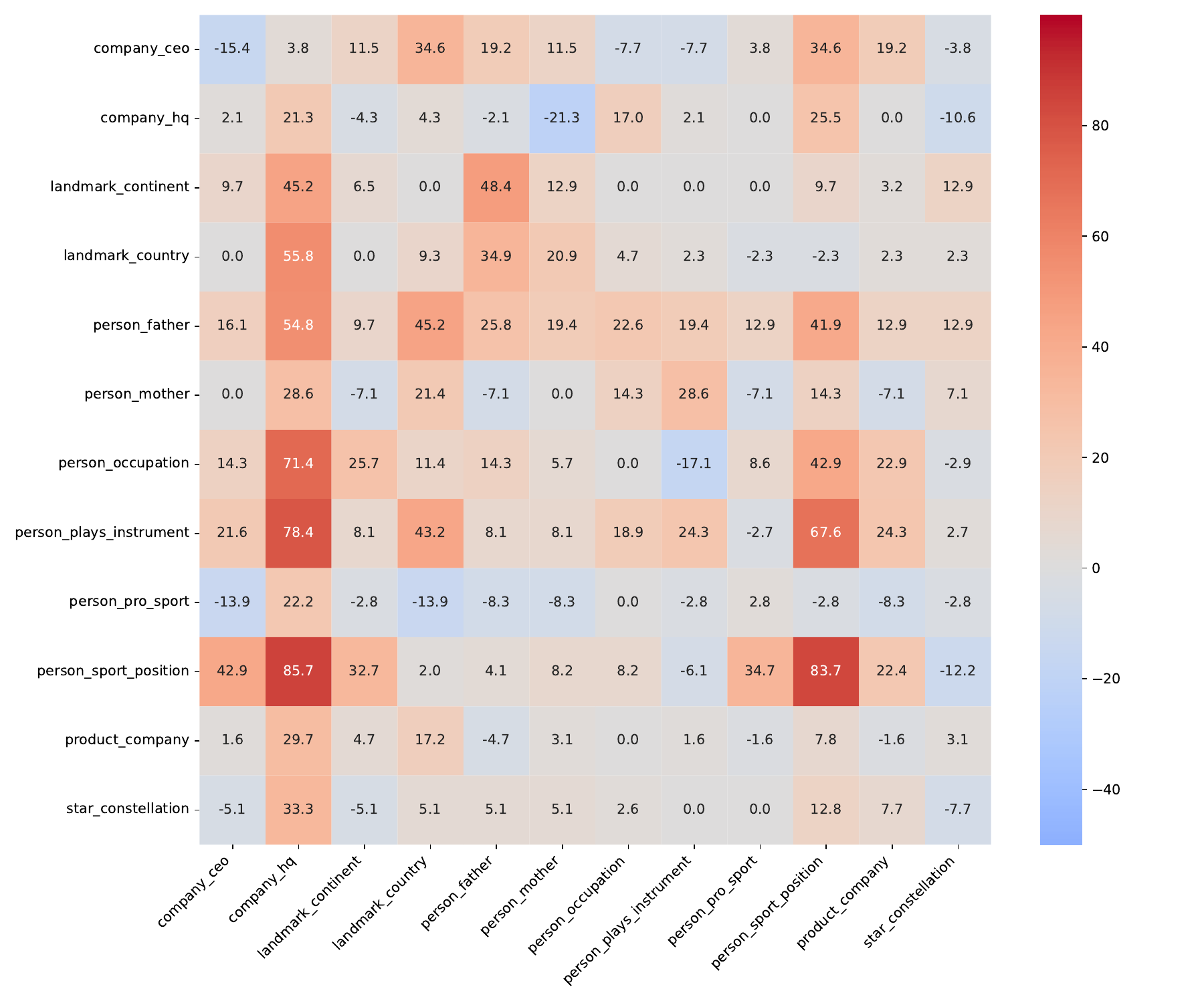}
    \caption{Inter-relation results of the 7B model when considering the neuron type variety as \textbf{\texttt{self\_attn}}.}
    \label{fig:neuron_type_self_attn}
\end{figure}

\begin{figure}
    \centering
    \setlength{\abovecaptionskip}{-0.1cm}
    \includegraphics[width=0.48\textwidth]{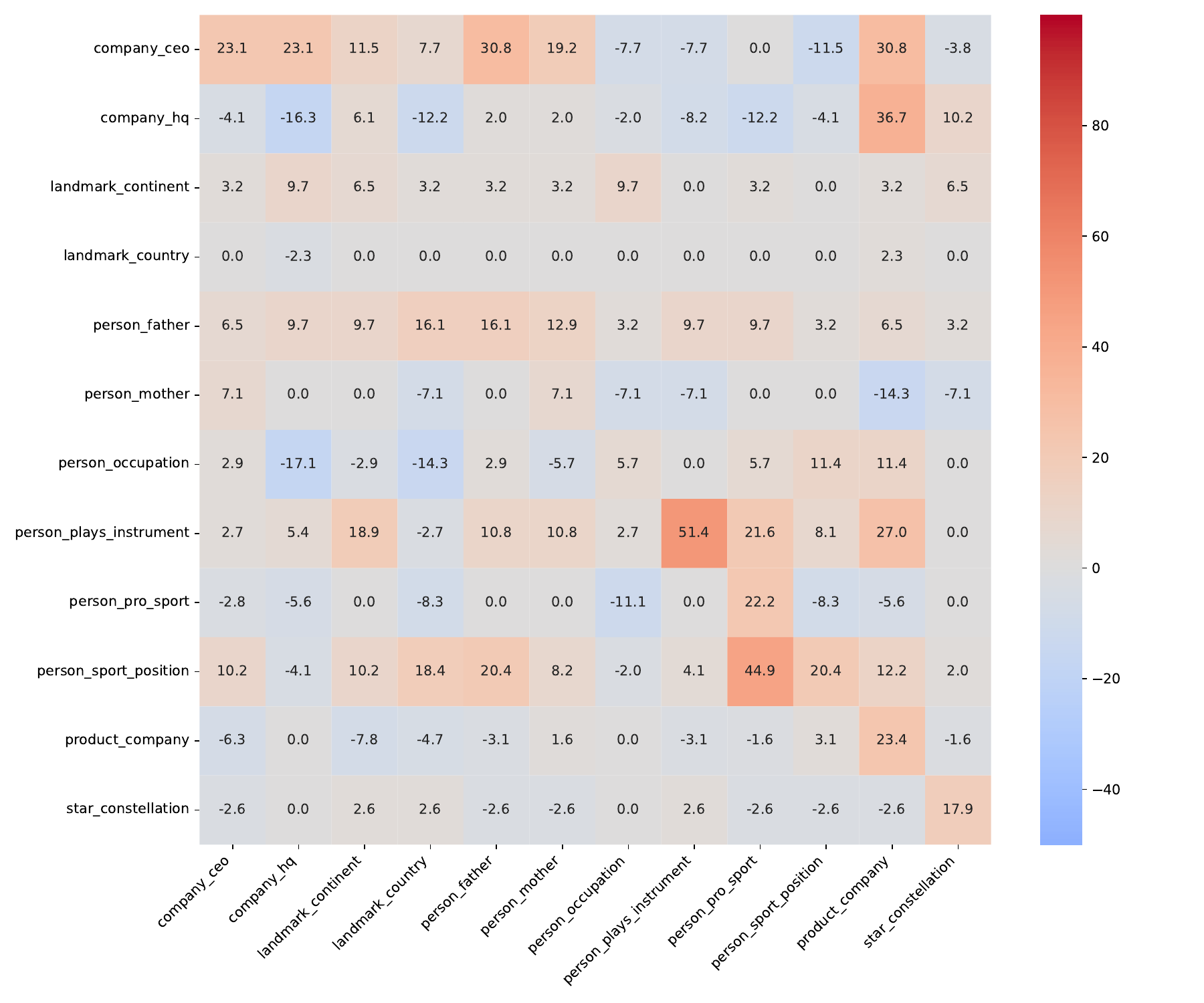}
    \caption{Inter-relation results of the 7B model when considering the neuron type variety as \textbf{\texttt{up\_proj}}.}
    \label{fig:neuron_type_up_proj}
\end{figure}

\begin{figure}
    \centering
    \setlength{\abovecaptionskip}{-0.1cm}
    \includegraphics[width=0.48\textwidth]{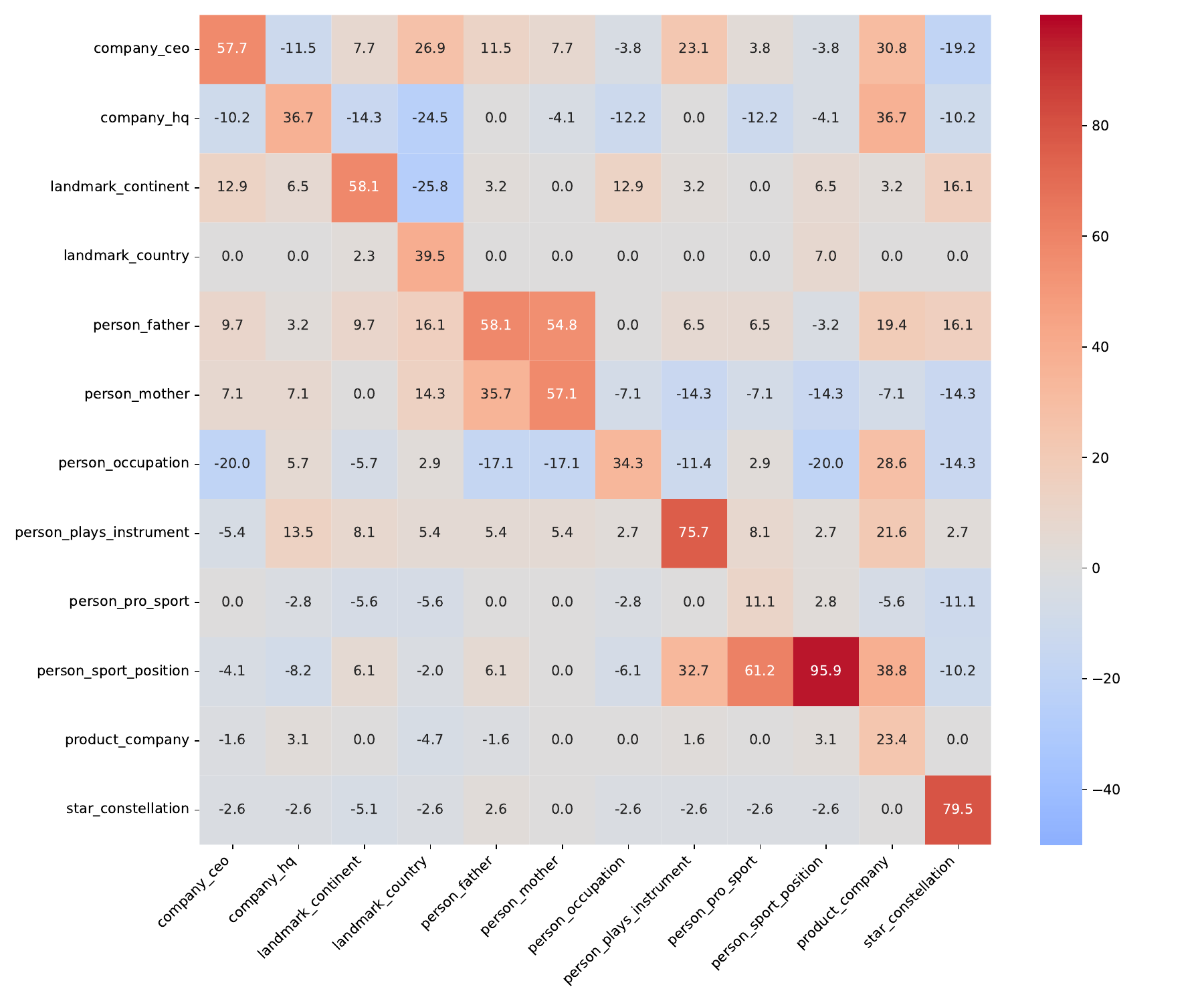}
    \caption{Inter-relation results of the 7B model when considering the neuron type variety as \textbf{\texttt{gate\_proj}}.}
    \label{fig:neuron_type_gate_proj}
\end{figure}

\begin{figure}
    \centering
    \setlength{\abovecaptionskip}{-0.1cm}
    \includegraphics[width=0.48\textwidth]{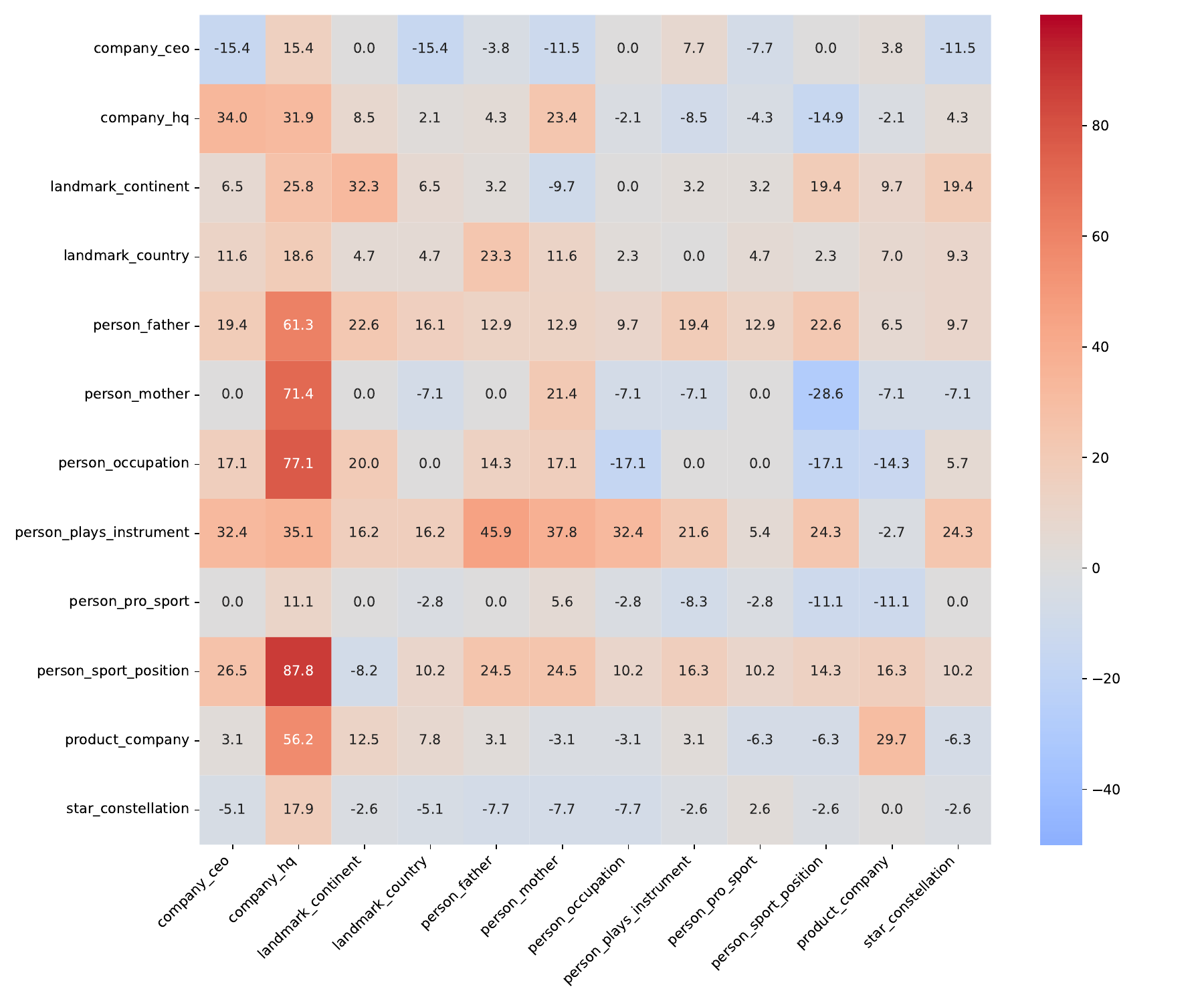}
    \caption{Inter-relation results of the 7B model when considering the neuron type variety as \textbf{\texttt{down\_proj}}.}
    \label{fig:neuron_type_down_proj}
\end{figure}

We consider the neurons in the FFNs (including up\_proj, gate\_proj, and down\_proj matrices) as our major setup. In this section, we explore the individual effects of different types of neurons. Specifically,
we consider five additional different varieties when selecting the top 3,000 neurons for the 7B model: \textbf{\texttt{all}} (neurons in any matrices), \textbf{\texttt{self\_attn}} (neurons in self-attention matrices), \textbf{\texttt{up\_proj}} (neurons in up\_proj matrices), \textbf{\texttt{gate\_proj}} (neurons in gate\_proj matrices), \textbf{\texttt{down\_proj}} (neurons in down\_proj matrices). We first draw the distribution of the neuron types across relations for variety \textbf{\texttt{all}} in Figure \ref{fig:neuron_type_dist} and report the inter-relation results in Figure \ref{fig:neuron_type_all} (\textbf{\texttt{all}}), \ref{fig:neuron_type_self_attn} (\textbf{\texttt{self\_attn}}), \ref{fig:neuron_type_up_proj} (\textbf{\texttt{up\_proj}}), \ref{fig:neuron_type_gate_proj} (\textbf{\texttt{gate\_proj}}), and \ref{fig:neuron_type_down_proj} (\textbf{\texttt{down\_proj}}).

According to the results, we observe that simply considering \textbf{\texttt{self\_attn}} does not offer a consistent accuracy drop for the relation itself (by looking at the diagonal: some relations are not influenced too much). This can be explained by the fact that \textbf{\texttt{self\_attn}} is shared across relations (as shown by \citet{elhelo2024inferring}), and facts are mainly stored in the FFNs. Only considering \textbf{\texttt{down\_proj}} offer similar results as \textbf{\texttt{self\_attn}}. Interestingly, deactivating \textbf{\texttt{up\_proj}} neurons does not influence all relations much in general, indicating it does not make sense to consider \textbf{\texttt{up\_proj}} alone. Considering \textbf{\texttt{all}} or \textbf{\texttt{gate\_proj}} neurons offer similar results compared to considering neurons in FFNs (shown in Figure \ref{fig:intra-relation}). However, by considering neurons in FFNs (i.e., \textbf{\texttt{up\_proj}}, \textbf{\texttt{gate\_proj}} and \textbf{\texttt{down\_proj}}), we see a more obvious inter-relation accuracy drop as shown on the diagonal in Figure \ref{fig:intra-relation}. Therefore, our additional analysis supports our choice of considering neurons in FFNs.

\section{Concept-Specific Neurons}
\begin{figure}
    \centering
    \setlength{\abovecaptionskip}{-0.1cm}
    \setlength{\belowcaptionskip}{-0.4cm}
\includegraphics[width=0.46\textwidth]{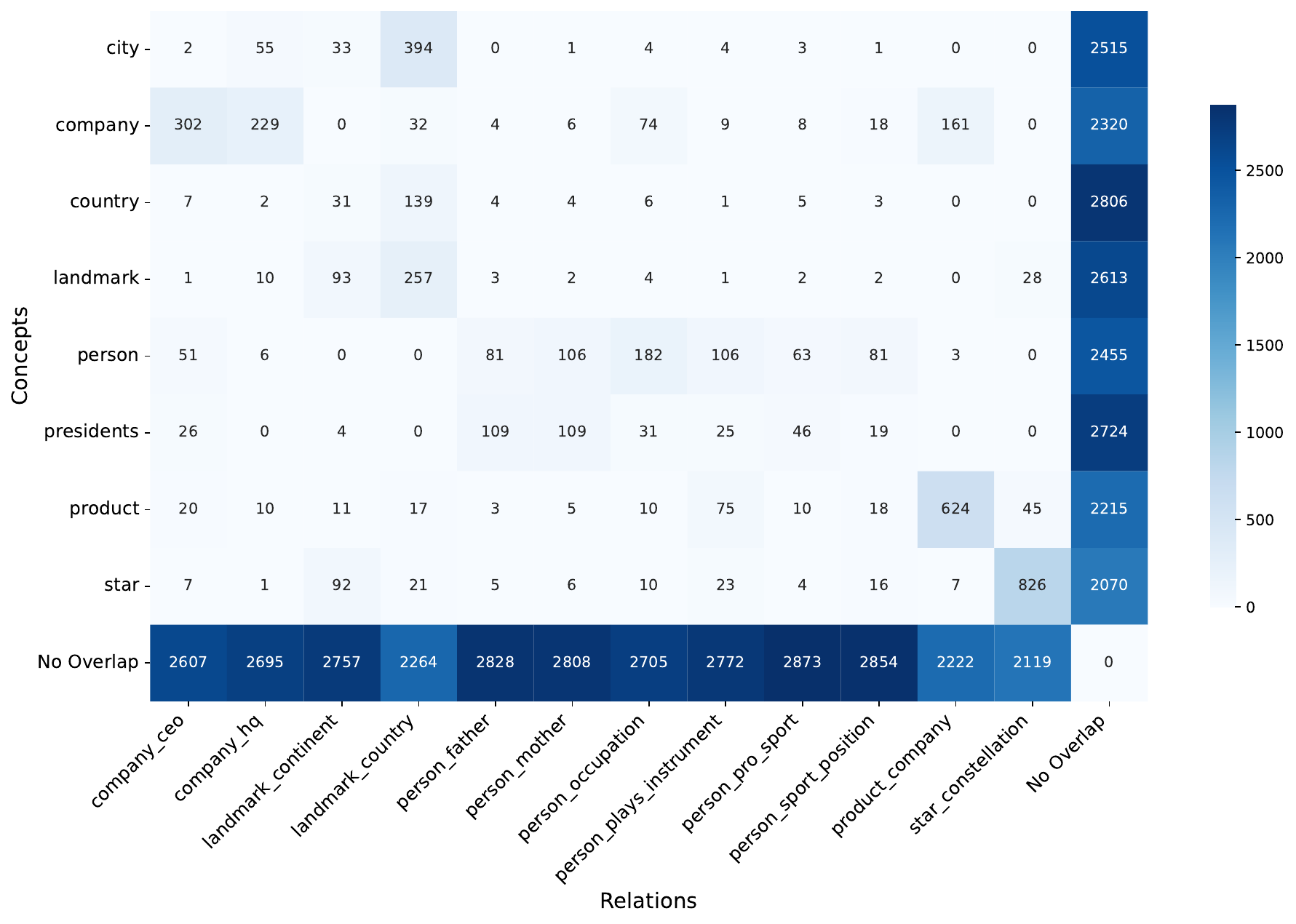}
    \caption{Overlap between the top 3000 identified neurons for each relation and concept in the 7B model. }
    \label{fig:concept_neuron_overlap}
\end{figure}

\seclabel{concepts}
\paragraph{Concept-Relation Overlap in the 7B Model}
Figure \ref{fig:concept_neuron_overlap} illustrates the overlap between individual relation- and concept-specific neurons in the 7b model. 
There, the overlap of concepts connected to the abstract notion of ``location'' and the relations are mostly concentrated on the \texttt{landmark\_country} relation in comparison to the 13b model, where they are spread over \texttt{company\_hq}, \texttt{landmark\_continent} and \texttt{landmark\_country}. 
This aligns with the difference between the 7B and 13B models in terms of their patterns of inter-relation results (cf. Figure \ref{fig:inter-relation}): deactivating the \texttt{landmark\_country} neurons results in a significant accuracy drop in other relations concerning ``location'' in the 13B model while not in the 7B model.
Another difference between both models is that there is more distributed neuron overlap in the 7b model between the subject concept \texttt{person} and all corresponding relations.

\paragraph{Validation of Concept-Specific Neurons}
The top neurons on a concept are evaluated on a random selection of 100 prompts from the LRE dataset that include the specified concept as a subject. Examples for the concept \texttt{person} are "Tom Hanks's father is named? Answer:", "Hilary Hahn plays the instrument of? Answer:", or "Thomas Mann went to university at? Answer:".

\begin{figure}
    \centering
    \setlength{\abovecaptionskip}{-0.1cm}
    \setlength{\belowcaptionskip}{-0.4cm}
\includegraphics[width=0.46\textwidth]{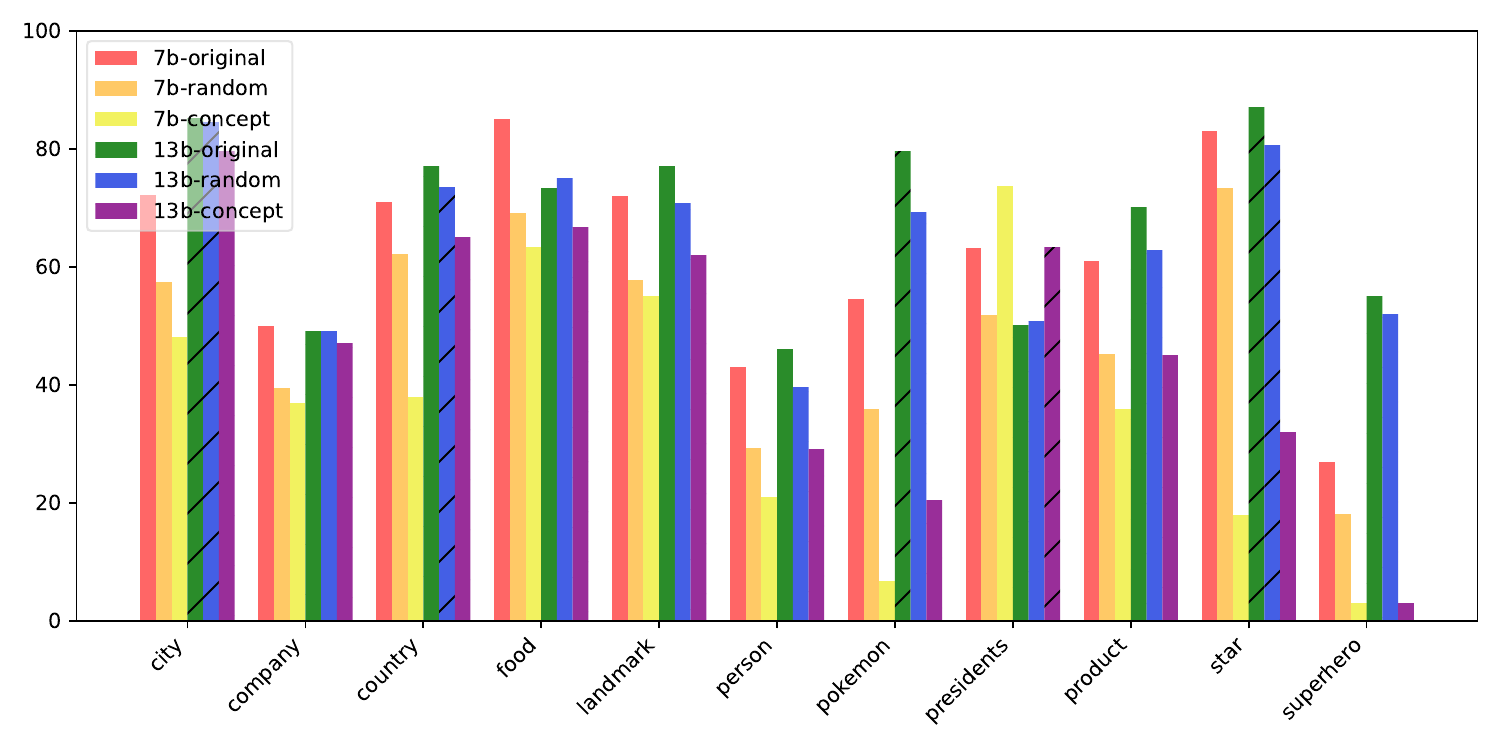}
    \caption{Accuracy results of evaluation prompts for 11 concepts in the 7b and 13b model. We report the performance of the original model (without any deactivation), e.g., \texttt{7b-original}, the model with 3000 randomly deactivated neurons, e.g., \texttt{7b-random}, and the model with deactivating the top 3000 identified concept-specific neurons, e.g., \texttt{7b-concept}.}
    \label{fig:concept_neuron_validation}
\end{figure}

Figure \ref{fig:concept_neuron_validation} shows the results for the validation on these validation prompts for both models with the original accuracy score, a baseline that ablates 3000 neurons randomly, and the ablation of 3000 concept-specific neurons. Note that the impact of ablating a certain amount of expert neurons varies between concepts. The observed drop in performance due to the ablation of 3000 neurons for concepts like \texttt{pokemon}, \texttt{superhero}, and \texttt{star} is very large, 
while accuracy scores of other concepts in the 13b model, such as \texttt{person} appear stable, or even improve, e.g., \texttt{presidents}. 
We assume the \textbf{neuron cumulativity} also applies to the concept-specific neurons. 
That is, the knowledge on a specific concept is distributed over a much larger population of neurons, and further accuracy drop can be observed once more concept-specific neurons are deactivated -- similar to what we observe for \RelationSpecificNeurons (cf. Figure \ref{fig:neuron_num}). 
As only partial knowledge is withheld from the deactivation of 3000 concept-specific neurons, 
this might be too little knowledge to affect the facts concerning that concept (substantial knowledge on the concept is stored in the remaining neurons), resulting in only a small accuracy drop.
Or, the 3000 concept-specific neurons store knowledge, though concerning the concept, unrelated to the prompts.
For instance, the validation prompts of the concept \texttt{presidents} all demand \textbf{historical dates} as predicted answers, which is only one kind of knowledge that might be expected in connection with presidents.
This phenomenon actually aligns with our neuron interference hypothesis: deactivating neurons that store unhelpful knowledge can less confuse the model, therefore improving the performance.

\section{Experimental Environment}\seclabel{env}

We run all experiments on NVIDIA RTX A6000 GPUs. The Python environment we use is the same as \citet{kojima-etal-2024-multilingual}.\footnote{\citet{kojima-etal-2024-multilingual}'s GitHub repository is available at \url{https://github.com/kojima-takeshi188/lang_neuron}} 

\section{Error Analysis}\seclabel{error_analysis}

We manually verified the prompts in each relation that the model could answer correctly originally, but failed to answer correctly when 3,000 \RelationSpecificNeurons were deactivated (cf.\ \secref{generation}). 
The three most common incorrect responses (regarded as \emph{systematic errors}) are listed in Table~\ref{tab:error_freq}.

\begin{table*}
    \centering
    \small
    \begin{tabular}{lrrrr}
        \toprule
         \textbf{Relation} & \textbf{Repeat Prompt} & \textbf{Answer with ``\emph{The}''} & \textbf{Answer with ``\emph{A.}''} & \textbf{Total Number}\\
         \midrule
         
         \texttt{company\_ceo} & \textbf{47.8\%} & 8.7\% & 34.8\% & 23\\
         \texttt{company\_hq} & \textbf{46.2\%} & \textbf{46.2\%} & 0\% & 26\\
         \texttt{landmark\_continent}& \textbf{17.6\%} & 5.9\% & 0\% & 17\\
         \texttt{landmark\_country}& \textbf{69.3\%} & 0\% & 0\% & 13\\
         \texttt{person\_father}& \textbf{84.2\%} & 5.3\% & 0\% & 19\\
         \texttt{person\_mother}& \textbf{70\%} & 20\% & 0\% & 10\\
         \texttt{person\_occupation}& \textbf{93.3\%} & 0\% & 0\% & 15\\
         \texttt{person\_plays\_instrument}& \textbf{51.6\%} & 29\% & 0\% & 31\\
         \texttt{person\_pro\_sport}& \textbf{25\%} & 15\% & 0\% & 20\\
         \texttt{person\_sport\_position}& 18.6\% & 11.6\% & \textbf{44.2\%} & 43\\
         \texttt{product\_company}& \textbf{70.1\%} & 11.8\% & 0\% & 17\\
         \texttt{star\_constellation}& \textbf{88.6\%} & 5.7\% & 0\% & 35\\ 
         \bottomrule
    \end{tabular}
    \caption{Most common incorrect answers generated by LLama-7b after deactivating 3,000 \RelationSpecificNeurons.}
    \label{tab:error_freq}
\end{table*}

After we deactivate the \RelationSpecificNeurons, we can see that the model appears to lose its ability to recall the correct object.
Instead, the model frequently answers with meaningless answers that start with tokens such as ``\emph{A.}'' or ``\emph{The}'', or simply repeats the given prompt.
We showcase representative examples of each phenomenon in Table~\ref{tab:repeat_prompt}, Table~\ref{tab:answer_the}, and Table~\ref{tab:answer_a}.
The results strongly indicate that the model loses its ability to capture relational semantics, resulting in increasingly noisy outputs after the deactivation of \RelationSpecificNeurons.

\begin{table*}[htbp]
    \centering
    \resizebox{\textwidth}{!}{
    \small
    \renewcommand{\arraystretch}{1.2}
    \setlength{\tabcolsep}{6pt}
    \rowcolors{2}{white}{gray!8}
    \begin{tabular}{
        >{\raggedright\arraybackslash}p{0.23\linewidth}
        >{\centering\arraybackslash}p{0.22\linewidth}
        >{\centering\arraybackslash}p{0.15\linewidth}
        >{\raggedright\arraybackslash}p{0.2\linewidth}
        >{\centering\arraybackslash}p{0.1\linewidth}
    }
    \toprule
    \rowcolor{gray!15}
    \textbf{\textcolor{forestgreen}{Subject}-\textcolor{darkmagenta}{Object} Pair} &
    \textbf{Prompt} &
    \textbf{Expected Output} &
    \textbf{Model Response} &
    \textbf{Deactivation} \\
    \midrule
    \multirow{2}{=}{(\textcolor{forestgreen}{Panasonic Corporation}, \textcolor{darkmagenta}{Kazuhiro Tsuga})} &
    \multirow{2}{=}{\textcolor{forestgreen}{Panasonic Corporation}'s CEO is? Answer:} &
    \multirow{2}{=}{\textcolor{darkmagenta}{Kazuhiro Tsuga}} &
    Kazuhiro Tsuga & \cellcolor{red!20}\textcolor{red}{No} \\
    \cdashline{4-5}[1.2pt/1pt]
    \addlinespace[2pt]  
    & & & \textcolor{dodgerblue}{[Pan]}asonic Corporation's CEO is:\texttt{\textbackslash n}Panasonic Corporation's CEO & \cellcolor{green!20}\textcolor{yellowgreen}{Yes} \\
    \bottomrule
    \end{tabular}
    }
    \caption{Model answers by repeating the prompt after deactivating \RelationSpecificNeurons. 
    We changed the output length from \textbf{2} tokens to \textbf{20} tokens to observe the complete output.
    The part enclosed in ``[]'' is the first 2 tokens of the output.  
    The triple (\texttt{Panasonic}, \texttt{company\_ceo}, \texttt{Kazuhiro Tsuga}) is selected for demonstration.}
    \label{tab:repeat_prompt}
\end{table*}

\begin{table*}[htbp]
    \centering
    \resizebox{\textwidth}{!}{
    \small
    \renewcommand{\arraystretch}{1.2}
    \setlength{\tabcolsep}{6pt}
    \rowcolors{2}{white}{gray!8}
    \begin{tabular}{
        >{\raggedright\arraybackslash}p{0.23\linewidth}
        >{\centering\arraybackslash}p{0.22\linewidth}
        >{\centering\arraybackslash}p{0.15\linewidth}
        >{\raggedright\arraybackslash}p{0.2\linewidth}
        >{\centering\arraybackslash}p{0.1\linewidth}
    }
    \toprule
    \rowcolor{gray!15}
    \textbf{\textcolor{forestgreen}{Subject}-\textcolor{darkmagenta}{Object} Pair} &
    \textbf{Prompt} &
    \textbf{Expected Output} &
    \textbf{Model Response} &
    \textbf{Deactivation} \\
    \midrule
    \multirow{2}{=}{(\textcolor{forestgreen}{Pagan Federation}, \textcolor{darkmagenta}{London})} &
    \multirow{2}{=}{\textcolor{forestgreen}{Pagan Federation} is headquartered in the city of? Answer:} &
    \multirow{2}{=}{\textcolor{darkmagenta}{London}} &
    London & \cellcolor{red!20}\textcolor{red}{No} \\
    \cdashline{4-5}[1.2pt/1pt]
    \addlinespace[2pt]  
    & & & \textcolor{dodgerblue}{[The]} Pagan Federation is a British organisation that represents the interests of Pagans and other Ne& \cellcolor{green!20}\textcolor{yellowgreen}{Yes} \\
    \bottomrule
    \end{tabular}
    }
    \caption{Model answers with ``\emph{The}'' after deactivating \RelationSpecificNeurons. 
    We changed the output length from \textbf{2} tokens to \textbf{20} tokens to observe the complete output.
    The part enclosed in ``[]'' is the first 2 tokens of the output.  
    The triple (\texttt{Pagan Federation}, \texttt{company\_hq}, \texttt{London}) is selected for demonstration.}
    \label{tab:answer_the}
\end{table*}

\begin{table*}[htbp]
    \centering
    \resizebox{\textwidth}{!}{
    \small
    \renewcommand{\arraystretch}{1.2}
    \setlength{\tabcolsep}{6pt}
    \rowcolors{2}{white}{gray!8}
    \begin{tabular}{
        >{\raggedright\arraybackslash}p{0.23\linewidth}
        >{\centering\arraybackslash}p{0.22\linewidth}
        >{\centering\arraybackslash}p{0.15\linewidth}
        >{\raggedright\arraybackslash}p{0.2\linewidth}
        >{\centering\arraybackslash}p{0.1\linewidth}
    }
    \toprule
    \rowcolor{gray!15}
    \textbf{\textcolor{forestgreen}{Subject}-\textcolor{darkmagenta}{Object} Pair} &
    \textbf{Prompt} &
    \textbf{Expected Output} &
    \textbf{Model Response} &
    \textbf{Deactivation} \\
    \midrule
    \multirow{2}{=}{(\textcolor{forestgreen}{Damon Huard}, \textcolor{darkmagenta}{quarterback})} &
    \multirow{2}{=}{\textcolor{forestgreen}{Damon Huard} plays in the position of a? Answer:} &
    \multirow{2}{=}{\textcolor{darkmagenta}{quarterback}} &
    Quarterback & \cellcolor{red!20}\textcolor{red}{No} \\
    \cdashline{4-5}[1.2pt/1pt]
    \addlinespace[2pt]  
    & & & \textcolor{dodgerblue}{[A.]}\texttt{\textbackslash n}Damon Huard plays in the position of a?\texttt{\textbackslash n}Answer: A.& \cellcolor{green!20}\textcolor{yellowgreen}{Yes} \\
    \bottomrule
    \end{tabular}
    }
    \caption{Model answers with ``\emph{A.}'' after deactivating \RelationSpecificNeurons. 
    We changed the output length from \textbf{2} tokens to \textbf{20} tokens to observe the complete output.
    The part enclosed in ``[]'' is the first 2 tokens of the output. 
    The triple (\texttt{Damon Huard}, \texttt{person\_sport\_position}, \texttt{quarterback}) is selected for demonstration..}
    \label{tab:answer_a}
\end{table*}

\section{Prompt Templates}\seclabel{prompts}

We show the actual prompt templates (with an object-subject example) we use for each relation across 6 considered languages:
\texttt{company\_ceo} in Table \ref{tab:prompts company_ceo}, 
\texttt{company\_hq} in Table \ref{tab:prompts company_hq},
\texttt{landmark\_continent} in Table \ref{tab:prompts_landmark_continent},
\texttt{landmark\_country} in Table \ref{tab:prompts_landmark_country},
\texttt{person\_father} in Table \ref{tab:prompts_person_father},
\texttt{person\_mother} in Table \ref{tab:prompts_person_mother},
\texttt{person\_occupation} in Table \ref{tab:prompts_person_occupation},
\texttt{person\_plays\_instrument} in Table \ref{tab:prompts_person_plays_instrument},
\texttt{person\_pro\_sport} in Table \ref{tab:prompts_person_pro_sport},
\texttt{person\_sport\_position} in Table \ref{tab:prompts_person_sport_position},
\texttt{product\_company} in Table \ref{tab:prompts_product_company}, and
\texttt{star\_constellation} in Table \ref{tab:prompts_star_constellation}.

Additionally, we show the templates used for measuring the effect on the general language modeling capability before and after ablating \RelationSpecificNeurons (cf. \secref{ppl}) in Table~\ref{tab:ppl_template_sentences}.

\begin{table*}[h]
\centering
\small
\resizebox{\textwidth}{!}{
\begin{tabular}{ll}
\toprule
\textbf{Relation} & \textbf{Template (Object inserted at \texttt{<object>})} \\
\midrule
\texttt{person\_sport\_position} & The jersey was displayed in honor of the legendary player known for their role as \texttt{<object>}. \\
\texttt{person\_pro\_sport} & The documentary explored the early years of the athlete’s rise in the world of \texttt{<object>}. \\
\texttt{person\_occupation} & After years of study and dedication, they finally earned recognition as a respected \texttt{<object>}. \\
\texttt{company\_hq} & The company's annual report was mailed directly from its headquarters in \texttt{<object>}. \\
\texttt{product\_company} & Consumers often associate the iconic design of the device with the brand \texttt{<object>}. \\
\texttt{person\_mother} & The award speech concluded with heartfelt thanks to someone very special, \texttt{<object>}. \\
\texttt{person\_father} & The biography ended with a touching story about a life lesson from \texttt{<object>}. \\
\texttt{landmark\_continent} & The travel show featured stunning views from across the diverse landscapes of \texttt{<object>}. \\
\texttt{landmark\_country} & Every year, thousands of tourists visit the historic site located in \texttt{<object>}. \\
\texttt{company\_ceo} & The major shift in strategy was attributed to the leadership of \texttt{<object>}. \\
\texttt{person\_plays\_instrument} & The final movement of the piece was written specifically for the sound of a \texttt{<object>}. \\
\texttt{star\_constellation} & Ancient civilizations once navigated the seas by charting the stars in \texttt{<object>}. \\
\bottomrule
\end{tabular}
}
\caption{Templates used to construct synthetic sentences for evaluating general language modeling of object tokens in \secref{ppl}. Each sentence includes the object in a natural context unrelated to the original subject or relation.}
\label{tab:ppl_template_sentences}
\end{table*}

\begin{table*}[htbp]
\scriptsize
\centering
\setlength{\tabcolsep}{1.0mm}{}
\begin{center}
\begin{tabular}{m{0.10\linewidth} p{0.30\linewidth} p{0.30\linewidth} m{0.15\linewidth}}
\textbf{Language} & \textbf{\textcolor{forestgreen}{Subject}-\textcolor{darkmagenta}{Object} Pair} & \textbf{Prompt} & \textbf{Expected Output} \\
\toprule
\multirow{3}{*}{\centering English} & \multirow{3}{*}{\raggedright (\textcolor{forestgreen}{Panasonic Corporation}, \textcolor{darkmagenta}{Kazuhiro Tsuga})} 
& \textcolor{forestgreen}{Panasonic Corporation}'s CEO is? Answer: & \multirow{3}{*}{\centering \textcolor{darkmagenta}{Kazuhiro Tsuga}} \\
& & The CEO of \textcolor{forestgreen}{Panasonic Corporation} is? Answer: & \\
\cdashline{3-3}[1.2pt/1pt]
\addlinespace[2pt]  
& & The name of the CEO of \textcolor{forestgreen}{Panasonic Corporation} is? Answer: & \\
& & Who is the CEO of \textcolor{forestgreen}{Panasonic Corporation}? Their name is? Answer: & \\
\midrule
\multirow{2}{*}{\centering German} & \multirow{2}{*}{\raggedright (\textcolor{forestgreen}{Panasonic}, \textcolor{darkmagenta}{Kazuhiro Tsuga})} & Der Name des CEO von \textcolor{forestgreen}{Panasonic} lautet & \multirow{2}{*}{\centering \textcolor{darkmagenta}{Kazuhiro Tsuga}} \\
& & Wer ist der CEO von \textcolor{forestgreen}{Panasonic}? Ihr Name ist & \\
\midrule
\multirow{2}{*}{\centering Spanish} & \multirow{2}{*}{\raggedright (\textcolor{forestgreen}{Panasonic}, \textcolor{darkmagenta}{Kazuhiro Tsuga})} & Por favor, responda directamente por su nombre. El nombre del director general de \textcolor{forestgreen}{Panasonic} es& \multirow{2}{*}{\centering \textcolor{darkmagenta}{Kazuhiro Tsuga}} \\
& & Por favor, responda directamente por su nombre. ¿Quién es el director general de \textcolor{forestgreen}{Panasonic}? Su nombre es & \\
\midrule
\multirow{2}{*}{\centering French} & \multirow{2}{*}{(\raggedright \textcolor{forestgreen}{Panasonic}, \textcolor{darkmagenta}{Kazuhiro Tsuga})} & Veuillez répondre directement avec le nom. Le nom du président-directeur général de \textcolor{forestgreen}{Panasonic} est& \multirow{2}{*}{\centering\textcolor{darkmagenta}{Kazuhiro Tsuga}} \\
& & Veuillez répondre directement avec le nom. Le PDG de \textcolor{forestgreen}{Panasonic} est nommé& \\
\midrule
\multirow{2}{*}{\centering Japanese} & \multirow{2}{*}{(\raggedright \textcolor{forestgreen}{\begin{CJK}{UTF8}{min}パナソニック株式会社\end{CJK}}, \textcolor{darkmagenta}{\begin{CJK}{UTF8}{min}津賀一宏\end{CJK}})} & 
\begin{CJK}{UTF8}{min}名前で直接お答えください。\end{CJK} \textcolor{forestgreen}{\begin{CJK}{UTF8}{min}パナソニック株式会社\end{CJK}} \begin{CJK}{UTF8}{min}のCEOの名前は\end{CJK} & \multirow{2}{*}{\centering
\textcolor{darkmagenta}{\begin{CJK}{UTF8}{min}津賀一宏\end{CJK}}} \\
& & \begin{CJK}{UTF8}{min}名前で直接お答えください。\end{CJK} \textcolor{forestgreen}{\begin{CJK}{UTF8}{min}パナソニック株式会社\end{CJK}} \begin{CJK}{UTF8}{min}のCEOは誰ですか？彼らの名前は\end{CJK} & \\
\midrule
\multirow{2}{*}{\centering Chinese} & \multirow{2}{*}{(\raggedright \textcolor{forestgreen}{\begin{CJK}{UTF8}{gbsn}松下公司\end{CJK}}, \textcolor{darkmagenta}{\begin{CJK}{UTF8}{gbsn}津贺一宏\end{CJK}})} & \textcolor{forestgreen}{\begin{CJK}{UTF8}{gbsn}松下公司\end{CJK}} \begin{CJK}{UTF8}{gbsn}的首席执行官名字叫做\end{CJK} & \multirow{2}{*}{\centering
\textcolor{darkmagenta}{\begin{CJK}{UTF8}{gbsn}津贺一宏\end{CJK}}} \\
& & \textcolor{forestgreen}{\begin{CJK}{UTF8}{gbsn}松下公司\end{CJK}} \begin{CJK}{UTF8}{gbsn}的CEO名字叫做\end{CJK} & \\
\bottomrule
\end{tabular}
\end{center}
\caption{Prompts for \textbf{\texttt{company\_ceo}} in different languages. We use the triple (\texttt{Panasonic}, \texttt{company\_ceo}, \texttt{Kazuhiro Tsuga}) as an example. The subject-object pair is represented in the respective language. The prompt shown below the dashed line is the new template introduced for the experiment described in \secref{effect_prompt}.}
\label{tab:prompts company_ceo}
\end{table*}

\begin{table*}[htbp]
\scriptsize
\centering
\setlength{\tabcolsep}{1.0mm}{}
\begin{center}
\begin{tabular}{m{0.10\linewidth} p{0.30\linewidth} p{0.30\linewidth} m{0.15\linewidth}}
\textbf{Language} & \textbf{\textcolor{forestgreen}{Subject}-\textcolor{darkmagenta}{Object} Pair} & \textbf{Prompt} & \textbf{Expected Output} \\
\toprule
\multirow{3}{*}{\centering English} 
    & \multirow{3}{*}{\raggedright (\textcolor{forestgreen}{Cadillac}, \textcolor{darkmagenta}{Detroit})}
    & The headquarters of \textcolor{forestgreen}{Cadillac} are in the city of? \newline Answer:
    & \multirow{3}{*}{\centering \textcolor{darkmagenta}{Detroit}} \\
    & 
    & The headquarters of \textcolor{forestgreen}{Cadillac} are in the city of? \newline Answer:
    & \\
    \cdashline{3-3}[1.2pt/1pt]
    \addlinespace[2pt]  
    & & Where are the headquarters of \textcolor{forestgreen}{Cadillac} It is in the city of? Answer: &\\
\midrule
\multirow{2}{*}{\centering German} 
    & \multirow{2}{*}{\raggedright (\textcolor{forestgreen}{Cadillac}, \textcolor{darkmagenta}{Detroit})}
    & \textcolor{forestgreen}{Cadillac} hat seinen Hauptsitz in der Stadt von 
    & \multirow{2}{*}{\centering \textcolor{darkmagenta}{Detroit}} \\
    & 
    & Der Hauptsitz von \textcolor{forestgreen}{Cadillac} befindet sich in der Stadt von 
    & \\
\midrule
\multirow{2}{*}{\centering Spanish} 
    & \multirow{2}{*}{\raggedright (\textcolor{forestgreen}{Cadillac}, \textcolor{darkmagenta}{Detroit})}
    & \textcolor{forestgreen}{Cadillac} tiene su sede en la ciudad de 
    & \multirow{2}{*}{\centering \textcolor{darkmagenta}{Detroit}} \\
    & 
    & La sede de \textcolor{forestgreen}{Cadillac} se encuentra en la ciudad de 
    & \\
\midrule
\multirow{2}{*}{\centering French} 
    & \multirow{2}{*}{\raggedright (\textcolor{forestgreen}{Cadillac}, \textcolor{darkmagenta}{Détroit})}
    & Le nom de la ville où se trouve le siège social de \textcolor{forestgreen}{Cadillac} est 
    & \multirow{2}{*}{\centering \textcolor{darkmagenta}{Détroit}} \\
    & 
    & La ville où se trouve le siège social de \textcolor{forestgreen}{Cadillac} s'appelle 
    & \\
\midrule
\multirow{2}{*}{\centering Japanese} 
    & \multirow{2}{*}{\raggedright (\textcolor{forestgreen}{\begin{CJK}{UTF8}{min}「キャデラック」\end{CJK}}, \textcolor{darkmagenta}{\begin{CJK}{UTF8}{min}デトロイト\end{CJK}})}
    & \textcolor{forestgreen}{\begin{CJK}{UTF8}{min}「キャデラック」\end{CJK}}\begin{CJK}{UTF8}{min}の本社がある都市はどこですか\end{CJK} 
    & \multirow{2}{*}{\centering \textcolor{darkmagenta}{\begin{CJK}{UTF8}{min}デトロイト\end{CJK}}} \\
    & 
    & \textcolor{forestgreen}{\begin{CJK}{UTF8}{min}「キャデラック」\end{CJK}}\begin{CJK}{UTF8}{min}の本社はどの都市にありますか\end{CJK} 
    & \\
\midrule
\multirow{2}{*}{\centering Chinese} 
    & \multirow{2}{*}{\raggedright (\textcolor{forestgreen}{\begin{CJK}{UTF8}{gbsn}凯迪拉克\end{CJK}}, \textcolor{darkmagenta}{\begin{CJK}{UTF8}{gbsn}底特律\end{CJK}})}
    & \textcolor{forestgreen}{\begin{CJK}{UTF8}{gbsn}凯迪拉克\end{CJK}}\begin{CJK}{UTF8}{gbsn}总部所位于的城市名字叫做\end{CJK} 
    & \multirow{2}{*}{\centering \textcolor{darkmagenta}{\begin{CJK}{UTF8}{gbsn}底特律\end{CJK}}} \\
    & 
    & \textcolor{forestgreen}{\begin{CJK}{UTF8}{gbsn}凯迪拉克\end{CJK}}\begin{CJK}{UTF8}{gbsn}的总部所在的城市名字叫 \end{CJK} 
    & \\
    
\bottomrule
\end{tabular}
\end{center}
\caption{Prompts for \textbf{\texttt{company\_hq}} in all languages. We use the triple (\texttt{Cadillac}, \texttt{company\_hq}, \texttt{Detroit}) as an example. The subject-object pair is represented in the respective language. The prompt shown below the dashed line is the new template introduced for the experiment described in \secref{effect_prompt}.}
\label{tab:prompts company_hq}
\end{table*}

\begin{table*}[htbp]
\scriptsize
\centering
\setlength{\tabcolsep}{1.0mm}{}
\begin{center}
\begin{tabular}{m{0.10\linewidth} p{0.30\linewidth} p{0.30\linewidth} m{0.15\linewidth}}
\textbf{Language} & \textbf{\textcolor{forestgreen}{Subject}-\textcolor{darkmagenta}{Object} Pair} & \textbf{Prompt} & \textbf{Expected Output} \\
\toprule
\multirow{2}{*}{\centering English} 
    & \multirow{2}{*}{\raggedright (\textcolor{forestgreen}{Elbe}, \textcolor{darkmagenta}{Europe})}
    & {\textcolor{forestgreen}{Elbe} is on the continent of?  Answer:}
    & \multirow{2}{*}{\centering \textcolor{darkmagenta}{Europe}} \\
    
    \cdashline{3-3}[1.2pt/1pt]
    \addlinespace[2pt]  
    & & What continent is \textcolor{forestgreen}{Elbe} on? It is on? Answer: &\\
\midrule
{\centering German} 
    & {\raggedright (\textcolor{forestgreen}{Elbe}, \textcolor{darkmagenta}{Europa})}
    & Bitte geben Sie den Kontinentnamen direkt an, z.\,B. Europa, Afrika usw. Der Name des Kontinents, auf dem \textcolor{forestgreen}{Elbe} liegt, lautet
    & {\centering \textcolor{darkmagenta}{Europa}} \\
\midrule
{\centering Spanish} 
    & {\raggedright (\textcolor{forestgreen}{Elba}, \textcolor{darkmagenta}{Europa})}
    & El nombre del continente donde se encuentra \textcolor{forestgreen}{Elba} es
    & {\centering \textcolor{darkmagenta}{Europa}} \\
\midrule
{\centering French} 
    & {\raggedright (\textcolor{forestgreen}{Elbe}, \textcolor{darkmagenta}{Europe})}
    & Veuillez répondre directement avec le nom du continent. Le nom du continent où se trouve \textcolor{forestgreen}{Elbe} est
    & {\centering \textcolor{darkmagenta}{Europe}} \\
\midrule
{\centering Japanese} 
    & {\raggedright (\textcolor{forestgreen}{\begin{CJK}{UTF8}{min}エルベ川\end{CJK}}, \textcolor{darkmagenta}{\begin{CJK}{UTF8}{min}ヨーロッパ\end{CJK}})}
    & \textcolor{forestgreen}{\begin{CJK}{UTF8}{min}エルベ川\end{CJK}}\begin{CJK}{UTF8}{min}が所在する大陸の名前は\end{CJK}
    & {\centering \textcolor{darkmagenta}{\begin{CJK}{UTF8}{min}ヨーロッパ\end{CJK}}} \\
\midrule
{\centering Chinese} 
    & {\raggedright (\textcolor{forestgreen}{\begin{CJK}{UTF8}{gbsn}易北河\end{CJK}}, \textcolor{darkmagenta}{\begin{CJK}{UTF8}{gbsn}欧洲\end{CJK}})}
    & \textcolor{forestgreen}{\begin{CJK}{UTF8}{gbsn}易北河\end{CJK}}\begin{CJK}{UTF8}{gbsn}所位于的大洲/大陆名字叫做\end{CJK}
    & {\centering \textcolor{darkmagenta}{\begin{CJK}{UTF8}{gbsn}欧洲\end{CJK}}} \\
\bottomrule
\end{tabular}
\end{center}
\caption{Prompts for the \textbf{\texttt{landmark\_continent}} relation in all languages. We use the triple (\texttt{Elbe}, \texttt{landmark\_continent}, \texttt{Europe}) as an example. The subject-object pair is represented in the respective language. The prompt shown below the dashed line is the new template introduced for the experiment described in \secref{effect_prompt}.}
\label{tab:prompts_landmark_continent}
\end{table*}

\begin{table*}[htbp]
\scriptsize
\centering
\setlength{\tabcolsep}{1.0mm}{}
\begin{center}
\begin{tabular}{m{0.10\linewidth} p{0.30\linewidth} p{0.30\linewidth} m{0.15\linewidth}}
\textbf{Language} & \textbf{\textcolor{forestgreen}{Subject}-\textcolor{darkmagenta}{Object} Pair} & \textbf{Prompt} & \textbf{Expected Output} \\
\toprule
\multirow{2}{*}{\centering English} 
    & \multirow{2}{*}{\raggedright (\textcolor{forestgreen}{Namba Station}, \textcolor{darkmagenta}{Japan})}
    & \textcolor{forestgreen}{Namba Station} is in the country of? Answer:
    & \multirow{2}{*}{\centering \textcolor{darkmagenta}{Japan}} \\
    \cdashline{3-3}[1.2pt/1pt]
    \addlinespace[2pt] 
    && What country is \textcolor{forestgreen}{Namba Station} in? It is in? \newline Answer:&\\
\midrule
{\centering German} 
    & {\raggedright (\textcolor{forestgreen}{Namba Station}, \textcolor{darkmagenta}{Japan})}
    & In welchem Land liegt \textcolor{forestgreen}{Namba Station}? Es liegt in
    & {\centering \textcolor{darkmagenta}{Japan}} \\
\midrule
{\centering Spanish} 
    & {\raggedright (\textcolor{forestgreen}{Namba Station}, \textcolor{darkmagenta}{Japan})}
    & El nombre del país donde se encuentra \textcolor{forestgreen}{Namba Station} es
    & {\centering \textcolor{darkmagenta}{Japan}} \\
\midrule
{\centering French} 
    & {\raggedright (\textcolor{forestgreen}{Namba Station}, \textcolor{darkmagenta}{Japan})}
    & Le nom du pays où se trouve \textcolor{forestgreen}{Namba Station} est
    & {\centering \textcolor{darkmagenta}{Japan}} \\
\midrule
{\centering Japanese} 
    & {\raggedright (\textcolor{forestgreen}{\begin{CJK}{UTF8}{min}難波駅\end{CJK}}, \textcolor{darkmagenta}{\begin{CJK}{UTF8}{min}日本\end{CJK}})}
    & \textcolor{forestgreen}{\begin{CJK}{UTF8}{min}難波駅\end{CJK}}\begin{CJK}{UTF8}{min}が所在する国の名前は\end{CJK}
    & {\centering \textcolor{darkmagenta}{\begin{CJK}{UTF8}{min}日本\end{CJK}}} \\
\midrule
{\centering Chinese} 
    & {\raggedright (\textcolor{forestgreen}{\begin{CJK}{UTF8}{gbsn}难波站\end{CJK}}, \textcolor{darkmagenta}{\begin{CJK}{UTF8}{gbsn}日本\end{CJK}})}
    & \textcolor{forestgreen}{\begin{CJK}{UTF8}{gbsn}难波站\end{CJK}}\begin{CJK}{UTF8}{gbsn}所位于的国家名字叫做\end{CJK}
    & {\centering \textcolor{darkmagenta}{\begin{CJK}{UTF8}{gbsn}日本\end{CJK}}} \\
\bottomrule
\end{tabular}
\end{center}
\caption{Prompts for the \textbf{\texttt{landmark\_country}} relation in all languages. We use the triple (\texttt{Namba Station}, \texttt{landmark\_country}, \texttt{Japan}) as an example. The subject-object pair is represented in the respective language. The prompt shown below the dashed line is the new template introduced for the experiment described in \secref{effect_prompt}.}
\label{tab:prompts_landmark_country}
\end{table*}

\begin{table*}[htbp]
\scriptsize
\centering
\setlength{\tabcolsep}{1.0mm}{}
\begin{center}
\begin{tabular}{m{0.10\linewidth} p{0.30\linewidth} p{0.30\linewidth} m{0.15\linewidth}}
\textbf{Language} & \textbf{\textcolor{forestgreen}{Subject}-\textcolor{darkmagenta}{Object} Pair} & \textbf{Prompt} & \textbf{Expected Output} \\
\toprule
\multirow{2}{*}{\centering English} 
    & \multirow{2}{*}{\raggedright (\textcolor{forestgreen}{Ronald Reagan}, \textcolor{darkmagenta}{Jack Reagan})}
    & \textcolor{forestgreen}{Ronald Reagan}'s father is named? Answer:
    & \multirow{2}{*}{\centering \textcolor{darkmagenta}{Jack Reagan}} \\
    \cdashline{3-3}[1.2pt/1pt]
    \addlinespace[2pt] 
    & &Who is \textcolor{forestgreen}{Ronald Reagan}'s father? Their father is named? Answer:&\\ 
\midrule
{\centering German} 
    & {\raggedright (\textcolor{forestgreen}{Ronald Reagan}, \textcolor{darkmagenta}{Jack Reagan})}
    & Der Vater von \textcolor{forestgreen}{Ronald Reagan} heißt
    & {\centering \textcolor{darkmagenta}{Jack Reagan}} \\
\midrule
{\centering Spanish} 
    & {\raggedright (\textcolor{forestgreen}{Ronald Reagan}, \textcolor{darkmagenta}{Jack Reagan})}
    & El padre de \textcolor{forestgreen}{Ronald Reagan} se llama
    & {\centering \textcolor{darkmagenta}{Jack Reagan}} \\
\midrule
{\centering French} 
    & {\raggedright (\textcolor{forestgreen}{Ronald Reagan}, \textcolor{darkmagenta}{Jack Reagan})}
    & Le père de \textcolor{forestgreen}{Ronald Reagan} s'appelle
    & {\centering \textcolor{darkmagenta}{Jack Reagan}} \\
\midrule
{\centering Japanese} 
    & {\raggedright (\textcolor{forestgreen}{\begin{CJK}{UTF8}{min}ロナルド・レーガン\end{CJK}}, \textcolor{darkmagenta}{\begin{CJK}{UTF8}{min}ジャック・レーガン\end{CJK}})}
    & \begin{CJK}{UTF8}{min}名前で直接お答えください。\end{CJK}\textcolor{forestgreen}{\begin{CJK}{UTF8}{min}ロナルド・レーガン\end{CJK}}\begin{CJK}{UTF8}{min}の父親の名前は\end{CJK}
    & {\centering \textcolor{darkmagenta}{\begin{CJK}{UTF8}{min}ジャック・レーガン\end{CJK}}} \\
\midrule
{\centering Chinese} 
    & {\raggedright (\textcolor{forestgreen}{\begin{CJK}{UTF8}{gbsn}罗纳德·里根\end{CJK}}, \textcolor{darkmagenta}{\begin{CJK}{UTF8}{gbsn}杰克·里根\end{CJK}})}
    & \textcolor{forestgreen}{\begin{CJK}{UTF8}{gbsn}罗纳德·里根\end{CJK}}\begin{CJK}{UTF8}{gbsn}的父亲名字叫做\end{CJK}
    & {\centering \textcolor{darkmagenta}{\begin{CJK}{UTF8}{gbsn}杰克·里根\end{CJK}}} \\
\bottomrule
\end{tabular}
\end{center}
\caption{Prompts for the \textbf{\texttt{person\_father}} relation in all languages. We use the triple (\texttt{Ronald Reagan}, \texttt{person\_father}, \texttt{Jack Reagan}) as an example. The subject-object pair is represented in the respective language. The prompt shown below the dashed line is the new template introduced for the experiment described in \secref{effect_prompt}.}
\label{tab:prompts_person_father}
\end{table*}

\begin{table*}[htbp]
\scriptsize
\centering
\setlength{\tabcolsep}{1.0mm}{}
\begin{center}
\begin{tabular}{m{0.10\linewidth} p{0.30\linewidth} p{0.30\linewidth} m{0.15\linewidth}}
\textbf{Language} & \textbf{\textcolor{forestgreen}{Subject}-\textcolor{darkmagenta}{Object} Pair} & \textbf{Prompt} & \textbf{Expected Output} \\
\toprule
\multirow{3}{*}{\centering English} 
    & \multirow{3}{*}{\raggedright (\textcolor{forestgreen}{Demi Moore}, \textcolor{darkmagenta}{Virginia King})}
    & \textcolor{forestgreen}{Demi Moore}'s mother is named? Answer:
    & \multirow{3}{*}{\centering \textcolor{darkmagenta}{Virginia King}} \\
    \cdashline{3-3}[1.2pt/1pt]
    \addlinespace[2pt] 
    & & Name of mother of \textcolor{forestgreen}{Demi Moore} is? Answer:& \\
    & & Who is \textcolor{forestgreen}{Demi Moore}'s mother? Their mother is named? Answer: & \\
\midrule
{\centering German} 
    & {\raggedright (\textcolor{forestgreen}{Demi Moore}, \textcolor{darkmagenta}{Virginia King})}
    & Die Mutter von \textcolor{forestgreen}{Demi Moore} heißt
    & {\centering \textcolor{darkmagenta}{Virginia King}} \\
\midrule
{\centering Spanish} 
    & {\raggedright (\textcolor{forestgreen}{Demi Moore}, \textcolor{darkmagenta}{Virginia King})}
    & La madre de \textcolor{forestgreen}{Demi Moore} se llama
    & {\centering \textcolor{darkmagenta}{Virginia King}} \\
\midrule
{\centering French} 
    & {\raggedright (\textcolor{forestgreen}{Demi Moore}, \textcolor{darkmagenta}{Virginia King})}
    & Qui est la mère de \textcolor{forestgreen}{Demi Moore} ? Leur mère s'appelle
    & {\centering \textcolor{darkmagenta}{Virginia King}} \\
\midrule
{\centering Japanese} 
    & {\raggedright (\textcolor{forestgreen}{\begin{CJK}{UTF8}{min}デミ・ムーア\end{CJK}}, \textcolor{darkmagenta}{\begin{CJK}{UTF8}{min}ヴァージニア・キング\end{CJK}})}
    & \begin{CJK}{UTF8}{min}名前で直接お答えください。\end{CJK}\textcolor{forestgreen}{\begin{CJK}{UTF8}{min}デミ・ムーア\end{CJK}}\begin{CJK}{UTF8}{min}の母親の名前は\end{CJK}
    & {\centering \textcolor{darkmagenta}{\begin{CJK}{UTF8}{min}ヴァージニア・キング\end{CJK}}} \\
\midrule
{\centering Chinese} 
    & {\raggedright (\textcolor{forestgreen}{\begin{CJK}{UTF8}{gbsn}黛米·摩尔\end{CJK}}, \textcolor{darkmagenta}{\begin{CJK}{UTF8}{gbsn}维吉尼亚·金\end{CJK}})}
    & \textcolor{forestgreen}{\begin{CJK}{UTF8}{gbsn}黛米·摩尔\end{CJK}}\begin{CJK}{UTF8}{gbsn}的母亲名字叫做\end{CJK}
    & {\centering \textcolor{darkmagenta}{\begin{CJK}{UTF8}{gbsn}维吉尼亚·金\end{CJK}}} \\
\bottomrule
\end{tabular}
\end{center}
\caption{Prompts for the \textbf{\texttt{person\_mother}} relation in all languages. We use the triple (\texttt{Demi Moore}, \texttt{person\_mother}, \texttt{Virginia King}) as an example. The subject-object pair is represented in the respective language. The prompt shown below the dashed line is the new template introduced for the experiment described in \secref{effect_prompt}.}
\label{tab:prompts_person_mother}
\end{table*}

\begin{table*}[htbp]
\scriptsize
\centering
\setlength{\tabcolsep}{1.0mm}{}
\begin{center}
\begin{tabular}{m{0.10\linewidth} p{0.30\linewidth} p{0.30\linewidth} m{0.15\linewidth}}
\textbf{Language} & \textbf{\textcolor{forestgreen}{Subject}-\textcolor{darkmagenta}{Object} Pair} & \textbf{Prompt} & \textbf{Expected Output} \\
\toprule
\multirow{3}{*}{\centering English} 
    & \multirow{3}{*}{\raggedright (\textcolor{forestgreen}{Martin Burrell}, \textcolor{darkmagenta}{politician})}
    & \textcolor{forestgreen}{Martin Burrell} works as a? Answer:
    & \multirow{3}{*}{\centering \textcolor{darkmagenta}{politician}} \\
    & 
    & By profession, \textcolor{forestgreen}{Martin Burrell} is a? \newline Answer:
    & \\
    \cdashline{3-3}[1.2pt/1pt]
    \addlinespace[2pt] 
    & & \textcolor{forestgreen}{Martin Burrell} works professionally as a?\newline Answer: & \\
\midrule
\multirow{2}{*}{\centering German}
    & \multirow{2}{*}{\raggedright (\textcolor{forestgreen}{Martin Burrell}, \textcolor{darkmagenta}{Politiker})}
    & \textcolor{forestgreen}{Martin Burrell} arbeitet als
    & \multirow{2}{*}{\centering \textcolor{darkmagenta}{Politiker}} \\
    & 
    & Von Beruf ist \textcolor{forestgreen}{Martin Burrell} ein
    & \\
\midrule
\multirow{2}{*}{\centering Spanish}
    & \multirow{2}{*}{\raggedright (\textcolor{forestgreen}{Martin Burrell}, \textcolor{darkmagenta}{político})}
    & Por favor especifique el nombre de su ocupación. \textcolor{forestgreen}{Martin Burrell} trabaja profesionalmente como
    & \multirow{2}{*}{\centering \textcolor{darkmagenta}{político}} \\
    & 
    & Por favor especifique el nombre de su ocupación. Por profesión, \textcolor{forestgreen}{Martin Burrell} es un(a)
    & \\
\midrule
\multirow{2}{*}{\centering French}
    & \multirow{2}{*}{\raggedright (\textcolor{forestgreen}{Martin Burrell}, \textcolor{darkmagenta}{personnalité politique})}
    & Veuillez répondre directement par le nom de votre profession. Le nom de la profession de \textcolor{forestgreen}{Martin Burrell} est
    & \multirow{2}{*}{\centering \textcolor{darkmagenta}{personnalité politique}} \\
    & 
    & Veuillez répondre directement par le nom de votre profession. \textcolor{forestgreen}{Martin Burrell} travaille professionnellement comme
    & \\
\midrule
\multirow{2}{*}{\centering Japanese}
    & \multirow{2}{*}{\raggedright (\textcolor{forestgreen}{\begin{CJK}{UTF8}{min}マーティン・バレル\end{CJK}}, \textcolor{darkmagenta}{\begin{CJK}{UTF8}{min}政治家\end{CJK}})}
    & \textcolor{forestgreen}{\begin{CJK}{UTF8}{min}マーティン・バレル\end{CJK}}\begin{CJK}{UTF8}{min}さんの職業名は\end{CJK}
    & \multirow{2}{*}{\centering \textcolor{darkmagenta}{\begin{CJK}{UTF8}{min}政治家\end{CJK}}} \\
    & 
    & \textcolor{forestgreen}{\begin{CJK}{UTF8}{min}マーティン・バレル\end{CJK}}\begin{CJK}{UTF8}{min}さんの職業名は\end{CJK}
    & \\
\midrule
\multirow{2}{*}{\centering Chinese}
    & \multirow{2}{*}{\raggedright (\textcolor{forestgreen}{\begin{CJK}{UTF8}{gbsn}马丁·巴雷尔\end{CJK}}, \textcolor{darkmagenta}{\begin{CJK}{UTF8}{gbsn}政治人物\end{CJK}})}
    & \textcolor{forestgreen}{\begin{CJK}{UTF8}{gbsn}马丁·巴雷尔\end{CJK}}\begin{CJK}{UTF8}{gbsn}从事的职业是一个\end{CJK}
    & \multirow{2}{*}{\centering \textcolor{darkmagenta}{\begin{CJK}{UTF8}{gbsn}政治人物\end{CJK}}} \\
    & 
    & \begin{CJK}{UTF8}{gbsn}职业上来说, \end{CJK}\textcolor{forestgreen}{\begin{CJK}{UTF8}{gbsn}马丁·巴雷尔\end{CJK}}\begin{CJK}{UTF8}{gbsn}是一名\end{CJK}
    & \\
\bottomrule
\end{tabular}
\end{center}
\caption{Prompts for the \textbf{\texttt{person\_occupation}} relation in all languages. We use the triple (\texttt{Martin Burrell}, \texttt{person\_occupation}, \texttt{politician}) as an example. The subject-object pair is represented in the respective language. The prompt shown below the dashed line is the new template introduced for the experiment described in \secref{effect_prompt}.}
\label{tab:prompts_person_occupation}
\end{table*}

\begin{table*}[htbp]
\scriptsize
\centering
\setlength{\tabcolsep}{1.0mm}{}
\begin{center}
\begin{tabular}{m{0.10\linewidth} p{0.30\linewidth} p{0.30\linewidth} m{0.15\linewidth}}
\textbf{Language} & \textbf{\textcolor{forestgreen}{Subject}-\textcolor{darkmagenta}{Object} Pair} & \textbf{Prompt} & \textbf{Expected Output} \\
\toprule
\multirow{3}{*}{\centering English} 
    & \multirow{3}{*}{\raggedright (\textcolor{forestgreen}{Anson Funderburgh}, \textcolor{darkmagenta}{guitar})}
    & \textcolor{forestgreen}{Anson Funderburgh} plays the instrument of? \newline Answer:
    & \multirow{3}{*}{\centering \textcolor{darkmagenta}{guitar}} \\
    \cdashline{3-3}[1.2pt/1pt]
    \addlinespace[2pt] 
    && What instrument does \textcolor{forestgreen}{Anson Funderburgh} play? They play the? Answer:&\\
    && The instrument that \textcolor{forestgreen}{Anson Funderburgh} plays is called the? Answer: & \\
\midrule
{\centering German} 
    & {\raggedright (\textcolor{forestgreen}{Anson Funderburgh}, \textcolor{darkmagenta}{Gitarre})}
    & Bitte geben Sie den Namen des Instruments direkt an. Das Instrument, das \textcolor{forestgreen}{Anson Funderburgh} spielt, heißt
    & {\centering \textcolor{darkmagenta}{Gitarre}} \\
\midrule
{\centering Spanish} 
    & {\raggedright (\textcolor{forestgreen}{Anson Funderburgh}, \textcolor{darkmagenta}{guitarra})}
    & Por favor responda directamente el nombre del instrumento ¿Qué instrumento toca \textcolor{forestgreen}{Anson Funderburgh}? Tocan el
    & {\centering \textcolor{darkmagenta}{guitarra}} \\
\midrule
{\centering French} 
    & {\raggedright (\textcolor{forestgreen}{Anson Funderburgh}, \textcolor{darkmagenta}{guitare})}
    & Veuillez répondre directement au nom de l'instrument. De quel instrument joue \textcolor{forestgreen}{Anson Funderburgh} ? Ils jouent du
    & {\centering \textcolor{darkmagenta}{guitare}} \\
\midrule
{\centering Japanese} 
    & {\raggedright (\textcolor{forestgreen}{\begin{CJK}{UTF8}{min}アンソン・ファンダーバーグ\end{CJK}}, \textcolor{darkmagenta}{\begin{CJK}{UTF8}{min}ギター\end{CJK}})}
    & \textcolor{forestgreen}{\begin{CJK}{UTF8}{min}アンソン・ファンダーバーグ\end{CJK}}\begin{CJK}{UTF8}{min}はどの楽器を演奏しますか\end{CJK}
    & {\centering \textcolor{darkmagenta}{\begin{CJK}{UTF8}{min}ギター\end{CJK}}} \\
\midrule
{\centering Chinese} 
    & {\raggedright (\textcolor{forestgreen}{\begin{CJK}{UTF8}{gbsn}安森·芬德伯格\end{CJK}}, \textcolor{darkmagenta}{\begin{CJK}{UTF8}{gbsn}吉他\end{CJK}})}
    & \textcolor{forestgreen}{\begin{CJK}{UTF8}{gbsn}安森·芬德伯格\end{CJK}}\begin{CJK}{UTF8}{gbsn}所演奏的乐器名字叫做\end{CJK}
    & {\centering \textcolor{darkmagenta}{\begin{CJK}{UTF8}{gbsn}吉他\end{CJK}}} \\
\bottomrule
\end{tabular}
\end{center}
\caption{Prompts for the \textbf{\texttt{person\_plays\_instrument}} relation in all languages. We use the triple (\texttt{Anson Funderburgh}, \texttt{person\_plays\_instrument}, \texttt{guitar}) as an example. The subject-object pair is represented in the respective language. The prompt shown below the dashed line is the new template introduced for the experiment described in \secref{effect_prompt}.}
\label{tab:prompts_person_plays_instrument}
\end{table*}

\begin{table*}[htbp]
\scriptsize
\centering
\setlength{\tabcolsep}{1.0mm}{}
\begin{center}
\begin{tabular}{m{0.10\linewidth} p{0.30\linewidth} p{0.30\linewidth} m{0.15\linewidth}}
\textbf{Language} & \textbf{\textcolor{forestgreen}{Subject}-\textcolor{darkmagenta}{Object} Pair} & \textbf{Prompt} & \textbf{Expected Output} \\
\toprule
\multirow{3}{*}{\centering English} 
    & \multirow{3}{*}{\raggedright (\textcolor{forestgreen}{Frédéric Piquionne}, \textcolor{darkmagenta}{soccer})}
    & \textcolor{forestgreen}{Frédéric Piquionne} plays the sport of? \newline Answer:
    & \multirow{3}{*}{\centering \textcolor{darkmagenta}{soccer}} \\
    \cdashline{3-3}[1.2pt/1pt]
    \addlinespace[2pt] 
    && What sport does \textcolor{forestgreen}{Frédéric Piquionne} play? They play? Answer: & \\
    && \textcolor{forestgreen}{Frédéric Piquionne} plays professionally in the sport of? Answer: & \\
\midrule
{\centering German} 
    & {\raggedright (\textcolor{forestgreen}{Frédéric Piquionne}, \textcolor{darkmagenta}{Fußball})}
    & Welchen Sport betreibt \textcolor{forestgreen}{Frédéric Piquionne}? Sie betreiben
    & {\centering \textcolor{darkmagenta}{Fußball}} \\
\midrule
{\centering Spanish} 
    & {\raggedright (\textcolor{forestgreen}{Frédéric Piquionne}, \textcolor{darkmagenta}{fútbol})}
    & Por favor, responda directamente el nombre del deporte, como fútbol, baloncesto, etc. El nombre del deporte que juega \textcolor{forestgreen}{Frédéric Piquionne} es:
    & {\centering \textcolor{darkmagenta}{fútbol}} \\
\midrule
{\centering French} 
    & {\raggedright (\textcolor{forestgreen}{Frédéric Piquionne}, \textcolor{darkmagenta}{football})}
    & Veuillez répondre directement par le nom du sport, comme le football, le basket-ball, etc. \textcolor{forestgreen}{Frédéric Piquionne} joue professionnellement dans le sport de
    & {\centering \textcolor{darkmagenta}{football}} \\
\midrule
{\centering Japanese} 
    & {\raggedright (\textcolor{forestgreen}{\begin{CJK}{UTF8}{min}フレデリック・ピキオンヌ\end{CJK}}, \textcolor{darkmagenta}{\begin{CJK}{UTF8}{min}サッカー\end{CJK}})}
    & \begin{CJK}{UTF8}{min}サッカー、バスケットボールなど、スポーツの名前を直接答えてください。\end{CJK}\textcolor{forestgreen}{\begin{CJK}{UTF8}{min}フレデリック・ピキオンヌ\end{CJK}}\begin{CJK}{UTF8}{min}はどのスポーツをしますか？彼らは（スポーツ名）をしています。\end{CJK}
    & {\centering \textcolor{darkmagenta}{\begin{CJK}{UTF8}{min}サッカー\end{CJK}}} \\
\midrule
{\centering Chinese} 
    & {\raggedright (\textcolor{forestgreen}{\begin{CJK}{UTF8}{gbsn}费德历·比基安尼\end{CJK}}, \textcolor{darkmagenta}{\begin{CJK}{UTF8}{gbsn}足球\end{CJK}})}
    & \textcolor{forestgreen}{\begin{CJK}{UTF8}{gbsn}费德历·比基安尼\end{CJK}}\begin{CJK}{UTF8}{gbsn}从事的运动叫做\end{CJK}
    & {\centering \textcolor{darkmagenta}{\begin{CJK}{UTF8}{gbsn}足球\end{CJK}}} \\
\bottomrule
\end{tabular}
\end{center}
\caption{Prompts for the \textbf{\texttt{person\_pro\_sport}} relation in all languages. We use the triple (\texttt{Frédéric Piquionne}, \texttt{person\_pro\_sport}, \texttt{soccer}) as an example. The subject-object pair is represented in the respective language. The prompt shown below the dashed line is the new template introduced for the experiment described in \secref{effect_prompt}.}
\label{tab:prompts_person_pro_sport}
\end{table*}

\begin{table*}[htbp]
\scriptsize
\centering
\setlength{\tabcolsep}{1.0mm}{}
\begin{center}
\begin{tabular}{m{0.10\linewidth} p{0.30\linewidth} p{0.30\linewidth} m{0.15\linewidth}}
\textbf{Language} & \textbf{\textcolor{forestgreen}{Subject}-\textcolor{darkmagenta}{Object} Pair} & \textbf{Prompt} & \textbf{Expected Output} \\
\toprule
\multirow{4}{*}{\centering English} 
    & \multirow{4}{*}{\raggedright (\textcolor{forestgreen}{Ju Yingzhi}, \textcolor{darkmagenta}{midfielder})}
    & \textcolor{forestgreen}{Ju Yingzhi} plays in the position of a? \newline Answer:
    & \multirow{4}{*}{\centering \textcolor{darkmagenta}{midfielder}} \\
    & 
    & In their sport, \textcolor{forestgreen}{Ju Yingzhi} plays as a? \newline Answer:
    & \\
    \cdashline{3-3}[1.2pt/1pt]
    \addlinespace[2pt] 
    && Which position does \textcolor{forestgreen}{Ju Yingzhi} play? They play as a? Answer: & \\
    && In their sport, \textcolor{forestgreen}{Ju Yingzhi} plays in the position of a? Answer: & \\
\midrule
\multirow{2}{*}{\centering German}
    & \multirow{2}{*}{\raggedright (\textcolor{forestgreen}{Ju Yingzhi}, \textcolor{darkmagenta}{Mittelfeldspieler})}
    & \textcolor{forestgreen}{Ju Yingzhi} spielt auf der Position von a
    & \multirow{2}{*}{\centering \textcolor{darkmagenta}{Mittelfeldspieler}} \\
    & 
    & In ihrer Sportart spielt \textcolor{forestgreen}{Ju Yingzhi} als
    & \\
\midrule
\multirow{2}{*}{\centering Spanish}
    & \multirow{2}{*}{\raggedright (\textcolor{forestgreen}{Ju Yingzhi}, \textcolor{darkmagenta}{centrocampista})}
    & Por favor, responda directamente el nombre de la posición deportiva, como delantero, defensor, etc. La posición de \textcolor{forestgreen}{Ju Yingzhi} en el campo deportivo es:
    & \multirow{2}{*}{\centering \textcolor{darkmagenta}{centrocampista}} \\
    & 
    & Por favor responda directamente con el nombre de la posición deportiva, como delantero, defensor, etc. En su deporte, \textcolor{forestgreen}{Ju Yingzhi} juega en la posición de un:
    & \\
\midrule
\multirow{2}{*}{\centering French}
    & \multirow{2}{*}{\raggedright (\textcolor{forestgreen}{Ju Yingzhi}, \textcolor{darkmagenta}{milieu de terrain})}
    & \textcolor{forestgreen}{Ju Yingzhi} évolue au poste de
    & \multirow{2}{*}{\centering \textcolor{darkmagenta}{milieu de terrain}} \\
    & 
    & Dans son sport, \textcolor{forestgreen}{Ju Yingzhi} occupe le rôle de
    & \\
\midrule
\multirow{2}{*}{\centering Japanese}
    & \multirow{2}{*}{\raggedright (\textcolor{forestgreen}{\begin{CJK}{UTF8}{min}ジュ・インジー\end{CJK}}, \textcolor{darkmagenta}{\begin{CJK}{UTF8}{min}ミッドフィールダー\end{CJK}})}
    & \begin{CJK}{UTF8}{min}彼がプレーするスポーツでは、\end{CJK}\textcolor{forestgreen}{\begin{CJK}{UTF8}{min}ジュ・インジー\end{CJK}}\begin{CJK}{UTF8}{min}のポジションは\end{CJK}
    & \multirow{2}{*}{\centering \textcolor{darkmagenta}{\begin{CJK}{UTF8}{min}ミッドフィールダー\end{CJK}}} \\
    & 
    & \textcolor{forestgreen}{\begin{CJK}{UTF8}{min}ジュ・インジー\end{CJK}}\begin{CJK}{UTF8}{min}競技場のポジションは\end{CJK}
    & \\
\midrule
\multirow{2}{*}{\centering Chinese}
    & \multirow{2}{*}{\raggedright (\textcolor{forestgreen}{\begin{CJK}{UTF8}{gbsn}鞠盈智\end{CJK}}, \textcolor{darkmagenta}{\begin{CJK}{UTF8}{gbsn}中场\end{CJK}})}
    & \textcolor{forestgreen}{\begin{CJK}{UTF8}{gbsn}鞠盈智\end{CJK}}\begin{CJK}{UTF8}{gbsn}在运动场上的位置名字叫做\end{CJK}
    & \multirow{2}{*}{\centering \textcolor{darkmagenta}{\begin{CJK}{UTF8}{gbsn}中场\end{CJK}}} \\
    & 
    & \begin{CJK}{UTF8}{gbsn}在他/她从事的运动中,\end{CJK}\textcolor{forestgreen}{\begin{CJK}{UTF8}{gbsn}鞠盈智\end{CJK}}\begin{CJK}{UTF8}{gbsn}的位置是\end{CJK}
    & \\
\bottomrule
\end{tabular}
\end{center}
\caption{Prompts for the \textbf{\texttt{person\_sport\_position}} relation in all languages. We use the triple (\texttt{Ju Yingzhi}, \texttt{person\_sport\_position}, \texttt{midfielder}) as an example. The subject-object pair is represented in the respective language. The prompt shown below the dashed line is the new template introduced for the experiment described in \secref{effect_prompt}.}
\label{tab:prompts_person_sport_position}
\end{table*}

\begin{table*}[htbp]
\scriptsize
\centering
\setlength{\tabcolsep}{1.0mm}{}
\begin{center}
\begin{tabular}{m{0.10\linewidth} p{0.30\linewidth} p{0.30\linewidth} m{0.15\linewidth}}
\textbf{Language} & \textbf{\textcolor{forestgreen}{Subject}-\textcolor{darkmagenta}{Object} Pair} & \textbf{Prompt} & \textbf{Expected Output} \\
\toprule
\multirow{3}{*}{\centering English} 
  & \multirow{3}{*}{\raggedright (\textcolor{forestgreen}{Jeep Grand Cherokee}, \textcolor{darkmagenta}{Chrysler})} 
  & \textcolor{forestgreen}{Jeep Grand Cherokee} was created by which company? \newline Answer:
  & \multirow{3}{*}{\centering \textcolor{darkmagenta}{Chrysler}} \\
  & 
  & \textcolor{forestgreen}{Jeep Grand Cherokee} is a product of which company? \newline Answer:
  & \\
  \cdashline{3-3}[1.2pt/1pt]
  \addlinespace[2pt] 
  && Which company developed \textcolor{forestgreen}{Jeep Grand Cherokee}? It was developed by? Answer: & \\
\midrule
\multirow{2}{*}{\centering German} 
  & \multirow{2}{*}{\raggedright (\textcolor{forestgreen}{Jeep Grand Cherokee}, \textcolor{darkmagenta}{Chrysler})} 
  & Bitte geben Sie direkt den Firmen-/Ländernamen an. Das Unternehmen/Land, das \textcolor{forestgreen}{Jeep Grand Cherokee} entwickelt hat, ist
  & \multirow{2}{*}{\centering \textcolor{darkmagenta}{Chrysler}} \\
  & 
  & Bitte geben Sie direkt den Firmen-/Ländernamen an. Welches Unternehmen hat \textcolor{forestgreen}{Jeep Grand Cherokee} entwickelt? Es wurde entwickelt von
  & \\
\midrule
\multirow{2}{*}{\centering Spanish} 
  & \multirow{2}{*}{\raggedright (\textcolor{forestgreen}{Jeep Grand Cherokee}, \textcolor{darkmagenta}{Chrysler})} 
  & Por favor, responda directamente el nombre de la empresa/país. ¿Qué empresa desarrolló \textcolor{forestgreen}{Jeep Grand Cherokee}? Fue desarrollado por
  & \multirow{2}{*}{\centering \textcolor{darkmagenta}{Chrysler}} \\
  & 
  & Por favor responda directamente con el nombre de la empresa/país. La empresa que desarrolló \textcolor{forestgreen}{Jeep Grand Cherokee} se llama
  & \\
\midrule
\multirow{2}{*}{\centering French} 
  & \multirow{2}{*}{\raggedright (\textcolor{forestgreen}{Jeep Grand Cherokee}, \textcolor{darkmagenta}{Chrysler})} 
  & \textcolor{forestgreen}{Jeep Grand Cherokee} a été développé(e) par
  & \multirow{2}{*}{\centering \textcolor{darkmagenta}{Chrysler}} \\
  & 
  & \textcolor{forestgreen}{Jeep Grand Cherokee} est un produit de l'entreprise
  & \\
\midrule
\multirow{2}{*}{\centering Japanese} 
  & \multirow{2}{*}{\raggedright (\textcolor{forestgreen}{\begin{CJK}{UTF8}{min}ジープ・グランドチェロキー\end{CJK}}, \textcolor{darkmagenta}{\begin{CJK}{UTF8}{min}クライスラー\end{CJK}})} 
  & \begin{CJK}{UTF8}{min}会社名/国名を直接お答えください。\end{CJK}\textcolor{forestgreen}{\begin{CJK}{UTF8}{min}ジープ・グランドチェロキー\end{CJK}}\begin{CJK}{UTF8}{min}を開発したのはどの会社ですか? 開発したのは次の会社は\end{CJK}
  & \multirow{2}{*}{\centering \textcolor{darkmagenta}{\begin{CJK}{UTF8}{min}クライスラー\end{CJK}}} \\
  & 
  & \begin{CJK}{UTF8}{min}会社名/国名を直接お答えください。\end{CJK}\textcolor{forestgreen}{\begin{CJK}{UTF8}{min}ジープ・グランドチェロキー\end{CJK}}\begin{CJK}{UTF8}{min}を開発した会社は\end{CJK}
  & \\
\midrule
\multirow{2}{*}{\centering Chinese} 
  & \multirow{2}{*}{\raggedright (\textcolor{forestgreen}{\begin{CJK}{UTF8}{gbsn}吉普大切诺基\end{CJK}}, \textcolor{darkmagenta}{\begin{CJK}{UTF8}{gbsn}克莱斯勒\end{CJK}})} 
  & \begin{CJK}{UTF8}{gbsn}开发了\end{CJK}\textcolor{forestgreen}{\begin{CJK}{UTF8}{gbsn}吉普大切诺基\end{CJK}}\begin{CJK}{UTF8}{gbsn}的公司名字叫做\end{CJK}
  & \multirow{2}{*}{\centering \textcolor{darkmagenta}{\begin{CJK}{UTF8}{gbsn}克莱斯勒\end{CJK}}} \\
  & 
  & \begin{CJK}{UTF8}{gbsn}开发产品\end{CJK}\textcolor{forestgreen}{\begin{CJK}{UTF8}{gbsn}吉普大切诺基\end{CJK}}\begin{CJK}{UTF8}{gbsn}的公司名字叫\end{CJK}
  & \\
\bottomrule
\end{tabular}
\end{center}
\caption{Prompts for the \textbf{\texttt{product\_company}} relation in all languages. We use the triple (\texttt{Jeep Grand Cherokee}, \texttt{product\_company}, \texttt{Chrysler}) as an example. The subject-object pair is represented in the respective language. The prompt shown below the dashed line is the new template introduced for the experiment described in \secref{effect_prompt}.}
\label{tab:prompts_product_company}
\end{table*}

\begin{table*}[htbp]
\scriptsize
\centering
\setlength{\tabcolsep}{1.0mm}{}
\begin{center}
\begin{tabular}{m{0.10\linewidth} p{0.30\linewidth} p{0.30\linewidth} m{0.15\linewidth}}
\textbf{Language} & \textbf{\textcolor{forestgreen}{Subject}-\textcolor{darkmagenta}{Object} Pair} & \textbf{Prompt} & \textbf{Expected Output} \\
\toprule
\multirow{3}{*}{\centering English} 
    & \multirow{3}{*}{\raggedright (\textcolor{forestgreen}{50 Persei E}, \textcolor{darkmagenta}{Perseus})}
    & \textcolor{forestgreen}{50 Persei E} is part of the constellation named? \newline Answer:
    & \multirow{3}{*}{\centering \textcolor{darkmagenta}{Perseus}} \\
    \cdashline{3-3}[1.2pt/1pt]
    \addlinespace[2pt] 
    && What is the name of the constellation that \textcolor{forestgreen}{50 Persei E} is part of? It is part of? Answer: & \\
    && What is the name of the constellation that \textcolor{forestgreen}{50 Persei E} belongs to? It belongs to? Answer: & \\
\midrule
{\centering German} 
    & {\raggedright (\textcolor{forestgreen}{50 Persei E}, \textcolor{darkmagenta}{Perseus})}
    & Bitte geben Sie den Namen des Sternbildes direkt an. Das Sternbild, zu dem \textcolor{forestgreen}{50 Persei E} gehört, heißt
    & {\centering \textcolor{darkmagenta}{Perseus}} \\
\midrule
{\centering Spanish} 
    & {\raggedright (\textcolor{forestgreen}{50 Persei E}, \textcolor{darkmagenta}{Perseus})}
    & \textcolor{forestgreen}{50 Persei E} forma parte de la constelación denominada
    & {\centering \textcolor{darkmagenta}{Perseus}} \\
\midrule
{\centering French} 
    & {\raggedright (\textcolor{forestgreen}{50 Persei E}, \textcolor{darkmagenta}{Persée})}
    & Le nom de la constellation dans laquelle se trouve \textcolor{forestgreen}{50 Persei E} est
    & {\centering \textcolor{darkmagenta}{Persée}} \\
\midrule
{\centering Japanese} 
    & {\raggedright (\textcolor{forestgreen}{\begin{CJK}{UTF8}{min}50 ペルセウス座 E\end{CJK}}, \textcolor{darkmagenta}{\begin{CJK}{UTF8}{min}ペルセウス座\end{CJK}})}
    & \textcolor{forestgreen}{\begin{CJK}{UTF8}{min}50 ペルセウス座 E\end{CJK}}\begin{CJK}{UTF8}{min}はどの星座に属していますか？それは（星座名）という星座の一部です。\end{CJK}
    & {\centering \textcolor{darkmagenta}{\begin{CJK}{UTF8}{min}ペルセウス座\end{CJK}}} \\
\midrule
{\centering Chinese} 
    & {\raggedright (\textcolor{forestgreen}{\begin{CJK}{UTF8}{gbsn}50 英仙座E\end{CJK}}, \textcolor{darkmagenta}{\begin{CJK}{UTF8}{gbsn}英仙座\end{CJK}})}
    & \textcolor{forestgreen}{\begin{CJK}{UTF8}{gbsn}50 英仙座E\end{CJK}}\begin{CJK}{UTF8}{gbsn}所位于的星座名字叫做\end{CJK}
    & {\centering \textcolor{darkmagenta}{\begin{CJK}{UTF8}{gbsn}英仙座\end{CJK}}} \\
\bottomrule
\end{tabular}
\end{center}
\caption{Prompts for the \textbf{\texttt{star\_constellation}} relation in all languages. We use the triple (\texttt{50 Persei E}, \texttt{star\_constellation}, \texttt{Perseus}) as an example. The subject-object pair is represented in the respective language. The prompt shown below the dashed line is the new template introduced for the experiment described in \secref{effect_prompt}.}
\label{tab:prompts_star_constellation}
\end{table*}

\end{document}